\documentclass{article}

\usepackage{microtype}
\usepackage{graphicx}
\usepackage{wrapfig}
\usepackage{subcaption}
\usepackage{booktabs} 

\usepackage{hyperref}

\usepackage[accepted]{icml2024}

\usepackage{amsmath,amsfonts,bm}

\def\eqref#1{equation~\ref{#1}}

\def\Algref#1{Algorithm~\ref{#1}}

\def\1{\bm{1}}

\DeclareMathAlphabet{\mathsfit}{\encodingdefault}{\sfdefault}{m}{sl}
\SetMathAlphabet{\mathsfit}{bold}{\encodingdefault}{\sfdefault}{bx}{n}

\usepackage{amsmath}
\usepackage{amssymb}
\usepackage{mathtools}
\usepackage{amsthm}

\usepackage[capitalize,noabbrev]{cleveref}

\usepackage{float}
\usepackage{array}
\usepackage{makecell}
\usepackage{bbm}
\usepackage{enumitem}
\usepackage{multirow}
\usepackage[export]{adjustbox}
\usepackage{mdframed}
\usepackage{tocloft}
\usepackage{colortbl}
\definecolor{mylightblue}{rgb}{0.78,0.90, 0.96}
\definecolor{mydarkblue}{rgb}{0,0.08,0.45}
\definecolor{myred}{rgb}{0.84,0.17,0.11}
\definecolor{myyellow}{rgb}{0.95,0.69,0.10}
\definecolor{mygreen}{rgb}{0.17,0.63,0.17}
\definecolor{myorange}{rgb}{0.99,0.50,0.05}
\definecolor{myblue}{rgb}{0.07,0.43,0.69}
\definecolor{mypurple}{rgb}{0.58,0.40,0.71}
\definecolor{mypink}{rgb}{0.85, 0.38, 0.69}

\theoremstyle{plain}
\newtheorem{theorem}{Theorem}[section]
\newtheorem{proposition}[theorem]{Proposition}

\theoremstyle{definition}
\newtheorem{definition}[theorem]{Definition}

\theoremstyle{remark}

\usepackage[textsize=tiny]{todonotes}

\newcommand{\Revise}[1]{#1}
\newcommand{\New}[1]{#1}

\icmltitlerunning{Exploiting the Value of Past Success in Off-Policy
Actor-Critic}

\begin{document}

\twocolumn[
\icmltitle{Seizing Serendipity: Exploiting the Value of Past Success in Off-Policy Actor-Critic}



\icmlsetsymbol{equal}{}

\begin{icmlauthorlist}
\icmlauthor{Tianying Ji}{thucst}
\icmlauthor{Yu Luo}{thucst}
\icmlauthor{Fuchun Sun}{thucst}
\icmlauthor{Xianyuan Zhan}{air,sai}
\icmlauthor{Jianwei Zhang}{ham}
\icmlauthor{Huazhe Xu}{sai,iis,qizhi}
\end{icmlauthorlist}

\icmlaffiliation{thucst}{Department of Computer Science and Technology, Tsinghua University}
\icmlaffiliation{ham}{Department of Informatics, University of Hamburg}
\icmlaffiliation{air}{Institute for AI Industry Research, Tsinghua University}
\icmlaffiliation{sai}{Shanghai Artificial Intelligence Laboratory}
\icmlaffiliation{iis}{Institute for Interdisciplinary Information Sciences, Tsinghua University}
\icmlaffiliation{qizhi}{Shanghai Qi Zhi Institute}
\icmlcorrespondingauthor{Fuchun Sun}{fcsun@tsinghua.edu.cn}

\icmlkeywords{Machine Learning, ICML}

\vskip 0.3in
]



\printAffiliationsAndNotice{}  

\newcommand{\ourshort}{BAC}
\newcommand{\mbshort}{MB-BAC}
\newcommand{\our}{BEE Actor-Critic}

\begin{abstract}
Learning high-quality $Q$-value functions plays a key role in the success of many modern off-policy deep reinforcement learning~(RL) algorithms. 
Previous works primarily focus on addressing the value overestimation issue, an outcome of adopting function approximators and off-policy learning. 
Deviating from the common viewpoint, we observe that $Q$-values are often underestimated in the latter stage of the RL training process, potentially hindering policy learning and reducing sample efficiency. We find that such a long-neglected phenomenon is often related to the use of inferior actions from the current policy in Bellman updates as compared to the more optimal action samples in the replay buffer.
To address this issue, our insight 
is to incorporate sufficient exploitation of past successes while maintaining exploration optimism.
We propose the Blended Exploitation and Exploration~(BEE) operator, a simple yet effective approach that updates $Q$-value using both historical best-performing actions and the current policy.
Based on BEE, the resulting practical algorithm BAC outperforms state-of-the-art methods in \textbf{over 50} continuous control tasks and achieves strong performance in failure-prone scenarios and \textbf{real-world robot} tasks. Benchmark results and videos are available at \textcolor{blue}{\url{https://jity16.github.io/BEE/}}.
\end{abstract}

\addtocontents{toc}{\protect\setcounter{tocdepth}{-1}}
\section{Introduction}
Reinforcement learning (RL) has achieved impressive progress in solving many complex decision-making problems in recent years~\citep{dqn, silver2016mastering,hutter2016anymal,ouyang2022training}. Many of these advances are obtained by off-policy deep RL algorithms, where the ability to leverage off-policy samples to learn high-quality value functions underpins their effectiveness.
Value overestimation~\citep{td3,top} has long been recognized as an important issue in off-policy RL algorithms, which is primarily associated with the function approximation errors~\citep{td3} and the side-effect of off-policy learning~\citep{auer2008near,jin2018q,azar2017minimax}, and can potentially lead to suboptimal policies.
To tackle this issue, techniques for alleviating value overestimations,
such as double-Q technique, have been ubiquitously adopted in modern off-policy RL algorithms~\citep{sac,laskin2020reinforcement,dac,top}.

\begin{figure}[t]
    \centering
    \includegraphics[height=3.8cm,keepaspectratio]{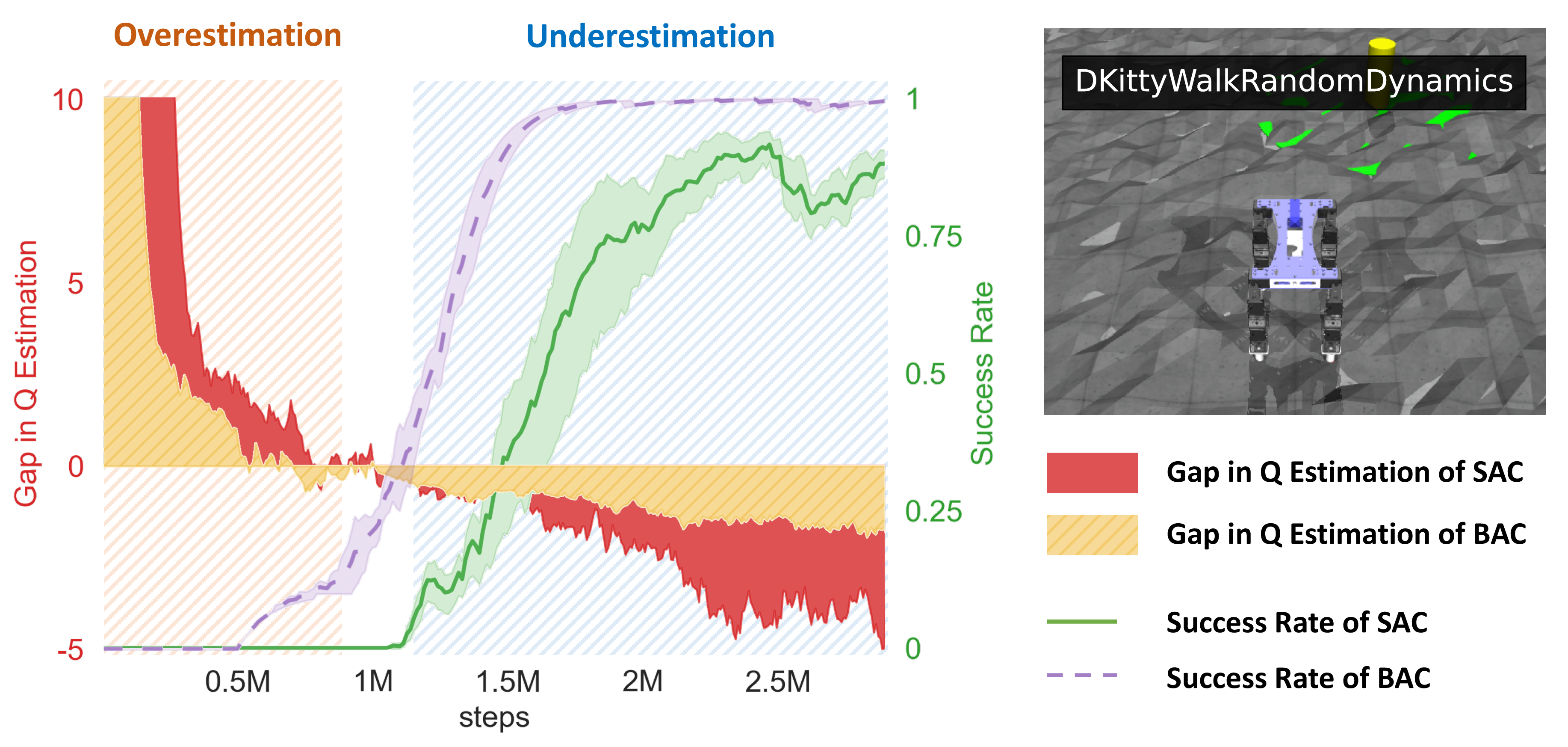} 
    \vspace{-3mm}
    \caption{\small \textbf{Example of the underestimation issue.}  In the DKittyWalkRandomDynamics task, when the current policy-generated action is inferior to the best action in the replay buffer, which usually occurs in the later stage of training, SAC is more prone to underestimation pitfalls than BAC.
    The gap in $Q$ estimation is evaluated by comparing the SAC $Q$-values and the Monte-Carlo $Q$ estimates.}
    \label{fig:visualization}
    \vspace{-5mm}
\end{figure}

\New{
Intriguingly, we find that in common online off-policy actor-critic~(AC) methods, the well-known value overestimation issue could disappear and be replaced by value underestimation in the later training stage. 
Figure~\ref{fig:visualization} shows an illustrative example of such a phenomenon in a robotic task, and we also observe similar patterns over a diverse range of RL tasks and different off-policy AC algorithms, as will be demonstrated in the following content of our paper.
Such a phenomenon does not simply result from the techniques used for alleviating value overestimation (e.g., double-Q trick), but more fundamentally, because 
the $Q$-value update could be negatively impacted by the actions $a'$ sampled from the current suboptimal policy during the Bellman Backup process.
Such suboptimality is inevitable since the policy optimization updates over the current $Q$-value function (i.e., $\pi\leftarrow\arg\max_{a\in\pi} Q(s,a)$) is often impossible to recover the ideal policy within only a few policy gradient updates, especially with an evolving $Q$ network.
This can potentially lead to underestimated $Q$ target update values, leading to inferior learning performance and sample efficiency.
}

\New{
To address this long-neglected phenomenon, we find that allowing the agent to fully exploit the best-performing actions stored in the replay buffer can be a natural cure. More concretely, if there exist more optimal actions in the replay buffer as compared to the ones generated by the current policy, then we can leverage them to perform more optimistic Bellman updates to resolve the underestimation issue. Such more optimal actions can be abundant 
especially in the later off-policy RL training stage, since the replay buffer is already filled by relatively diverse state-action pairs, which may be caused by exploration behaviors in online RL and the non-optimality optimization nature of the actor-critic framework.
}

In this paper,  we connect this intuition with Bellman operators: the Bellman Exploitation operator enables effective exploitation of high-quality historical samples while the Bellman Exploration operator targets maintaining exploration optimism. 
This gives rise to a remarkably simple and effective mechanism, the Blended Exploration and Exploitation~(BEE) operator, which combines the merits of both sides.
BEE operator provides superior $Q$-value estimation, effectively avoiding the value underestimation issue.  
Moreover, it can be flexibly integrated into existing off-policy AC frameworks, leading to two practical algorithm instantiations: \ourshort~(\our) for model-free settings and \mbshort~(Model-based BAC) for model-based settings.

\ourshort\ outperforms 
other popular online RL methods
on various \textbf{MuJoCo}, \textbf{DMControl}, \textbf{Meta-World}, \textbf{ManiSkill2}, \textbf{Adroit}, \textbf{Shadow Dexterous Hand}, \textbf{MyoSuite}, \textbf{ROBEL} benchmark tasks by a large margin. 
On many failure-prone tasks, \ourshort\ achieves over 2x the evaluation scores of the strongest baseline. 
Moreover, we conduct \textbf{real-world} experiments on complex quadruped robot locomotion tasks, and \ourshort\ achieves strong performance. 
Furthermore, in our experiments, we observed unanimously improved performance when applying the BEE operator to different backbone algorithms, highlighting its flexibility and generic nature.

\section{Related Works}\label{related_works}
\New{
Off-policy actor-critic methods, alternating between $Q$-value estimation and policy optimization w.r.t the $Q$-value, have been a cornerstone in reinforcement learning~(RL) research~\citep{ddpg, a3c, sac, td3,lee2020stochastic,zheng2022stackelberg}. A widely recognized challenge in these methods is the accuracy of value estimation~\cite{kimura1998analysis,grondman2012survey}, which is crucial for effective policy extraction. Inaccurate value estimations can significantly hinder policy updates and misguide exploration efforts.}

\New{
Overestimation could erroneously attribute high values to suboptimal states and actions~\citep{hasselt2010double, td3,cetin2023learning}. It has been a long-discussed problem. Previous works suggest overestimation is an outcome of the combination of off-policy learning and high-dimensional, nonlinear function approximation~\citep{hasselt2010double, td3,kuznetsov2020controlling}, also associated with optimistic exploration~\citep{jin2018q,laskin2020reinforcement,top}. Hence, attempts to correct for overestimation, \emph{e.g.}, taking the minimum of two separate critics, have been widely adopted in AC methods~\citep{td3, sac, dac, rrs}.}

\New{
While underestimation could hamper the reselection of potentially high-value state-action pairs and thus negatively impact policy optimization, it has received much less attention compared to overestimation.
Some existing efforts may blame the underestimation issues for directly taking the minimum value from an ensemble of $Q$-values~(a technique originally intended to combat overestimation), thus devising a less extreme form of the minimum value to avoid overly conservative estimates~\cite {oac, top, peer2021ensemble}. }

\New{
However, attributing underestimation solely to the clipped double-Q technique is an oversimplification. Our findings suggest that the inherent nature of actor-critic optimization in RL also contributes to underestimation. This indicates that previous works that focused solely on adjusting the double-Q technique may not fully address the issue.
A more comprehensive approach that considers the underlying interplay of actor-critic methods is necessary to effectively address underestimation in reinforcement learning.
}

\New{For extensive related works, please refer to Appendix~\ref{ap:extensive_related_works}.}

\section{Preliminaries}
\noindent\textbf{Markov decision process.}\quad 
We denote a discounted Markov decision process~(MDP) as $\mathcal{M} = ({\cal S},{\cal A}, P, r, \gamma)$, where ${\cal S}$  denotes the state space, ${\cal A}$  the action space,  $r: {\cal S\times A}\in [-R_{max}, R_{max}]$  the reward function, and $\gamma\in (0,1)$ the discount factor, and $P(\cdot \mid s,a)$ stands for transition dynamics. 

\noindent\textbf{Off-policy actor-critic RL.}\quad 
\New{In Q-learning, the policy is derived from the state-action value function $Q$ by selecting the maximizing action. The learning of the 
Q-value involves repeated application of the Bellman optimality operator, \emph{i.e.}, $\mathcal{T}^*Q(s,a)\triangleq r(s,a)+\gamma\cdot\max_{a'\in \mathcal{A}}\mathbb{E}_{s'\sim P(s'\vert s,a)}[Q(s',a')]$. However, it entails traversing all possible actions, being intractable in continuous action spaces~\citep{bear,garg2023extreme}. }
\New{Off-policy actor-critic methods tackle this issue by alternating between training a policy to maximize the $Q$ value, denoted as $\pi =\arg\max_{a\sim \pi}Q(s,a)$,
and repeatedly applying a Bellman evaluation operator $\mathcal{T}$~\citep{sutton1988learning,watkins1989learning}, defined as:}
\begin{equation*}
\mathcal{T} Q(s, a) \triangleq r(s, a) + \gamma \cdot \mathbb{E}_{s' \sim P(\cdot | s, a)} \left[ \mathbb{E}_{a' \sim \pi(\cdot | s')} [Q(s', a')] \right].
\end{equation*}

\noindent\textbf{In-sample learning.}\quad
In-sample learning paradigms have been widely explored in offline RL. In-sample learning methods~\citep{IQL,xuoffline,garg2023extreme} learn $V(s)$ and $Q(s,a)$ completely using dataset samples, following general learning objectives for $V(s)$ and $Q(s,a)$, depending on different choices of the $f$ function. 
$\min_V \mathbb{E}_{(s,a)\sim \mathcal{D}} \mathcal{L}_V^f (Q(s,a)-V(s))$ for updating $V$-function and $\max_Q \mathbb{E}_{(s,a,s')\sim \mathcal{D}}[r(s,a) +\gamma V(s') - Q(s,a)]^2$ for updating $Q$-function. 
The well-known offline RL algorithm IQL also belongs to this family with $\mathcal{L}_V^f(y) = \vert \tau - \mathbbm{1}(y<0)\vert y^2$, where $\tau\in (0,1)$ is the expectile hyperparameter. 
IQL can recover the optimal value function under the dataset support constraints. In particular, if $f=\log(x)$, it corresponds to EQL~\citep{xuoffline} and XQL~\citep{garg2023extreme} with
$\mathcal{L}_V^f(y)=\exp(y/ \alpha)-y/ \alpha$.
If $f=x-1$, it corresponds to SQL~\citep{xuoffline} with $\mathcal{L}_V^f(y)=\mathbbm{1} (1+y/ 2\alpha>0)(1+y/ 2 \alpha)^2-y\ /alpha$. These Insights from the good practices of in-sample learning methods in offline RL could be leveraged to enhance online RL $Q$-value estimation.

\begin{figure*}[t]
    \centering
    \includegraphics[height=6.5cm,keepaspectratio]{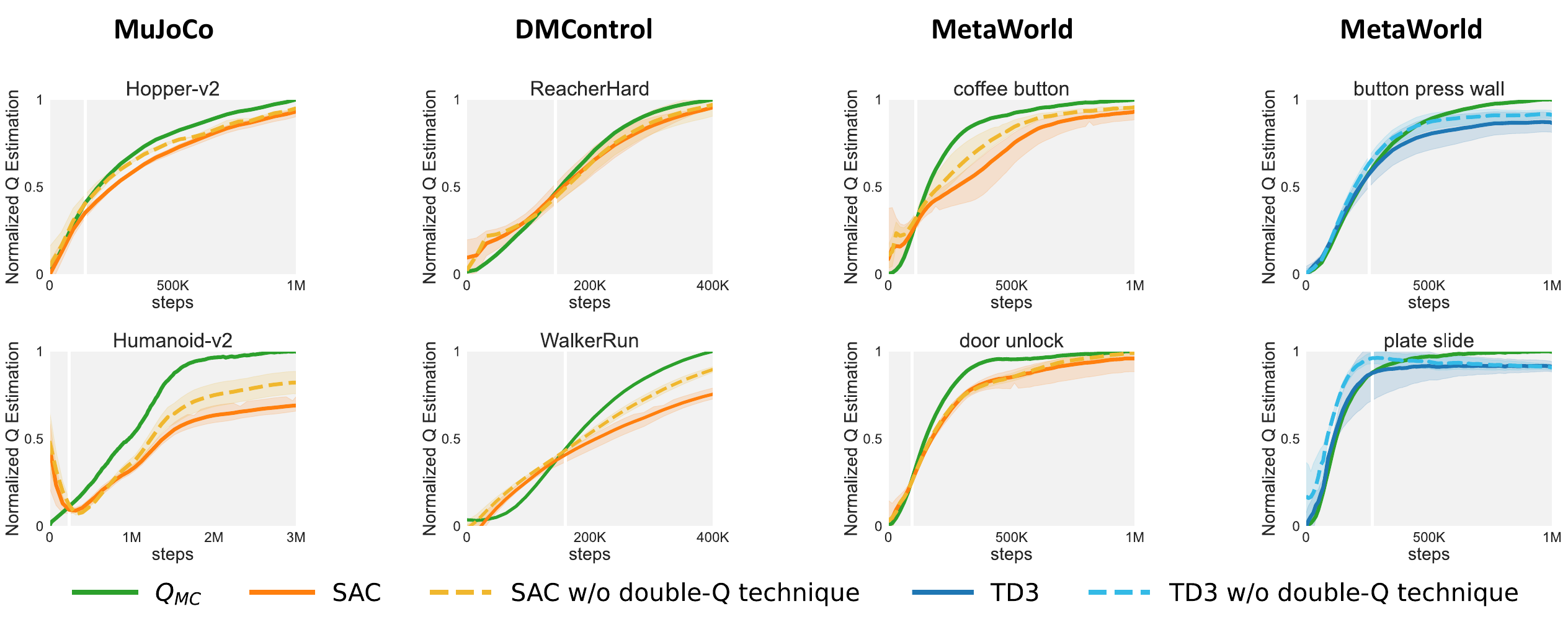}
    \vspace{-2.5mm}
    \caption{\small \textbf{Normalized $Q$ estimation value comparison.} 
    We plot the normalized $Q$-value estimates of two widely-used off-policy actor-critic algorithms, namely SAC and TD3, across various benchmark suites, spanning locomotion and manipulation tasks. The approximated true value $Q_{MC}$ is obtained by Monte-Carlo $Q$ estimates are derived from trajectories sampled using the current policy. The white vertical line marks the separation of the overestimation and underestimation stages during training.
    Moreover, we disabled the double-Q technique of SAC and TD3 and observed the underestimation issue still occurs.}
    \label{fig:withoutdoubleQtrick_4pages}
\end{figure*}

\section{Exploiting past success for off-policy optimization}
In this section, we delve into the long-neglected underestimation bias in the latter stage of the RL training process. 
We identify that this bias can be partially attributed to the inherent non-optimality of the current policy in the Actor-Critic framework\footnote{For comprehensive investigations of the underestimation issue, please see Appendix~\ref{section:appendix-underestimation}.}. 
These discoveries encourage us to exploit the more optimal actions in the replay buffer to shorten the gap to the optima, hence mitigating underestimation. Finally, we arrive at a simple, versatile, yet effective solution, the BEE operator, and demonstrate its effectiveness.

\subsection{Underestimation issue in Actor-Critic}
The underestimation bias has long been attributed to the double-Q-technique, as highlighted in previous studies~\citep{td3, top}. Yet, one must consider whether this technique is the sole cause of the bias.

In Figure~\ref{fig:withoutdoubleQtrick_4pages}, we plot the $Q$-value estimation for SAC and TD3, alongside their variants without the double-Q technique. Our observations indicate that both SAC and TD3 encounter an underestimation issue across various robotic tasks. 
Intriguingly, even when we eliminate the double-Q-technique from well-known algorithms SAC~\citep{sac} and TD3~\citep{td3}, the problem of underestimation still persists. This suggests the existence of other, less explored factors contributing to the underestimation issue.
We identify the underlying optimization procedure in the Actor-Critic framework itself may also contribute to underestimation. 

\noindent\textbf{The optimization procedure of the AC framework contributes to underestimation.}\quad 
Ideally, the Bellman update needs to solve $Q(s,a)\leftarrow r(s,a)+\gamma \mathbb{E}_{s'}[\max_{a'} Q(s',a')]$. However, as $\max_{a'} Q(s',a')$ operations are often impractical to calculate, so in the AC framework, we typically iteratively evaluate target Q-value as $\mathbb{E}_{\pi}[Q(s',a')]$, while implicitly conducting the $\max$-$Q$ operation in a separate policy improvement step to learn policy $\pi$. 
Note that the ideal $\pi=\arg\max_{a'\sim \pi} Q(s',a')$ is not possible to achieve practically within only a few policy gradient updates~\citep{BCQ,chan2022greedification}. Hence, the actual target value used in AC Bellman update $\mathbb{E}_{s',a'\sim \pi}Q(s',a')$ can have a high chance to be smaller than $\mathbb{E}_{s'}[\max_{a'} Q(s',a')]$, causing underestimation. 

In a nutshell, the inherent non-optimality of the current policy in the AC framework contributes to underestimation.  Specifically, the discrepancy between the theoretically optimal policy and the practically achievable policy, especially in the latter stages of learning,  would negatively affect the target update value. This is because the target-update actions $a'$ sampled from the current policy may be inferior compared to those generated by an ideally optimal policy.

\begin{figure}[t]
    \vspace{-0.2cm}
    \centering
    \includegraphics[height=2.7cm,keepaspectratio]{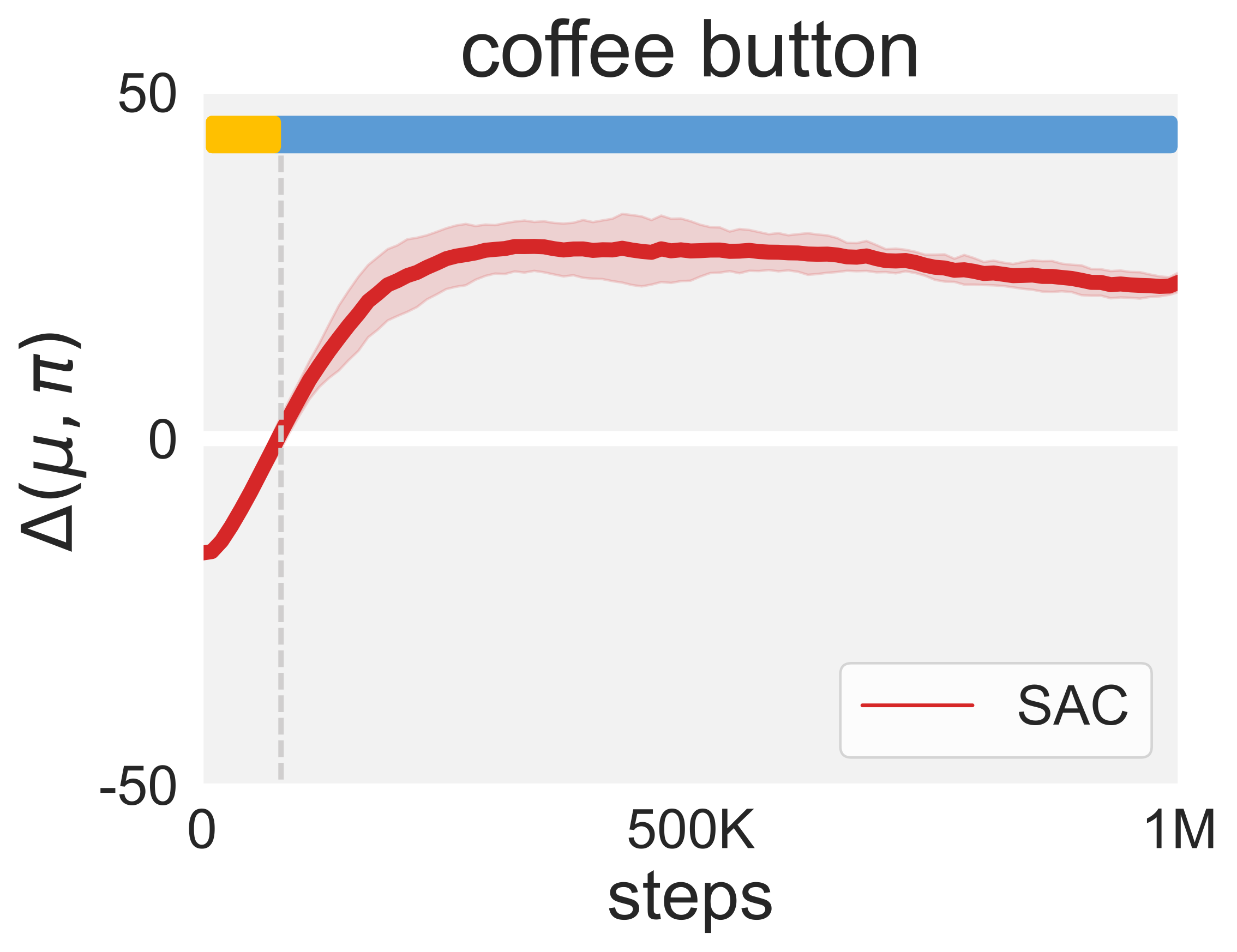}
        \includegraphics[height=2.7cm,keepaspectratio]{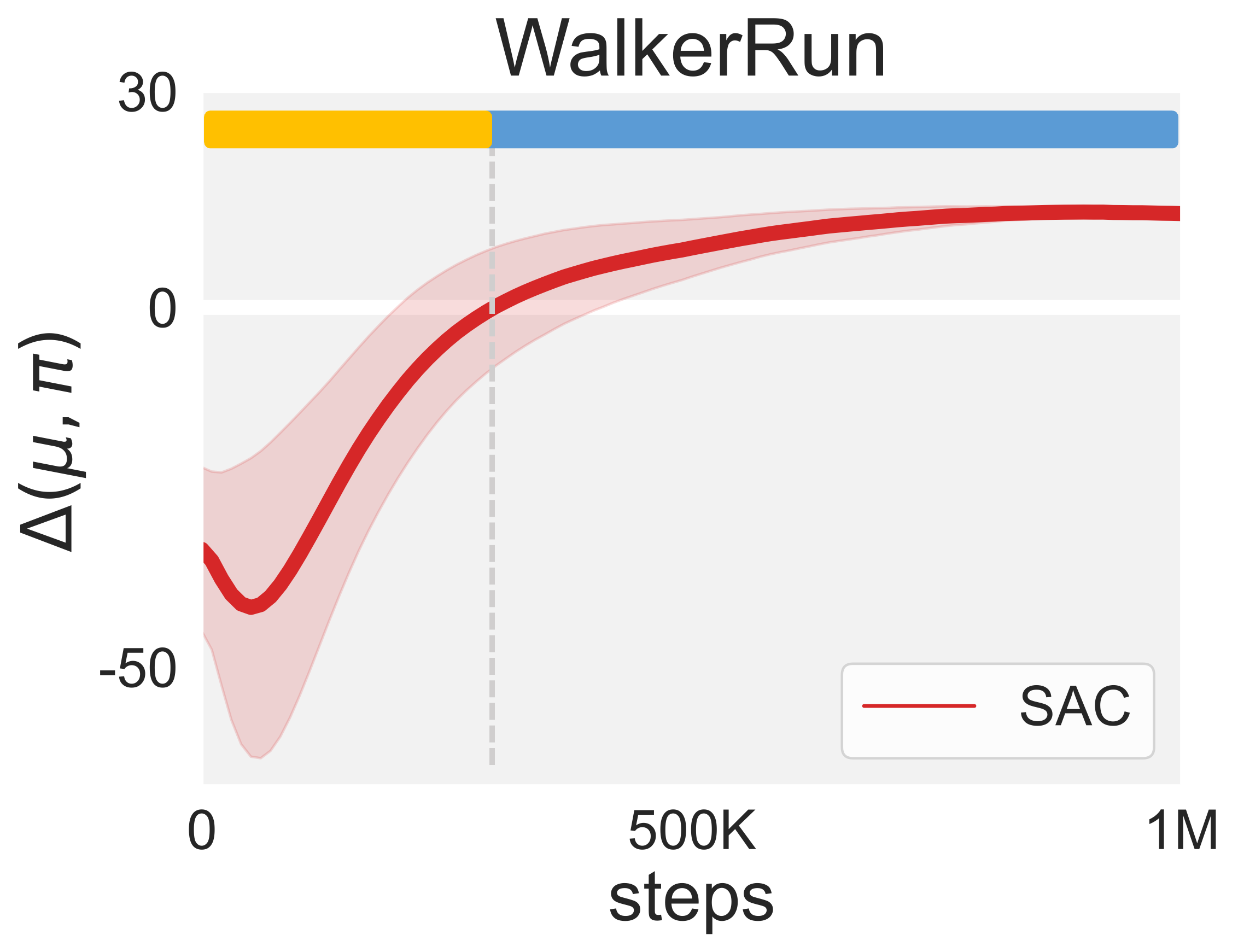}\\
        \includegraphics[height=2.7cm,keepaspectratio]{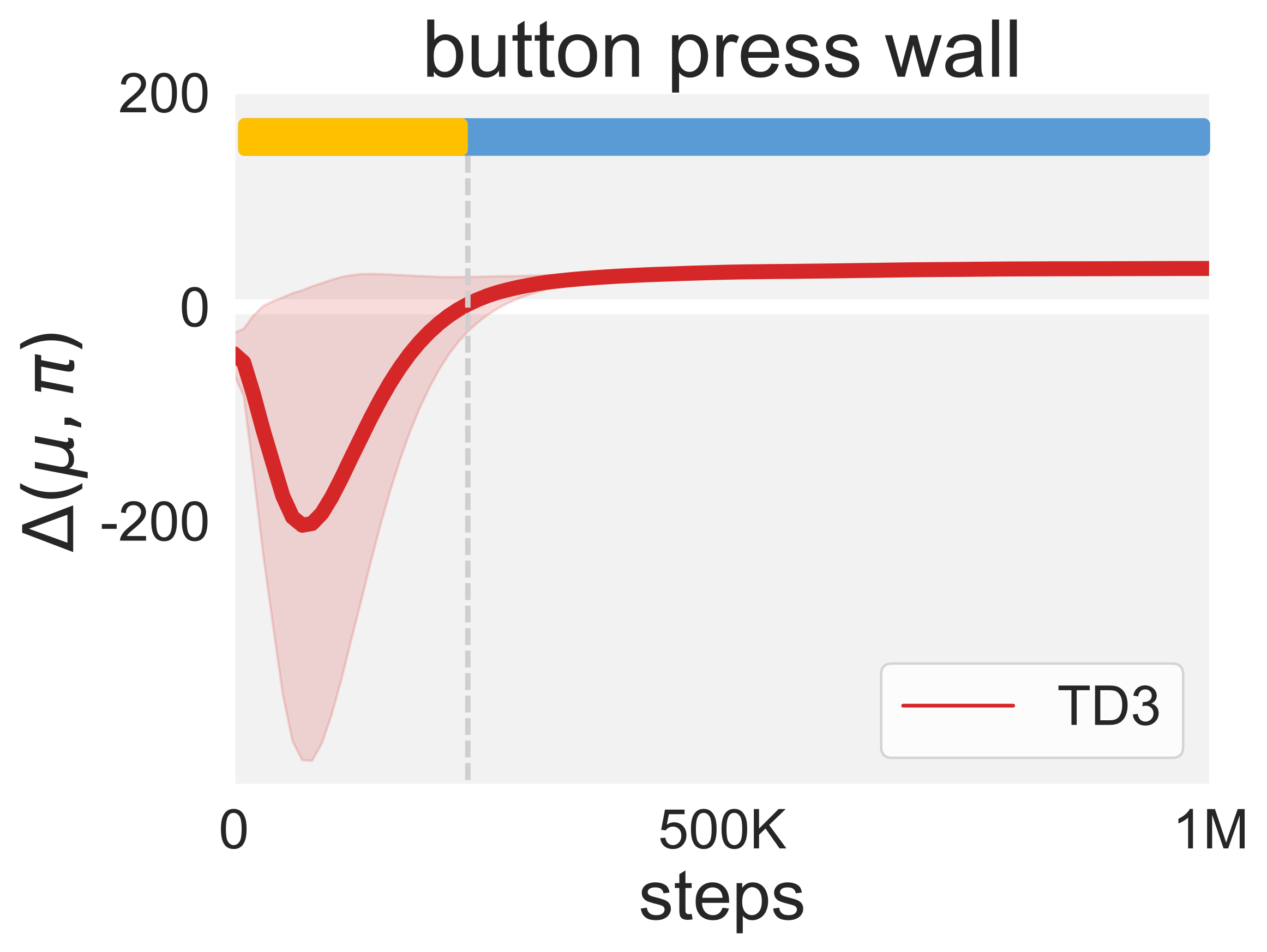}
        \includegraphics[height=2.7cm,keepaspectratio]{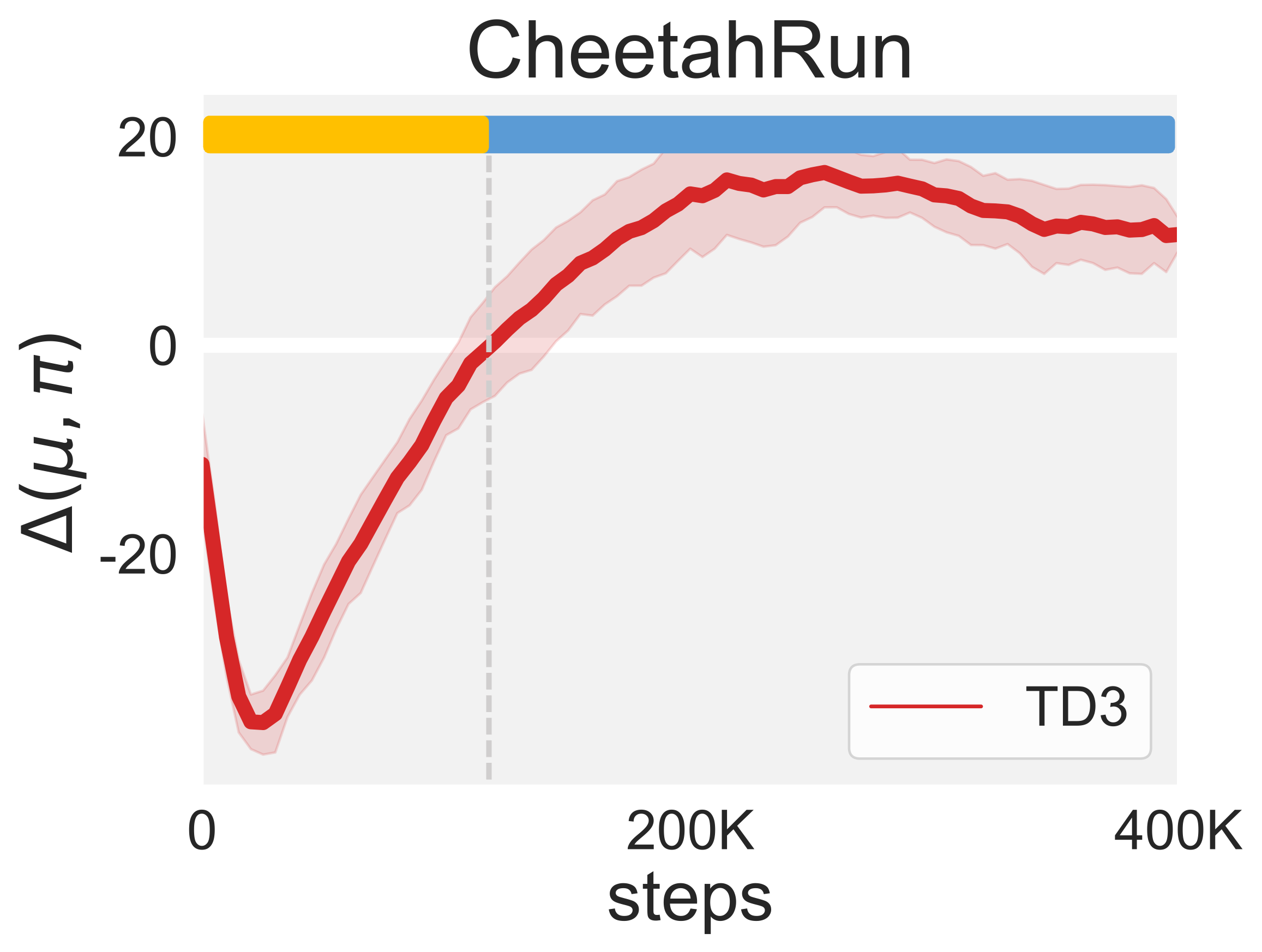}
        \vspace{-0.25cm}
        \caption{\small \textbf{$\Delta(\mu,\pi)$ with an SAC or TD3 agent.} Blue bars correspond to positive $\Delta(\mu,\pi)$, indicating an optimal policy derived from the replay buffer
would outperform the current policy.}
        \label{fig:delta-visualization-two}
    \vspace{-0.2cm}
\end{figure}
\begin{figure}[t]
    \centering
    \vspace{-0.12cm}
    \includegraphics[height=4.3cm,keepaspectratio]{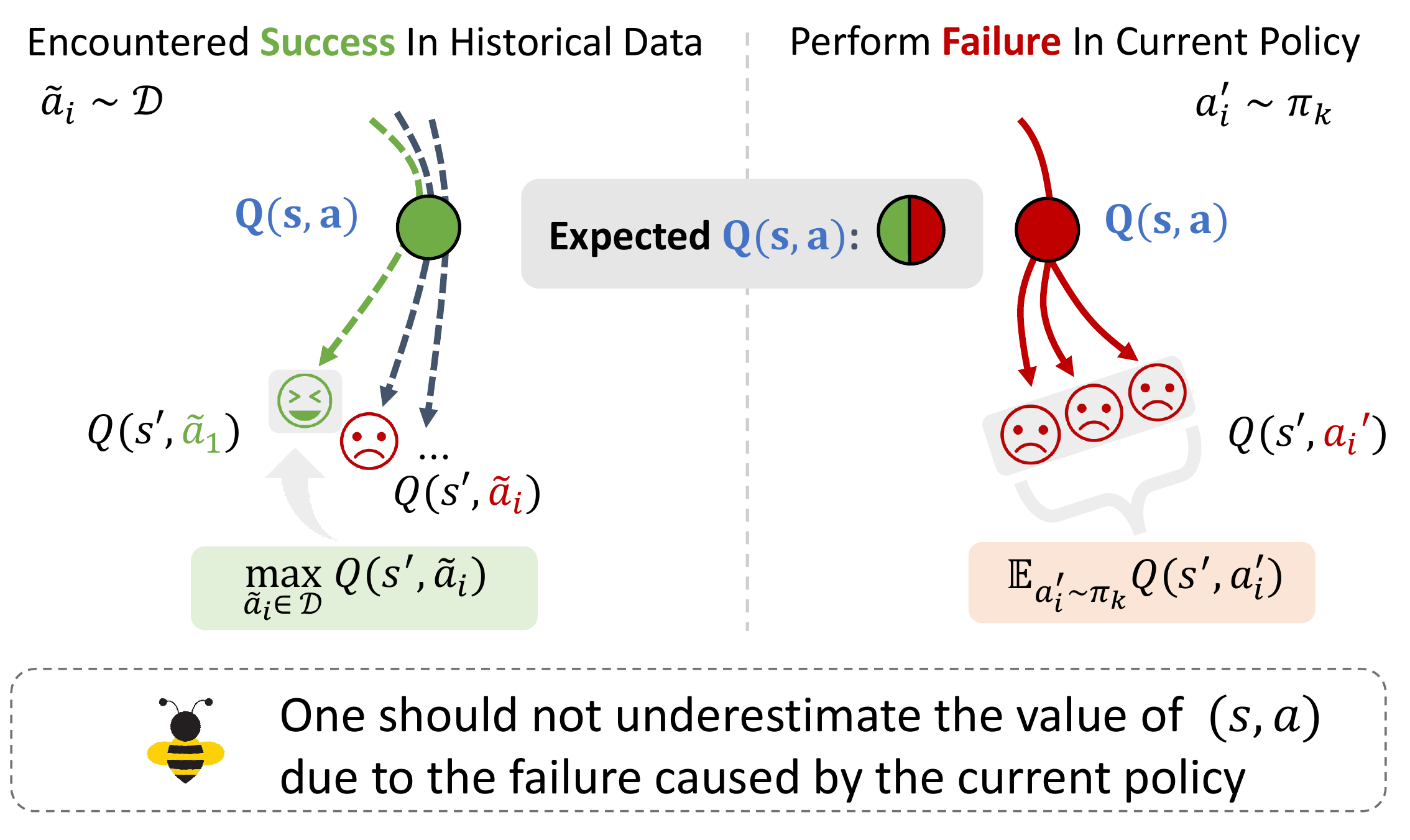} 
    \vspace{-0.12cm}
    \caption{\small 
    \textbf{Illustrative figure on target-update actions.} Leveraging the high-quality samples from the replay buffer would produce a more optimistic $Q$-value, thus mitigating value underestimation. Practically, the Bellman evaluation operator, whose target-update actions $a'$ are only sampled from the current policy, tends to underestimate it.}
    \label{fig:motivation}
    \vspace{-0.3cm}
\end{figure}

Hence, reducing the gap between target-update actions and those of an ideal optimal policy could lessen the underestimation issue. 
Now, the critical challenge lies in identifying a more optimal source for these target-update actions.

\begin{figure*}[t]
    \centering
    \begin{subfigure}[b]{0.24\textwidth}
        \hspace{7mm}
        \includegraphics[height=2.5cm,keepaspectratio]{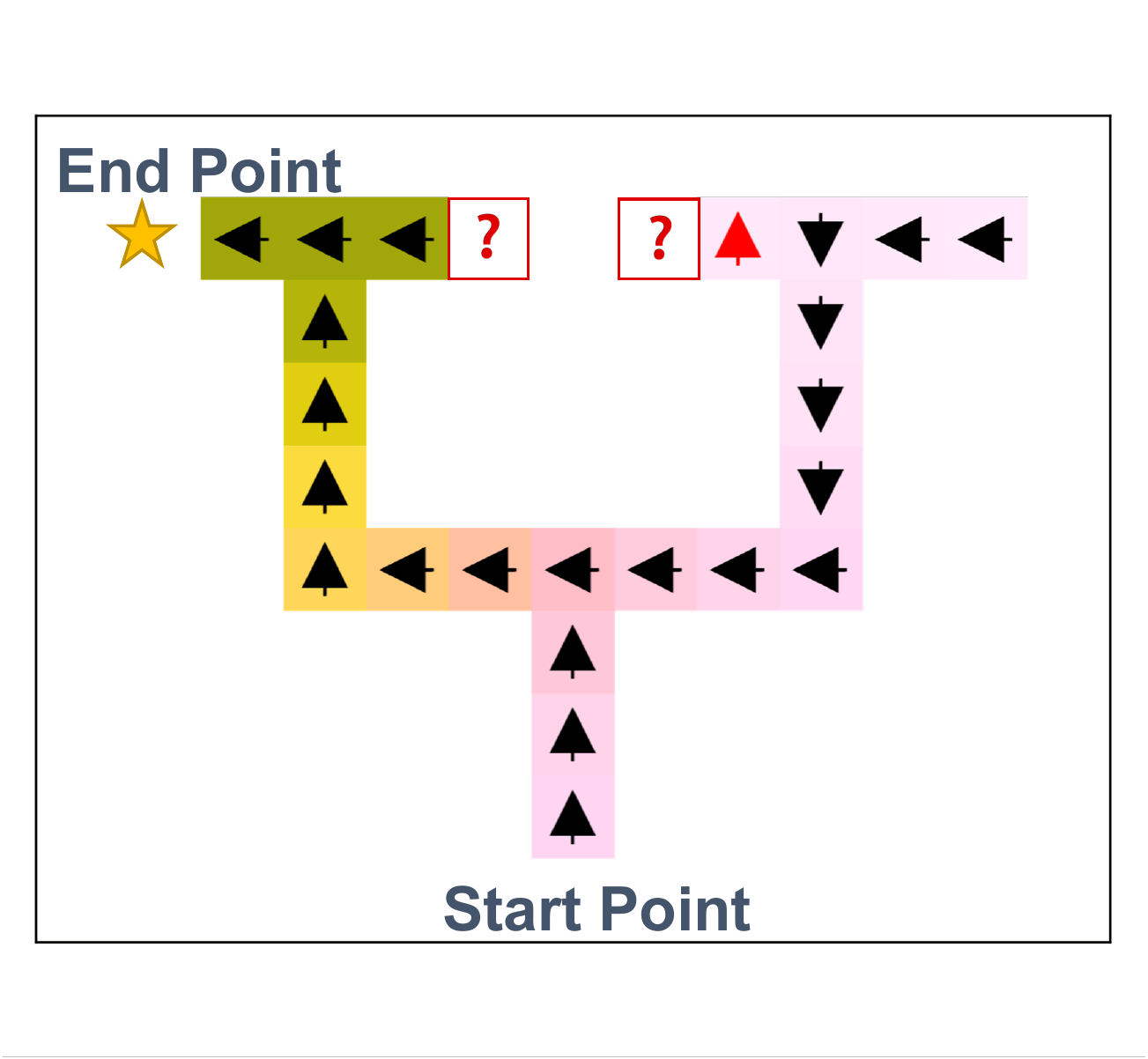}
        \vspace{-0.1cm}
        \caption{\small $Q$ from pure $\mathcal{T}_{exploit}$}
    \end{subfigure}
    \begin{subfigure}[b]{0.24\textwidth}
        \hspace{7mm}
        \includegraphics[height=2.5cm,keepaspectratio]{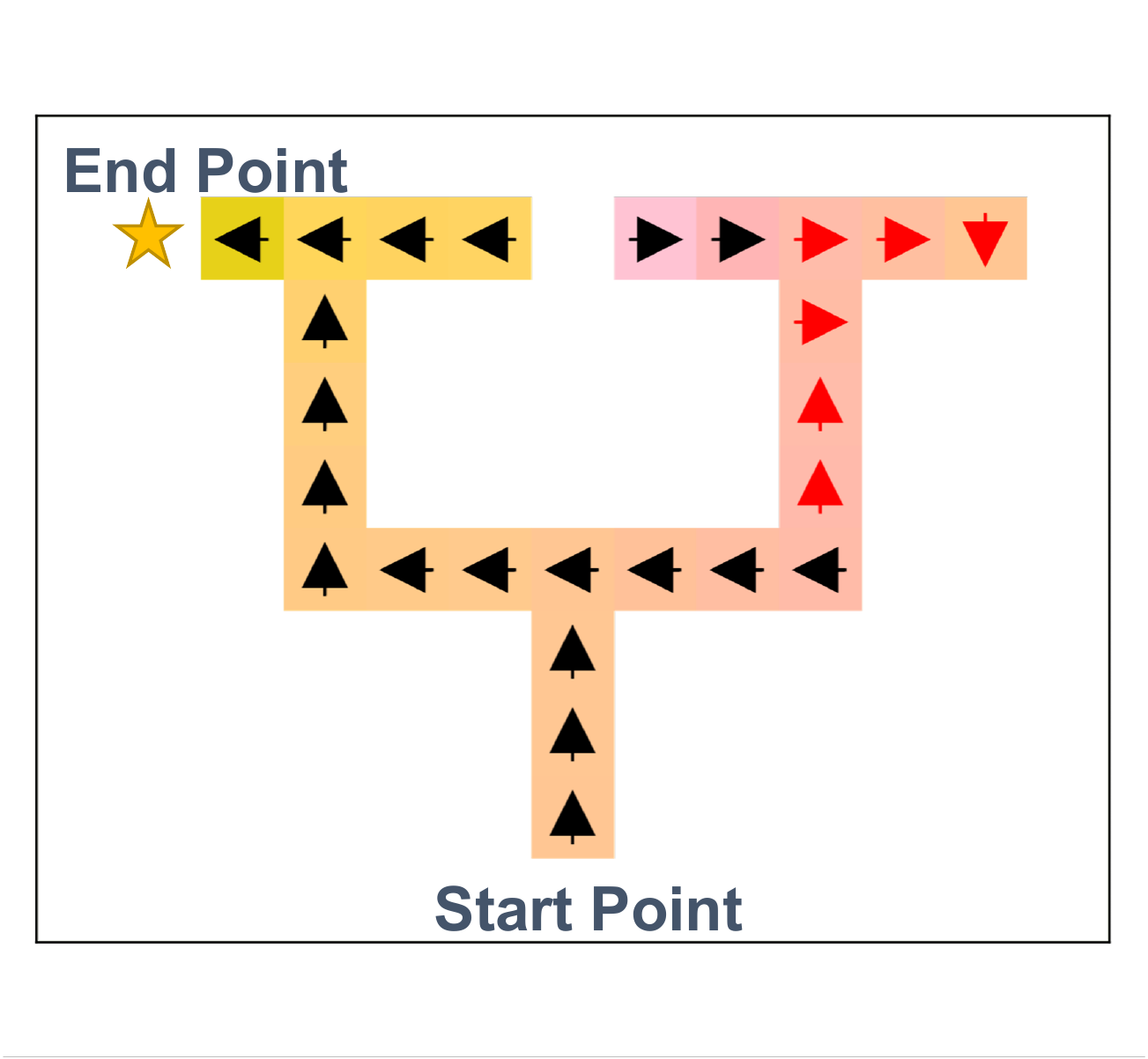}
        \vspace{-0.1cm}
        \caption{\small $Q$ from pure $\mathcal{T}_{explore}$}
    \end{subfigure}
    \begin{subfigure}[b]{0.24\textwidth}
        \hspace{7mm}
        \includegraphics[height=2.5cm,keepaspectratio]{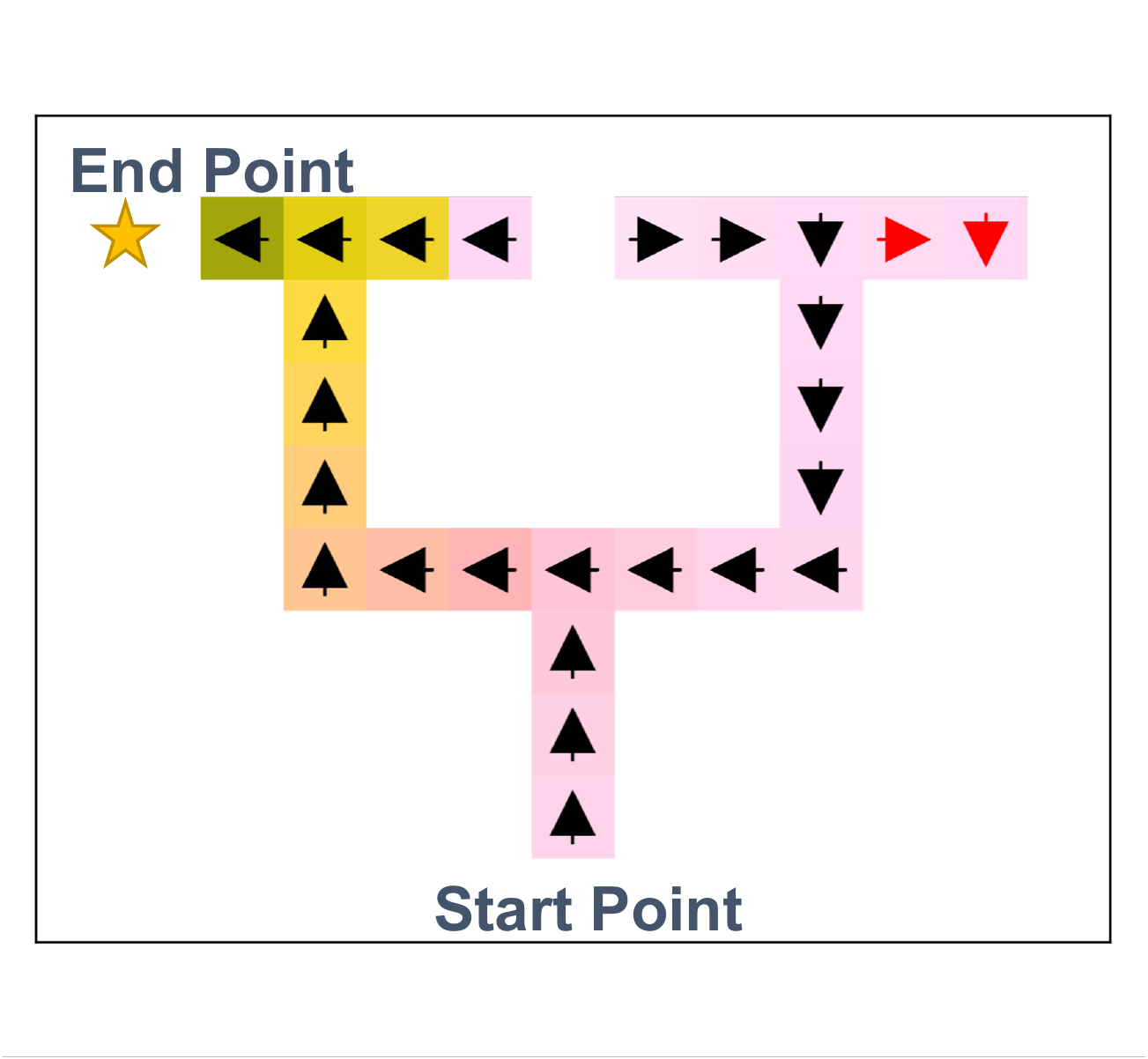}
        \vspace{-0.1cm}
        \caption{\small $Q$ from our $\mathcal{B}$}
    \end{subfigure}
    \begin{subfigure}[b]{0.24\textwidth}
        \hspace{6mm}
        \includegraphics[height=2.5cm,keepaspectratio]{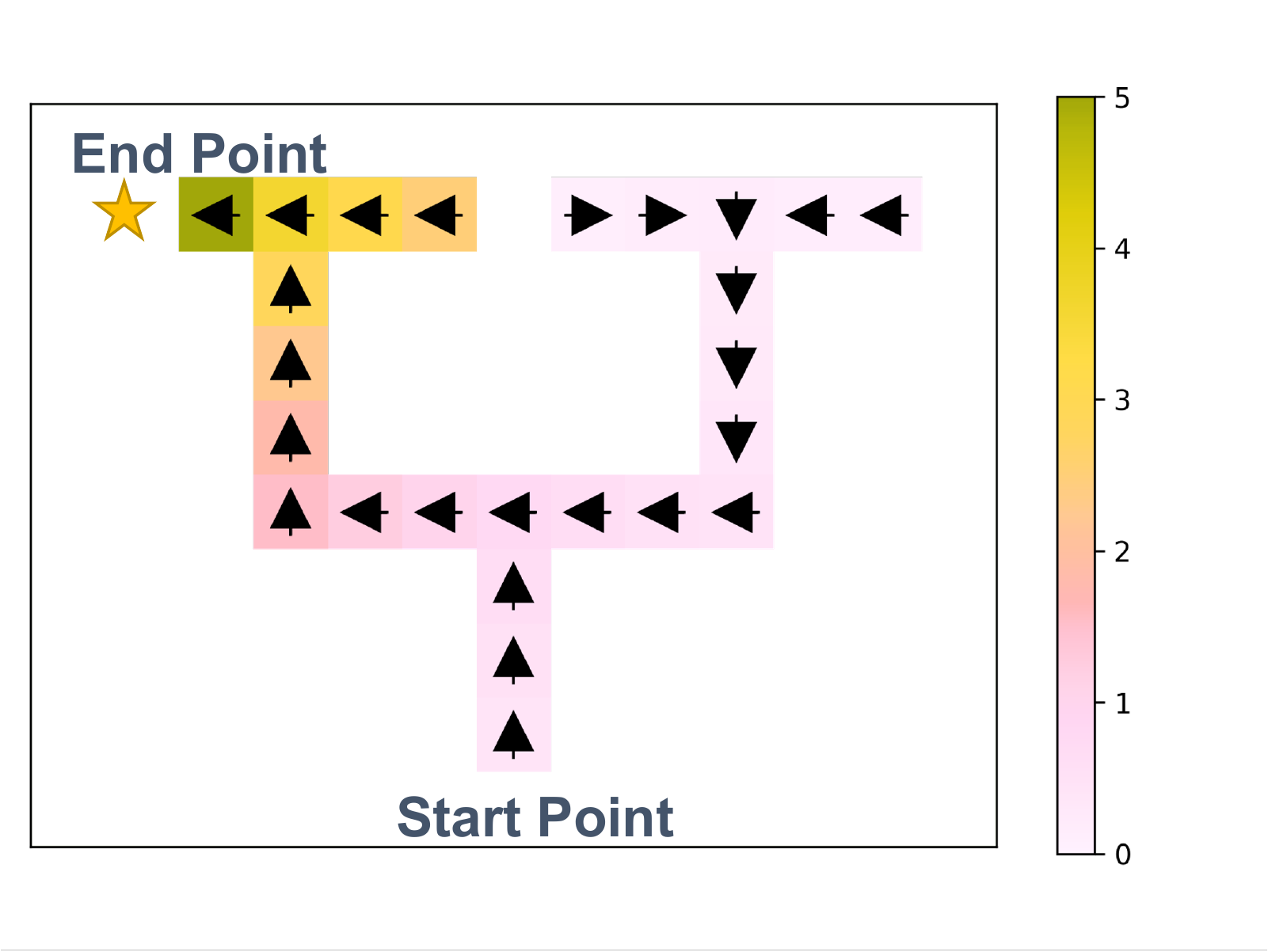}
        \vspace{-0.1cm}
        \caption{\small Optimal $Q$ and actions}
    \end{subfigure}
    \vspace{-0.1cm}
    \caption{\small \textbf{Comparison of three operators on a toy grid world.} The agent's goal is to navigate from the bottom of the maze to the top left. The color of each square shows the learned value, red arrows reveal incorrect actions, and question marks indicate unencountered states. (a) employing a pure exploitation operator may lead to insufficient exploration, causing the agent to miss out on blocks. Conversely, (b) a pure exploration operator $\mathcal{T}_{explore}^\pi$ tends to overestimate the value of less frequently visited areas while underestimating those on the optimal path, resulting in increased sample complexity, (c) Our BEE operator, however, balances between exploitation and exploration, improving convergence and accuracy. And (d) reveals the optimal Q and actions.}
    \label{grid-world-visualization}
    \vspace{-0.3cm}
\end{figure*}
\subsection{More optimal source for target-update actions}
Actually, actions sampled from the non-optimality of the current policy may fall short of the optimal ones stored in the replay buffer. Naturally, exploiting the optimal actions from the replay buffer to bootstrap $Q$ would shorten the gap to the optima, hence mitigating underestimation.

\New{To discern the optimal actions from the replay buffer, it's essential to characterize the implicit policy that can be inferred from it. During policy learning, the replay buffer accumulates samples from a series of historical policies, denoted as $\Pi^k = \{\pi_0, \pi_1,\ldots,\pi_k\}$, each associated with a mixture distribution weight $\alpha_k$. 
Hence, the state-action visitation density in replay buffer is defined as $d^{\mu_k}(s,a) = \sum_{i=1}^{k}\alpha_i^k d^{\pi_i}(s,a)$, where $\mu_k$ is a mixed policy derived from the replay buffer~\citep{made,wang2022live}.}

To quantify the discrepancy between the value of the optimal actions derived from the replay buffer and those from the current policy, we 
calculate the expected difference $\Delta(\mu_k, \pi_k)$ between the maximum $Q$-value over the policy mixture $\mu_k$ and the expected $Q$-value under the existing policy $\pi_k$ at policy iteration $k$, stated as, 
\begin{equation*}
\Delta(\mu_k, \pi_k) = \mathbb{E}_{s} \left[ \max_{a \sim \mu_k} Q^{\mu_k}(s, a) - \mathbb{E}_{a \sim \pi_k}\left[Q^{\pi_k}(s, a)\right]\right].
\end{equation*}
A positive $\Delta(\mu_k, \pi_k)$ indicates that the value of optimal target-update actions in the replay buffer exceeds that of the actions generated by the current policy. This suggests that an optimal policy derived from the replay buffer would outperform the current policy, implying a potential under-exploitation of valuable historical data.

Figure~\ref{fig:delta-visualization-two} empirically shows that during the course of training, the replay buffer could contain more optimal actions as compared to the ones generated by the current policy. This becomes prevalent, especially in the latter training stage, when the replay buffer is filled with high-performance samples. Such an observation indicates a notable shortfall of existing methods in exploiting the good samples in the replay buffer. 
\New{
In light of this, allowing the RL agent to swiftly seize the serendipities, i.e., luckily, successful experiences can be a natural cure to resolve the underestimation issue, as illustrated in Figure~\ref{fig:motivation}. 
}
Then, we shift our focus to devise a method for utilizing the optimal actions from the replay buffer as target-update actions to boost $Q$-value estimation.

\subsection{Blended Exploitation and Exploration operator}
To extract the best-performing actions from the replay buffer for updating the $Q$-target value, we consider the policy mixture $\mu$ induced by the replay buffer, which contains many samples and varies per policy iteration. Based on $\mu$, we introduce the Bellman Exploitation operator $\mathcal{T}_{exploit}$:
\begin{align}\label{T-exploit} 
    \nonumber
    &\quad \mathcal{T}_{exploit}^{\mu} Q(s,a) =   r(s,a) \\
    & + \gamma\cdot\max_{a'\in \mathcal{A},\mu(a'\vert s')>0} \mathbb{E}_{s'\sim P(s'\vert s,a)}[Q(s',a')].
\end{align}
It yields a $Q$-value estimation that is less affected by the optimality level of the current policy. Several offline RL methods~\cite{IQL,xuoffline,garg2023extreme} have also focused on computing $\max Q$ constrained to the support of a pre-collected dataset for Bellman update, yet rely on a stationary behavior policy, which could be viewed as a reduced form of the $\mathcal{T}_{exploit}$ operator.

Meanwhile, to maintain the exploration optimism, 
we utilize the general Bellman Exploration operator. Here, $\omega(s',a'\vert \pi)$ refers to a chosen exploration term. 
\begin{align}\label{T_explore}
    \nonumber
    \mathcal{T}_{explore}^{\pi} Q(s,a) = & r(s,a) + \gamma\cdot \mathbb{E}_{s'\sim P(s'\vert s,a)}\mathbb{E}_{a'\sim \pi(a'\vert s')} \\
    & [Q(s',a')-\omega(s',a'\vert\pi)]
\end{align}
With the Bellman Exploitation and Bellman Exploration operators, which respectively capitalize on past successes and promote the exploration of uncertain regions, we aim to take the merits of both.  

\New{
While there are numerous ways to blend the two operators, a simple yet efficient algorithm has long been advocated~\citep{td3-bc}.  We propose the BEE operator, employing a linear combination to regulate the balance, aligns well with the requirement. 
The $\mathcal{T}_{exploit}$ component is particularly vital, as it provides a reference $Q$-value, thereby improving $Q$-value estimation. Later in the paper, extensive experiments showcase the efficiency of the BEE operator.}

\begin{definition}
The  Blended  Exploitation and Exploration~(\textbf{BEE}) Bellman operator  $\mathcal{B}$ is defined as:
\begin{align}\label{BEE_Q}
    \nonumber
    \mathcal{B}^{\{\mu,\pi\}} Q(s,a) = & \lambda\cdot \mathcal{T}_{exploit}^{\mu} Q(s,a) \\ 
    & + (1-\lambda)\cdot \mathcal{T}_{explore}^{\pi} Q(s,a)
\end{align}
Here, $\mu$ is the policy mixture,  $\pi$ is the current policy, and $\lambda\in (0,1)$ is a trade-off hyperparameter.
\end{definition}

The choice of $\lambda$ in Eq.(\ref{BEE_Q}) impacts the exploitation-exploration trade-off, as shown in Figure~\ref{grid-world-visualization}.
Besides choosing a fixed number, $\lambda$ can also be autonomously and adaptively tuned with multiple methods as detailed in Appendix~\ref{section:auto-lambda-mechanism}.
The single-operator design 
incurs comparable computational costs to general-purpose algorithms such as SAC~\citep{sac}, and is relatively lightweight compared to other methods that require training a large number of $Q$-networks to tackle the exploration-exploitation dilemma~\citep{oac, redq}.

\subsection{Superior $Q$-value estimation using BEE operator}
For a better understanding of the BEE operator, we conduct a theoretical analysis of its dynamic programming properties in the tabular MDP setting, covering policy evaluation, policy improvement, and policy iteration. All proofs are included in Appendix~\ref{ap-omittedproofs}.

\begin{proposition}[\textbf{Policy evaluation}]
Consider an initial $Q_0:\mathcal{S}\times\mathcal{A}\rightarrow\mathbb{R}$ with $\vert \mathcal{A}\vert < \infty$, and define $Q_{k+1}= \mathcal{B}^{\{\mu,\pi\}}Q_{k}$. Then the sequence $\{Q_{k}\}$ converges to a fixed point $Q^{\{\mu,\pi\}}$ as $k\rightarrow \infty$.
\end{proposition}

\begin{proposition}[\textbf{Policy improvement}]\label{policy-improvement}
Let $\{\mu_{k},\pi_{k}\}$ be the policies at iteration $k$, and $\{\mu_{k+1},\pi_{k+1}\}$ be the updated policies, where  $\pi_{k+1} $ is the greedy policy of the $Q$-value. 
Then for all $(s,a)\in \mathcal{S}\times \mathcal{A}$, $\vert \mathcal{A}\vert < \infty$, we have $Q^{\{\mu_{k+1},\pi_{k+1}\}}(s,a) \geq Q^{\{\mu_k,\pi_k\}}(s,a)$.
\end{proposition}
\begin{figure*}[t]
    \centering \includegraphics[height=2.6cm,keepaspectratio]{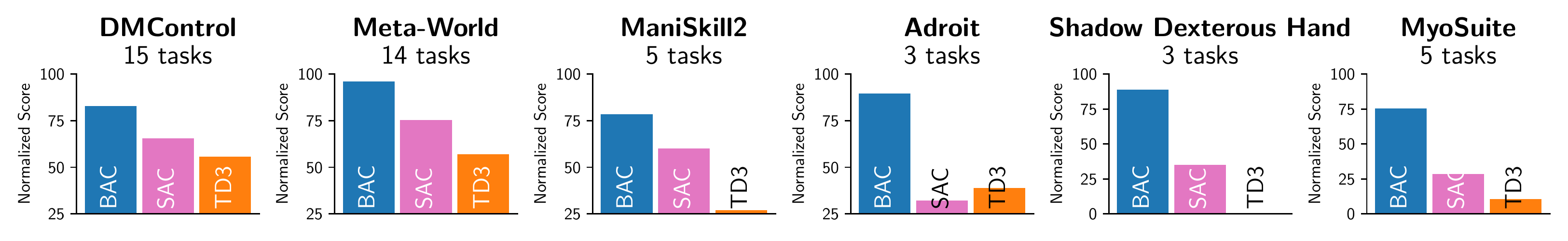} 
    \vspace{-5mm}
    \caption{\small \textbf{Performance overview over six benchmarks.} BAC compares favorably to popular model-free methods across DMControl, Meta-World, ManiSkill2, Adroit, Shadow Dexterous Hand, and MyoSuite benchmark tasks with a single set of hyperparameters. }
    \label{fig:summary}
\end{figure*}

\begin{proposition}
[\textbf{Policy iteration}]
Assume $\vert \mathcal{A}\vert < \infty$, by repeating iterations of the policy evaluation and policy improvement, any initial policies converge to the optimal policies $\{\mu^*, \pi^*\}$, s.t. $Q^{\{\mu^*,\pi^*\}}(s_t,a_t)\geq Q^{\{\mu',\pi'\}}(s_t,a_t), \forall \mu' \in \Pi, \pi'\in \Pi, \forall (s_t,a_t) \in \mathcal{S}\times \mathcal{A} $.
\end{proposition}
With the approximate dynamic programming properties established, our BEE operator is well-defined and flexible and could be integrated into various off-policy actor-critic algorithms.
\New{In Appendix~\ref{section:superiorQ}, we show that the BEE operator would alleviate underestimation without inciting additional overestimation, thus facilitating the estimation of $Q$ and improving learning efficiency.}

\begin{algorithm}[t]
    \caption{BEE Actor-Critic (\ourshort)}
    \begin{algorithmic}
    \STATE \textbf{initialize:} $Q$ networks $Q_{\phi}$, policy $\pi_\theta$, replay buffer $\mathcal{D}$;
    \FOR{policy training steps $t = 1, 2, \cdots, T$}
        \STATE Sample $N$ transitions $(s,a,r,s')$ from $\mathcal{D}$
        \STATE Compute $\mathcal{T}_{exploit}Q_{\phi}$ by Eq.(\ref{T-exploit}) \COMMENT{\textcolor{myblue}{multiple design choices available}}
        \STATE Compute $\mathcal{T}_{explore}Q_{\phi}$ by Eq.(\ref{T_explore}) \COMMENT{\textcolor{myblue}{with a chosen exploration term $\omega(\cdot\vert \pi_\theta)$}}
        \STATE Calculate the target Q value: $\mathcal{B}Q_{\phi} \leftarrow\lambda \mathcal{T}_{exploit}Q_{\phi} + (1-\lambda)\mathcal{T}_{explore}Q_{\phi}$
        \FOR{each environment step}
            \STATE Collect $(s,a,s',r)$ with $\pi_\theta$ from real environment; add to $\mathcal{D}$
        \ENDFOR
        \FOR{each gradient step}
            \STATE Update $Q_{\phi}$ by $\min_\phi\left(\mathcal{B}Q_{\phi}-Q_{\phi}\right)^2$
            \STATE Update $\pi_\theta$ by $\max_{\theta} Q_\phi(s,\pi_{\theta})$
        \ENDFOR 
    \ENDFOR
    \label{general-bac-algorithm}
    \end{algorithmic}
\end{algorithm}

\subsection{Algorithmic instantiation}
We now describe two practical algorithmic instantiations based on the BEE operator $\mathcal{B}$ for both model-free and model-based RL paradigms, namely BEE Actor-Critic (\ourshort) and Model-Based BAC (\mbshort), respectively. 
The implementation of our methods requires the specification of two main design choices: 1) a practical way to optimize the objective value on the Bellman Exploitation operator, and 2) a specific choice on the exploration term $\omega(\cdot\vert \pi)$ in the Bellman Exploration operator.

To effectively compute the $\max Q$-target value in Eq.(\ref{T-exploit}) subject to the samples in the replay buffer, we utilize the in-sample learning objectives~\citep{IQL,garg2023extreme, xuoffline} to learn the maximum $Q$-value over actions in the replay buffer. This treatment not only avoids the explicit computation of the policy mixture $\mu$ of replay buffer but also promotes the stability of $Q$ estimation by only extracting actions that have been previously encountered for the Bellman update.

For the exploration term $\omega(\cdot\vert \pi_{\theta})$, numerous options have been extensively explored in prior off-policy actor-critic methods~\citep{sac,dac,eberhard2022pink}.
Here, we employ the entropy regularization term from SAC to compute  $\mathcal{T}_{explore}Q_{\phi}(s,a)$, where actions $a'$ for target updating are extracted from $\pi_\theta$.  
For extensive design choices for \ourshort\ see Appendix~\ref{section:designchoice}.

\noindent\textbf{Integration into Dyna-style model-based RL.}\quad
Our method could be invoked into the Dyna-style model-based RL (MBRL) framework~\citep{sutton1990integrated, sutton1991dyna,metrpo, steve, slbo}.
As observed in~\citep{slbo,objectivemismatch,ghugare2023simplifying}, a better policy optimizer could potentially further enhance the algorithm's performance, this motivates us to incorporate the BEE operator in existing model-based approaches.
We propose a modification to the general Dyna-style algorithm, where the standard $Q$-value update rule is replaced with our BEE operator, resulting in the Model-based BAC~(\mbshort) algorithm.

In contrast to previous methods that utilize SAC as policy optimization backbone~\citep{mbpo, autombpo, m2ac, ji2022update}, \mbshort\ treats real and model-generated data differently. 
It applies the $\mathcal{T}_{exploit}$ to real data $\mathcal{D}_e$, capitalizing on optimal real experiences while employing the $\mathcal{T}_{explore}$ on model rollout data $\mathcal{D}_{m}$ to explore new possibilities. 
This approach enhances the effective use of valuable real data and fosters exploration in new regions of the state space. 
The practical implementation builds upon MBPO~\citep{mbpo} by integrating the \ourshort\ as policy optimizer, with the pseudocode in Appendix~\ref{section:appendix-mbbac}.

\begin{figure*}[t]
    \centering
    \begin{minipage}[t]{1.0\textwidth}
        \centering
        \includegraphics[height=3.5cm,keepaspectratio]{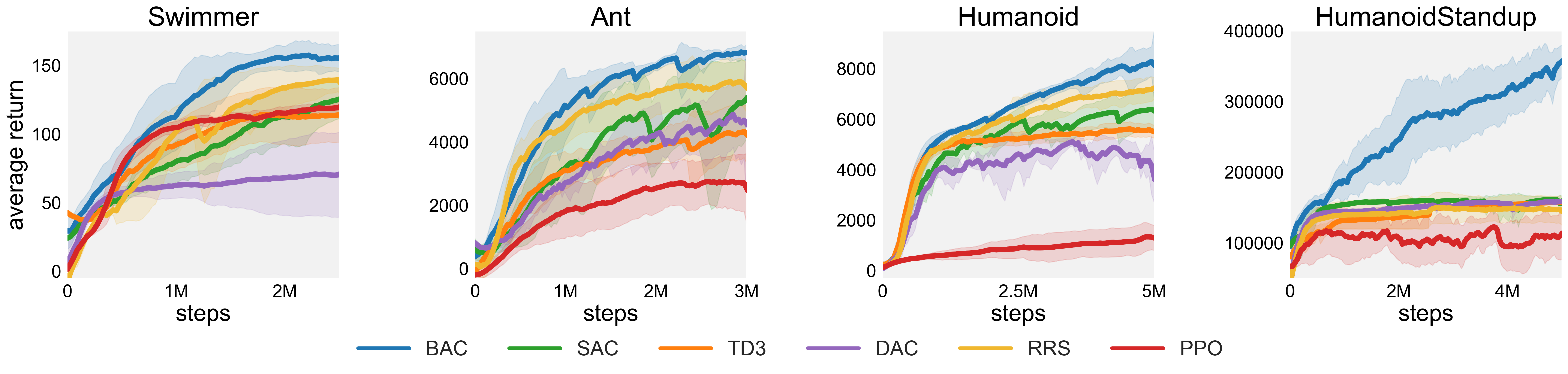} 
        \vspace{-2mm}
        \caption{\small Training curves of BAC and five baselines on four continuous control benchmarks.  Solid curves depict the mean of ten trials, and shaded regions correspond to the one standard deviation. To ensure a fair comparison, a uniform discount factor of $\gamma=0.99$ is used for all baselines and tasks. Notably, for the Swimmer task, a significantly higher discount factor of $\gamma=0.9999$ greatly enhances performance; these results are provided in the Appendix~\ref{ap:swimmer}.}
        \label{fig:main-res-mf}
        \vspace{3pt}
    \end{minipage}
    \vspace{3mm}
    \begin{minipage}[t]{1.0\textwidth}
        \centering
        \includegraphics[height=3.5cm,keepaspectratio]{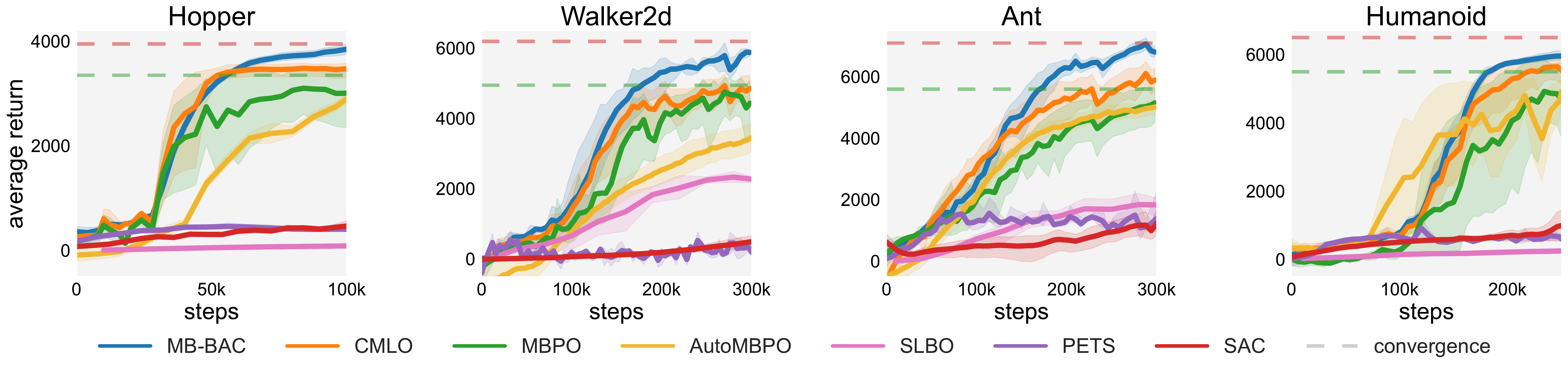} 
        \vspace{-2mm}
        \caption{\small Training curves of MB-BAC and six baselines on four continuous control benchmarks, averaged over ten trials. The dashed lines are the asymptotic performance of  SAC (up to 3M) and MBPO.}
        \label{fig:main-res-mb}
        \vspace{-2mm}
    \end{minipage}
\end{figure*}
\begin{figure*}[t]
    \centering
    \begin{minipage}[t]{0.48\textwidth}
            \centering
                \includegraphics[height=1.5cm,keepaspectratio]{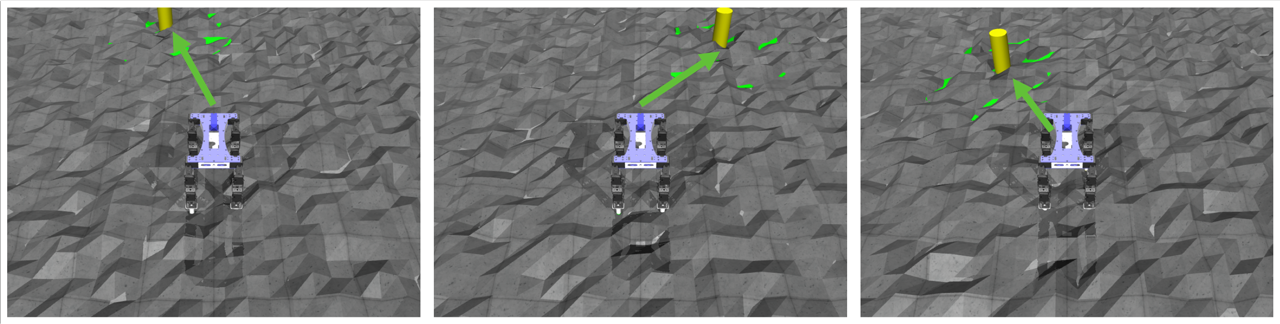}
                \includegraphics[height=3.0cm,keepaspectratio]{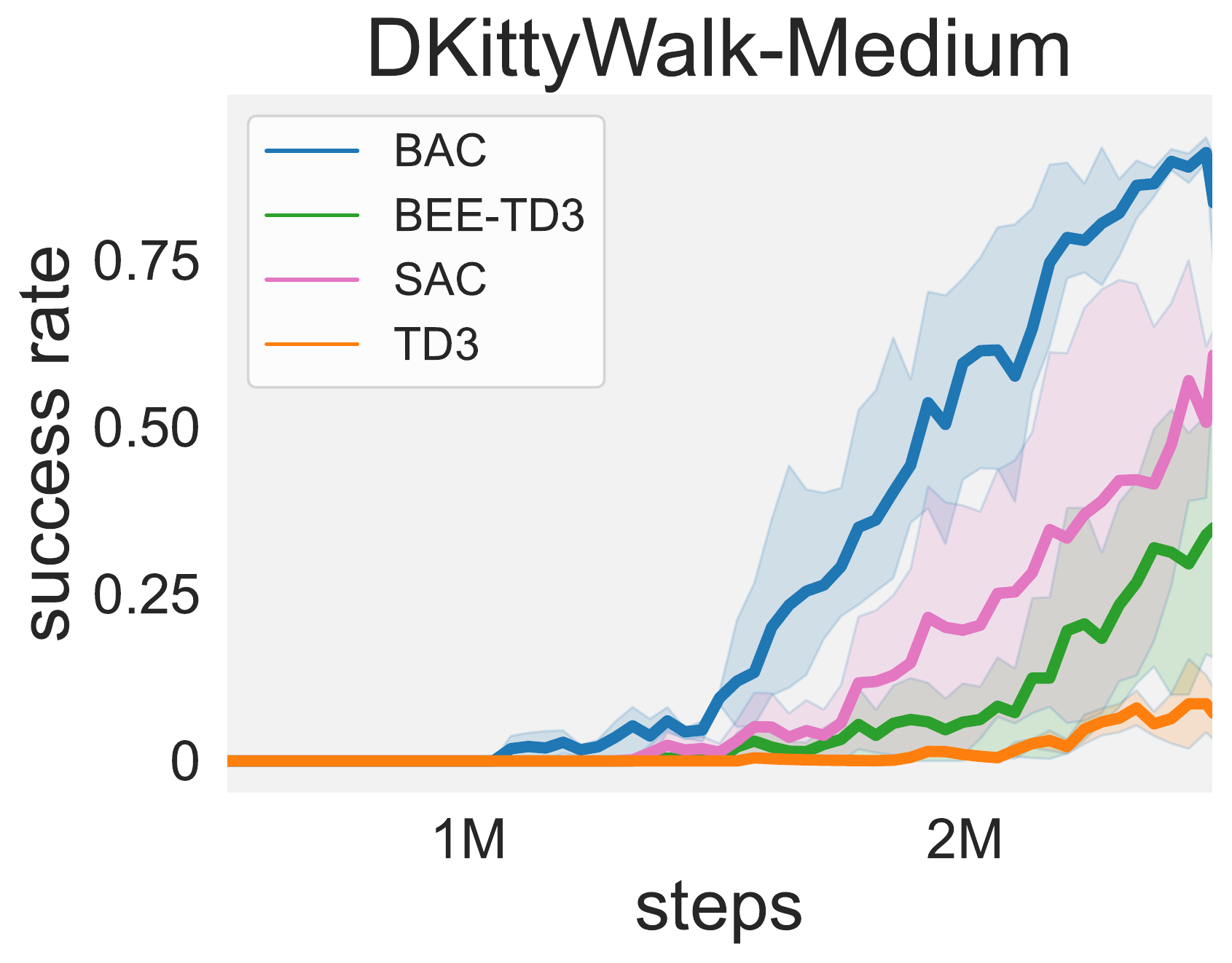}
                \includegraphics[height=3.0cm,keepaspectratio]{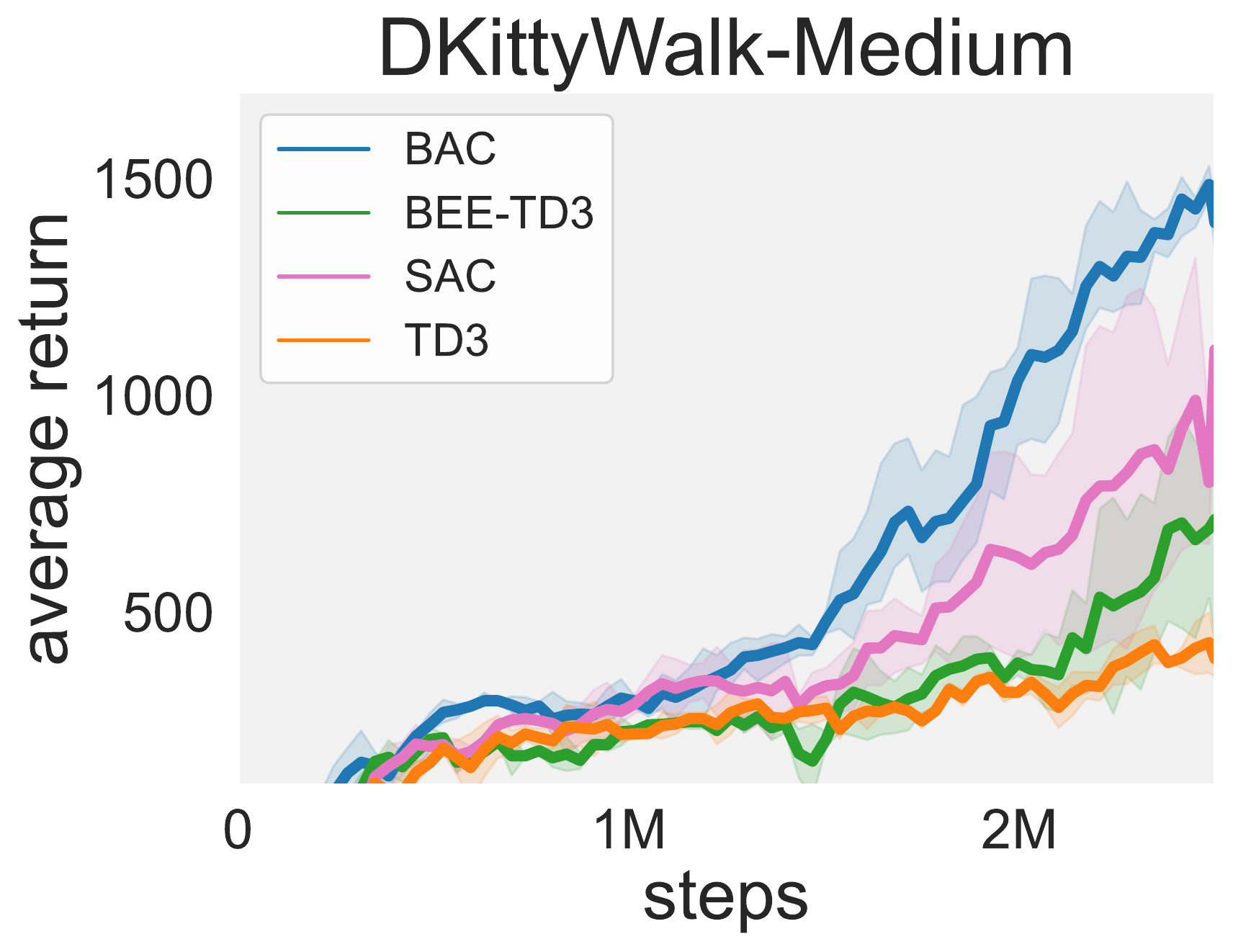}
            \vspace{-0.2cm}
            \caption{\small Success rate and average return in DKittyWalk-Medium.}
            \label{fig:hard}
             \vspace{0.2cm}
        \end{minipage}
    \hfill
    \begin{minipage}[t]{0.48\textwidth}
        \centering
            \includegraphics[height=1.5cm,keepaspectratio]{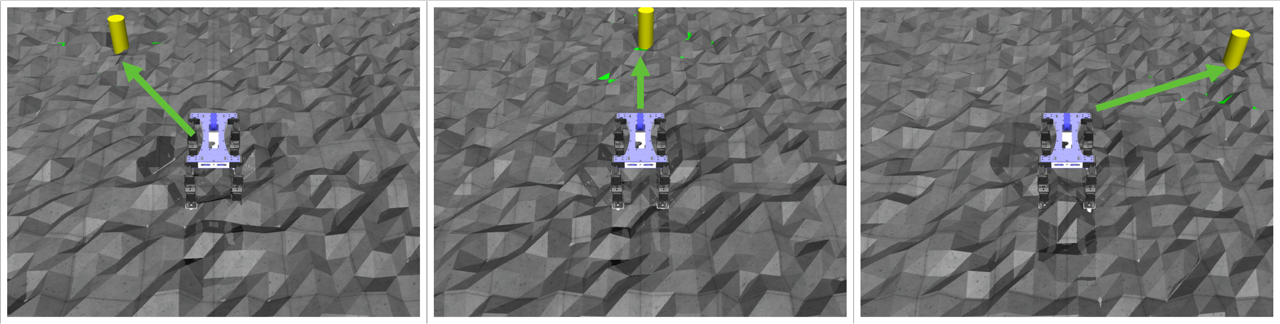}
            \includegraphics[height=3.0cm,keepaspectratio]{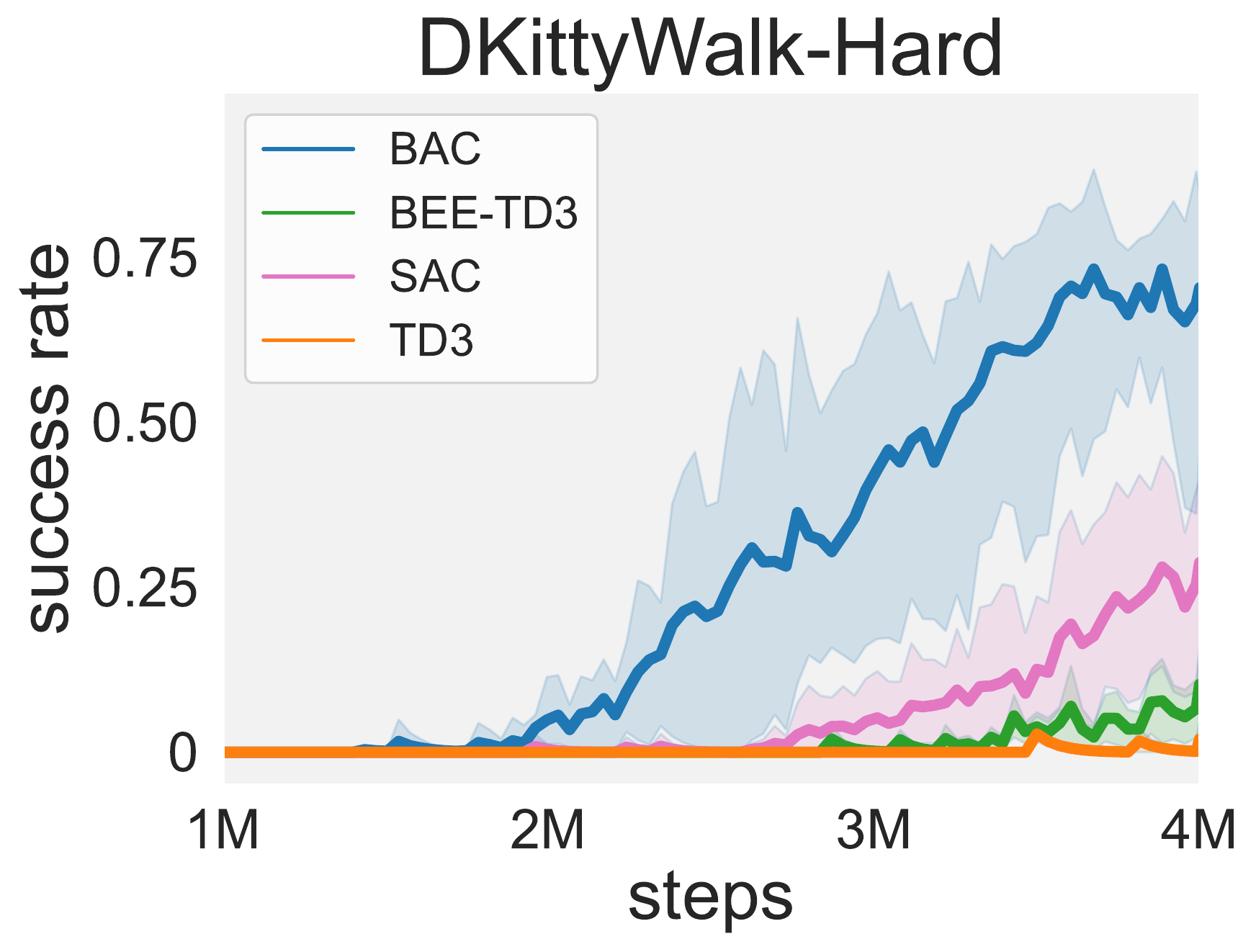}
            \includegraphics[height=3.0cm,keepaspectratio]{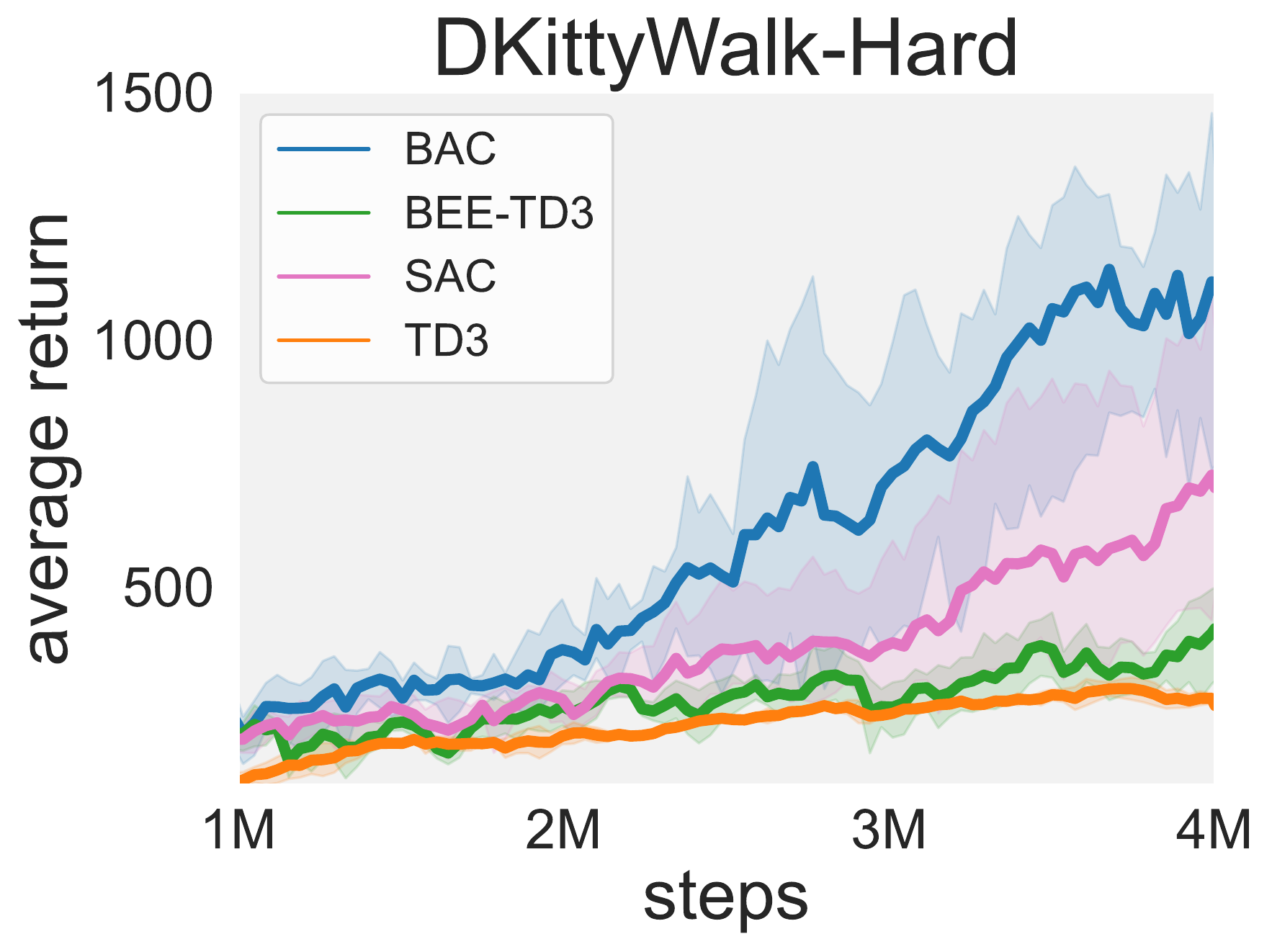}
        \vspace{-0.2cm}
        \caption{\small Success rate and average return in  DKittyWalk-Hard.}
        \label{fig:challenging}
        \vspace{0.2cm}
    \end{minipage}
    \begin{minipage}[t]{0.48\textwidth}
            \centering
            \begin{subfigure}[t]{0.49\textwidth}
                \centering
                \includegraphics[height=2.4cm,keepaspectratio]{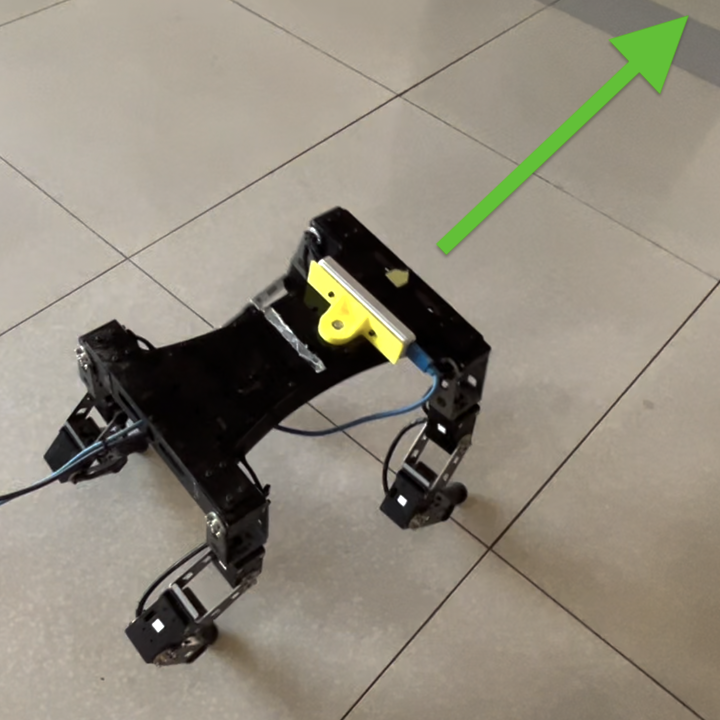}
                \includegraphics[width=0.99\textwidth]{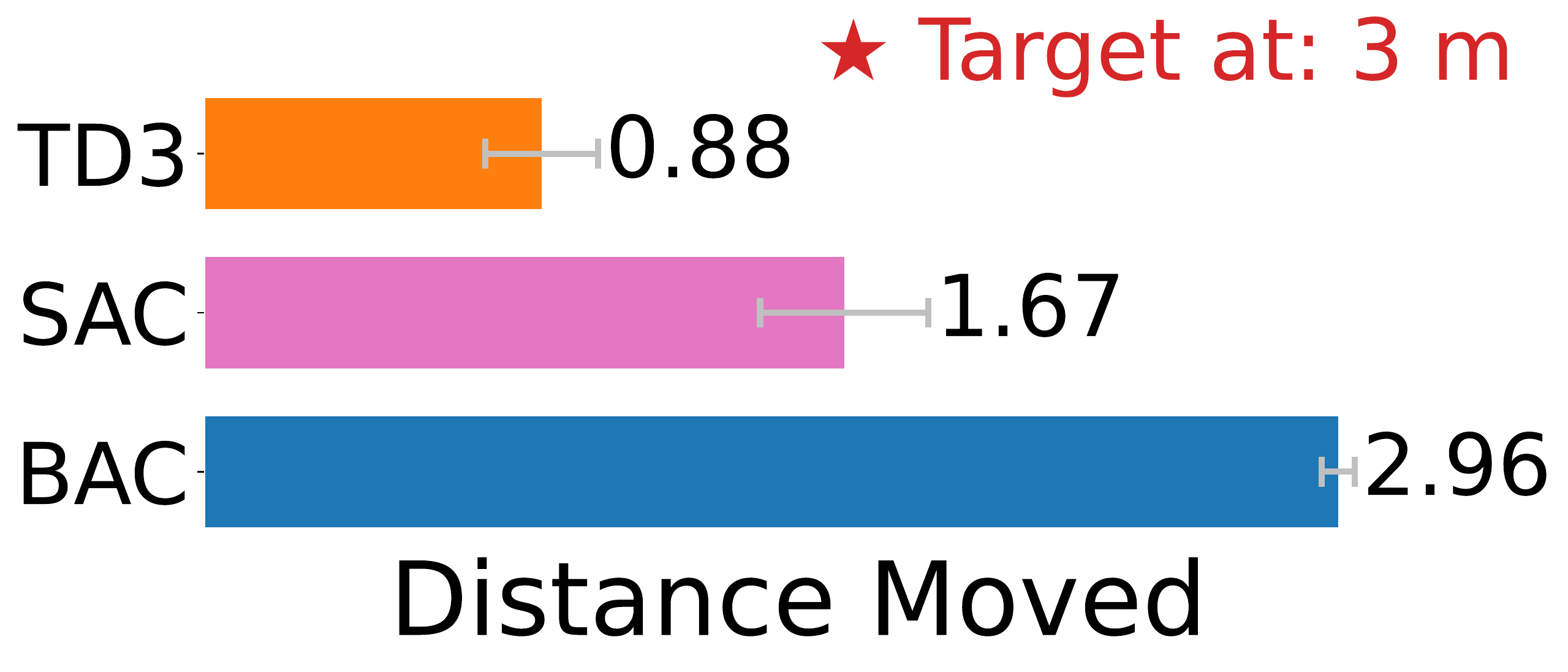}                
                \caption{\small Smooth road}
                \label{fig:ab-return_1}
            \end{subfigure}
            \begin{subfigure}[t]{0.49\textwidth}
                \centering                \includegraphics[height=2.4cm,keepaspectratio]{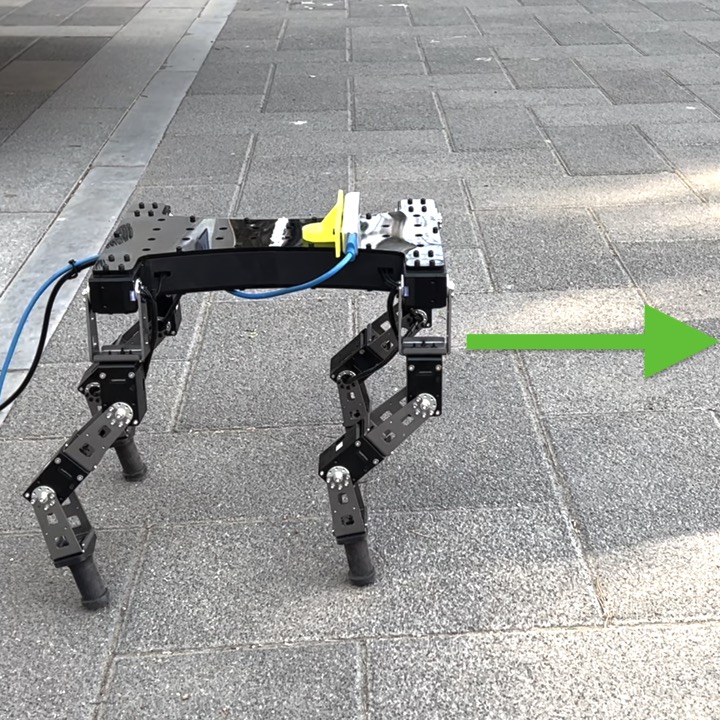}
                \includegraphics[width=0.99\textwidth]{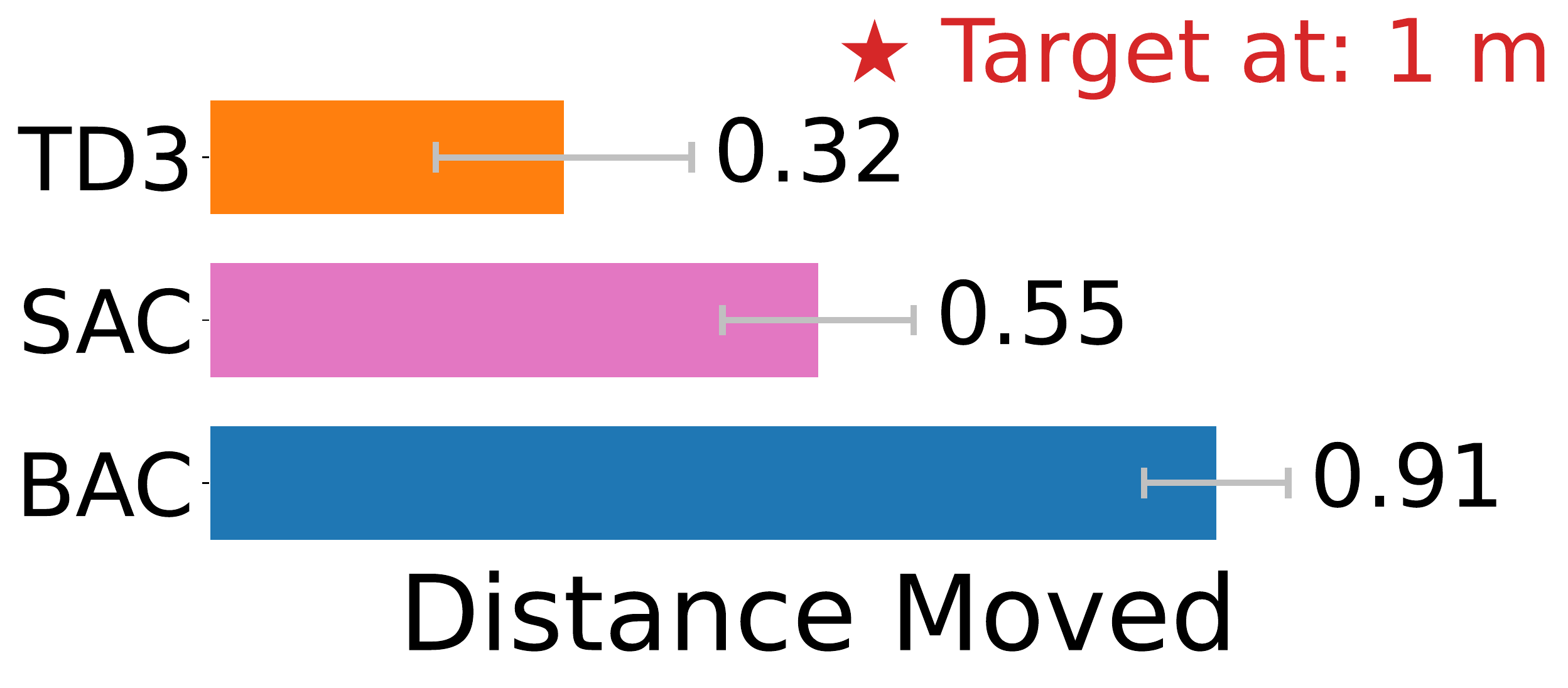}
                \caption{\small Rough stone road}
                \label{fig:ab-return_2}
            \end{subfigure}
            \vspace{3pt}
    \end{minipage}
    \hfill
    \begin{minipage}[t]{0.48\textwidth}
        \centering
        \begin{subfigure}[t]{0.49\textwidth}
            \centering
            \includegraphics[height=2.4cm,keepaspectratio]{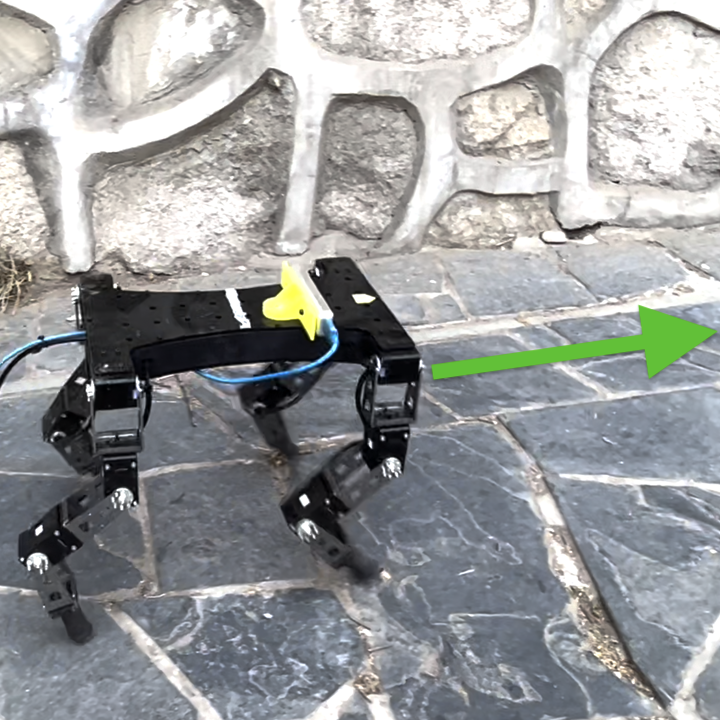}
            \includegraphics[width=0.99\textwidth]{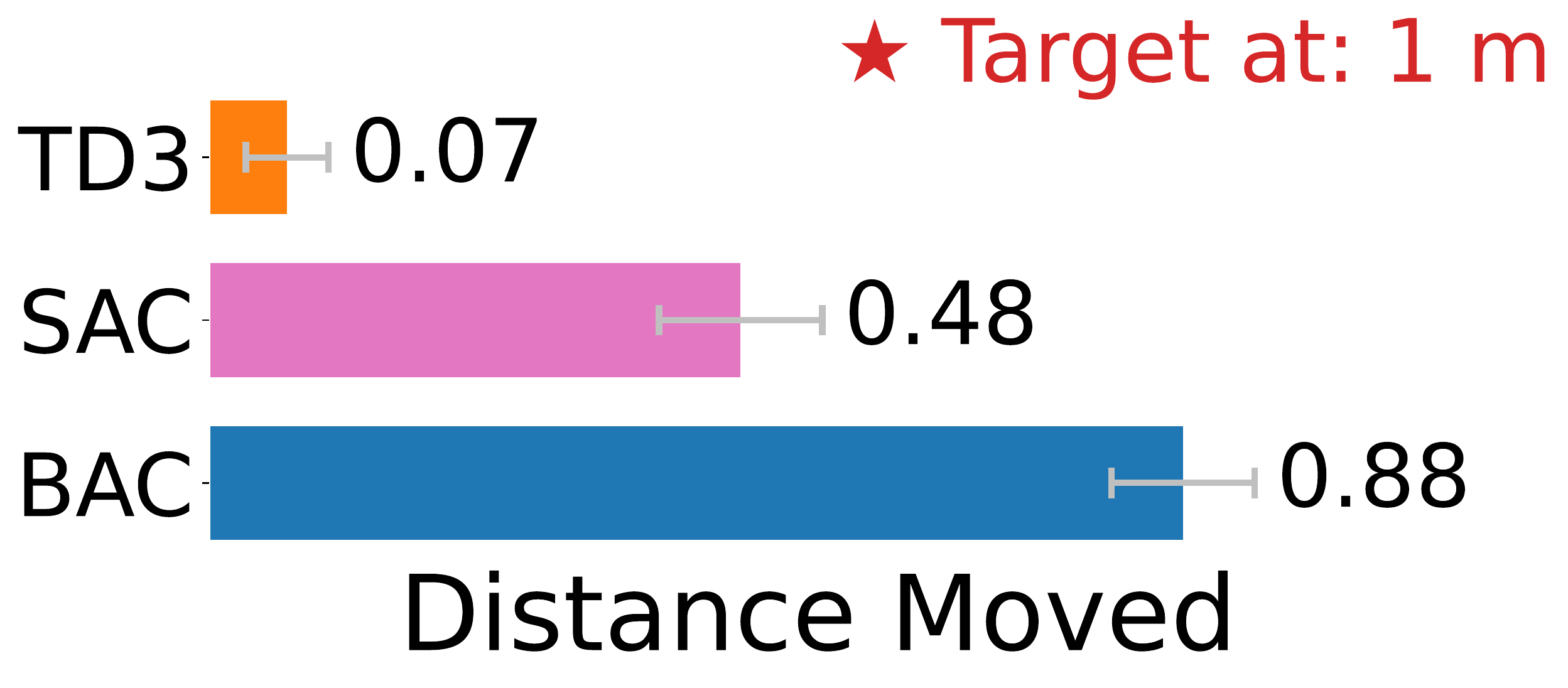}
            \caption{\small Uphill stone road}
            \label{fig:ab-return_3}
        \end{subfigure}
        \begin{subfigure}[t]{0.49\textwidth}
            \centering
            \includegraphics[height=2.4cm,keepaspectratio]{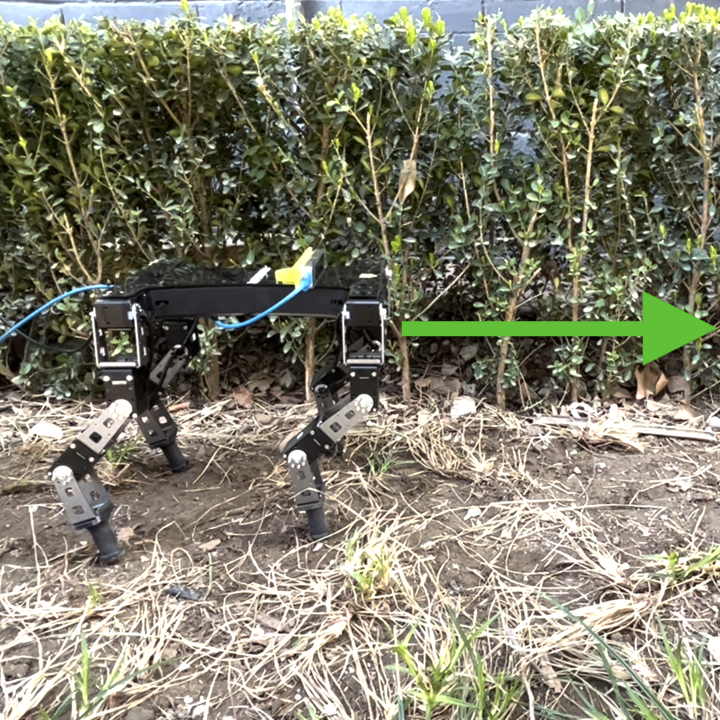}
            \includegraphics[width=0.99\textwidth]{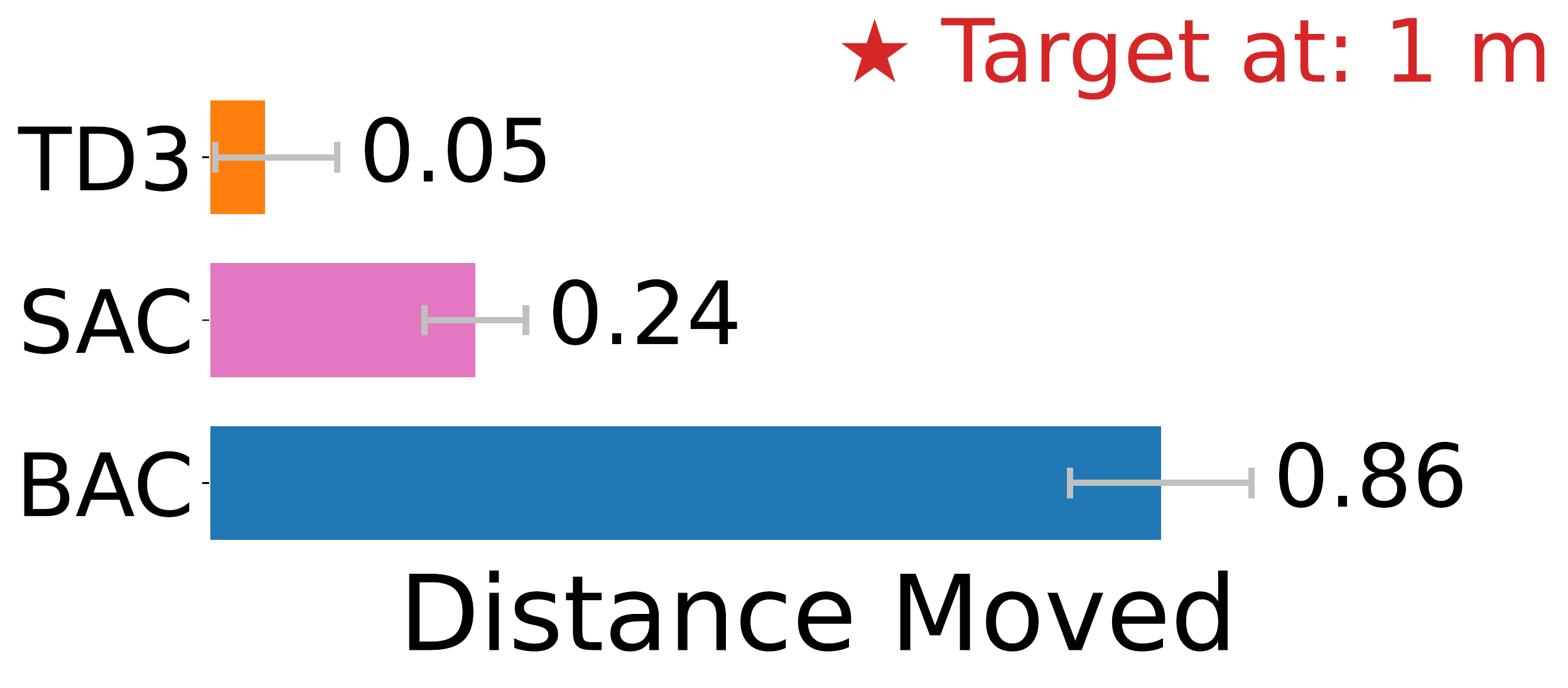}
            \caption{\small Grassland}
            \label{fig:ab-return_4}
        \end{subfigure}
    \end{minipage}
    \vspace{-0.2cm}
    \caption{\small \textbf{Comparisons on four challenging real-world tasks.} The bar plots show how far the agent walks toward the goal for each algorithm averaged over ten runs. For (a) and (b), we employ the policy trained in the -Medium task, and for (c) and (d) use the policy trained in the -Hard task.}
    \label{fig:real-world}
    \vspace{-0.2cm}
\end{figure*}

\section{Experiments}
Our experimental evaluation aims to investigate the following questions: 
1) How effective is the proposed BEE operator in model-based and model-free paradigms?
2) How effectively does \ourshort\ perform in failure-prone scenarios, that highlight the ability to seize serendipity and fleeting successes, particularly in various real-world tasks?

\subsection{Evaluation on various standard benchmarks}
To illustrate the effectiveness of the BEE operator across both model-based and model-free paradigms, we evaluate \ourshort\ and \mbshort\ on various continuous control tasks. 

Notably, BAC demonstrates superior performance compared to popular model-free methods across \textbf{MuJoCo}~\citep{Mujoco}, \textbf{DMControl}~\citep{dmcontrol}, \textbf{Meta-World}~\citep{yu2019meta}, \textbf{ManiSkill2}~\citep{gu2023maniskill2}, \textbf{Adroit}~\citep{adroit}, \textbf{MyoSuite}~\citep{MyoSuite2022}, \textbf{ROBEL}~\citep{dkitty} benchmark tasks, and even show effectiveness in sparse reward tasks and noisy environments. We summarize the performance over \New{six} benchmarks in Figure~\ref{fig:summary}.  Detailed performance curves on these benchmark suites are in Appendix~\ref{section:more-benchmark-results}.

\noindent\textbf{Comparison of model-free methods.}\quad
We compare \ourshort\ to several popular model-free baselines, including: 
1) SAC, regarded as the most popular off-policy actor-critic method;
2) TD3, which introduces the Double $Q$-learning trick to reduce function approximation error;
3) Diversity Actor-Critic~(DAC)~\citep{dac}, a variant of SAC, using a sample-aware entropy regularization instead, which is a potential choice for our $\omega(\cdot\vert s,a)$;
4) Random Reward Shift~(RRS)~\citep{rrs}, which learns multiple value functions~(seven double-$Q$ networks) with different shifting constants for the exploration and exploitation trade-off;
5) PPO~\citep{ppo}, a stable on-policy algorithm that discards historical policies.

We evaluate \ourshort\ and the baselines on a set of MuJoCo continuous control tasks.
\ourshort\ surpasses all baselines in terms of eventual performance, coupled with better sample efficiency, as shown in Figure~\ref{fig:main-res-mf}. 
Notably, the HumanoidStandup task, known for its high action dimension and susceptibility to failure~\citep{dac}, requires the algorithms to be able to seize and value serendipity.  
In this task, \ourshort\ gains a significantly better performance, with average returns up to 280,000 at 2.5M steps and 360,000 at 5M steps, which is 1.5x and 2.1x higher than the strongest baseline, respectively.
This echoes the hypothesis that BAC exploits past serendipities in failure-prone environments. Trajectory visualizations in Figure~\ref{fig:standup_visualization} show that \ourshort\ agent could swiftly reach a stable standing, while the SAC agent ends up with a wobbling kneeling posture, the DAC agent sitting on the ground, and the RRS agent rolling around.

Experimental results on the more failure-prone tasks are in Appendix~\ref{section:failure-prone}.  We note that \ourshort\ is the \textbf{\textit{ﬁrst documented model-free method}} of solving the complex Dog tasks of DMControl.  Additionally, we integrate our BEE into the TD3 algorithm and find that the ad-hoc BEE-TD3 also outperforms the original TD3 in fifteen DMControl tasks, refer to Appendix~\ref{section:dmc_benchmark}.

\noindent\textbf{Comparison of model-based methods.}\quad
We evaluate the performance of \mbshort,  
which integrates the BEE operator into the MBPO algorithm, 
against several model-based and model-free baselines. Among the Dyna-style counterparts, MBPO~\citep{mbpo}, CMLO~\citep{ji2022update}, and AutoMBPO~\citep{autombpo} use SAC as the policy optimizer, while SLBO~\citep{slbo} employs TRPO~\citep{trpo}. 
PETS~\citep{pets} is a planning-based method that utilizes CEM~\citep{botev2013cross} as the planner. Figure~\ref{fig:main-res-mb} showcases that \mbshort\ learns faster than other modern model-based RL methods and yields promising asymptotic performance compared with model-free counterparts.  Moreover, the result highlights the universality of the BEE operator.

\subsection{Evaluation in real-world quadruped robot tasks}
We evaluate \ourshort\ on a real quadruped robot D'Kitty~\citep{dkitty}. We follow the sim2real paradigm as in previous legged locomotion works~\citep{agarwal2023legged, hwangbo2019learning, tan2018sim} where the agent is trained in simulated environments with randomized terrains and then deployed in the real world without further training. The task is challenging, as agents are prone to falling due to fluctuating terrain.  As for real-world scenarios, the D'Kitty robot is required to traverse various complex terrains, contending with unpredictable environmental factors.

Firstly, we construct two simulation task variants, DKittyWalk-Medium and DKittyWalk-Hard.
The -Medium variant features a random height region of 0.07m, while the -Hard variant has a height of 0.09m, which is 1.4 times and 1.8 times higher than the base task DKittyWalkRandomDynamics, respectively. 
Given D'Kitty's leg length of around 0.3m when standing, navigating uneven terrain with height variations of over 0.2x to 0.3x the leg length poses a significant challenge, as a deviation of 0.02m would lead to a considerable shift in the center of gravity.
Figure~\ref{fig:hard} and~\ref{fig:challenging} demonstrate that \ourshort\ outperforms other algorithms in both tasks with clearer advantages. 
\ourshort\ achieves a success rate surpassing SAC by approximately 50\%. 
The ad-hoc BEE-TD3 also outperforms the TD3.

More crucially, \ourshort\ achieves superior performance when deployed in the real world across various terrains, as shown in Figure~\ref{fig:real-world}.
The policy learned in the -Medium variant is deployed on two terrains — smooth road and rough stone road, with target points positioned at distances of 3m and 1m, respectively. For more challenging terrains — uphill stone roads and grasslands, we employ the policy trained in the -Hard variant, with a target point located 1m ahead.
Specifically, the \ourshort\ algorithm outperformed the TD3 and SAC agents in achieving stable movement across a variety of terrains and displaying natural gaits. In contrast, the TD3 agent prefers lower postures, such as knee walking, which makes it prone to falling on uneven terrain, while the SAC agent suffers from more oscillatory gait patterns, as shown in the supplementary videos. 
The empirical results also shed light on the necessity of algorithmic improvement for real-world robotics in addition to building better environments and designing informative rewards.

\subsection{Ablation studies}

\begin{figure}[t]
    \begin{tabular}{c}
        \begin{minipage}{\linewidth}
                \centering
                \includegraphics[height=2.6cm,keepaspectratio]{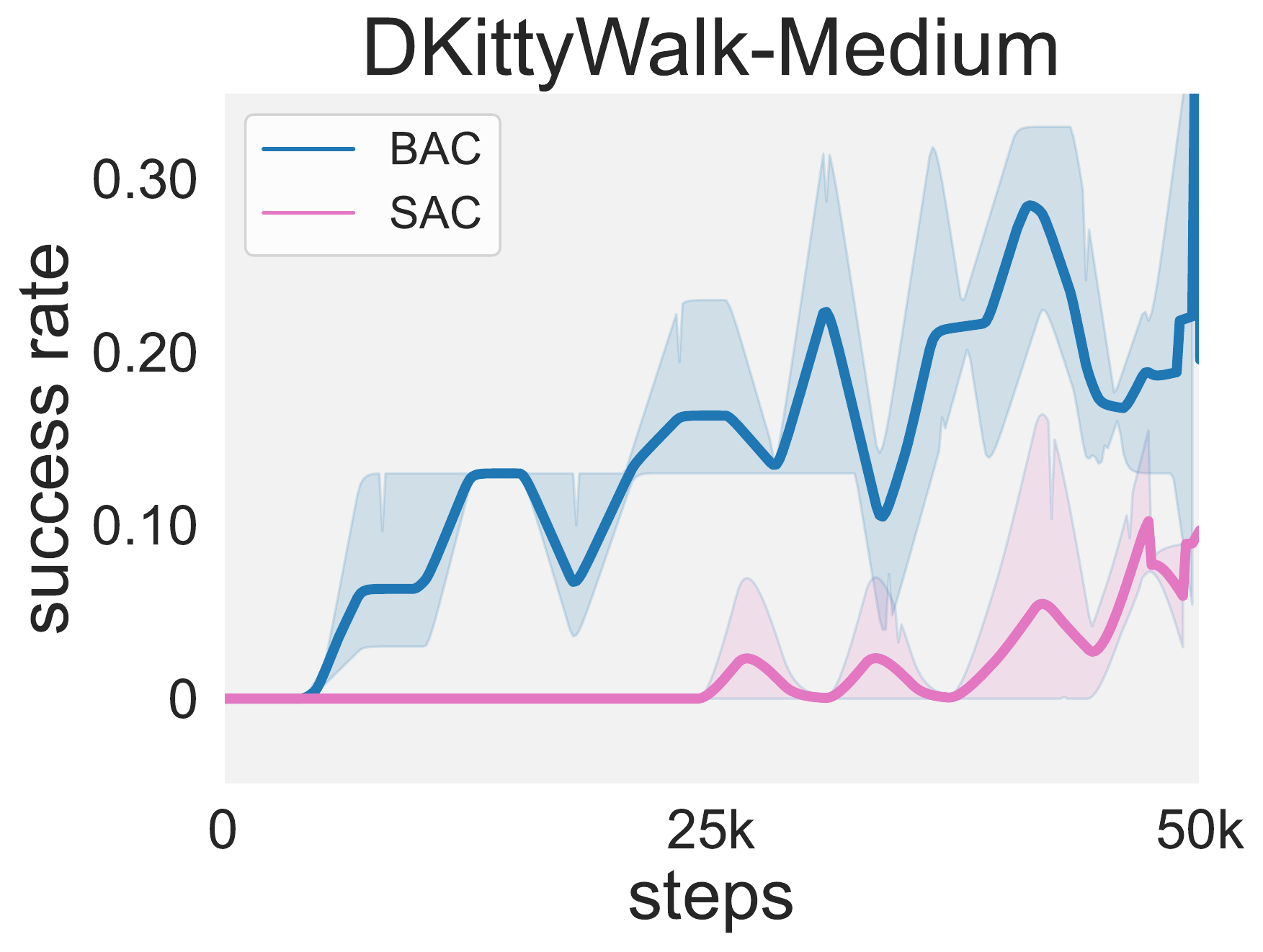} 
                \hspace{5mm}
                \includegraphics[height=2.6cm,keepaspectratio]{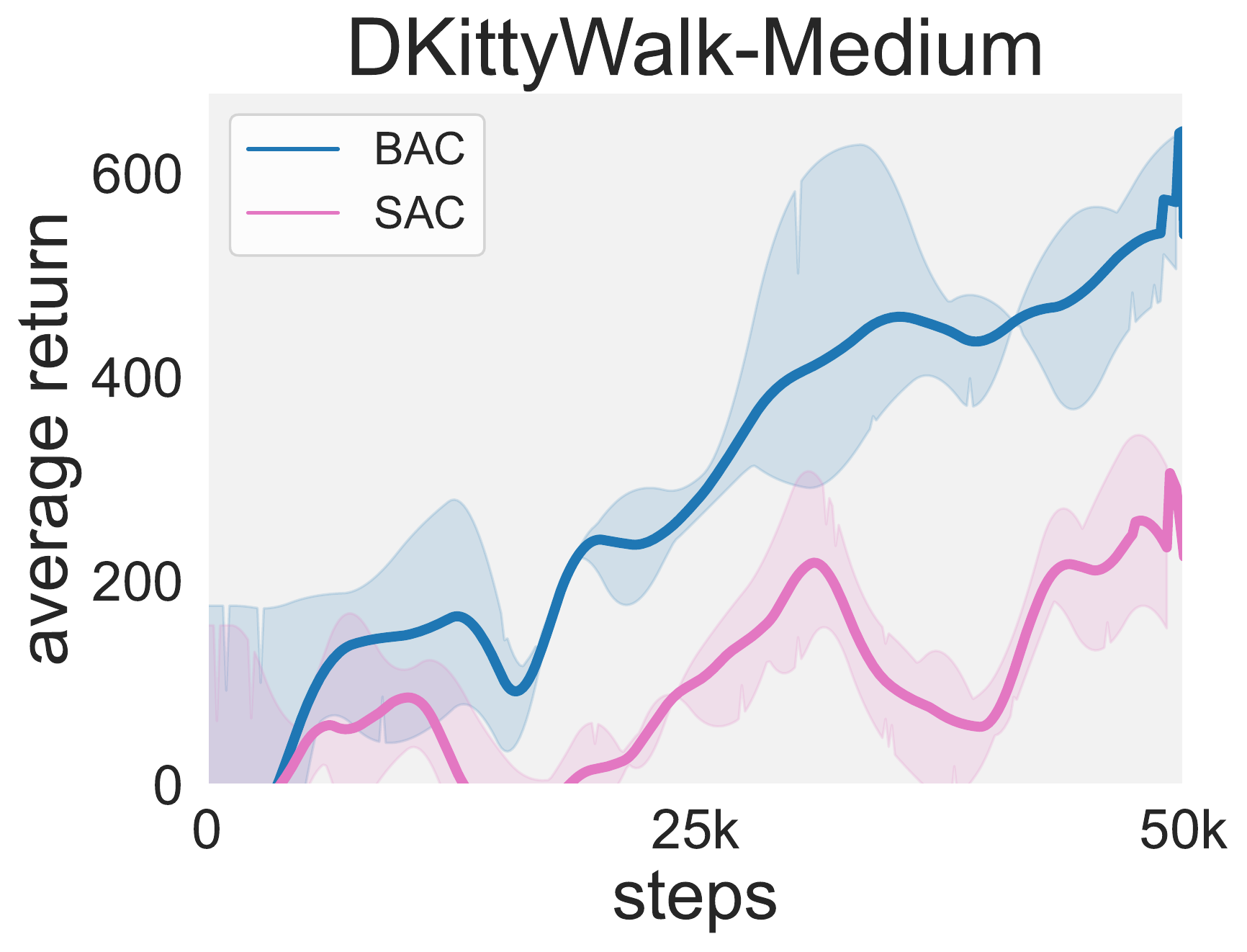}
                \vspace{-2.5mm}
                \caption{\small
                Comparison of the ability to seize serendipity in DKittyWalk-Medium.\textit{Left}: success rate; \textit{Right}: average return. 
                }
                \label{fig:mix_training}
                \vspace{3mm}
        \end{minipage}%
    \\
    \begin{minipage}{\linewidth}
                \centering
               \includegraphics[height=2.6cm,keepaspectratio]{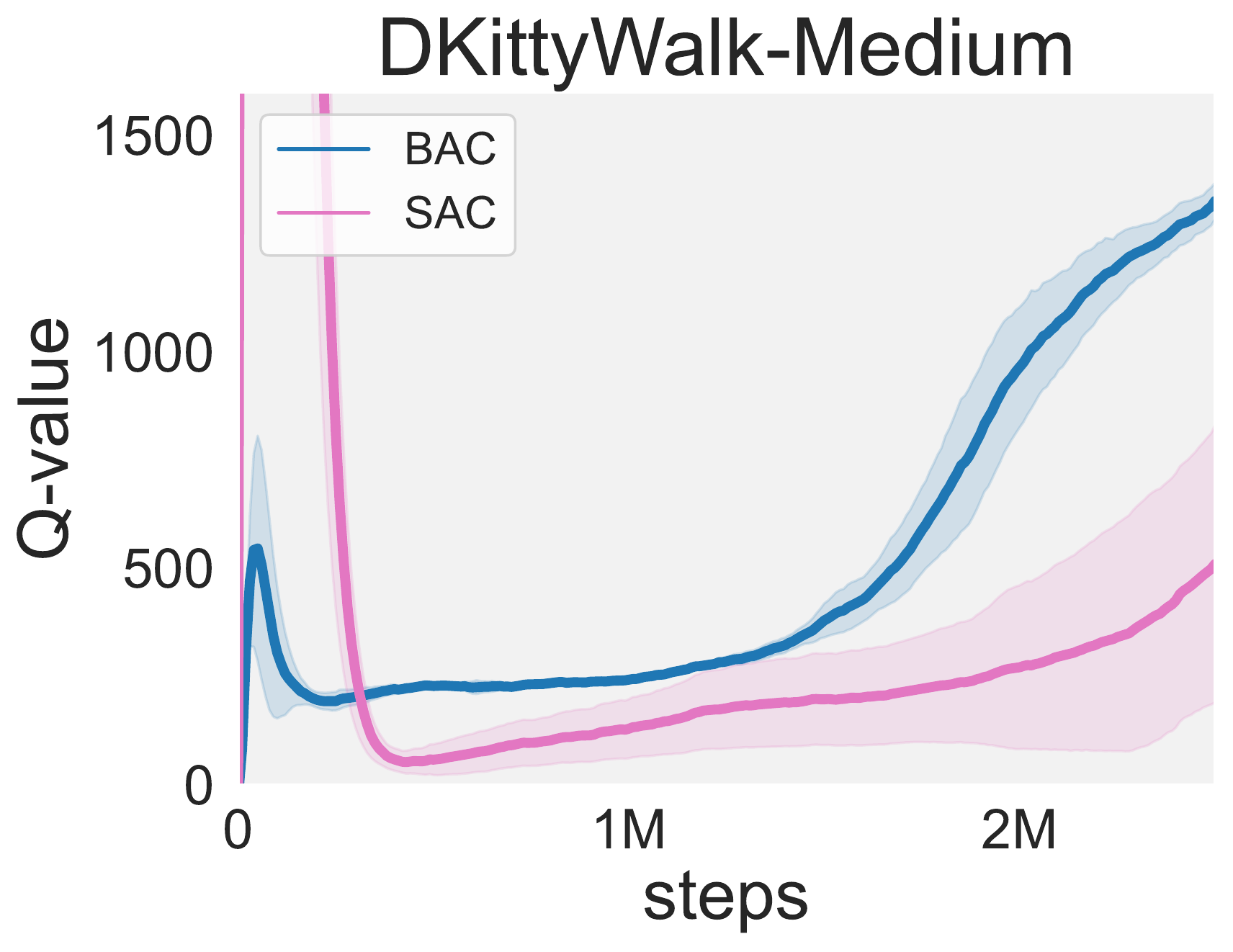}
                \hspace{5mm} 
                \includegraphics[height=2.6cm,keepaspectratio]{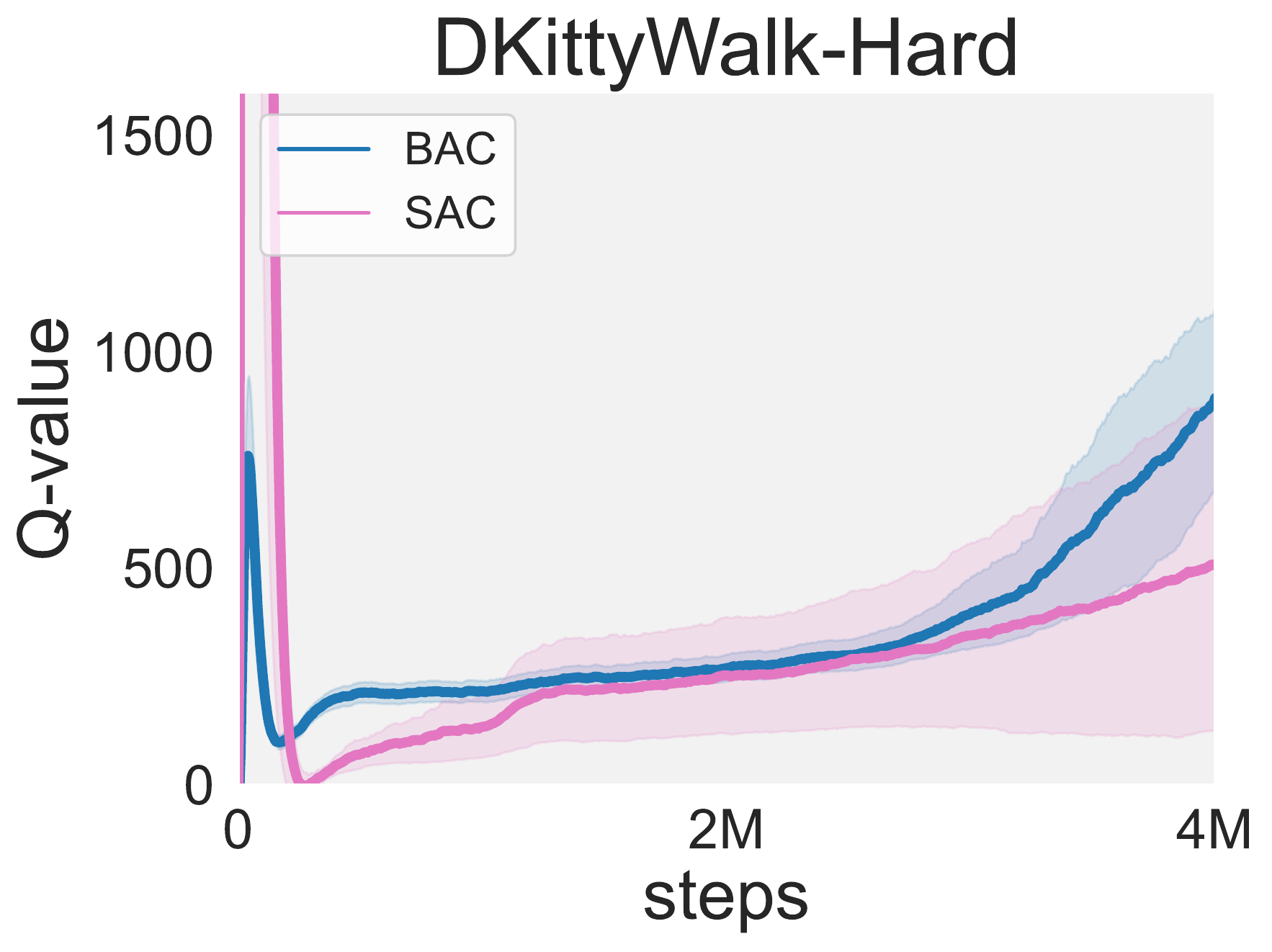}
                \vspace{-2.5mm}
                \caption{\small $Q$-value learning stability comparison. 10 seeds. The lower variance observed with BAC indicates better learning stability across runs. }
                \vspace{3mm}
                \label{fig:stableQ}
        \end{minipage}%
    \\
    \begin{minipage}{\linewidth}
        \centering
      \includegraphics[height=2.6cm,keepaspectratio]{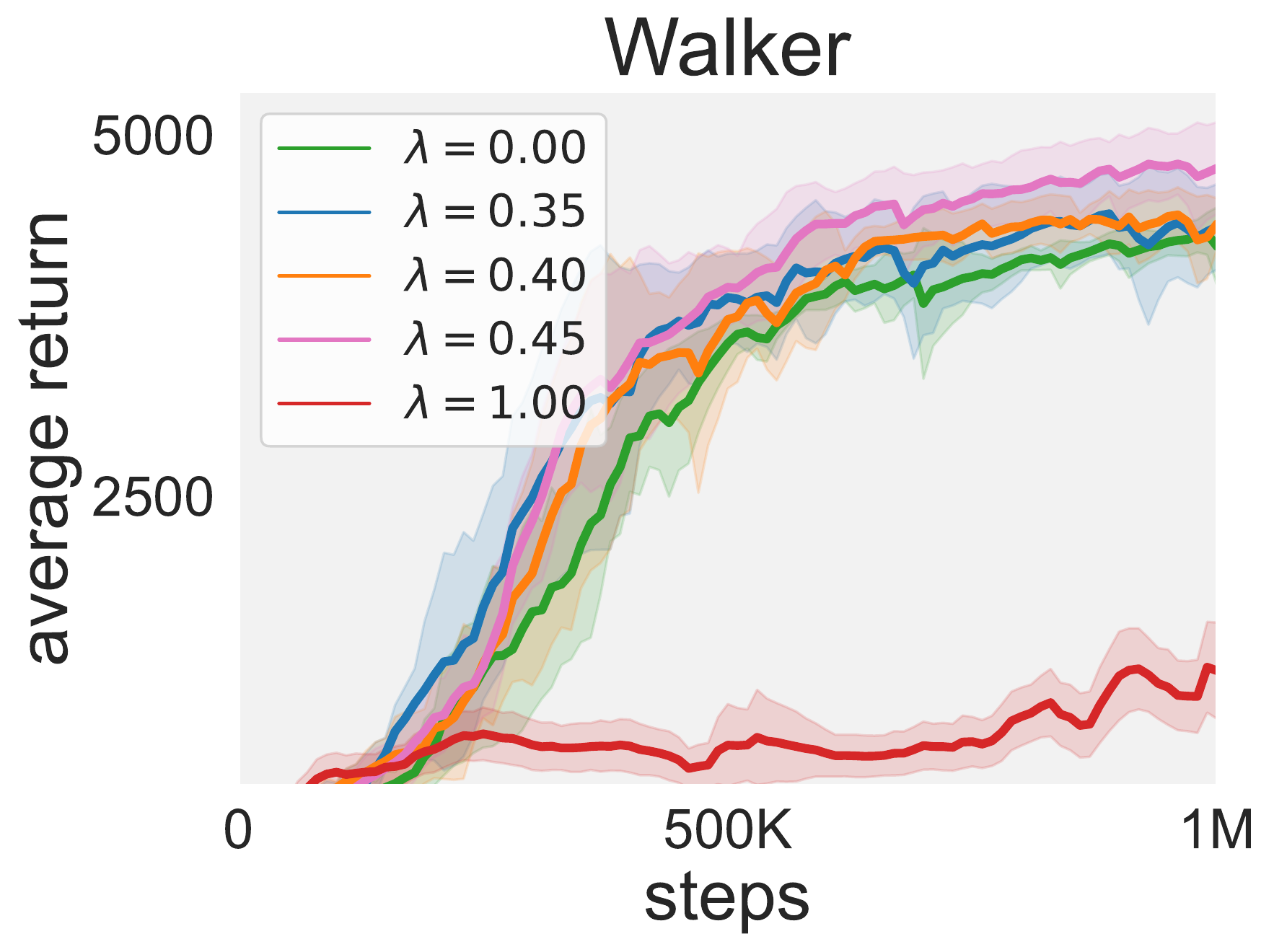} 
      \hspace{5mm} \includegraphics[height=2.6cm,keepaspectratio]{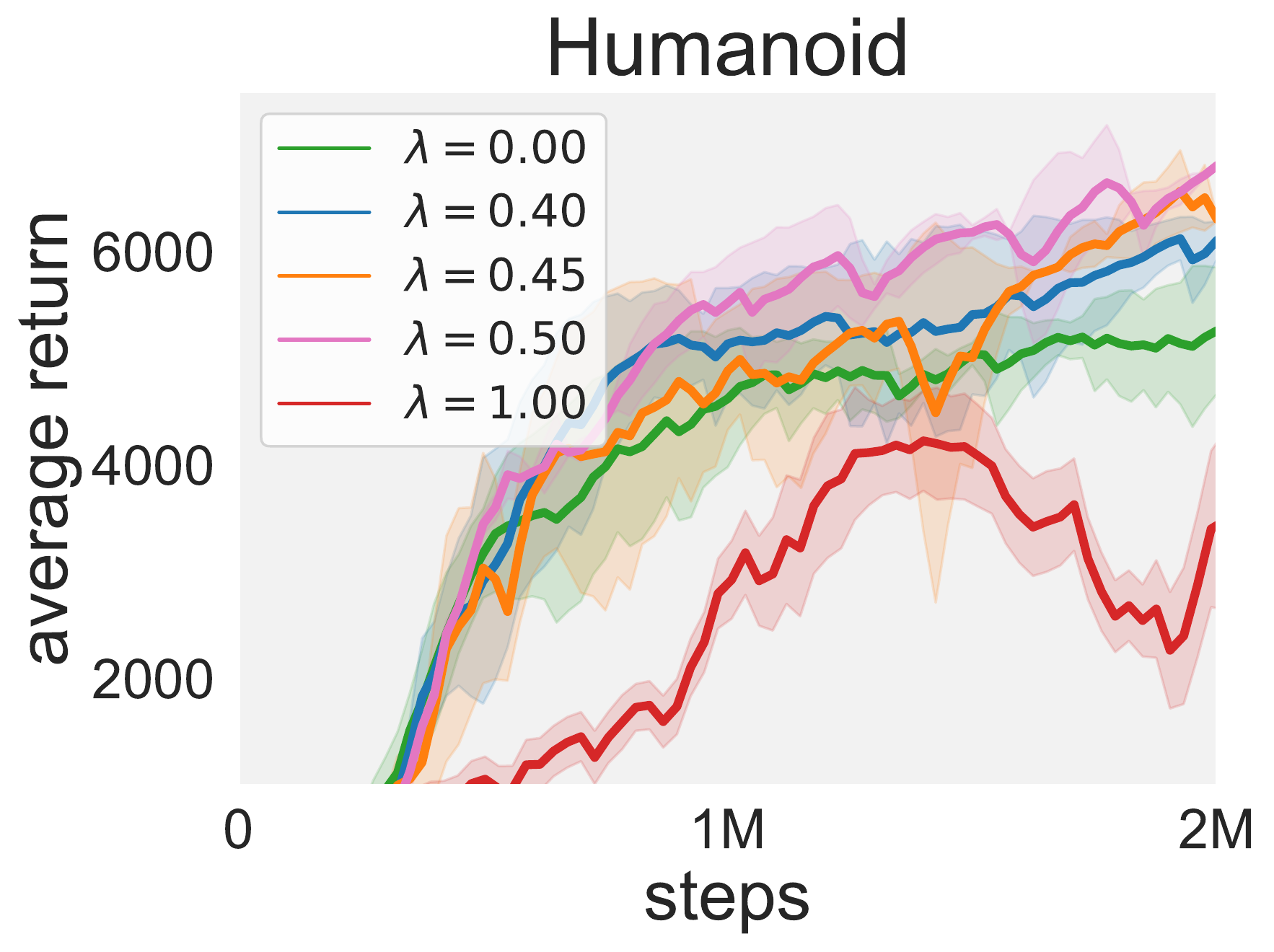}
        \vspace{-2.5mm}
        \caption{\small Hyperparameter study on $\lambda$.}
                \label{fig:lambda_param}
        \end{minipage}%
    \end{tabular}
\end{figure}
\noindent\textbf{Ability to seize serendipity.}\quad
To better understand how well the BEE operator captures past well-performing actions, we conduct experiments on the DKittyWalk-Medium task. We initialize SAC and \ourshort\ with the identical $Q$ network, random policy, and replay buffer. Next, we collected 15 trajectories~(2400 transitions in total)
using an expert policy whose success rate is 100\% and adding them to the replay buffer.  Keeping all components and parameters the same as in the main experiment, we train \ourshort\ and SAC on the blended buffer harboring several successful actions.  Figure~\ref{fig:mix_training} suggests that \ourshort\ recovers success faster than SAC, indicating its supposed ability to seize serendipity.

\noindent\textbf{More stable $Q$-value in practice.}\quad
In failure-prone scenarios, policy performance typically severely oscillates across iterations due to easily encountered failure samples from the current policy in $Q$-value update if using the Bellman evaluation operator. The $Q$-value learned by the BEE operator is less affected by the optimality level of the current policy, thus it might be expected to have better learning stability. The smaller error bar in Figure~\ref{fig:stableQ} supports it. To be specific, this stability is especially noteworthy in environments prone to failure, where policy oscillations are significantly influenced by environmental randomness (including seed variations). The Q-values derived from BAC are less susceptible to the immediate quality of the current policy, thus exhibiting lower variance across multiple runs.

\noindent\textbf{Hyperparameter study.}\quad
Setting an appropriate weighted coefficient $\lambda$, \ourshort\ could balance the exploitation and exploration well. We may note that the algorithm is reduced to the online version of IQL~\citep{IQL} for an extreme value $\lambda=0$. According to Figure~\ref{fig:lambda_param}, and the detailed settings and hyperparameter studies in Appendix~\ref{section:appendix-hyper}, we find that a moderate choice of $\lambda$ around 0.5 is sufficient to achieve the desired performance across all 35 locomotion and manipulation tasks we have benchmarked. This underscores that BAC does not need heavy tuning for strong performance.

\New{In-depth ablation studies in terms of the ability to counteract failure, effectiveness in noisy environments, and performance in sparse-reward tasks,  are presented in Appendix~\ref{section:failure-prone}.}

\section{Conclusion}
In this paper, we investigate the overlooked issue of value underestimation in off-policy actor-critic methods and find that incorporating sufficient exploitation might mitigate this issue.
These observations motivate us to propose the Blended Exploitation and Exploration~(BEE) operator, which leverages the value of past successes to enhance $Q$-value estimation and policy learning.
The proposed algorithms \ourshort\ and \mbshort\ outperform both model-based and model-free methods across various continuous control tasks. 
Remarkably, without further training,  \ourshort\  shines in real-robot tasks, emphasizing the need for improved general-purpose algorithms in real-world robotics.
Finally, our work sheds light on future work on fully fusing exploitation and exploration techniques, \emph{e.g.}, incorporating up-to-date design choices for computing $\max Q$ or exploration term,  in building strong RL methods.

\section*{Limitations and Broader Impact}
BAC is almost as lightweight as SAC, with only a slight increase in computation time. This increase is acceptable, especially since BAC often requires fewer interactions to achieve similar performance, potentially reducing overall computation time. Besides, as with any off-policy RL algorithm, safety measures are necessary to prevent unsafe behavior during real-world exploration.

This research advances both the cognition and the application of Reinforcement Learning, particularly in the domain of off-policy actor-critic framework. 

It sheds light on a critical yet previously underappreciated factor that underpins the underestimation bias in the latter stages of online RL training. 
Through extensive research, this work provides new insights into this previously overlooked aspect of RL. Moreover, this work demystifies the misperception that offline RL is over-conservative and incompatible with the online RL setting. It suggests a new paradigm that incorporates exploitation ingredients from offline RL to enhance pure online RL.

The elegance of the proposed algorithm lies in its simplicity and adaptability. It demonstrates exceptional performance across a broad spectrum of tasks with just one simple set of hyperparameters.
This positions it as a strong candidate for a backbone algorithm in various RL applications, showing promising possibilities for deployment in real-world contexts, notably in the field of robotics.

\addtocontents{toc}{\protect\setcounter{tocdepth}{3}}

\bibliography{Includes/IEEEabrv, ICML2024/ref}
\bibliographystyle{icml2024}

\newpage
\appendix
\onecolumn

{
\hypersetup{hidelinks}
\tableofcontents
}

\clearpage
\section{Extensive Related Works}\label{ap:extensive_related_works}
Off-policy actor-critic methods leverage a replay buffer to update the $Q$-function and policy~\citep{ddpg, a3c}, providing higher sample efficiency than on-policy RL methods. The prior works commonly rely on the standard policy gradient formulation~\citep{peters2008reinforcement} for policy improvement.
Various attempts have been devoted to modifying the policy evaluation procedure, primarily pursuing a high-quality value function to tackle the exploration or exploitation issue — central concerns in online RL~\citep{rnd, ecoffet2019go}.

Despite the ongoing interest in exploration and exploitation, most previous works devoted to exploration design following the optimism principle in the face of uncertainty~\citep{auer2008near, fruit2018efficient,szita2008many}, but view exploitation as merely maximizing $Q$-function. 
The Bellman evaluation operator, $\mathcal{T}Q(s,a) = r(s,a)+\gamma\mathbb{E}_{s'\sim P}\mathbb{E}_{a'\sim \pi}Q(s',a')$, underpins the critic learning. 
Existing efforts can be summarized into modifying this operator $\mathcal{T}$ in three main ways:
1) perturbing action $a'$ with techniques such as $\epsilon$-greedy, target policy smoothing~\citep{td3}, and pink noise~\citep{eberhard2022pink}; 
2) augmenting reward $r$ to foster exploration~\cite{ostrovski2017count, rnd, badianever, made};
3) directly adjusting $Q$ values such as max-entropy RL methods~\citep{ zhang2021exploration,hazan2019provably,lee2019efficient,islam2019marginalized,sac,dac} that infuse the operator with an entropy term, and optimistic exploration methods that learn Upper Confidence Bound (UCB)~\citep{ishfaq2021randomized,auer2002using,nikolovinformation} of ensemble $Q$-value networks~\citep{ob2i, oac,top}.  

Reducing value estimation bias, both underestimation and overestimation, has been widely explored in value-based reinforcement learning (RL) methods~\citep{lan2020maxmin, anschel2017averaged, yu2018historical, lee2013bias, zhang2017weighted}. However, value estimation bias remains a significant issue in off-policy actor-critic methods.

In essence, value overestimation might be associated with optimistic exploration~\citep{jin2018q,laskin2020reinforcement,top}, alongside factors such as off-policy learning and high-dimensional, nonlinear function approximation.
Hence, attempts to correct for overestimation, \emph{e.g.}, taking the minimum of two separate critics, have been widely adopted in the above exploration-driven methods~\citep{td3, sac, dac, rrs}. 
Yet directly applying such a minimum may cause underestimation~\citep{hasselt2010double}. To mitigate it, prior methods~\citep{oac, top} seek for a milder form, assuming the epistemic uncertainty as the standard deviation of ensemble $Q$ values.
We identify the value underestimation that particularly occurs in the latter training stages and uncover its long-neglected culprit.
Our findings suggest that incorporating sufficient exploitation into current exploration-driven algorithms would be a natural solution and lead to an improved algorithm.

\Revise{Experience Replay~(ER)~\citep{dqn} boosts exploitation in off-policy RL by enabling data reuse. Recent works in prioritized replay ~\citep{schaul2015prioritized,liu2021regret,sinha2022experience} propose various metrics to replay or reweight important transitions more frequently, benefiting sample efficiency. We primarily implement BAC with the vanilla ER method for simplicity, yet more advanced ER techniques could be integrated for further enhancement.}
Outside the online RL paradigm, imitation learning~\citep{pomerleau1988alvinn,schaal1996learning,ross2011reduction} and offline RL algorithms~\citep{BCQ, bear,kumar2020conservative,IQL,zhan2022offline} are known for their effective exploitation of provided datasets. 
Although the prospect of integrating these techniques to enhance online RL is attractive, offline learning is often considered overly conservative and requires a reasonable-quality dataset for high performance~\citep{li2022data}, leading to limited success in improving online learning~\citep{niu2022trust}.
In standard online RL, we only have access to a dynamic and imperfect replay buffer, rather than a well-behaved dataset. 
As a result, recent efforts are mainly under a two-stage paradigm, integrating these techniques as policy pre-training for subsequent online training, such as initializing the policy with behavior cloning~\citep{hester2018deep, shah2021rrl,wangvrl3, baker2022video}  or performing offline RL followed by online fine-tuning~\citep{nair2020awac, lee2022offline,hansen2023idql}. 
By contrast, our work suggests a new paradigm that incorporates exploitation ingredients from offline RL to enhance pure online RL, as demonstrated in our proposed framework.

\clearpage
\section{Omitted Proofs} \label{ap-omittedproofs}

\begin{proposition}[\textbf{Policy evaluation}]\label{policy-evaluation}
Consider an initial $Q_0:\mathcal{S}\times\mathcal{A}\rightarrow\mathbb{R}$ with $\vert \mathcal{A}\vert < \infty$, and define $Q_{k+1}= \mathcal{B}^{\{\mu,\pi\}}Q_{k}$. Then the sequence $\{Q_{k}\}$ converges to a fixed point $Q^{\{\mu,\pi\}}$ as $k\rightarrow \infty$.
\end{proposition}
\begin{proof}
First, let us show that the BEE operator $\mathcal{B}$ is a $\gamma$-contraction operator in the $\mathcal{L}_\infty$ norm.

Let $Q_1$ and $Q_2$ be two arbitrary $Q$ functions, for the Bellman Exploitation operator $\mathcal{T}_{exploit}$, since target-update actions $a'$ are extracted from $\mu$, we have that,
\[
\begin{aligned}
\Vert \mathcal{T}_{exploit}^{\mu}Q_1 - \mathcal{T}_{exploit}^{\mu}Q_2\Vert_\infty
=&\max_{s,a}\vert (r(s,a)+ \gamma\mathbb{E}_{s'}\max_{a'\sim \mu}[Q_1(s',a')]) -(r(s,a)+ \gamma\mathbb{E}_{s'}\max_{a'\sim \mu}[Q_2(s',a')])\vert\\
=&\gamma\max_{s,a} \vert \mathbb{E}_{s'}[\max_{a'\sim \mu}Q_1(s',a')- \max_{a'\sim \mu}Q_2(s',a')]\vert\\
\leq& \gamma \max_{s,a}\mathbb{E}_{s'}\vert \max_{a'\sim \mu} Q_1(s',a')- \max_{a'\sim \mu}Q_2(s',a')\vert\\\leq & \gamma \max_{s,a} \Vert Q_1 - Q_2\Vert_\infty= \gamma\Vert Q_1 - Q_2\Vert_\infty
\end{aligned}
\]
Also, for the Bellman Exploration Operator $\mathcal{T}_{explore}$, as $a'\sim \pi$, we have, 
\[
\begin{aligned}
\Vert \mathcal{T}_{explore}^{\pi} Q_1 - \mathcal{T}_{explore}^{\pi} Q_2\Vert_\infty=
& \max_{s,a} \vert \gamma \mathbb{E}_{s'}\big[\mathbb{E}_{a'\sim \pi}Q_1(s',a') - \gamma \mathbb{E}_{a'\sim \pi}Q_2(s',a')\big]\vert \\
\leq &\gamma \max_{s,a} \mathbb{E}_{s'}\vert \mathbb{E}_{a'\sim \pi}Q_1(s',a') - \mathbb{E}_{a'\sim \pi}Q_2(s',a')\vert\\
\leq & \gamma\max_{s,a} \mathbb{E}_{s'}\mathbb{E}_{a'\sim \pi}\vert Q_1(s',a') - Q_2(s',a')\vert\\
\leq & \gamma \max_{s,a}\Vert Q_1- Q_2\Vert_\infty
=  \gamma \Vert Q_1 -Q_2\Vert_\infty
\end{aligned}
\]
Combining the results, we have that the BEE operator satisfies $\gamma$-contraction property: 
\[
\begin{aligned}
\Vert \mathcal{B}^{\{\mu, \pi\}}Q_1 - \mathcal{B}^{\{\mu, \pi\}}Q_2\Vert_\infty
=& \Vert \lambda (\mathcal{T}_{exploit}^{\mu}Q_1- \mathcal{T}_{exploit}^{\mu}Q_2) + (1-\lambda)(\mathcal{T}_{explore}^{\pi}Q_1- \mathcal{T}_{explore}^{\pi}Q_2)\Vert_\infty\\
\leq & \lambda\Vert \mathcal{T}_{exploit}^{\mu}Q_1- \mathcal{T}_{exploit}^{\mu}Q_2\Vert_\infty + (1-\lambda)\Vert \mathcal{T}_{explore}^{\pi}Q_1- \mathcal{T}_{explore}^{\pi}Q_2\Vert_\infty\\\leq& \lambda \gamma\Vert Q_1-Q_2\Vert_\infty + (1-\lambda)\gamma\Vert Q_1-Q_2\Vert_\infty=\gamma\Vert Q_1- Q_2\Vert_\infty
\end{aligned}
\]
we conclude that the BEE operator is a $\gamma$ -contraction, which naturally leads to the conclusion that any initial Q function will converge to a unique ﬁxed point by repeatedly applying $\mathcal{B}^{\{\mu,\pi\}}$.

\end{proof}

\begin{proposition}[\textbf{Policy improvement}]\label{proof-policy-improvement}
Let $\{\mu_{k},\pi_{k}\}$ be the policies at iteration $k$, and $\{\mu_{k+1},\pi_{k+1}\}$ be the updated policies, where  $\pi_{k+1} $ is the greedy policy of the $Q$-value. 
Then for all $(s,a)\in \mathcal{S}\times \mathcal{A}$, $\vert \mathcal{A}\vert < \infty$, we have $Q^{\{\mu_{k+1},\pi_{k+1}\}}(s,a) \geq Q^{\{\mu_k,\pi_k\}}(s,a)$.
\end{proposition}
\begin{proof}
At iteration $k$, $\mu_k$ denotes the policy mixture and $\pi_k$ the current policy, and the corresponding value function is $Q^{\{\mu,\pi\}}$.
We firstly update the policies from $\{\mu_k,\pi_k\}$  to $\{\mu_{k},\pi_{k+1}\}$, where $\pi_{k+1}$  is the greedy policy w.r.t $J_{\pi_k,\mu_k}(\mu_k,\pi)$, \emph{i.e.}, $\pi_{k+1} = \arg\max_{\pi}\mathbb{E}_{a\sim \pi}[Q^{\{\mu_k,\pi_k\}}(s,a)-\omega(s,a\vert \pi)]$. 

We commence with the proof that $Q^{\{\mu_k,\pi_{k+1}\}} (s,a) \geq Q^{\{\mu_k, \pi_{k}\}}(s,a)$ for all $(s,a) \in \mathcal{S}\times\mathcal{A}$.
Since $\pi_{k+1} = \arg\max_{\pi}J_{\pi_k,\mu_k}(\mu_k,\pi)$, we have that 
$J_{\pi_k,\mu_k}(\mu_k,\pi_{k+1}) \geq J_{\pi_k,\mu_k}(\mu_k,\pi_k)$. 
Expressing $J_{\pi_k,\mu_k}(\mu_k,\pi_{k+1})$ and $J_{\pi_k,\mu_k}(\mu_k,\pi_k)$ by their definition, we have 
$\mathbb{E}_{a\sim \pi_{k+1}}[Q^{\{\mu_k,\pi_k\}}(s,a)-\omega(s,a\vert \pi_{k+1})] \geq \mathbb{E}_{a\sim \pi_k}[Q^{\{\mu_k,\pi_k\}}(s,a) -\omega(s,a\vert \pi_{k})]$.

In a similar way to the proof of the soft policy improvement~\citep{sac}, we come to the following inequality:

\[
\begin{aligned}
Q^{\{\mu_k,\pi_k\}}(s_t,a_t) 
 =& r(s_t,a_t) + \gamma\mathbb{E}_{s_{t+1}}
 \big\{
 \lambda\cdot \max_{\tilde{a}_{t+1}\sim \mu_k} Q^{\{\mu_k,\pi_k\}}(s_{t+1},\tilde{a}_{t+1}) \\&+ (1-\lambda)\cdot \mathbb{E}_{a_{t+1}\sim\pi_k}[Q^{\mu_k,\pi_k}(s_{t+1},a_{t+1})-\omega(s_{t+1},a_{t+1}\vert\pi_{k})]
 \big\} 
 \\
\leq & r(s_t,a_t) + \gamma\mathbb{E}_{s_{t+1}}\{\lambda\cdot \max_{\tilde{a}_{t+1}\sim \mu_k} Q^{\{\mu_k,\pi_{k}\}}(s_{t+1},\tilde{a}_{t+1}) + \\
&(1-\lambda)\cdot \mathbb{E}_{a_{t+1}\sim\pi_{k+1}}[Q^{\{\mu_k,\pi_{k}\}}(s_{t+1},a_{t+1})-\omega(s_{t+1},a_{t+1}\vert\pi_{k+1})]\}\\
& \vdots\\
\leq& Q^{\{\mu_k, \pi_{k+1}\}}(s_t,a_t)
\end{aligned}
\]

Here, the inequality is obtained by repeatedly expanding $Q^{\{\mu_k,\pi_k\}}$ on the RHS through 
$Q^{\{\mu_k,\pi_k\}}(s,a) 
 =r(s,a) + \gamma\mathbb{E}_{s'}
 \big\{
 \lambda\cdot \max_{\tilde{a}'\sim \mu_k} Q^{\{\mu_k,\pi_k\}}(s',\tilde{a}') + (1-\lambda)\cdot \mathbb{E}_{a'\sim\pi_k}[Q^{\{\mu_k,\pi_k\}}(s',a')-\omega(s',a'\vert\pi_{k})]
 \big\}$ 
and applying the inequality $\mathbb{E}_{a\sim \pi_{k+1}}[Q^{\{\mu_k,\pi_k\}}(s,a)-\omega(s,a\vert \pi_{k+1})] \geq \mathbb{E}_{a\sim \pi_k}[Q^{\{\mu_k,\pi_k\}}(s,a) -\omega(s,a\vert \pi_{k})]$.  Finally, we arrive at convergence to $Q^{\{\mu_k, \pi_{k+1}\}}(s_t,a_t)$.

Then, we expand the historical policy sequence $\Pi_k=\{\pi_0, \pi_1, \cdots, \pi_{k-1}\}$ by adding the policy $\pi_k$, and obtain  $\Pi_{k+1}=\{\pi_0, \pi_1, \cdots, \pi_{k}\}$. 
Next, we consider to prove $Q^{\{\mu_{k+1},\pi_{k+1}\}} (s,a) \geq Q^{\{\mu_k, \pi_{k+1}\}}(s,a), \forall (s,a) \in \mathcal{S}\times\mathcal{A}$. 
Recall that $\mu_{k+1}$ is the stationary policy mixture of $\Pi_{k+1}$, if the state-action visitation density $d^{\pi_i}(s,a) > 0, i = 0, \ldots k$, then the corresponding mixture distribution $d^{\mu}(s,a) > 0$, hence the support region of $\mu_{k}$ is a subset of the support region of $\mu_{k+1}$, \emph{i.e.}, $\text{supp}(\mu_k) \in \text{supp}(\mu_{k+1})$. Since $\max_{a\sim \mu_{i}} Q(s,a) = \max_{a\in\text{supp}(\mu_{i})} Q(s,a)$, then for any $Q: \mathcal{S}\times \mathcal{A} \rightarrow \mathbb{R}$, the following inequality can be established:
\[
\max_{a\sim \mu_{k+1}} Q(s,a) \geq \max_{a\sim \mu_k} Q(s,a), \forall s\in \mathcal{S}
\]
Hence, we expand the $Q^{\{\pi,\mu\}}$ and utilize the above inequality repeatedly, then we obtain
\[
\begin{aligned}
Q^{\{\mu_k,\pi_{k+1}\}}(s,a)=&r(s,a)+\lambda\gamma\cdot \mathbb{E}_{s'}[\max_{a'\sim \mu_{k}(\cdot\vert s')} Q^{\{\mu_k,\pi_{k+1}\}}(s',a')] \\
&+(1-\lambda)\gamma \mathbb{E}_{s'}\mathbb{E}_{a'\sim\pi_{k+1}} [Q^{\{\mu_k,\pi_{k+1}\}}(s',a')]\\
\leq& r(s,a)+\lambda\gamma\cdot \mathbb{E}_{s'}[\max_{a'\sim \mu_{k+1}(\cdot\vert s')} Q^{\{\mu_k,\pi_{k+1}\}}(s',a')]\\
&+ (1-\lambda)\gamma \mathbb{E}_{s'}\mathbb{E}_{a'\sim\pi_{k+1}} [Q^{\{\mu_k,\pi_{k+1}\}}(s',a')]\\
& \vdots\\
\leq&Q^{\{\mu_{k+1},\pi_{k+1}\}}(s,a)
\end{aligned}
\]

With the inequalities of these two stages, the policy improvement property is satisfied, $Q^{\{\mu_{k+1},\pi_{k+1}\}}(s,a) \geq Q^{\{\mu_k,\pi_k\}}(s,a), \forall (s,a)\in \mathcal{S}\times\mathcal{A}, \vert \mathcal{A}\vert < \infty$. 

\end{proof}

\begin{proposition}
[\textbf{Policy iteration}]
Assume $\vert \mathcal{A}\vert < \infty$, by repeating iterations of the policy evaluation and policy improvement, any initial policies converge to the optimal policies $\{\mu^*, \pi^*\}$, s.t. $Q^{\{\mu^*,\pi^*\}}(s_t,a_t)\geq Q^{\{\mu',\pi'\}}(s_t,a_t), \forall \mu' \in \Pi, \pi'\in \Pi, \forall (s_t,a_t) \in \mathcal{S}\times \mathcal{A} $.
\end{proposition}

\begin{proof}
Let $\Pi$ be the space of policy distributions and let $\{\mu_i, \pi_i\}$ be the policies at iteration $i$. 
By the policy improvement property in Proposition~\ref{policy-improvement}, the sequence $Q^{\{\mu_i,\pi_i\}}$ is monotonically increasing. Also, for any state-action pair $(s_t,a_t)\in \mathcal{S}\times \mathcal{A}$, each $Q^{\mu_i,\pi_i}$ is bounded due to the discount factor $\gamma$. Thus, the sequence of $\{\mu_i, \pi_i\}$ converges to some $\{\mu^*, \pi^*\}$ that are local optimum. 
We will still need to show that $\{\mu^*, \pi^*\}$ are indeed optimal, we assume finite MDP, as typically assumed for convergence proof in usual policy iteration~\citep{sutton1988learning}. 
At convergence, we get $J_{\mu^*,\pi^*}(\mu^*,\pi^*)[s]\geq J_{\mu^*,\pi^*}(\mu',\pi')[s], \forall \pi'\in \Pi, \mu' \in \Pi$. Using the same iterative augument as in the proof of Proposition~\ref{policy-improvement}, we get $Q^{\{\mu^*, \pi^*\}}(s,a) \geq Q^{\{\mu', \pi'\}}(s,a)$ for all $(s,a)\in \mathcal{S}\times \mathcal{A}$. Hence, $\{\mu^*,\pi^*\}$ are optimal in $\Pi$.

\end{proof}

\clearpage
\section{Implementation Details and Extensive Design Choices}

\subsection{Primary implementation details on \ourshort}
Instantiating \ourshort\ amounts to specifying two main components: the use of in-sample learning for calculating the Bellman Exploitation operator $\mathcal{T}_{exploit}$, and the application of entropy regularization in the Bellman Exploration operator $\mathcal{T}_{explore}$. Here we provide the details for our primary implementation. For a broader discussion of potential design choices and extensions refer to Section~\ref{section:designchoice}. 

\paragraph{In-sample learning for $\mathcal{T}_{exploit}$.}
We leverage a simple and efficient approach for policy extraction using expectile regression~\citep{IQL} to learn the value function, where only a hyperparameter $\tau$ is introduced. Considering that some large $Q$-values potentially are a result of ``lucky'' samples, we introduce a state value function $V$ which approximates a high expectile of $Q(s,a)$ on the replay buffer $\mathcal{D}$. In this way, we can better account for the potential variance in $Q$-values, reducing overestimation error risk and ensuring that our algorithm is not relying solely on ``lucky'' samples.

To be specific, we initialize a state value $V$ network to capture the maximum of $Q$ value. Given the replay buffer $\mathcal{D}$, we can update the $V$ network by a high expectile $\tau$ of $Q(s,a)$,
\begin{equation*}
V(s)\leftarrow \arg\min_{V}\mathbb{E}_{(s,a)\sim \mathcal{D}}\left[|\tau-\mathbbm{1}(Q(s,a)-V(s)<0)|(Q(s,a)-V(s))\right]^2
\end{equation*}
Given $\tau>0.5$, this asymmetric loss function would downweight the contributions of $Q(s,a)$ when $Q(s,a)<V(s)$ while giving more weights to larger values. If $\tau\rightarrow 1$, we have $V(s)\rightarrow \max_{a\sim\mu_k}Q(s,a)$. Hence, the target value of $\mathcal{T}_{exploit}$ can be calculated by
\begin{equation*}
\mathcal{T}_{exploit}Q(s,a)=r(s,a)+\gamma\mathbb{E}_{s'\sim\mathcal{D}}\left[V(s')\right].
\end{equation*}

\paragraph{Entropy regularization in $\mathcal{T}_{explore}$.}
Based on the follow-up actions $a'$ derived from fresh policy $\pi_\theta$, we compute $\mathcal{T}_{explore}Q(s,a)$, employing the entropy regularization $\alpha\log\pi(a_t\vert s_t)$ from SAC~\citep{sac} as the $\omega(\cdot\vert \pi)$.  
To ease the computational burden of learning a separate $V$-function for $\mathcal{T}_{explore}$, we opt to directly compute the expectation of the $Q$-value. Thus, the target value of the Bellman Exploration operator $\mathcal{T}_{exploit}$ can be calculated as follows:
\begin{equation*}
    \mathcal{T}_{explore}Q(s,a) = r(s,a)  + \gamma\mathbb{E}
_{s'\sim \mathcal{D}} \big[\mathbb{E}_{a'\sim \pi}Q(s',a') - \alpha\log\pi(a'\vert s')\big]
\end{equation*}

\paragraph{Algorithm overview on \ourshort.}
The pseudocode of our proposed \ourshort\ is provided in \Algref{algorithm_bac}.

\begin{algorithm}[htb]
    \caption{Primary Implementation of BEE Actor-Critic (\ourshort)}
    \label{algorithm_bac}
    \begin{algorithmic}
    \STATE \textbf{initialize:} Q networks $Q_{\phi}$, V network $V_{\psi}$, policy $\pi_\theta$, replay buffer $\mathcal{D}$
    \STATE Sample $n$ tuples from random policy and add to $\mathcal{D}$
    \REPEAT
        \FOR{each gradient step}
            \STATE Sample a mini-batch of $N$ transitions $(s, a, r, s')$ from $\mathcal{D}$
            \STATE Update $V_{\psi}$ by $\min_\psi \mathbb{E}_{s,a} \left[ |\tau - \mathbbm{1}(Q_\phi(s, a) < V_\psi(s))| (Q_\phi(s, a) - V_\psi(s))^2 \right]$
        \ENDFOR
        \FOR{each environment step}
            \STATE Collect data with $\pi_\theta$ from real environment; add to $\mathcal{D}$
        \ENDFOR
        \FOR{each gradient step}
            \STATE Compute $\mathcal{T}_{exploit}Q_{\phi}(s, a) \leftarrow r + \gamma \mathbb{E}_{s'}[V_{\psi}(s')]$
            \STATE Compute $\mathcal{T}_{explore}Q_{\phi}(s, a) \leftarrow r + \gamma \mathbb{E}_{s'} \mathbb{E}_{a' \sim \pi_\theta}[Q_{\phi}(s', a') - \alpha \log \pi_{\theta}(a' \vert s')]$
            \STATE Calculate the target Q value: $\mathcal{B}Q_{\phi} \leftarrow \lambda \mathcal{T}_{exploit}Q_{\phi} + (1 - \lambda) \mathcal{T}_{explore}Q_{\phi}$
            \STATE Update $Q_{\phi}$ by $\min_\phi \left(\mathcal{B}Q_{\phi} - Q_{\phi}\right)^2$
            \STATE Update $\pi_\theta$ by $\max_{\theta} Q_\phi(s, a)$
        \ENDFOR
    \UNTIL{the policy performs well in the environment}
    \end{algorithmic}
\end{algorithm}

\subsection{Primary implementation details on \mbshort\ algorithm}\label{section:appendix-mbbac}
\paragraph{Modeling and learning the dynamics models.} 
We adopt the widely used model learning technique in our baseline methods~\citep{mbpo, autombpo, ji2022update, zhang2024fine}. To be specific, 
\mbshort\ uses a bootstrap ensemble of dynamics models $\{ \hat {f}_{\phi_1},\hat {f}_{\phi_2}, \ldots, \hat {f}_{\phi_K}\}$. 
They are fitted on a shared replay buffer ${\cal D}_e$, with the data shuffled differently for each model in the ensemble. The objective is to optimize the Negative Log Likelihood (NLL), 
\[
{\cal L}^H(\phi) = \sum\limits_{t}^{H}[\mu_{\phi}(s_t,a_t)-s_{t+1}]^T\Sigma_{\phi}^{-1}(s_t,a_t)[\mu_{\phi}(s_t,a_t)-s_{t+1}] + \log \det \Sigma_{\phi}(s_t,a_t).
\] 
The prediction for these ensemble models is, $\hat{s}_{t+1} = \frac{1}{K}\sum_{i = 1}^{K} \hat{f}_{\phi_i}(s_t, a_t)$. More details on network settings are presented in Table~\ref{mbbac-hyperparameters}.

\paragraph{Policy optimization and model rollouts.} 
We employ \ourshort\ as the policy optimization oracle in \mbshort. Using the truncated short model rollouts strategy~\citep{mbpo, autombpo, m2ac, ji2022update}, we generate model rollouts from the current fresh policy. 
In the policy evaluation step, we repeatedly apply the BEE operator to the $Q$-value.  We compute the $V$-function for the Bellman Exploitation operator on the environment buffer ${\cal D}_e$, which contains real environment interactions collected by historical policies, and we compute the $\mathcal{T}_{explore}Q$ operation to the model buffer ${\cal D}_m$ generated by the current policy $\pi$.

\paragraph{Algorithm overview on \mbshort.}
We give an overview of  \mbshort\ in \Algref{algorithm_mbbac}.

\begin{algorithm}[ht]
\caption{Primary Implementation of Model-based BAC~(\mbshort)}
\begin{algorithmic}
    \STATE \textbf{initialize:} Q networks $Q_{\phi}$, V network $V_{\psi}$, policy $\pi_\theta$, ensemble models $\{ \hat {f}_{\phi_1},\hat {f}_{\phi_2}, \ldots, \hat {f}_{\phi_K}\}$, environment buffer ${\cal D}_e$ and model buffer ${\cal D}_m$;

    \REPEAT
        \FOR{each environment step}
            \STATE Collect data with $\pi_\theta$ from real environment; add to $\mathcal{D}_e$. \textcolor{gray}{// Interactions with real env}
        \ENDFOR
        \FOR{each gradient step}
            \STATE Train all models $\{ \hat {f}_{\phi_1},\hat {f}_{\phi_2}, \ldots, \hat {f}_{\phi_K}\}$ on  ${\cal D}_e$. 
            \textcolor{gray}{// Model learning}
        \ENDFOR
    
        \FOR{each model rollout step}
            \STATE Perform $h$-step model rollouts using policy $\pi_\theta$; add to ${\cal D}_m$. \textcolor{gray}{// Model rollouts}
        \ENDFOR
        
        \STATE
        \begin{minipage}{0.95\linewidth}
            \begin{mdframed}[linewidth=0.5pt, linecolor=myblue]
                \STATE \textcolor{myblue}{// Policy optimization}
                \STATE Update $V_{\psi}$ by $\min_\psi\mathbb{E}_{{s,a}\sim \mathcal{D}_e}\vert \tau - \mathbbm{1}\left(Q_\phi(s,a)<V_\psi(s)\right)\vert\left(Q_\phi(s,a)-V_\psi(s)\right)^2$
                \STATE Compute $\mathcal{T}_{exploit}Q_{\phi}(s,a)=r(s,a)+\gamma\mathbb{E}_{s'\sim \mathcal{D}_m} [V_\phi(s')]$.
                \STATE Compute $\mathcal{T}_{explore}Q_{\phi}(s,a)=r(s,a)+\gamma\mathbb{E}_{s'\sim \mathcal{D}_m}\mathbb{E}_{a'\sim \pi}[Q(s',a')-\alpha\log\pi_\theta(a'\vert s')]$.
                \STATE Calculate the target Q value: $\mathcal{B}Q_{\phi} \leftarrow\lambda \mathcal{T}_{exploit}Q_{\phi} + (1-\lambda)\mathcal{T}_{explore}Q_{\phi}$
            \end{mdframed}
        \end{minipage}
        \STATE
        
        \FOR{each gradient step}
            \STATE Update $Q_{\phi}$ by $\min_\phi\left(\mathcal{B}Q_{\phi}-Q_{\phi}\right)^2$
            \STATE Update $\pi_\theta$ by $\max_{\theta} Q_\phi(s,a)$. 
            \textcolor{gray}{// Policy optimization}
        \ENDFOR
    \UNTIL{the policy performs well in the environment}
\end{algorithmic}
\label{algorithm_mbbac}
\end{algorithm}

\clearpage
\subsection{Possible design choices and extensions}\label{section:designchoice}
\subsubsection{More design choices on computing $\mathcal{T}_{exploit}Q$}\label{section:maxQ}
Towards computing $\mathcal{T}_{exploit}Q$ based on the policy mixture $\mu$, 
a direct solution might be using an autoencoder to model $\mu$~\citep{BCQ, MCB}. Unfortunately, in the online setting, learning $\mu$ would be computationally expensive as it varies dynamically with policy iterations.
In our main implementation, we use the expectile regression, an in-sample approach, for the computation of $\max Q$. Beyond this, here we introduce two other in-sample techniques that can be used to calculate $\max Q$.

\paragraph{Sparse Q-learning.}
Sparse Q-learning~\citep{xuoffline} considers an implicit value regularization framework by imposing a general behavior regularization term. When applied Neyman $\mathcal{X}^2$-divergence as the regularization term, the state value function can be trained by
\begin{equation*}
V(s)\leftarrow \arg\min_{V}\mathbb{E}_{(s,a)\sim \mathcal{D}}\left[\mathbbm{1}\left(1+\frac{Q(s,a)-V(s)}{2\alpha}>0\right)\left(1+\frac{Q(s,a)-V(s)}{2\alpha}\right)^2+\frac{V(s)}{2\alpha}\right].
\end{equation*}

\paragraph{Exponential Q-learning.} Similar to sparse Q-learning, exponential Q-learning~\citep{xuoffline} utilizes Reverse KL divergence as the regularization term and the state value function $V(s)$ can be updated by
\begin{equation*}
    V(s)\leftarrow \arg \min_{V} \mathbb{E}_{(s,a)\sim \mathcal{D}}\left[
    \exp\left(\frac{Q(s,a)-V(s)}{\alpha}\right) + \frac{V(s)}{\alpha}
    \right].
\end{equation*}

Based on the state value function $V(s)$ learned by sparse Q-learning or exponential Q-learning, we can compute the $\mathcal{T}_{exploit}Q$ by, 
\begin{equation*}
    \mathcal{T}_{exploit}Q(s,a) = r(s,a) + \gamma \mathbb{E}_{s'\sim \mathcal{D}}\left[V(s')\right].
\end{equation*}

\begin{wrapfigure}[13]{r}{0.38\textwidth}
    \centering
    \includegraphics[height=3.45cm,keepaspectratio]{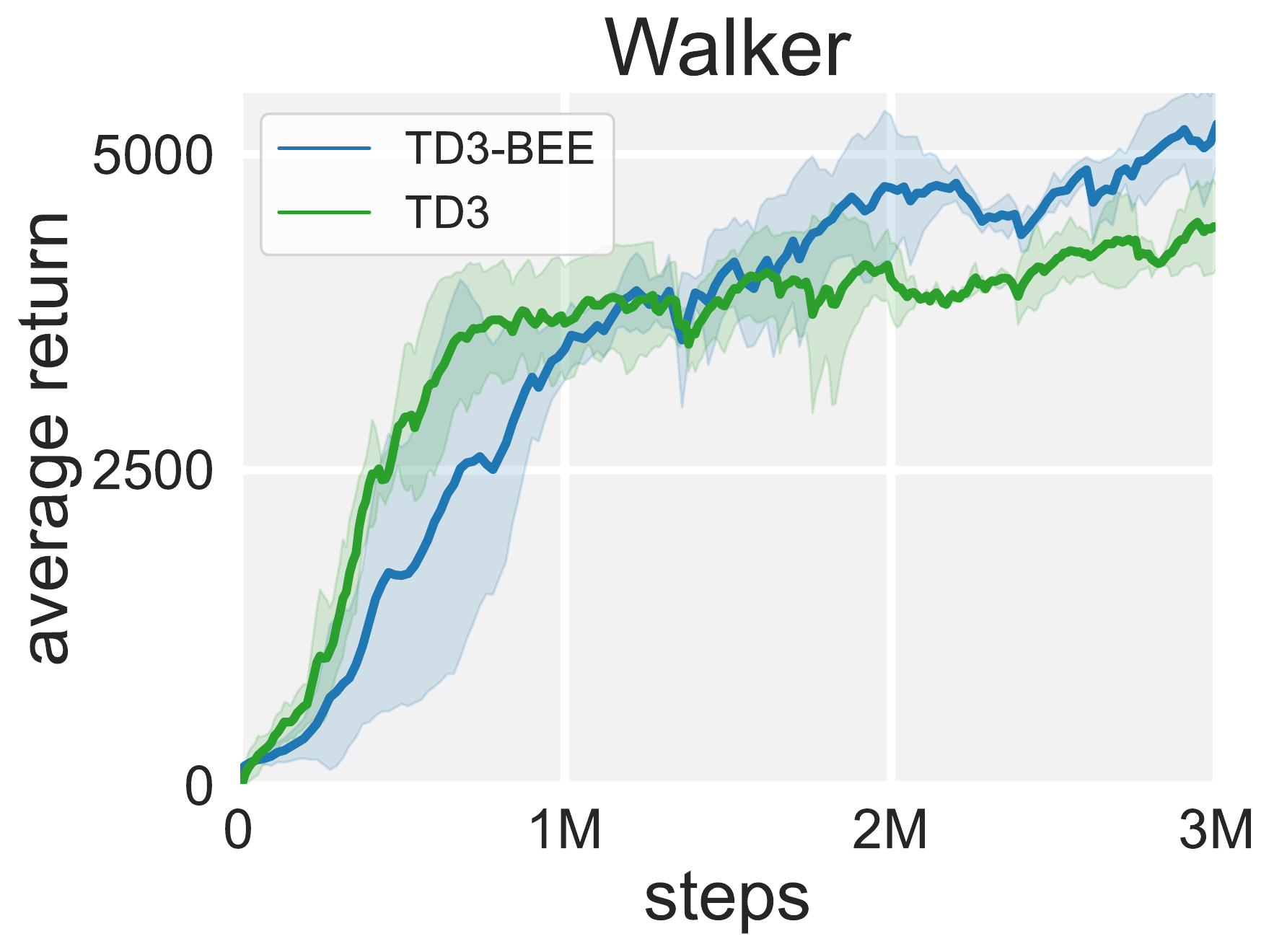}
    \caption{Ablation on the target policy smoothing regularization variant.}
    \label{fig:TD3-BEE}
\end{wrapfigure}

\subsubsection{More design choices on exploration term $\omega(\cdot\vert \pi)$}

In our primary implementation, we adopt the widely-used entropy regularization proposed in SAC~\citep{sac}. 
Various exploration terms $\omega(s,a\vert \pi)$, which have been extensively explored in previous off-policy actor-critic methods~\citep{sac, td3, dac}, could be adopted in our algorithm.  

\textbf{Variant on target policy smoothing regularization.} Here we conduct an ablation study upon adopting the target policy smoothing regularization introduced by TD3~\citep{td3}, we term the variant of our algorithm TD3-BEE. Compared to TD3, our method exhibits improvements, as demonstrated in two experiments on D'Kitty in the main paper, as well as the Walker2d-v2 experiment in Figure~\ref{fig:TD3-BEE}.

\textbf{Other possible extensions.} Various up-to-date advances in exploration term designs can be incorporated into our algorithm. For instance, pink noise~\citep{eberhard2022pink} could be utilized to replace target policy smoothing regularization. Additionally, specially designed entropy terms, such as state or state-action occupancy entropy based on Shannon entropy~\citep{hazan2019provably, islam2019marginalized,lee2019efficient} or R{'e}nyi entropy~\citep{zhang2021exploration}, could be considered. In certain ``hard-exploration'' scenarios~\citep {aytar2018playing, ecoffet2019go,ecoffet2021first}, it may be beneficial to use specially tailored exploration terms, such as sample-aware entropy regularization~\citep{dac}, particularly in sparse-reward or delayed-reward scenarios.

\subsubsection{Extensions: automatic adaptive $\lambda$ mechanisms}
\label{section:auto-lambda-mechanism}
In our main experiments, we use a fixed $\lambda$ value for simplicity. Although the value of $\lambda$ does not fluctuate significantly in most of the environments we tested, specific scenarios may necessitate some tuning effort to find an appropriate $\lambda$. 

To circumvent this $\lambda$ search, we present three possible automatic adaptation methods for $\lambda$. 
The first two mechanisms involve using a binary value for $\lambda$, allowing the agent to freely switch between exploration and exploitation. 

\begin{itemize}[leftmargin=16pt]
\vspace{-5pt}
  \item $\min(\lambda)$. The insight here is to choose the smaller of the target update values induced by the Bellman Exploration operator and Bellman Exploitation operator, which might aid in alleviating the overestimation issue and enhance learning stability. The possible drawback is that it might prefer to exclusively choose the conservative $Q$-value. We formulate this mechanism as,
  \begin{equation*}
    \lambda=\mathbbm{1}\left(\mathcal{T}_{exploit}Q(s,a)-\mathcal{T}_{explore}Q(s,a)\leq 0\right).
  \end{equation*}
where $\mathbbm{1}(x\leq0)$ is an indicator function 
\begin{equation*}
\mathbbm{1}(x\leq0)=
\begin{cases}
0 & x>0, \\
1 & x\leq0. 
\end{cases}
\end{equation*}
  
  \item  $\max(\lambda)$. This mechanism, conversely, selects the larger of the two values. This method might yield unstable results due to the influence of function approximation error. We formulate this mechanism as
  \begin{equation*}
      \lambda=\mathbbm{1}\left(\mathcal{T}_{exploit}Q(s,a)-\mathcal{T}_{explore}Q(s,a)\geq 0\right).
  \end{equation*}
\end{itemize}

We also design a third mechanism for suggesting a continuous value of $\lambda$.
\begin{itemize}[leftmargin=16pt]
    \vspace{-5pt}
    \item ${\rm ada}(\lambda)$.  
    Upon integrating new data into the replay buffer, the Bellman error variation would be small if the data is well exploited, and larger if not. Hence, when the Bellman error on the new-coming data is small, we may curtail reliance on executing $\mathcal{T}_{exploit}$ in the replay buffer and allocate more weight towards exploration. Motivated by this insight, we could adjust the value of $\lambda$ according to the Bellman error. In practice, we divide the current Bellman error $\delta_{k}$ by the prior Bellman error $\delta_{k-1}$ to focus more on the Bellman error caused by the introduction of new-coming data. This way, $\lambda$ can be automatically adapted during training as follows:
    \begin{equation*}
        \lambda={\rm clip}\left(\frac{\delta_{k}}{\delta_{k-1}},0,1\right).
    \end{equation*}
    Here, ${\rm clip}(\cdot,0,1)$ clips the $\lambda$ by removing the value outside of the interval $[0, 1]$.
\end{itemize}

\begin{figure}[b]
    \centering
    \begin{minipage}[t]{0.65\textwidth}
        \includegraphics[width=0.47\linewidth]{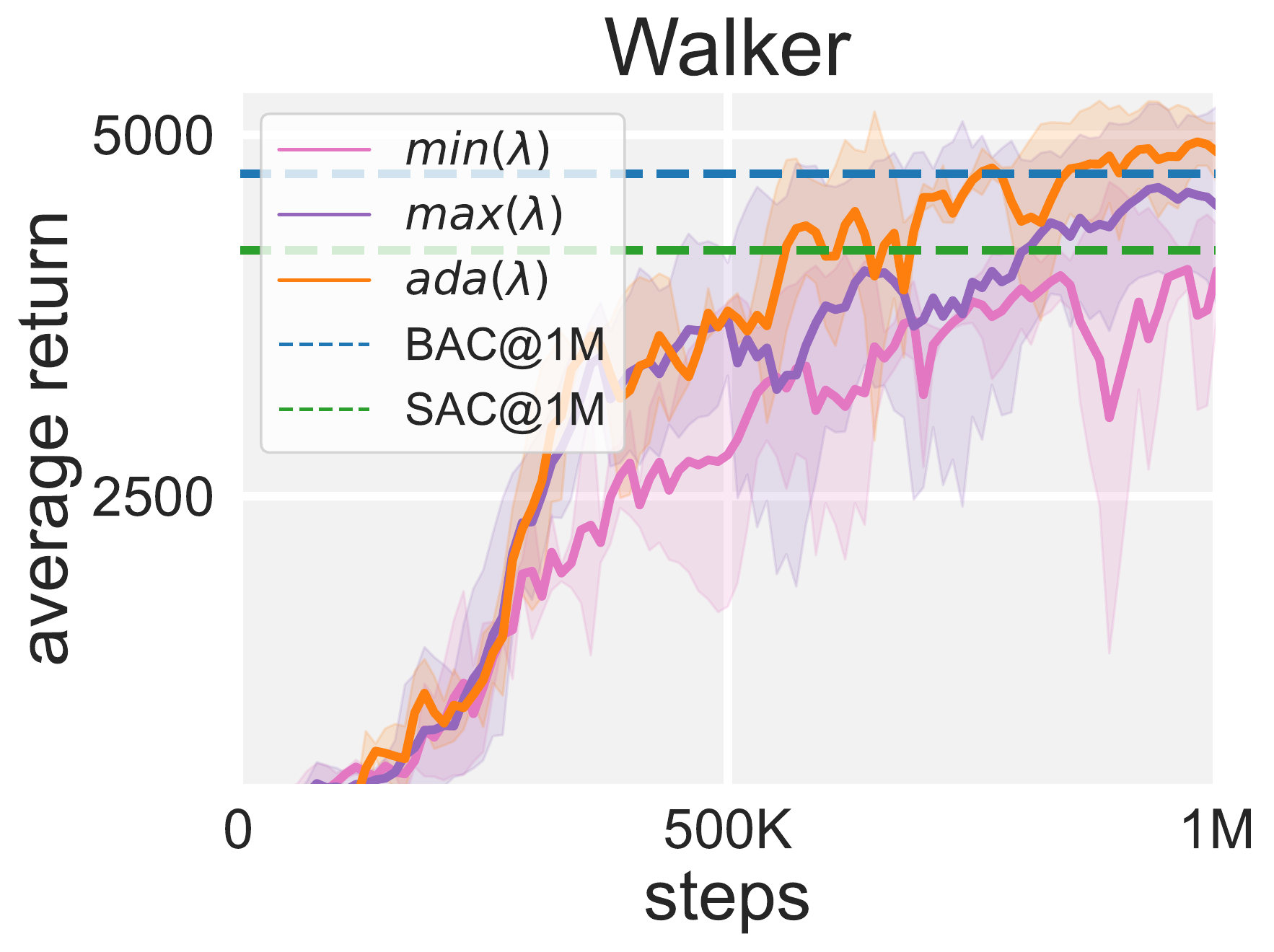}
        \hfill
        \includegraphics[width=0.47\linewidth]{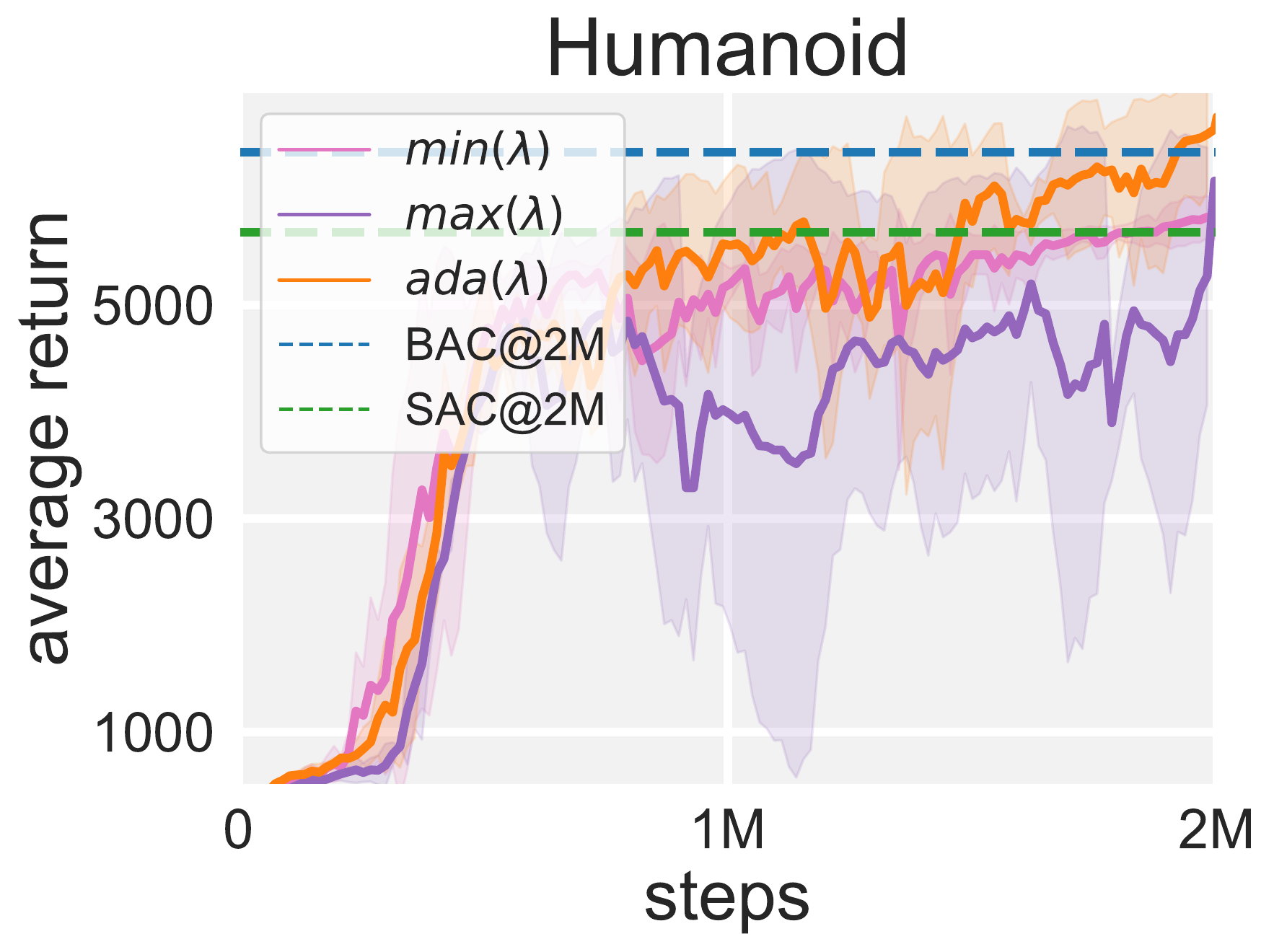}
        \caption{Learning curves with different lambda mechanisms in Walker2d and Humanoid tasks, where the dotted line indicates the eventual performance of BAC and SAC.}
        \label{fig:differentlambda}
    \end{minipage}
    \hfill
    \begin{minipage}[t]{0.32\textwidth}
        \includegraphics[width=0.94\linewidth]{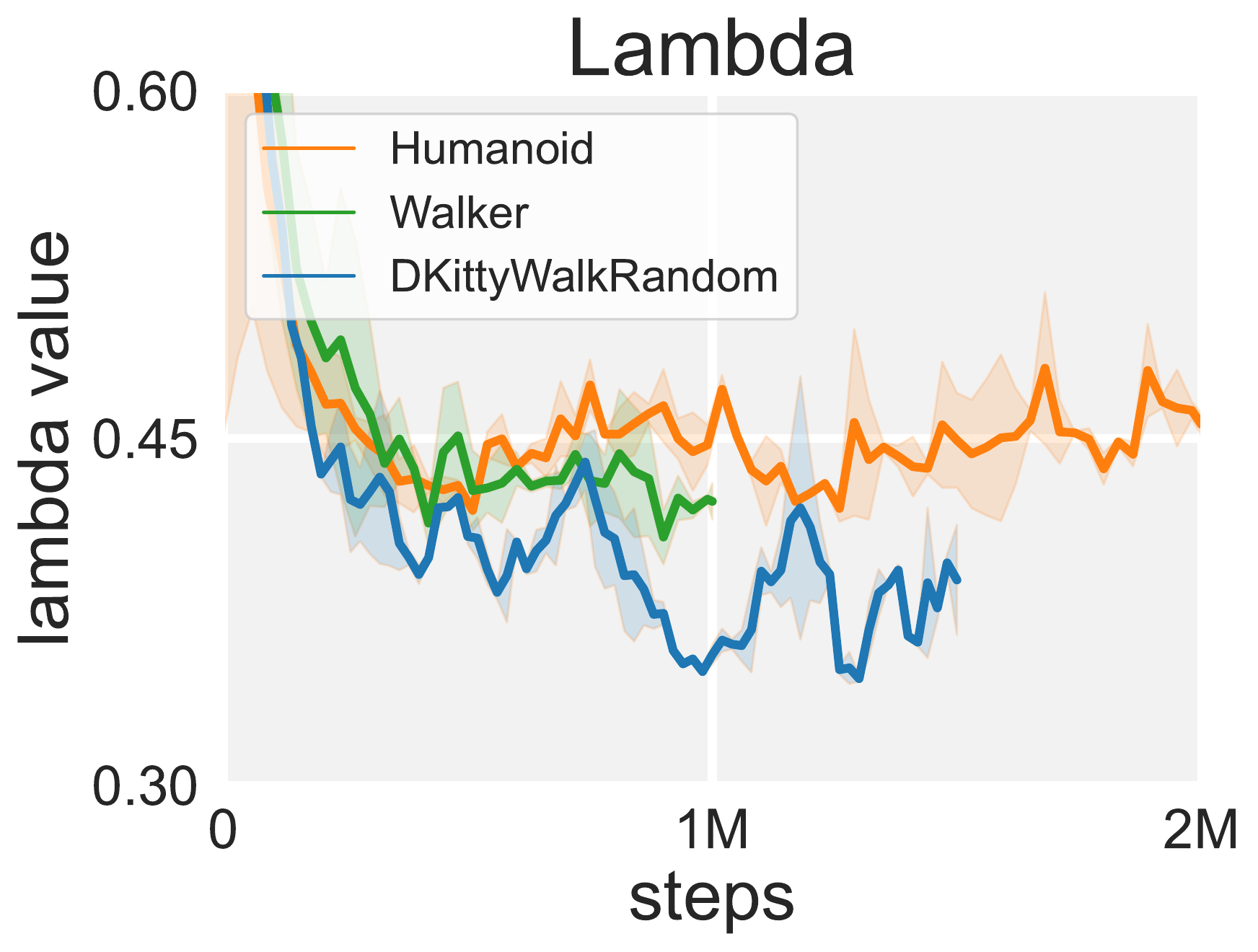}
        \caption{Curve of $\lambda$ with ${\rm ada}(\lambda)$ mechanism in different environments.}
        \label{fig:autolambda}
    \end{minipage}
\end{figure}
\textcolor{mydarkblue}{\textit{Remark 1:}} In Figure~\ref{fig:differentlambda}, we depict the learning curves of these three automatic $\lambda$ adjustment mechanisms on Walker2d and Humanoid tasks, along with the eventual performance of SAC and the primary BAC. In these two settings, the ada($\lambda$) mechanism generally yields competitive eventual performance, while min($\lambda$) and max($\lambda$) are more influenced by the environment settings. For instance, in the Humanoid task, we observed that the $\min(\lambda)$ mechanism almost entirely selects 0 after 1M iterations, thus could be considered as reducing to SAC in the later stages, and its final performance matches that of SAC; however, in the Walker2d environment, $\min(\lambda)$ results in a $\lambda$ that switches more frequently.

\textcolor{mydarkblue}{\textit{Remark 2:}} Additionally, the third mechanism ${\rm ada}(\lambda)$ often yields promising results. Although it might introduce some oscillation, its advantage lies in providing guidelines for choosing $\lambda$, such as setting it to a fixed constant. As shown in Figure~\ref{fig:autolambda}, the final fixed values of $\lambda$ chosen for these three environments fall within the range of 0.4 to 0.5.

\clearpage
\subsection{Hyperparameter settings}\label{section:appendix-hyper}
\paragraph{A simple setting of hyperparameters $\lambda$  and $\tau$.}

The primary implementation of \ourshort\ introduces two extra hyperparameters, the balance factor $\lambda$ and the quantile $\tau$. 
\textcolor{myblue}{A single set of hyperparameters ($\lambda=0.5, \tau=0.7$) is sufficient for our algorithm to achieve strong performance throughout all experiments across MuJoCo, DMControl, Meta-World, ROBEL, Panda-gym, Adroit, maniskill2, and Shadow Dexterous Hand benchmark suites.} 

Notably, the BEE operator is not bound to hyperparameter $\tau$. As detailed in Appendix~\ref{section:designchoice}, implementing BEE with other in-sample learning offline techniques, such as SQL and EQL instead of IQL would not have $\tau$ at all.

\paragraph{Heuristics for selecting $\lambda$ and $\tau$.}
\begin{itemize}[leftmargin=16pt]
    \item $\lambda$: We initiated from $\lambda=0.5$ as a balanced weight for $\mathcal{T}_{exploit}$ and $\mathcal{T}_{explore}$. Figure~\ref{fig:differentlambda} depicts that moderate values around 0.5 obtain good performance. Besides, the automatic adaptive mechanisms we provided in Appendix~\ref{section:auto-lambda-mechanism} may suffice and circumvent tuning.
    
    \item $\tau$: Our choice to primarily use 0.7 comes from the IQL paper~\citep{IQL} which uses 0.7 for MuJoCo tasks. And $\tau=0.7$ already suffices for expected performance, thus we mostly use 0.7 in DMControl and Meta-World tasks.  
\end{itemize}

\paragraph{Hyperparameters Tables.}
Hyperparameters for \ourshort\ are outlined in Table~\ref{bac-hyperparameters} and Table~\ref{mbbac-hyperparameters}, respectively. 

In \mbshort, we follow the hyperparameters specified in MBPO~\citep{mbpo}. The symbol ``$x\rightarrow y$ over epochs $a\rightarrow b$'' denotes a linear function for establishing the rollout length. That is, at epoch $t$, $f(t) = \min(\max(x + \frac{t-a}{b-a}\cdot (y-x), x),y)$. And we set $\lambda=0.5$ and $\tau=0.7$ for each task.

\begin{figure}[H]
    \begin{minipage}{0.35\textwidth}
        \centering
        \includegraphics[height=3.0cm,keepaspectratio]{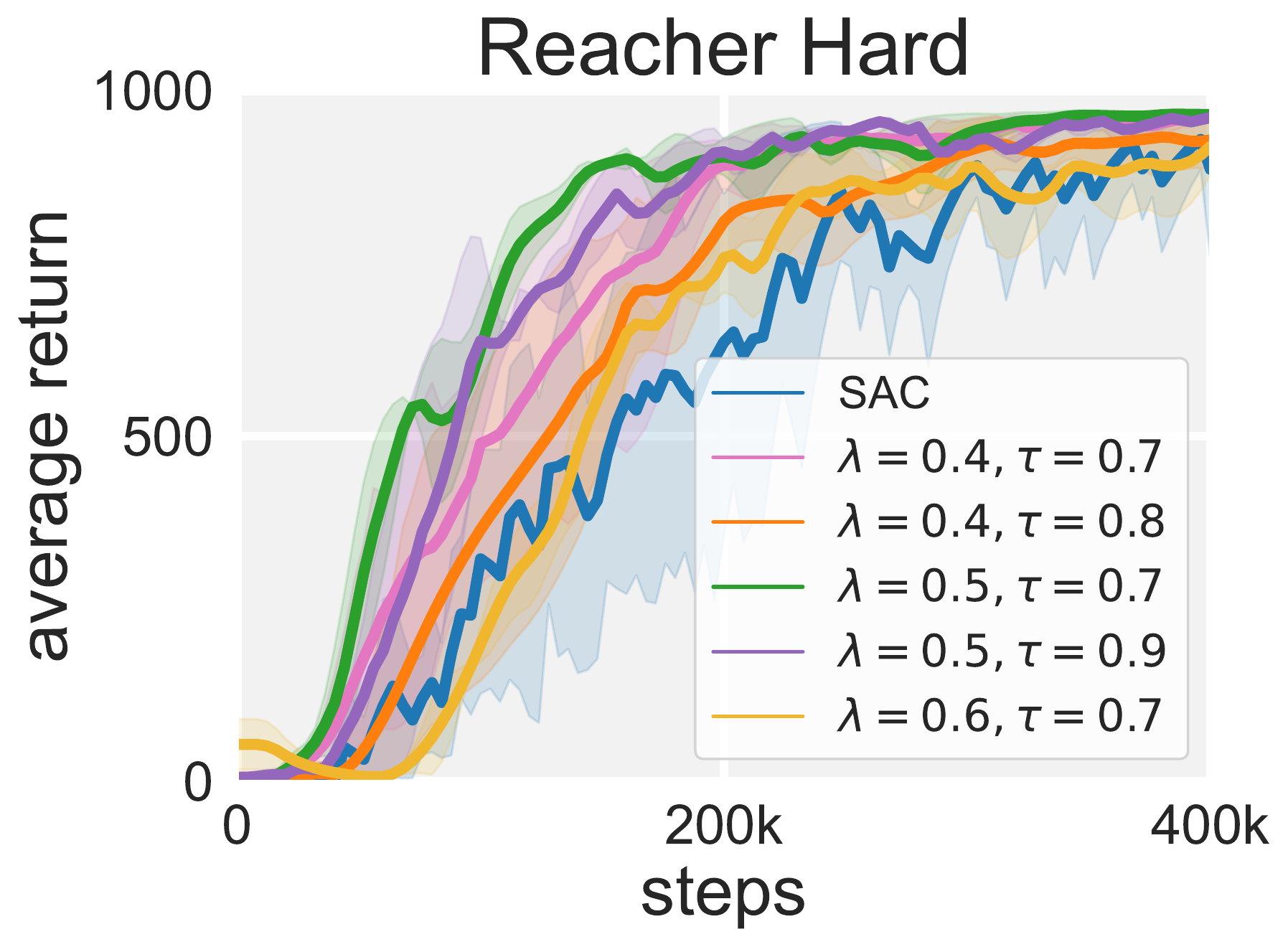}
        \caption{Learnings curves of BAC with a wider set of hyperparameters on ReacherHard.}
        \label{fig:wider-HPstudy}
    \end{minipage}
    \begin{minipage}{0.62\linewidth}
        \centering
        \includegraphics[height=3.0cm,keepaspectratio]{./Figures/lambda-Walker-v2.pdf}
        \hspace{5mm}
        \includegraphics[height=3.0cm,keepaspectratio]{./Figures/lambda-Humanoid-v2.pdf}
        \caption{\small Parameter study on $\lambda$. The experiments are run over 4 random seeds.}
        \label{fig:lambda}
    \end{minipage}%
    \vspace{-4mm}
\end{figure}

\paragraph{Hyperparameter study.} 
For practical use, $\lambda=0.5$, $\tau=0.7$ may suffice.  If slightly tuned, we find that BAC can achieve even better performance.
We provide the performance comparison of a wider set of hyperparameters in Figure~\ref{fig:wider-HPstudy}. The results reveal that utilizing a higher $\tau = 0.9$ is not problematic and, in fact, enhances performance in ReacherHard in comparison to $\tau = 0.7$. Each BAC instance with varied hyperparameters surpasses the SAC in ReacherHard. 

\textcolor{mydarkblue}{\textit{Remark 3:}}
A high (e.g., 0.9) $\tau$ may not be problematic in the online setting. This differs from offline RL.  In the offline setting, the peak distributions will occur in various quantiles for different datasets, thus an unsuitable $\tau$ may cause erroneous estimation. However, in online settings, ongoing interactions could enrich peak data. As policy improves, the replay buffer accumulates high-value data, thus reducing sensitivity to $\tau$.

Setting an appropriate weighted coefficient $\lambda$, \ourshort\ could balance the exploitation and exploration well. We may note that the algorithm is reduced to the online version of IQL~\citep{IQL} for an extreme value $\lambda=0$. According to Figure~\ref{fig:lambda}, and the detailed settings and hyperparameter studies in Appendix~\ref{section:appendix-hyper}, we find that a moderate choice of $\lambda$ around 0.5 is sufficient to achieve the desired performance across all locomotion and manipulation tasks we have benchmarked. This underscores that BAC does not need heavy tuning for strong performance.

\begin{table}[h]
    \caption{Hyperparameter settings for \ourshort\ in benchmark tasks.}
    \begin{center}
    \resizebox{0.65\textwidth}{!}{
        \begin{tabular}{
            >{\centering}m{0.3\textwidth}
            | c
            | c
            | c
            | c
            | c
            | c
        }
            \toprule
            \rowcolor{mylightblue}\textbf{Hyper-parameter} & \multicolumn{6}{c}{
                \textbf{Value}
            }\\
            \midrule
            $Q$-value network & \multicolumn{6}{c}{
                MLP with hidden size 512
            } \\
            \midrule
            $V$-value network & \multicolumn{6}{c}{
                MLP with hidden size 512
            } \\
            \midrule
            policy network & \multicolumn{6}{c}{
                Gaussian MLP with hidden size 512 
            } \\
            \midrule
            discounted factor  & \multicolumn{6}{c}{
                0.99
            }\\
            \midrule
            soft update factor & \multicolumn{6}{c}{
                0.005
            }\\
            \midrule
            learning rate & \multicolumn{6}{c}{
                0.0003
            }\\
            \midrule
            batch size & \multicolumn{6}{c}{
                512
            }\\
            \midrule
            policy updates per step & \multicolumn{6}{c}{
                1
            }\\
            \midrule
            value updates per step & \multicolumn{6}{c}{
                1
            }\\
            \midrule
            $\lambda$  
            & \multicolumn{6}{c}{
                0.5
            }
            \\
            \midrule
            $\tau$  
            &\multicolumn{6}{c}{
                0.7
            }
            \\   
            \bottomrule
        \end{tabular}
        }
    \end{center}
    \label{bac-hyperparameters}
\end{table}
\begin{table}[h]
    \caption{Hyperparameter settings \mbshort\ in benchmark tasks.}
    \begin{center}
    \resizebox{0.8\textwidth}{!}{
        \begin{tabular}{
            >{\centering}m{0.25\textwidth}
            | c
            | c
            | c
            | c
        }
            \toprule
            \rowcolor{mylightblue}\textbf{Hyper-parameter} 
            & \textbf{Hopper} & \textbf{Walker2d} & \textbf{Ant} &  \textbf{Humanoid}  \\
            \midrule
            dynamical models network & \multicolumn{4}{c}{
                Gaussian MLP with 4 hidden layers of size 200
            }\\
            \midrule
            ensemble size & \multicolumn{4}{c}{
                5
            }\\
            \midrule
            model rollouts per policy update & \multicolumn{4}{c}{
                400
            } \\
            \midrule
            rollout schedule 
            & \makecell[c]{{1 $\rightarrow$ 15} \\over epochs\\ {20 $\rightarrow$ 100}}
            &\makecell[c]{1}
            & \makecell[c]{{1 $\rightarrow$ 25} \\over epochs\\ {20 $\rightarrow$ 100}}
            & \makecell[c]{{1 $\rightarrow$ 25} \\over epochs\\ {20 $\rightarrow$ 300}}
            \\
            \midrule
            policy network
            &\multicolumn{2}{c|}{
                Gaussian  with hidden size 512
            }
            &\multicolumn{2}{c}{
                Gaussian  with hidden size 1024
            }
            \\
            \midrule
            policy updates per step
            & 40
            & 20
            & 20
            & 20
            \\            
            \bottomrule
        \end{tabular}
        }
    \end{center}
    \label{mbbac-hyperparameters}
\end{table}

\subsection{Computing infrastructure and computational time}
Table~\ref{computation} presents the computing infrastructure used for training our algorithm \ourshort\ and \mbshort\ on benchmark tasks, along with the corresponding computational time.  Compared to SAC, the total training time of \ourshort\ only increased by around 8\% for Humanoid (1.19 H for 5M steps).  Thus. we believe the additional costs are acceptable.  Further, for practical use, \ourshort\ requires fewer interactions for similar performance, which may lower the needed computation time.
\begin{table}[h]
    \caption{Computing infrastructure and the computational time for MuJoCo benchmark tasks.}
    
    \begin{center}
    \resizebox{0.8\textwidth}{!}{
        \begin{tabular}{
            >{\centering}m{0.15\textwidth}
            | c
            | c
            | c
            | c
            | c
            | c
        }
            \toprule
            \rowcolor{mylightblue}
            &  \textbf{Hopper} & \textbf{Walker}  
             &  \textbf{Swimmer} & \textbf{Ant}  & \textbf{HumanoidStandup} &  \textbf{Humanoid} \\
            
            \midrule
            CPU & \multicolumn{6}{c}{
                AMD EPYC 7763 64-Core Processor (256 threads)
            } \\
            \midrule
            GPU & \multicolumn{6}{c}{
                NVIDIA GeForce RTX 3090 $\times$ 4
            } \\
            \midrule
            
            \textbf{BAC} computation time in hours
            & -
            & -
            & 6.56 
            & 13.47 
            & 11.65 
            & 16.57 
            \\
            \midrule
            \textbf{SAC} computation time in hours 
            & -
            & -
            & 6.08 
            & 12.88 
            & 11.18 
            & 15.38 
            \\
            \midrule
            \textbf{MB-BAC} computation time in hours
            & 18.35
            & 19.51
            & -
            & 27.57
            & -
            & 30.86  \\ 
            \bottomrule
        \end{tabular}
    \label{computation}
        }
    \end{center}
\end{table}

\clearpage
\section{Environment Setup}

We evaluate the BEE operator across \textcolor{myblue}{over 50 diverse continuous control tasks}, spanning \textbf{MuJoCo, ROBEL, DMControl, Meta-World, Adroit, ManiSkill2, Panda-gym, Shadow Dexterous Hand, MyoSuite} benchmark suites. We find our algorithm \ourshort\ excels in both locomotion and manipulation tasks. 

Besides, we conduct experiments on \textcolor{myblue}{4 noisy environments} and \textcolor{myblue}{6 sparse reward tasks} to further showcase the effectiveness of the BEE operator.

As a versatile plugin, it seamlessly enhances performance with various policy optimization methods, shining in model-based and model-free paradigms.

We also validate BAC using a \textbf{cost-effective D’Kitty robot} to navigate various complex terrains and finally reach goal points and desired postures. The \textcolor{mypink}{4 real-world quadruped locomotion tasks} highlight BAC’s effectiveness in real-world scenarios.

\begin{figure}[H]
\centering
 \includegraphics[width=0.95\linewidth]{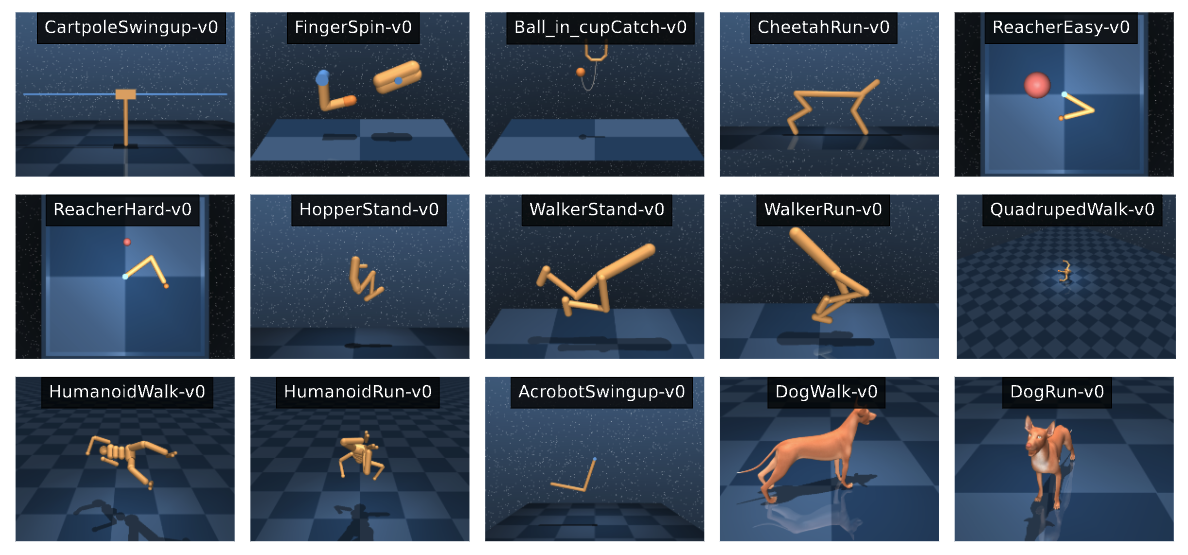}
\caption{Visualization of \textbf{\textcolor{myblue}{simulated tasks}} from DMControl.}
\label{fig:scenarios-dmcontrol}
\end{figure}

\begin{figure}[H]
\centering
 \includegraphics[width=0.95\linewidth]{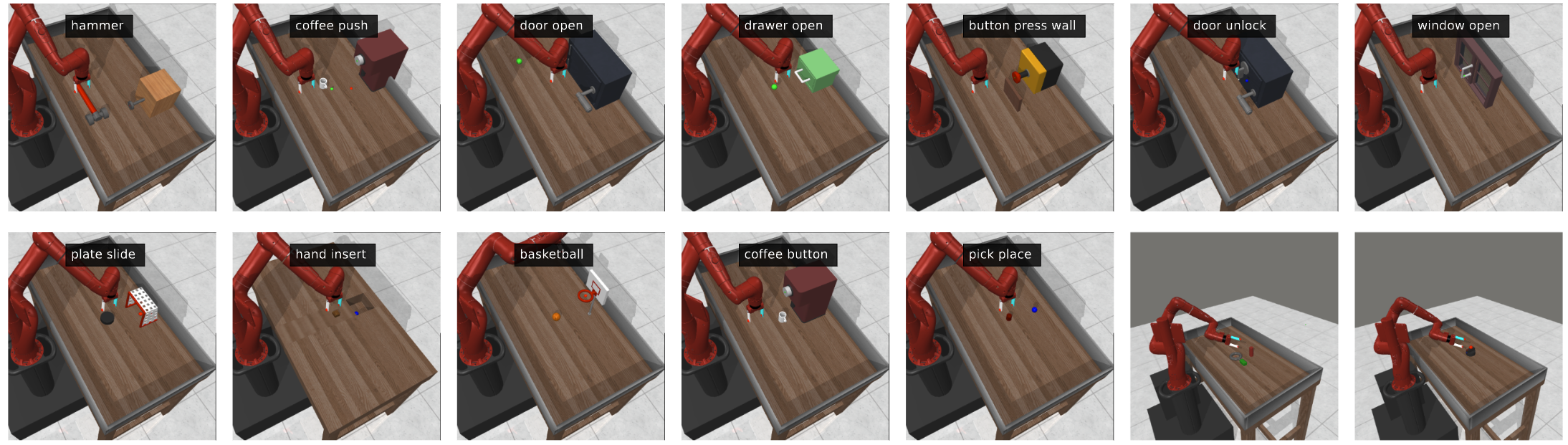}
 \caption{Visualization of \textbf{\textcolor{myblue}{simulated tasks}} from Meta-World.}
 \label{fig:scenarios-meta-world}
\end{figure}

\begin{figure}[H]
    \begin{minipage}{0.74\textwidth}
    \centering
            \includegraphics[height=3.5cm,keepaspectratio]{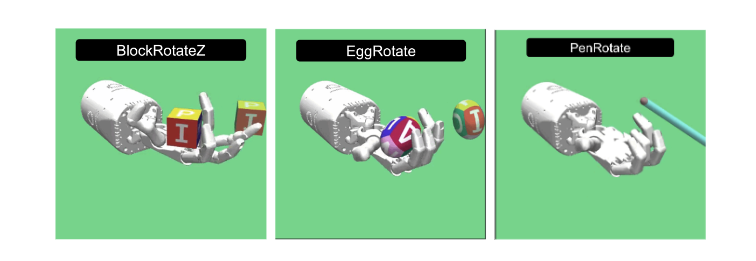}
            \caption{Visualization of \textbf{\textcolor{myblue}{simulated tasks}} from Shadow Dexterous Hand.}
            \label{fig:scenarios-shadow-dexterous-hand}
    \end{minipage}
    \begin{minipage}{0.25\textwidth}
    \centering
     \includegraphics[height=2.8cm,keepaspectratio]{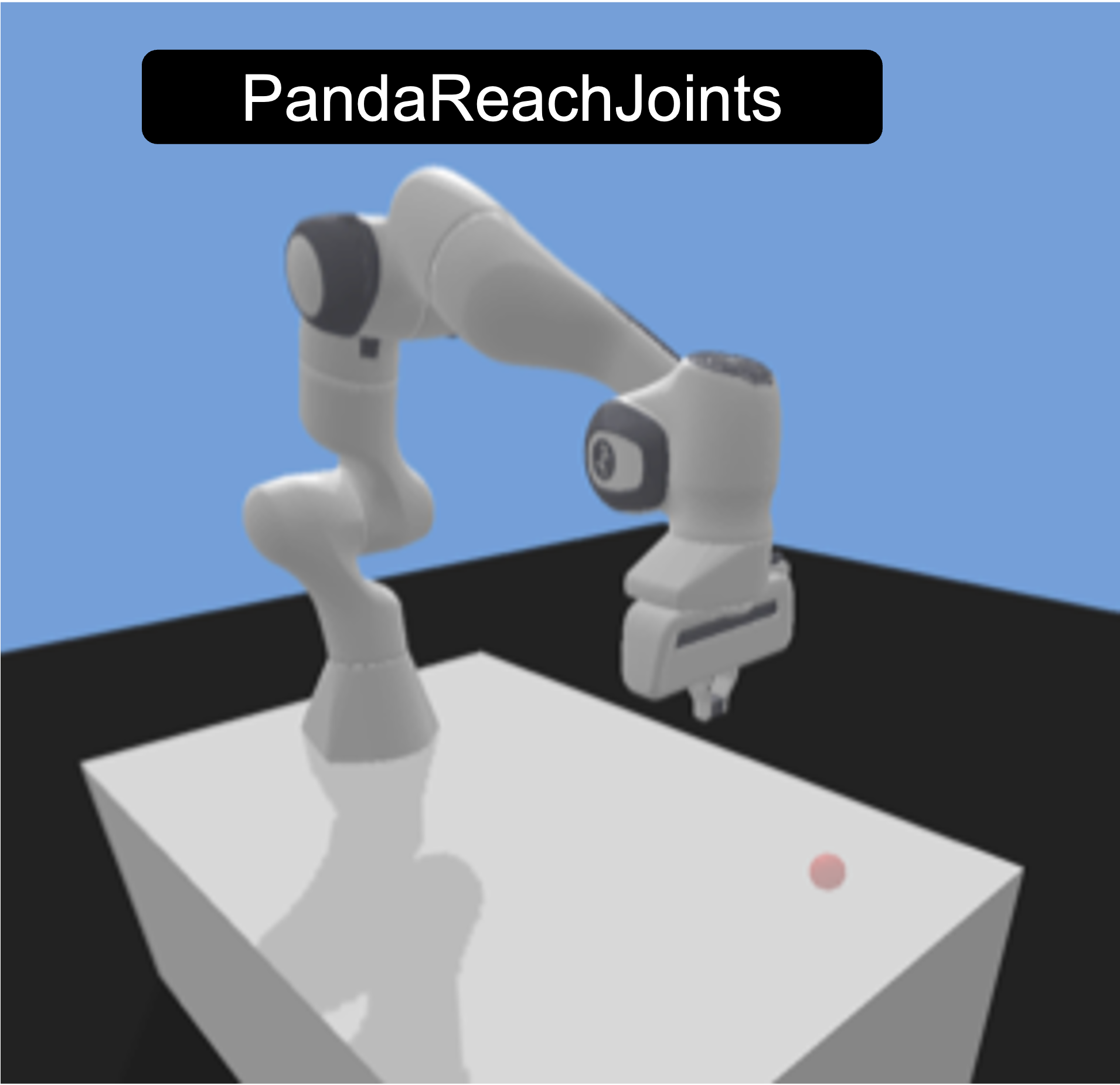}
            \caption{Visualization of \textbf{\textcolor{myblue}{simulated task}} from Panda-gym.}
            \label{fig:scenarios-panda-gym}
    \end{minipage}
\end{figure}

\begin{figure}[H]
\centering
 \includegraphics[width=0.8\linewidth]{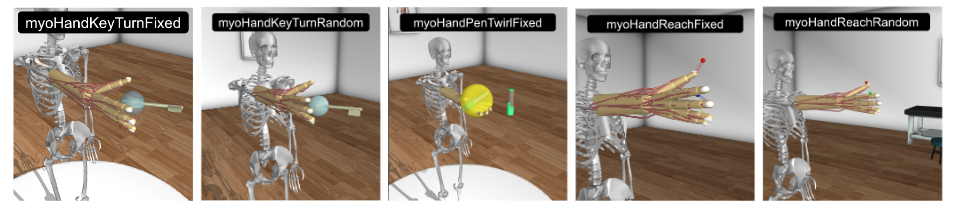}
\caption{Visualization of \textbf{\textcolor{myblue}{simulated tasks}} from MyoSuite.}
        \label{fig:scenarios-MyoSuite}
\end{figure}

$\bigstar$ Visualizations of these tasks are provided in Figure~\ref{fig:scenarios-mujoco}, \ref{fig:scenarios-dkitty}, \ref{fig:scenarios-dmcontrol}, \ref{fig:scenarios-meta-world}, \ref{fig:scenarios-adroit}, \ref{fig:scenarios-maniskill2}, \ref{fig:scenarios-shadow-dexterous-hand}, \ref{fig:scenarios-panda-gym} and \ref{fig:scenarios-MyoSuite}.

\subsection{Environment setup for evaluating \ourshort}
\paragraph{MuJoCo benchmark tasks.} 
We benchmark \ourshort\ on four continuous control tasks in OpenAI Gym~\citep{gym} with the  MuJoCo~\citep{Mujoco} physics simulator, including  Swimmer, Ant, Humanoid, HumanoidStandup, using their standard versions.

\begin{center}
    \begin{minipage}{0.9\textwidth}
    \begin{figure}[H]
        \centering
        \newcommand{\img}[1]{\hspace{0.05cm}\includegraphics[width=0.17\linewidth]{{#1}}\hspace{0.05cm}}
        \newcommand{\imgsmall}[1]{\hspace{0.04cm}\includegraphics[width=0.17\linewidth]{{#1}}\hspace{0.04cm}}
        \begin{tabular}{@{}c@{}c@{}c@{}c}
            \img{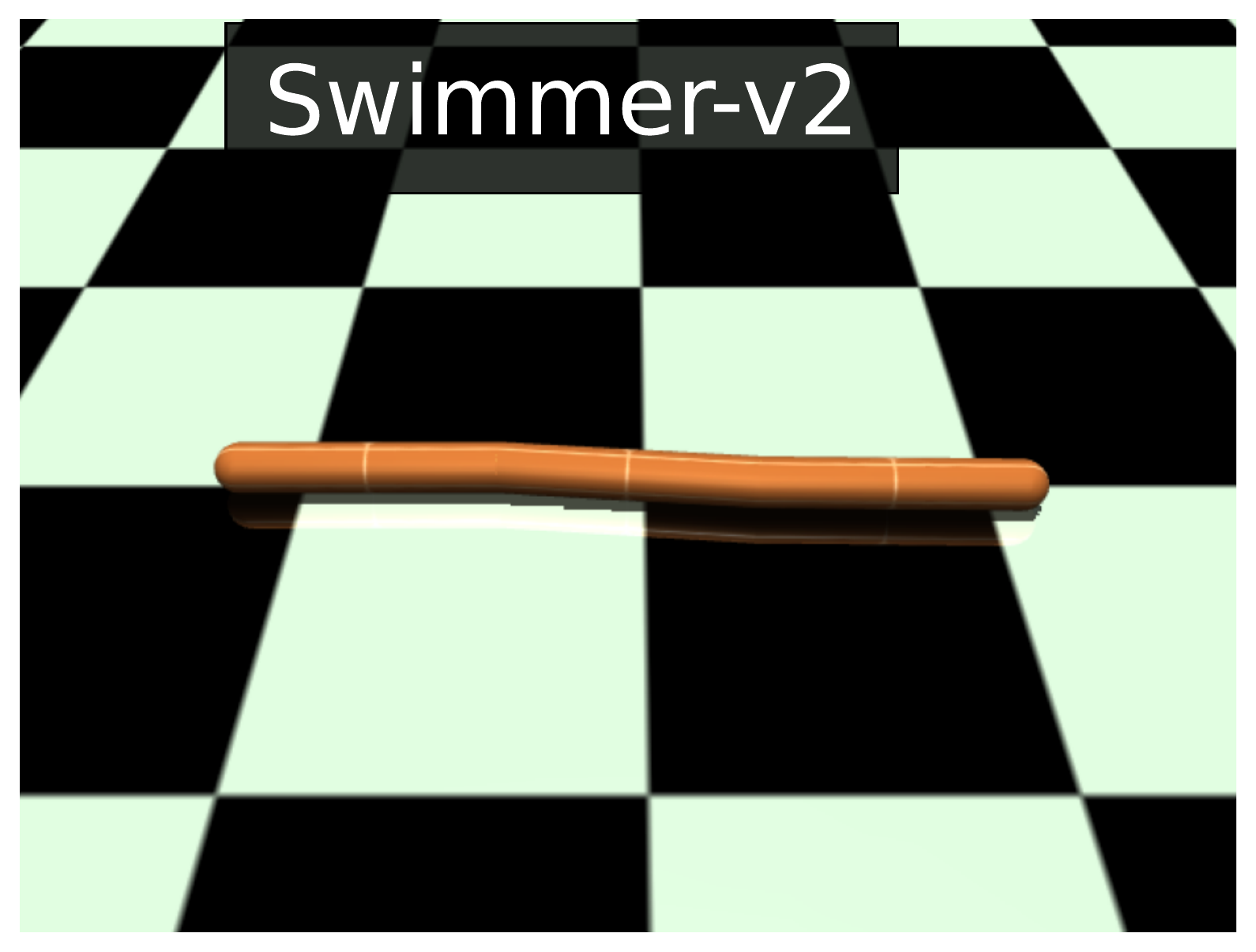} &
            \img{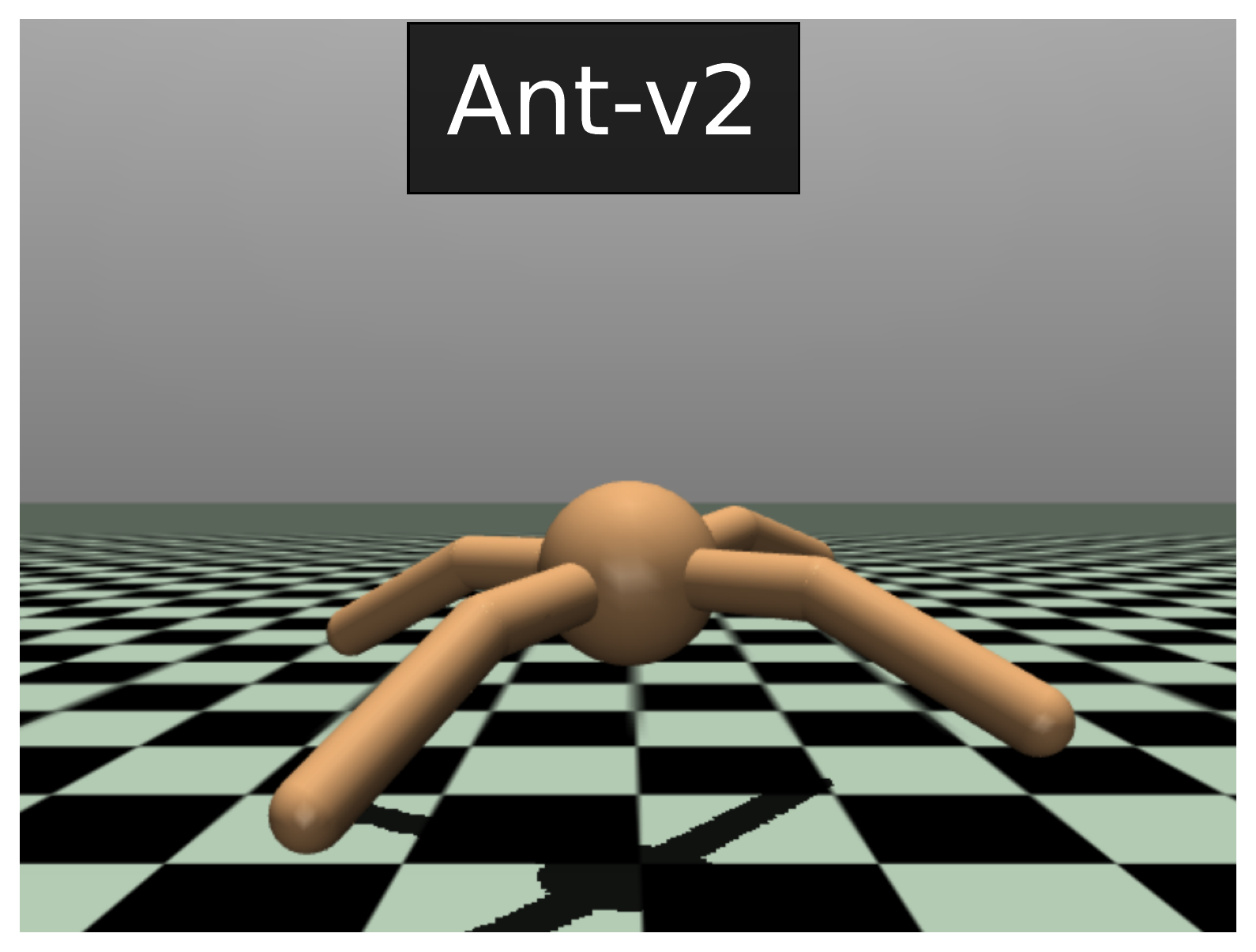} &
            \img{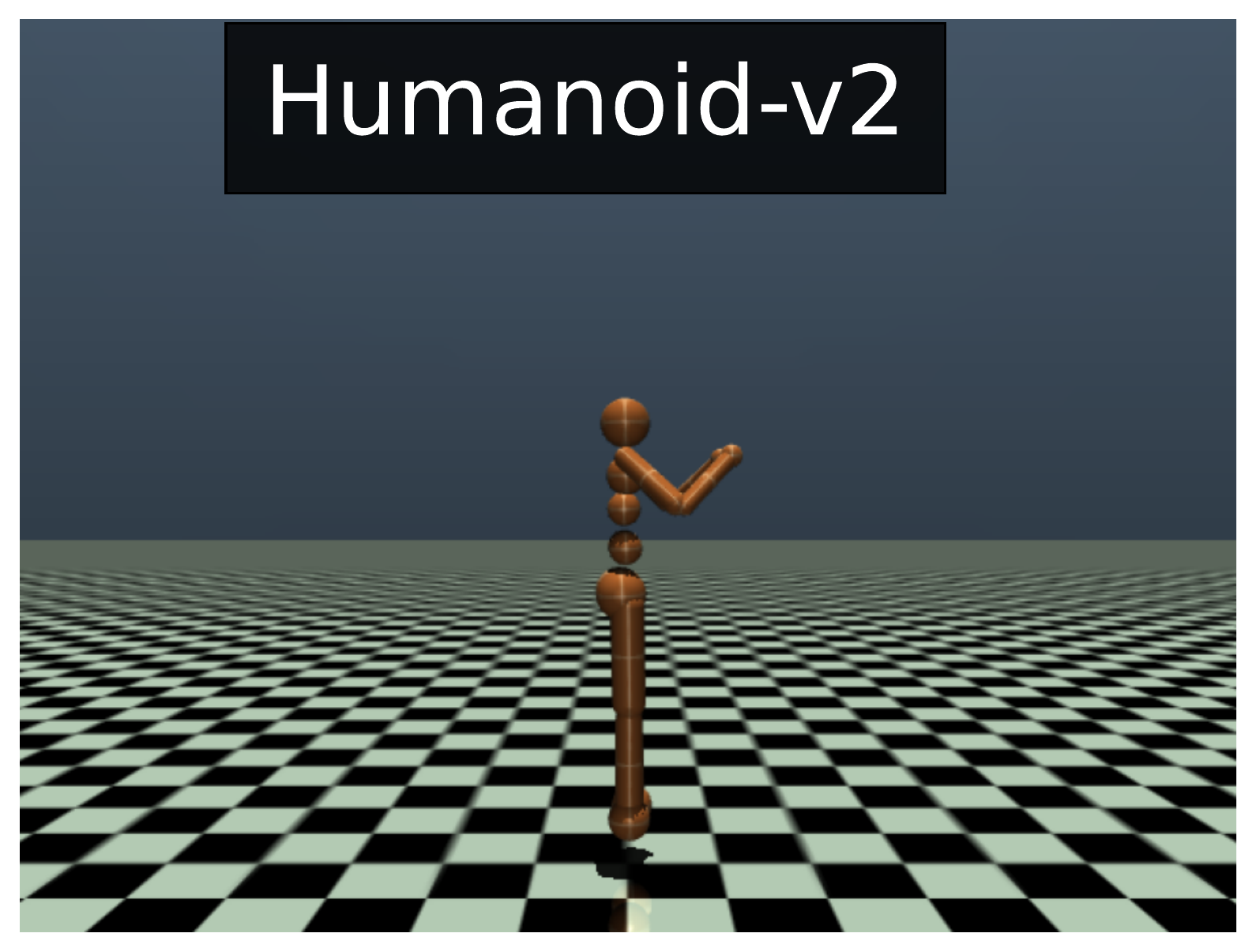} &
            \img{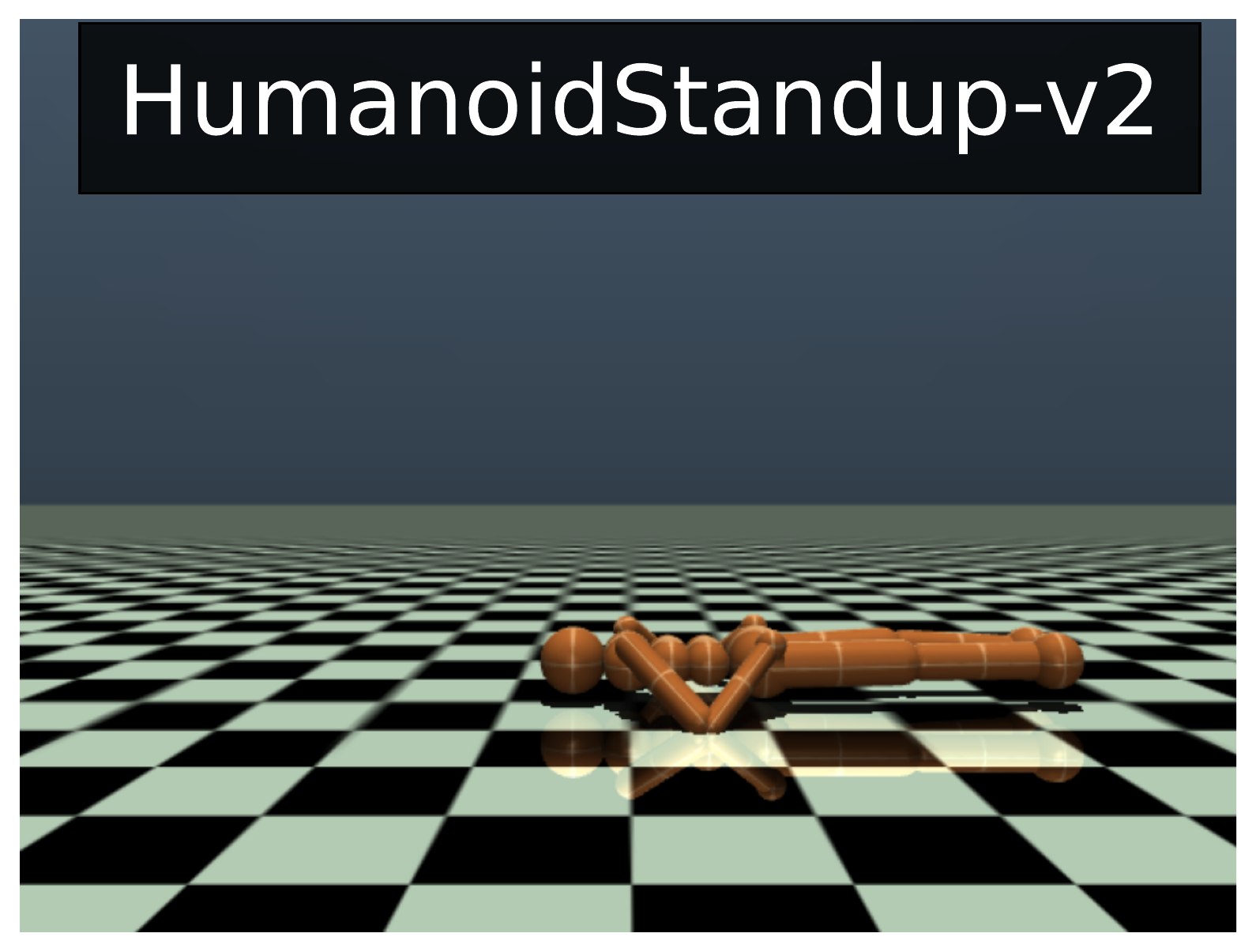} \\
        \end{tabular}
        \caption{Visualization of \textbf{\textcolor{myblue}{simulated tasks}} from MuJoCo.}
        \label{fig:scenarios-mujoco}
    \end{figure}
    \end{minipage}
\end{center}

\paragraph{D’Kitty simulated tasks.} 
ROBEL~\citep{dkitty} is an open-source platform of cost-effective robots designed for real-world reinforcement learning. The D’Kitty robot, with four legs, specializes in agile-legged locomotion tasks. 
ROBEL platform provides a set of continuous control benchmark tasks for the D’Kitty robot. Details for the base task~(DKittyWalkRandomDynamics): 
\begin{itemize}[leftmargin=16pt]
    \item Task: D’Kitty robot moves from an initial position $p_{t, kitty}$ to a desired one $p_{goal}$ while maintaining balance and facing direction.
    \item Setting: Randomized joint parameters, damping, friction, and terrain with heights up to $0.05{\rm m}$.
    \item Reward function: The reward function contains five parts, the upright standing reward $r_{t, upright}$, the distance penalty $d_{t, goal}=\Vert p_{goal}-p_{t, kitty}\Vert_2$, the heading alignment $h_{t, goal}=R_{y,t, kitty}(p_{goal}-p_{t, kitty})/d_{t, goal}$, a small task success bonus $r_{bonus\_small}$ and a big task success bonus $r_{bonus\_big}$. Thus, the reward function is defined as
    \begin{equation*}   r_t=r_{t,upright}-4d_{t,goal}+2h_{t,goal}+r_{bonus\_small}+r_{bonus\_big}.
    \end{equation*}
    \item Success indicator: The success is defined by meeting distance and orientation criteria. The formulation of the success indicator is: 
    \begin{equation*}
    \phi_{se}(\pi)=\mathbb{E}_{\tau\sim\pi}\left[\mathbbm{1}\left(d^{(\tau)}_{T,goal}<0.5\right)*\mathbbm{1}\left(u^{(\tau)}_{T,kitty}>\cos(25^\circ)\right)\right].
    \end{equation*}
    The ``Success Rate'' in our experiments refers to the success percentage over 10 runs.
\end{itemize}

To construct more challenging locomotion tasks, we modify the base task DKittyWalkRandomDynamics by increasing terrain unevenness: 
\begin{itemize}[leftmargin=16pt]
\item \textit{\textcolor{myblue}{DKittyWalk-Hard}}: the randomized height ﬁeld is generated with heights up to 0.07m.
\item \textit{\textcolor{myblue}{DKittyWalk-Medium}}: the randomized height ﬁeld is generated with heights up to 0.09m.
\end{itemize}

\paragraph{D’Kitty real-world tasks.}
Our real-world validation experiments are performed using a cost-effective D’Kitty robot. D’Kitty~\cite {dkitty} is a twelve-DOF quadruped robot capable of agile locomotion. It consists of four identical legs mounted on a square base. Its feet are 3D-printed parts with rubber ends. 

The D’Kitty robot is required to traverse various complex terrains, contending with unpredictable environmental factors, and finally reach a target point. 
We evaluate \ourshort\ and baseline methods on four different terrains: \textit{\textcolor{mypink}{smooth road}} (with a target point at 3m), \textit{\textcolor{mypink}{rough stone road}} (target point at 1m), \textit{\textcolor{mypink}{uphill stone road}} (target point at 1m), and \textit{\textcolor{mypink}{grassland}} (target point at 1m).

\begin{center}
    \begin{minipage}{0.9\textwidth}
    \begin{figure}[H]
        \centering
        \newcommand{\img}[1]{\hspace{0.05cm}\includegraphics[width=0.15\linewidth]{{#1}}\hspace{0.05cm}}
        \newcommand{\imgsmall}[1]{\hspace{0.04cm}\includegraphics[width=0.15\linewidth]{{#1}}\hspace{0.04cm}}
        \begin{tabular}{@{}c@{}c@{}c@{}c@{}c@{}c@{}}
            \img{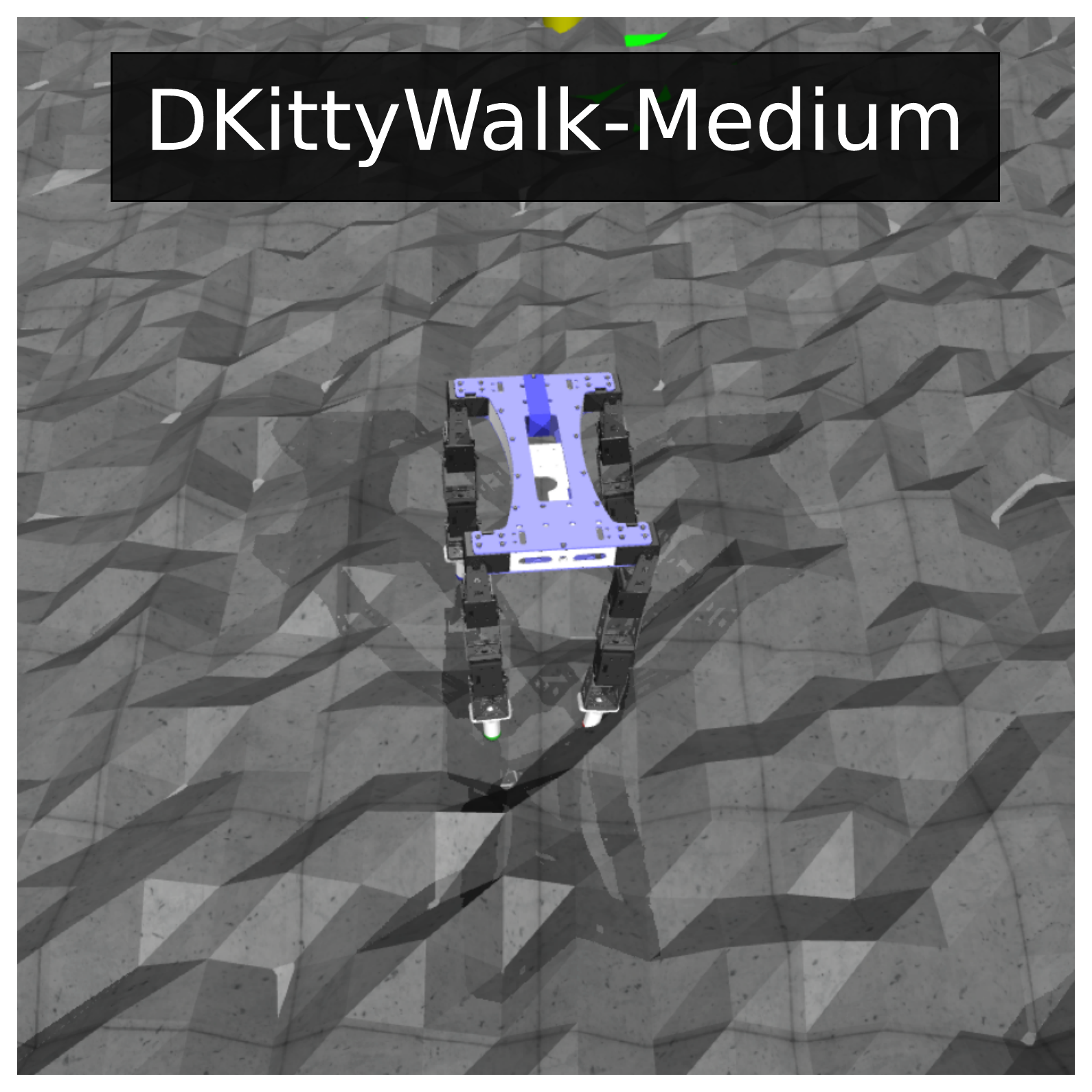} &
            \img{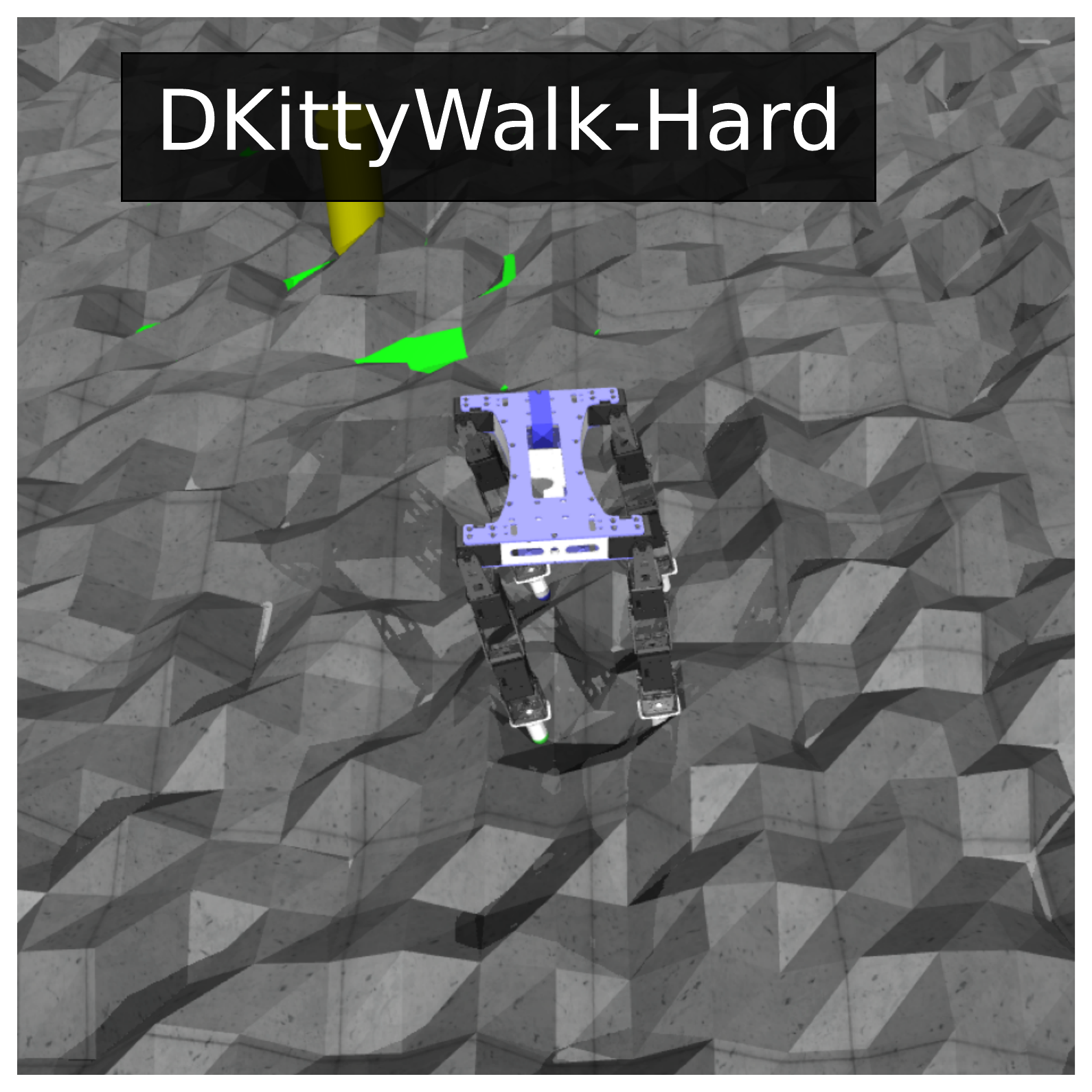} &
            \img{Figures/1-smooth-road.png} &
            \img{Figures/2-rough-stone-road.png} &
            \img{Figures/3-uphill-stone-path.png} &
            \img{Figures/4-grassland.png} \\
            \multicolumn{2}{c}{(a) {D’Kitty simulated tasks}} &
            \multicolumn{4}{c}{(b) {D’Kitty real-world tasks}} \\
        \end{tabular}
        \caption{Visualization of \textbf{\textcolor{myblue}{simulated tasks}} and \textbf{\textcolor{mypink}{real-world robot tasks}} from D’Kitty.}
        \label{fig:scenarios-dkitty}
    \end{figure}
    \end{minipage}
\end{center}

\paragraph{DMControl tasks.}
The DeepMind Control Suite (DMControl)~\citep{dmcontrol}, provides a set of continuous control tasks with standardized structures and interpretable rewards. 
We evaluate \ourshort\, BEE-TD3, SAC, and TD3 on \textcolor{myblue}{15 diverse benchmark tasks from DMControl}, including challenging high-dimensional tasks like Humanoid Walk, Humanoid Run, DogWalk, and DogRun.

\paragraph{Meta-World tasks.}
Meta-World~\citep{yu2019meta} provides a suite of simulated manipulation tasks with everyday objects, all of which are contained in a shared, tabletop environment with a simulated Sawyer arm. 
We evaluate BAC in \textcolor{myblue}{14 individual Meta-World tasks}.  Note that we conduct experiments on the goal-conditioned versions of the tasks from Meta-World-v2, which are considered harder than the
single-goal variant often used in other works.

\paragraph{Adroit tasks.}
Adroit~\citep{adroit} provides various dexterous manipulation tasks, consisting of a Shadow Dexterous Hand attached to a free arm. The system can have up to 30 actuated degrees of freedom. 

\begin{center}
    \begin{minipage}{0.9\textwidth}
    \begin{figure}[H]
        \centering
        \newcommand{\img}[1]{\hspace{0.05cm}\includegraphics[width=0.16\linewidth]{{#1}}\hspace{0.05cm}}
        \newcommand{\imgsmall}[1]{\hspace{0.04cm}\includegraphics[width=0.16\linewidth]{{#1}}\hspace{0.04cm}}
        \begin{tabular}{@{}c@{}c@{}c}
            \img{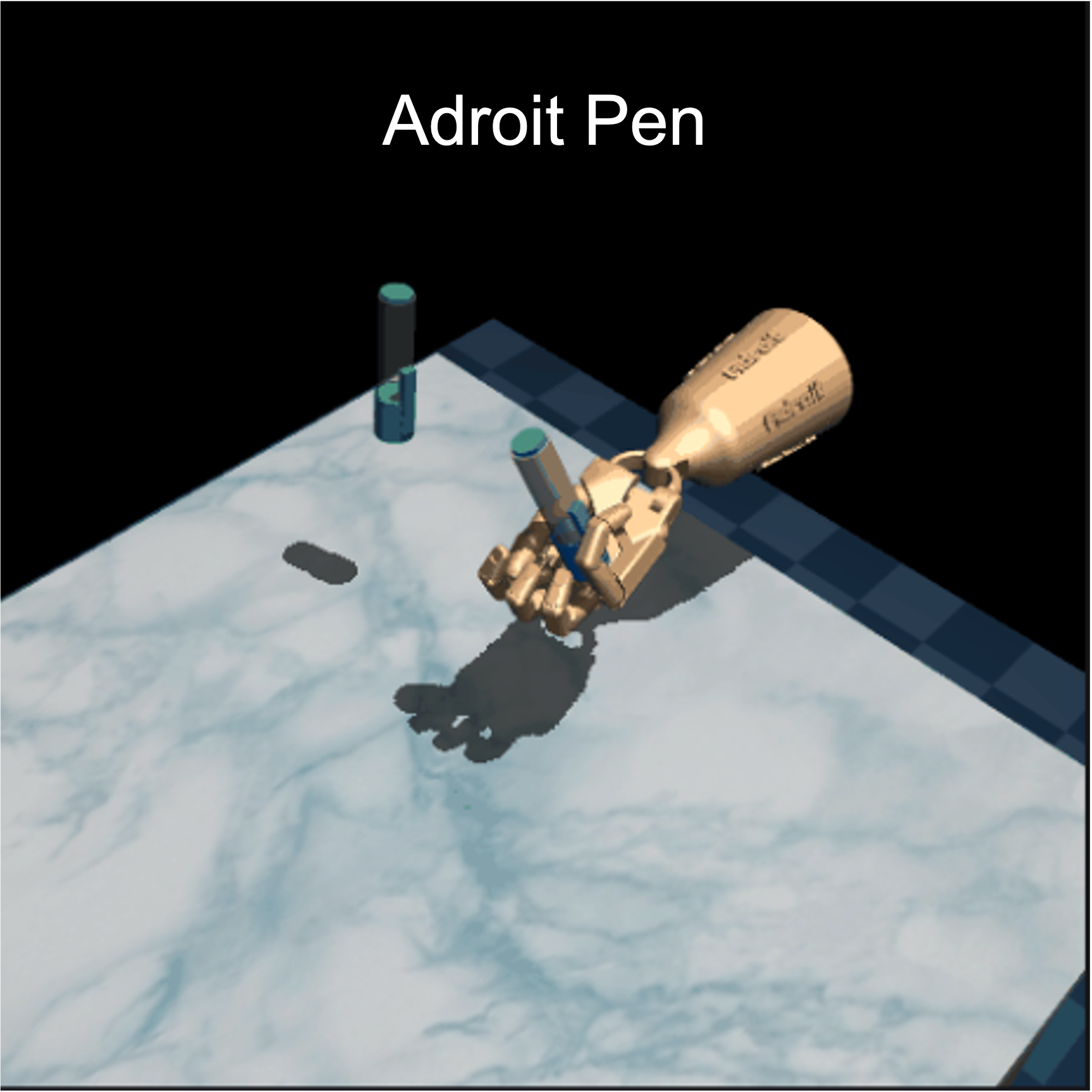} &
            \img{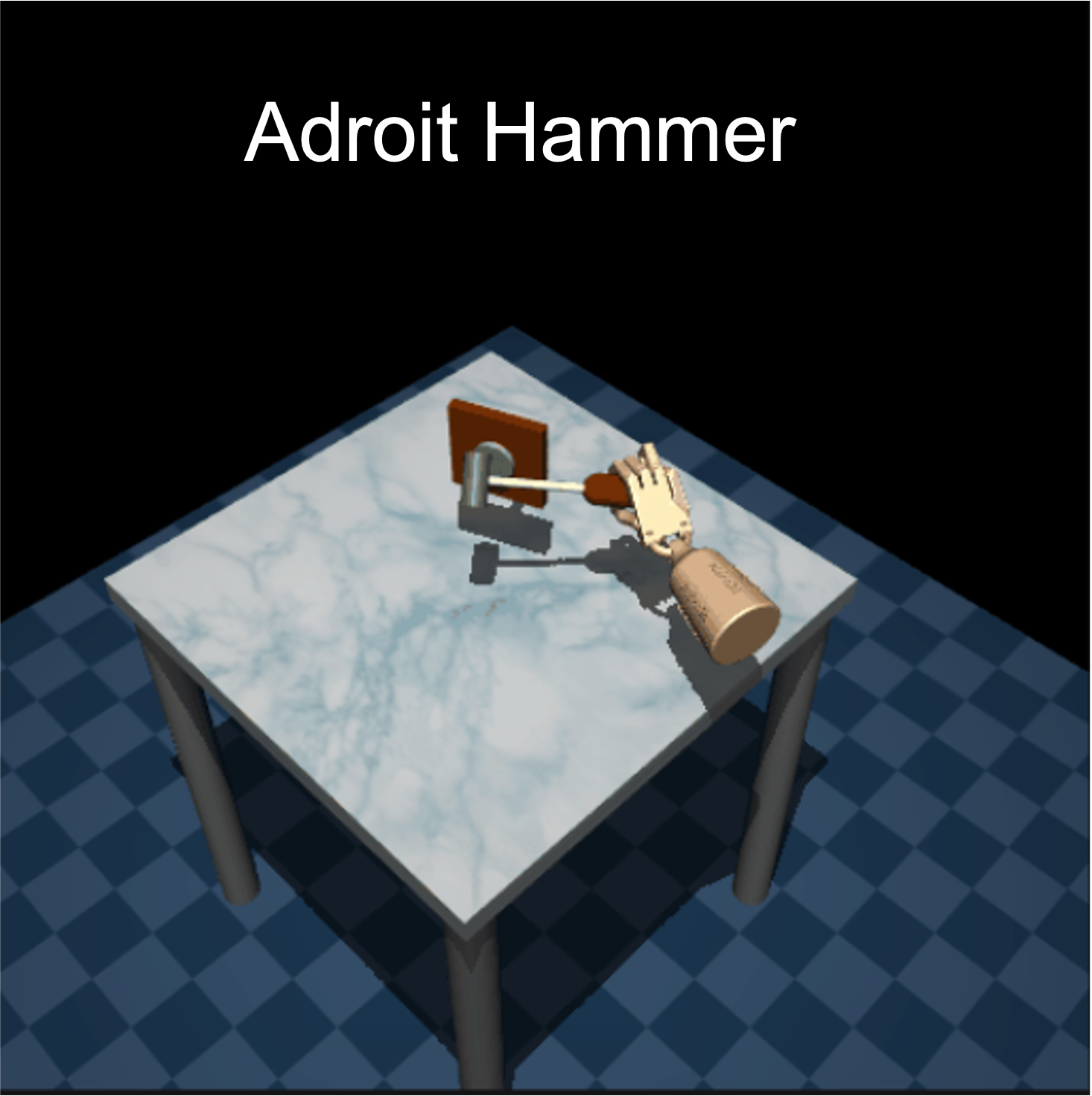} &
            \img{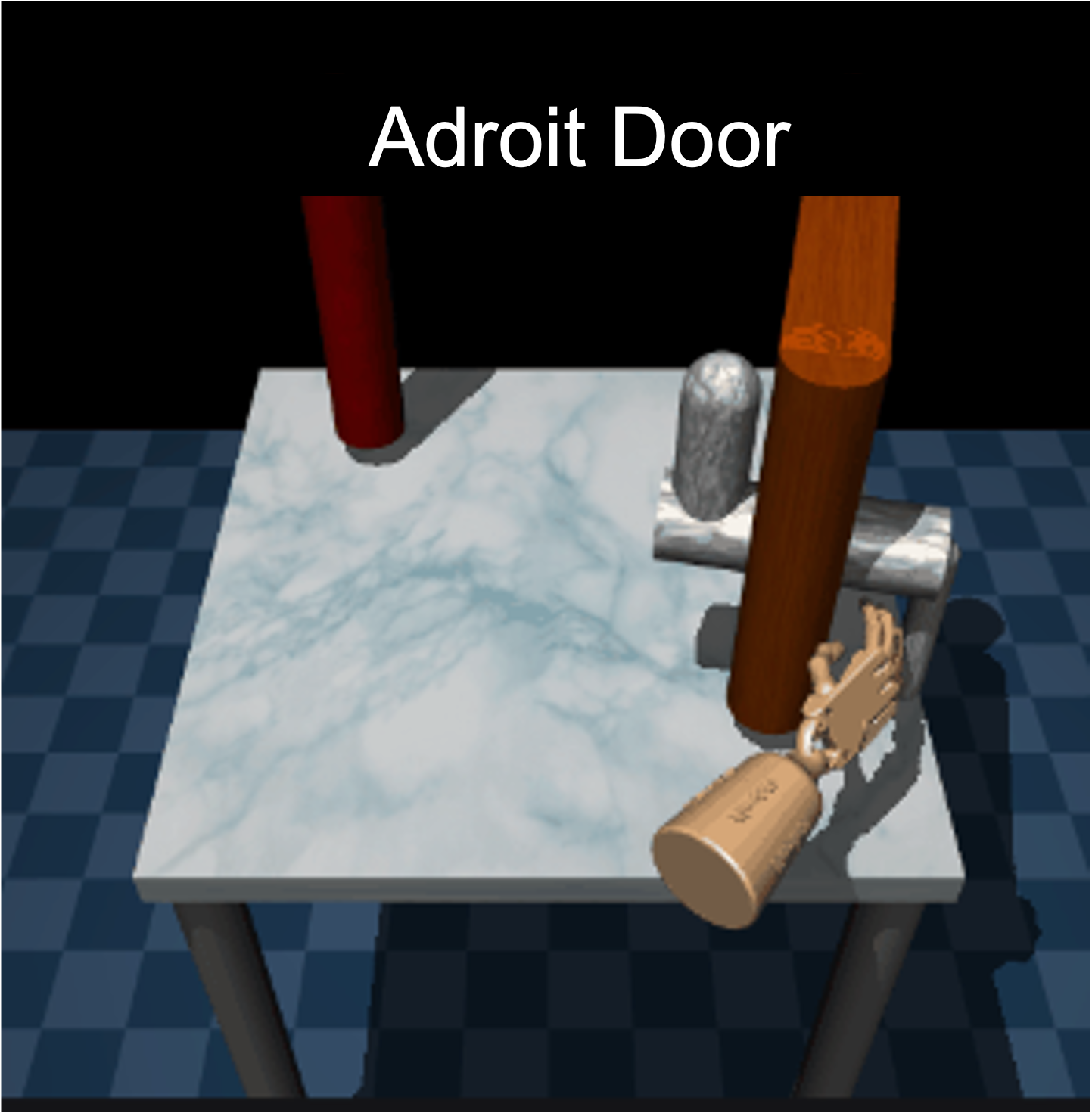} \\
        \end{tabular}
        \caption{Visualization of \textbf{\textcolor{myblue}{simulated tasks}} from Adroit.}
        \label{fig:scenarios-adroit}
    \end{figure}
    \end{minipage}
\end{center}

\paragraph{ManiSkill2 tasks.}
ManiSkill2~\citep{gu2023maniskill2} provides a suite of simulation tasks for learning generalizable manipulation skills and tackling long-horizon and complex daily chores.

\begin{center}
    \begin{minipage}{0.9\textwidth}
    \begin{figure}[H]
        \centering
        \newcommand{\img}[1]{\hspace{0.05cm}\includegraphics[width=0.16\linewidth]{{#1}}\hspace{0.05cm}}
        \newcommand{\imgsmall}[1]{\hspace{0.04cm}\includegraphics[width=0.16\linewidth]{{#1}}\hspace{0.04cm}}
        \begin{tabular}{@{}c@{}c@{}c@{}c@{}c}
            \img{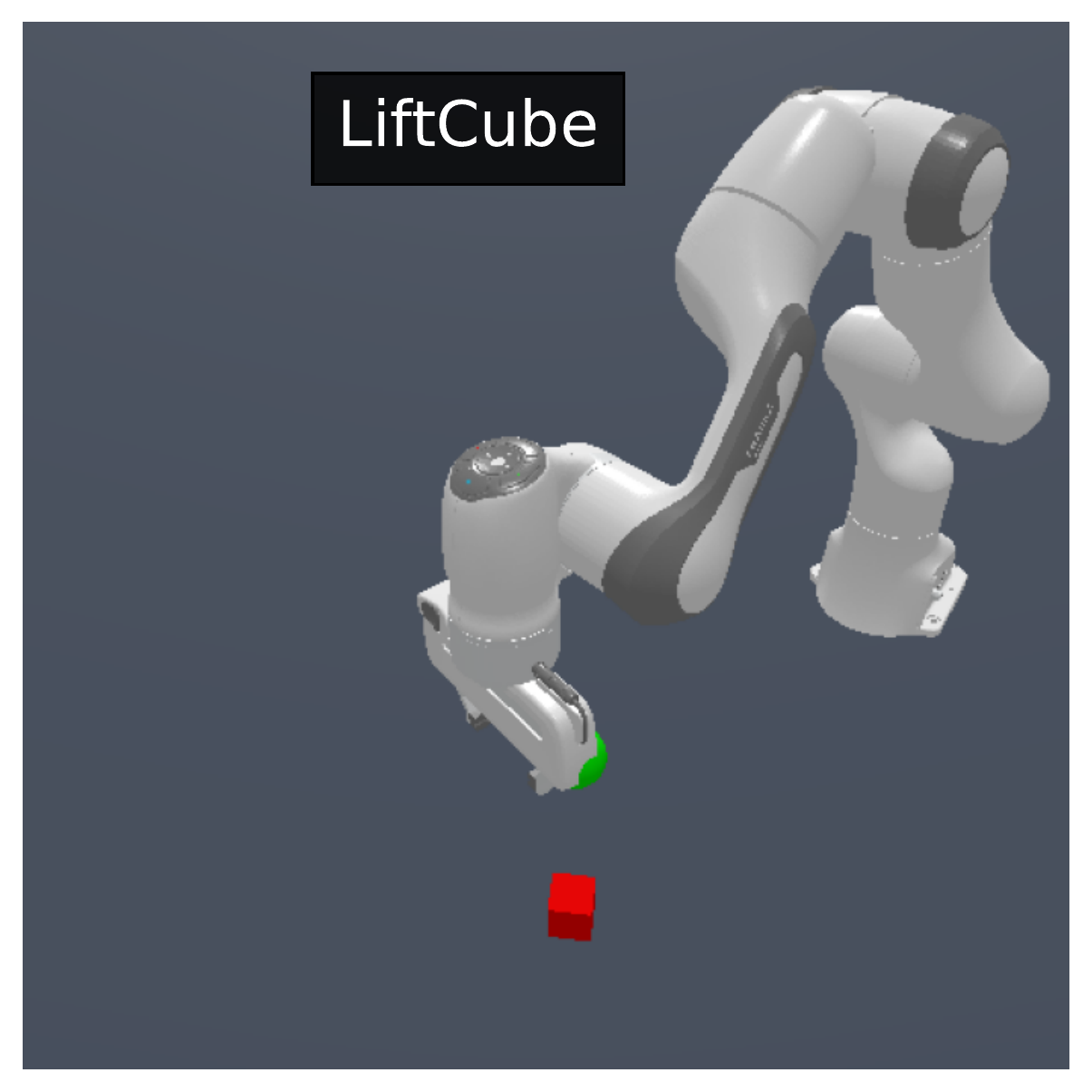} &
            \img{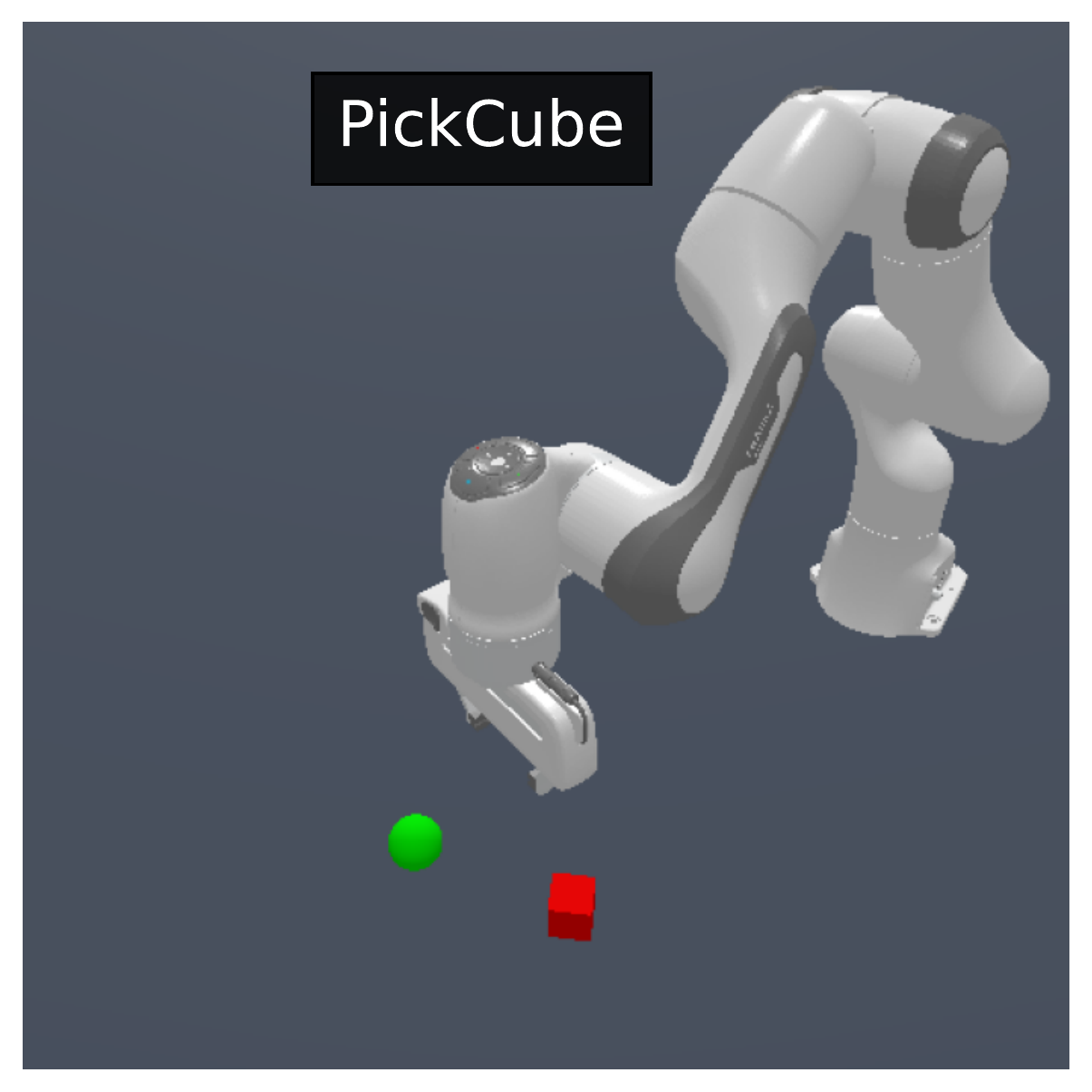} &
            \img{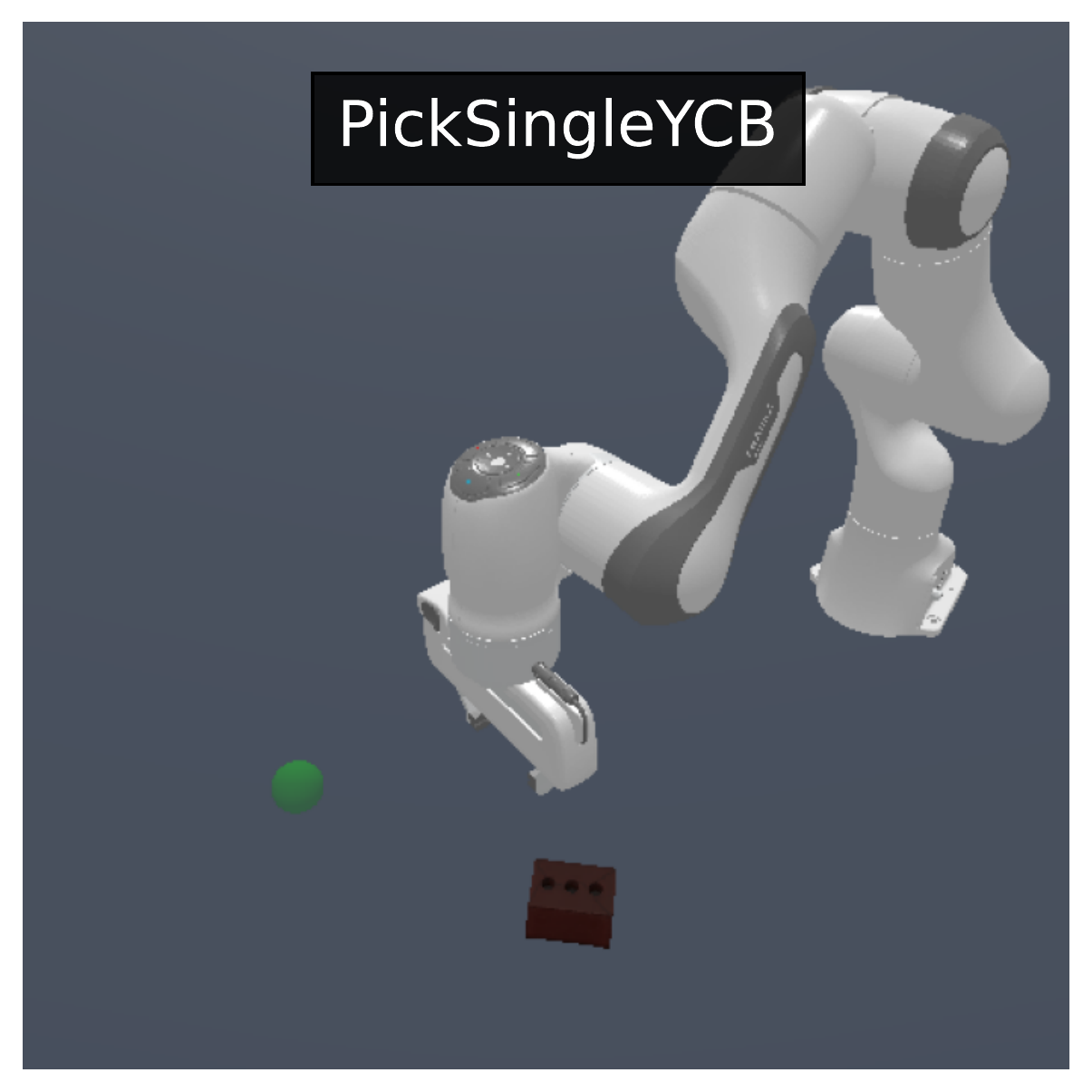} &
            \img{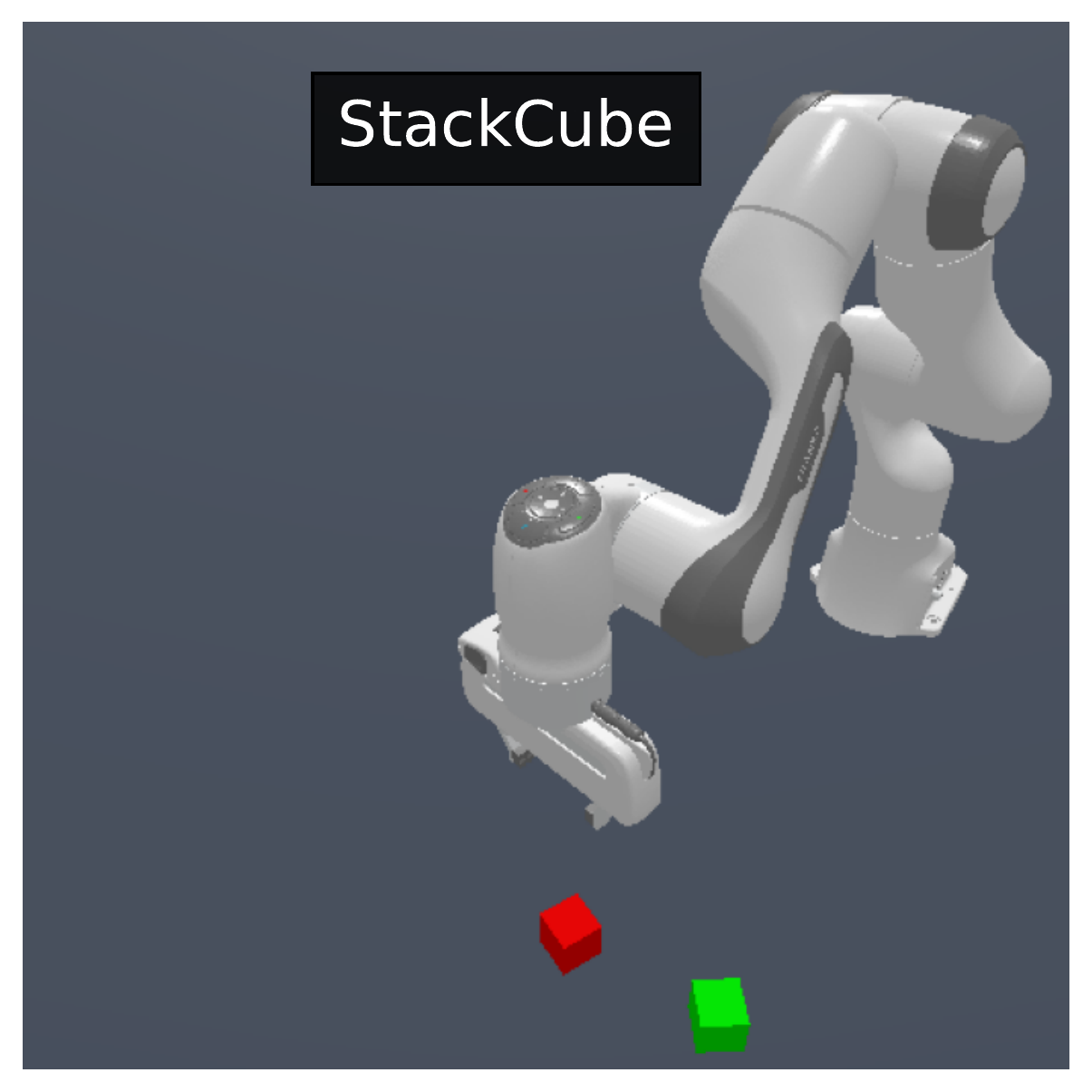} &
            \img{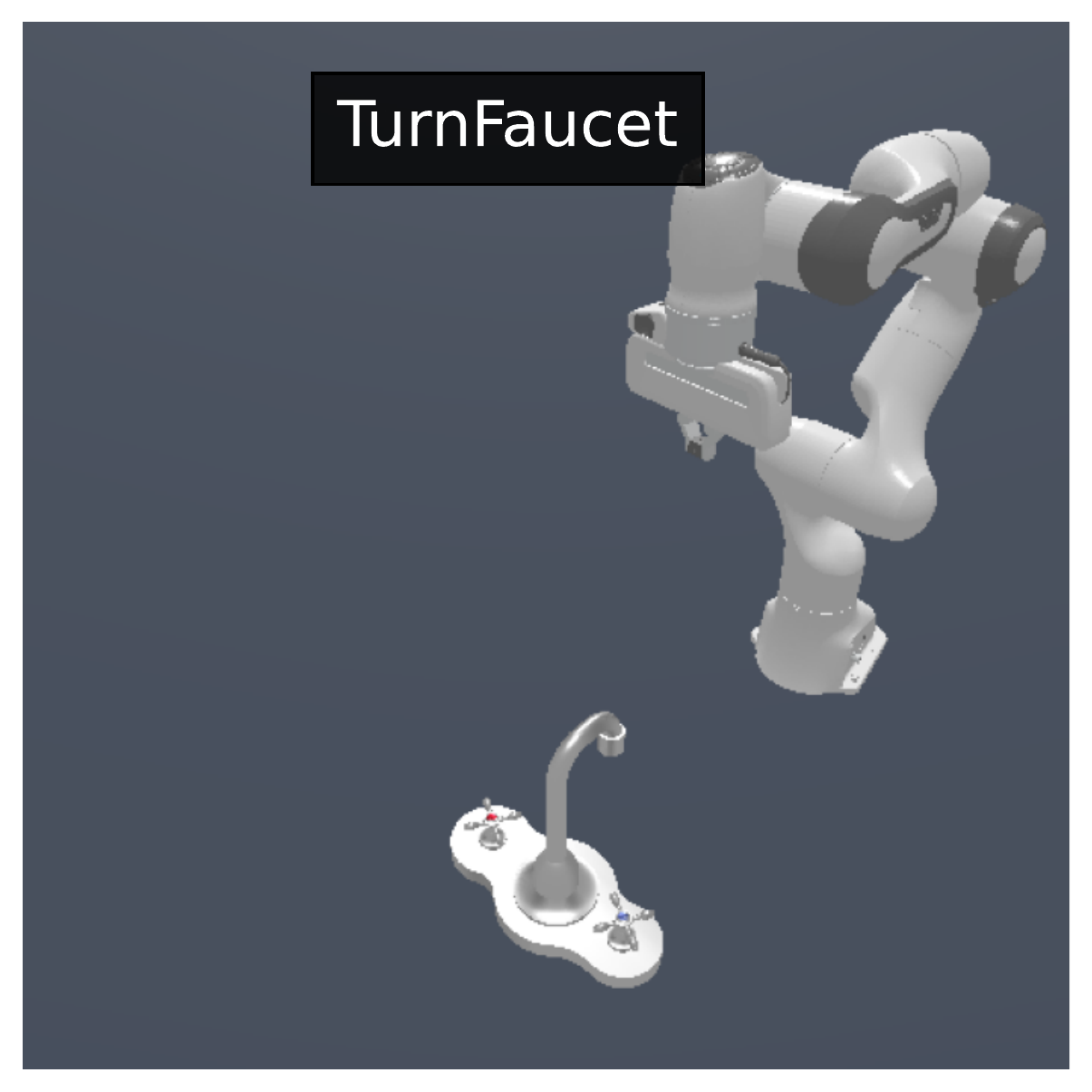} \\
        \end{tabular}
        \caption{Visualization of \textbf{\textcolor{myblue}{simulated tasks}} from ManiSkill2.}
        \label{fig:scenarios-maniskill2}
    \end{figure}
    \end{minipage}
\end{center}

\paragraph{Shadow Dexterous Hand tasks.}
The environments of Shadow Dexterous Hand~\citep{dexshadowhand} are based on the Shadow Dexterous Hand, 5 which is an anthropomorphic robotic hand with 24 degrees of freedom. Of those 24 joints, 20 can be controlled independently, whereas the remaining ones are coupled joints. This sophisticated design mirrors human hand movements, allowing for complex and nuanced robotic tasks.

\paragraph{Panda-gym.}
Panda-gym provides~\citep{panda-gym} a simulated environment of Franka Emika Panda robot for common tasks used to evaluate the RL algorithms.

\paragraph{MyoSuite}
MyoSuite~\citep{MyoSuite2022} is a set of challenging environments and tasks that help test how well reinforcement learning algorithms work in controlling muscles and bones in a realistic way. It includes detailed models of the elbow, wrist, and hand that can interact physically, making it possible to learn tasks that require complex and skillful movements. These tasks vary from simple body postures to more advanced actions like turning a key, spinning a pen, or rolling two balls in one hand.

\subsection{Environment setup for evaluating \mbshort}
We evaluate \mbshort\ and its counterparts on four continuous control tasks in MuJoCo~\citep{Mujoco}. To ensure a fair comparison, we follow the same settings as our model-based baselines (MBPO~\citep{mbpo}, AutoMBPO~\citep{autombpo}, CMLO~\citep{ji2022update}), in which observations are truncated.
The details of the experimental environments are provided in Table~\ref{mb-env-setting}. 

\begin{table}[h]
    \caption{Overview on environment settings for \mbshort\ and model-based baselines. Here, $\theta_t$ denotes the joint angle and $z_t$ denotes the height at time $t$.}
    \begin{center}
    \resizebox{0.7\textwidth}{!}{
        \begin{tabular}{
            >{\centering}m{0.15\textwidth}
            | c
            | c
            | c
            | c
        }
            \toprule
            & \makecell[c]{State Space \\Dimension} & \makecell[c]{Action Space \\Dimension} & Horizon & Terminal Function \\
            \midrule
            Hopper-v2
            & 11
            & 3
            & 1000
            & $z_t \leq 0.7$ or $\mathbb{\theta}_t \geq 0.2$
            \\
            \midrule
            Walker2d-v2
            & 17
            & 6
            & 1000
            & \makecell[c]{$z_t \geq 2.0$ or $z_t \leq 0.8$ or \\ $\mathbb{\theta}_t \leq -1.0$  or $\mathbb{\theta}_t\geq 1.0$}
            \\
            \midrule
            Ant-v2
            & 27
            & 8
            & 1000
            & $z_t < 0.2$ or $z_t > 1.0$
            \\
            \midrule
            Humanoid-v2
            & 45
            & 17
            & 1000
            & $z_t < 1.0$ or $z_t > 2.0$
            \\
            \bottomrule
        \end{tabular}
        }
    \end{center}
    \label{mb-env-setting}
\end{table}

\clearpage
\section{Baselines Implementation}\label{baseline-implementation}
\paragraph{Model-free RL algorithms.} We compare with five popular model-free baselines, Soft Actor-Critic~(SAC)~\citep{sac}, Diversity Actor-Critic~(DAC)~\citep{dac}, Random Reward Shift~(RRS)~\citep{rrs}, Twin Delayed DDPG~(TD3)~\citep{td3}, Proximal Policy Optimization~(PPO)~\citep{ppo}. 
For RRS, we use the RRS-7 0.5 version as it provides better performance across diverse environments compared to other alternatives~(RRS-3 0.5, RRS-3 1.0, RRS-7 1.0).
For MuJoCo tasks, the hyperparameters of DAC, RRS, TD3, and PPO are kept the same as the authors’ implementations. \Revise{We list the hyperparameters of TD3 in Table~\ref{td3-hyperparameters}. Note that we mostly follow the implementation of the original paper but improve upon certain hyperparameter choices for DMControl and Meta-World tasks.}

The implementation of SAC is based on the open-source repo (\citet{saclib}, MIT License). 
And we use automating entropy adjustment~\citep{sacapplication} for automatic $\alpha$ tuning. 
On MuJoCo benchmarks, we retain other parameters as used by the authors~\citep{sac}. On DMControl benchmarks, we followed the SAC hyperparameters suggested by TD-MPC paper~\citep{tdmpc}.  On Meta-World benchmarks, we followed the SAC hyperparameters suggested by Meta-World paper~\citep{yu2019meta}.  \Revise{We list the hyperparameters of SAC in Table~\ref{sac-hyperparameters}.}

\begin{table}[h]
    \caption{\Revise{Hyperparameter settings for SAC in MuJoCo and DMControl, Meta-World benchmark tasks.}}
    \begin{center}
    \resizebox{0.8\textwidth}{!}{
    {
        \begin{tabular}{
            >{\centering}m{0.35\textwidth}
            | c
            | c
            | c
        }
            \toprule
             & ~~~MuJoCo~~~ & ~~~DMControl~~~ & ~~~Meta-World~~~ \\
            \midrule
            optimizer for $Q$ & \multicolumn{3}{c}{
                Adam($\beta_1$=0.9, $\beta_2$=0.999)
            } \\
            \midrule
            optimizer for $\alpha$ & \multicolumn{3}{c}{
                Adam($\beta_1$=0.5, $\beta_2$=0.999)
            } \\
            \midrule
            
            learning rate & $3 \times 10^{-4}$ & \makecell{$1 \times 10^{-4}$ (otherwise)\\ $3 \times 10^{-4}$ (Dog)} & $3 \times 10^{-4}$ \\
            \midrule
            discount ($\gamma$) & \multicolumn{3}{c}{
                0.99
            }\\
            \midrule
            number of hidden units per layer & 256 & 1024 & 256\\
            \midrule
            number of samples per minibatch & 256& \makecell{ 512 (otherwise)\\2048 (Dog)} & 500\\
            \midrule
            target smoothing coefficient ($\tau$) & \multicolumn{3}{c}{
                0.005
            }\\
            \midrule
            target update interval & 1 & 2 & 1 \\
            \midrule
            gradient steps & \multicolumn{3}{c}{
                1
            }\\
            \bottomrule
        \end{tabular}
        }
    }
    \end{center}
    \label{sac-hyperparameters}
    \vspace{-5mm}
\end{table}

\begin{table}[h]
    \caption{\Revise{Hyperparameter settings for TD3 in MuJoCo and DMControl, Meta-World benchmark tasks.}}
    \begin{center}
    \resizebox{0.8\textwidth}{!}{
    {
        \begin{tabular}{
            >{\centering}m{0.35\textwidth}
            | c
            | c
            | c
        }
            \toprule
             & ~~~MuJoCo~~~ & ~~~DMControl~~~ & ~~~Meta-World~~~ \\
            \midrule
            optimizer for $Q$ & \multicolumn{3}{c}{
                Adam($\beta_1$=0.9, $\beta_2$=0.999)
            } \\
            \midrule
            exploration noise & \multicolumn{3}{c}{
                 $\mathcal{N}(0, 0.1)$
            } \\
            \midrule
            
            learning rate & $3 \times 10^{-4}$ & \makecell{$1 \times 10^{-4}$ (otherwise)\\ $3 \times 10^{-4}$ (Dog)} & $3 \times 10^{-4}$ \\
            \midrule
            discount ($\gamma$) & \multicolumn{3}{c}{
                0.99
            }\\
            \midrule
            hidden layers & (400, 300) & (512, 512) & (512, 512)\\
            \midrule
            number of samples per minibatch & 100 & \makecell{ 256 (otherwise)\\512 (Dog)} & 256\\
            \midrule
            target smoothing coefficient ($\tau$) & \multicolumn{3}{c}{
                0.005
            }\\
            \midrule
            target update interval & \multicolumn{3}{c}{1}  \\
            \bottomrule
        \end{tabular}
        }
    }
    \end{center}
    \label{td3-hyperparameters}
\end{table}

\paragraph{Model-based RL algorithms.}
As for model-based methods, we compare with four state-of-the-art model-based algorithms, MBPO~\citep{mbpo}, SLBO~\citep{slbo}, CMLO~\citep{ji2022update}, AutoMBPO~\citep{autombpo}. 
The implementation of SLBO is taken from an open-source MBRL benchmark~\citep{wang2019benchmarking}, while MBPO is implemented based on the \texttt{MBRL-LIB} toolbox~\citep{Pineda2021MBRL}. To facilitate a fair comparison, \mbshort\ and MBPO are run with identical network architectures and training configurations as specified by \texttt{MBRL-LIB}.

\clearpage
\section{Investigations on the Underestimation Issue}\label{section:appendix-underestimation}
The underestimation issue matters. While prior works focus more on reducing overestimation, our work shows that mitigating underestimation itself may improve both performance and sample efficiency. Let's delve deeper.

\subsection{Why underestimation and under-exploitation matters?}
Underestimation in the under-exploitation stage would negatively impact $Q$-value estimation.  Underestimating Q-values of $(s,a)$ due to suboptimal current policy successors, ignoring high-value replay buffer successors, hampers reselection of $(s,a)$ . Two issues might arise, 
\begin{itemize}[leftmargin=16pt]
\item  \textbf{Reduce sample efficiency}: The agent would require more samples to re-encounter such $(s,a)$.
\item  \textbf{Hinder policy learning}: Misleading $Q$ may trap the policy in ineffective exploration.  The issue is exacerbated in failure-prone scenarios where high-value tuples are serendipities and policy performance oscillates.
\end{itemize}

\subsection{What may cause the underestimation issue?} 
\begin{wrapfigure}[10]{r}{0.32\textwidth}
    \vspace{-1.5cm}
    \centering
    \includegraphics[height=4.0cm,keepaspectratio]{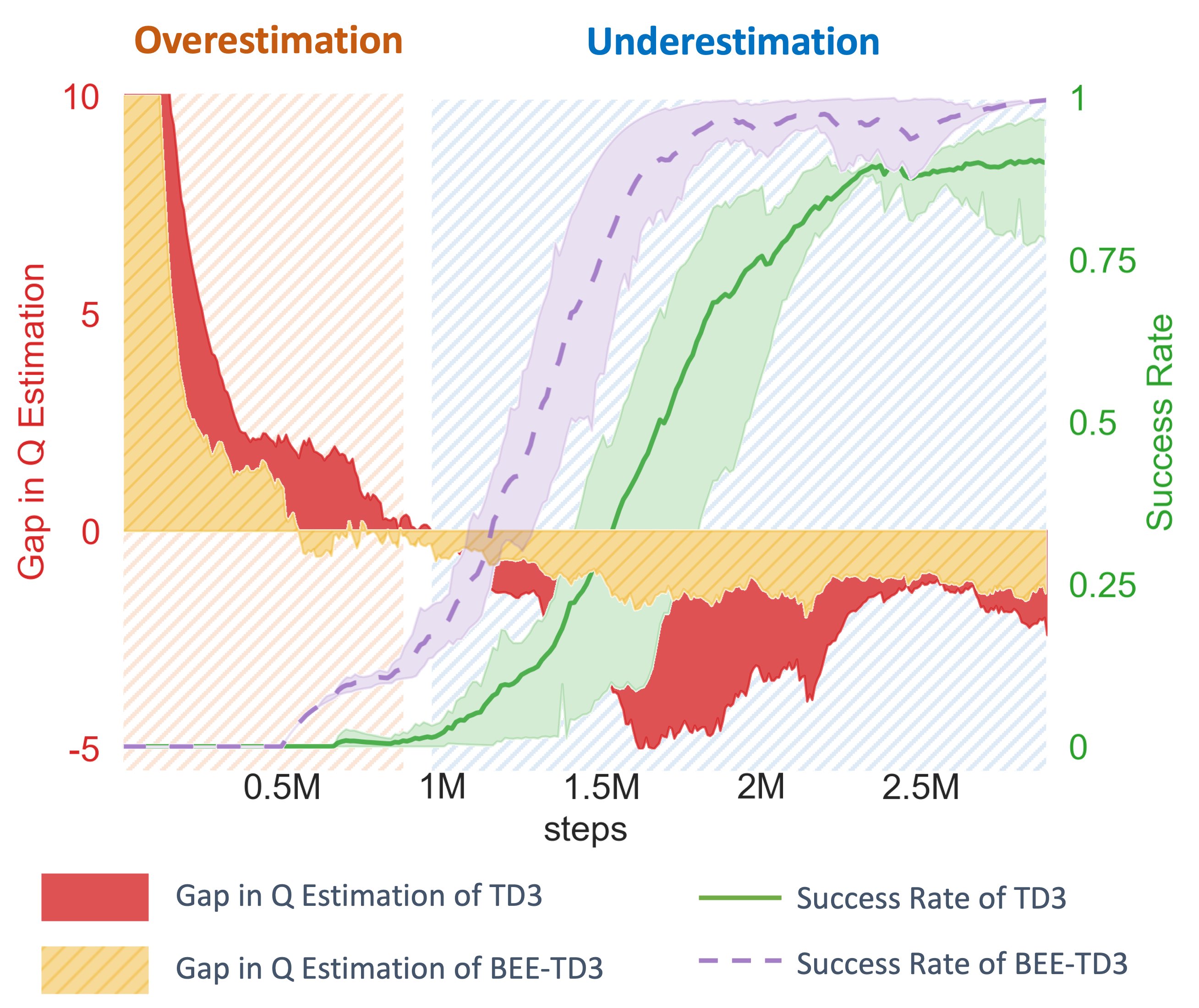} 
    \caption{TD3 is also prone to underestimation pitfalls in the latter stage of training. }
    \label{app-fig:td3underestimation}
     \vspace{-5pt}
\end{wrapfigure}
Underestimation bias may result in inefficient exploration, diminished sample efficiency, and, ultimately, reduced policy learning. The underestimation bias has been associated with the use of double-Q-technique, as highlighted in previous works~\citep{td3, top}. However, is the underestimation issue a sore outcome of the double-Q technique?

\paragraph{Underestimation issue in various off-policy actor-critic algorithms.} 

AC algorithms are susceptible to the underestimation issue,  as shown in Figure~\ref{fig:motivation}. To further illustrate this issue, we quantify the $Q$-value estimation gap of TD3 and BEE-TD3 in the DKittyWalkRandomDynamics task.  
The gap in Q estimation is evaluated by comparing the TD3/BEE-TD3's $Q$-values and the Monte-Carlo $Q$ estimates using the trajectories in the replay buffer.

As shown in Figure~\ref{app-fig:td3underestimation}, TD3 also experiences underestimation pitfalls during the later stages of training. Notably, we observe that the BEE operator helps to mitigate this underestimation issue and finally benefits performance.

\paragraph{Double-Q technique is not the only culprit.}
We identify that the underestimation problem also occurs in many off-policy Actor-Critic~(AC) algorithms independent of this technique. 
In this section, we investigate the causes of underestimation in the AC framework, irrespective of the double-Q technique’s application.   We also empirically show that various off-policy AC algorithms, with or without the double-Q trick, are prone to underestimation issues in many tasks.

\begin{figure}[t]
    \centering
    \begin{subfigure}[t]{0.63\textwidth}
        \centering
        \includegraphics[height=2.6cm,keepaspectratio]{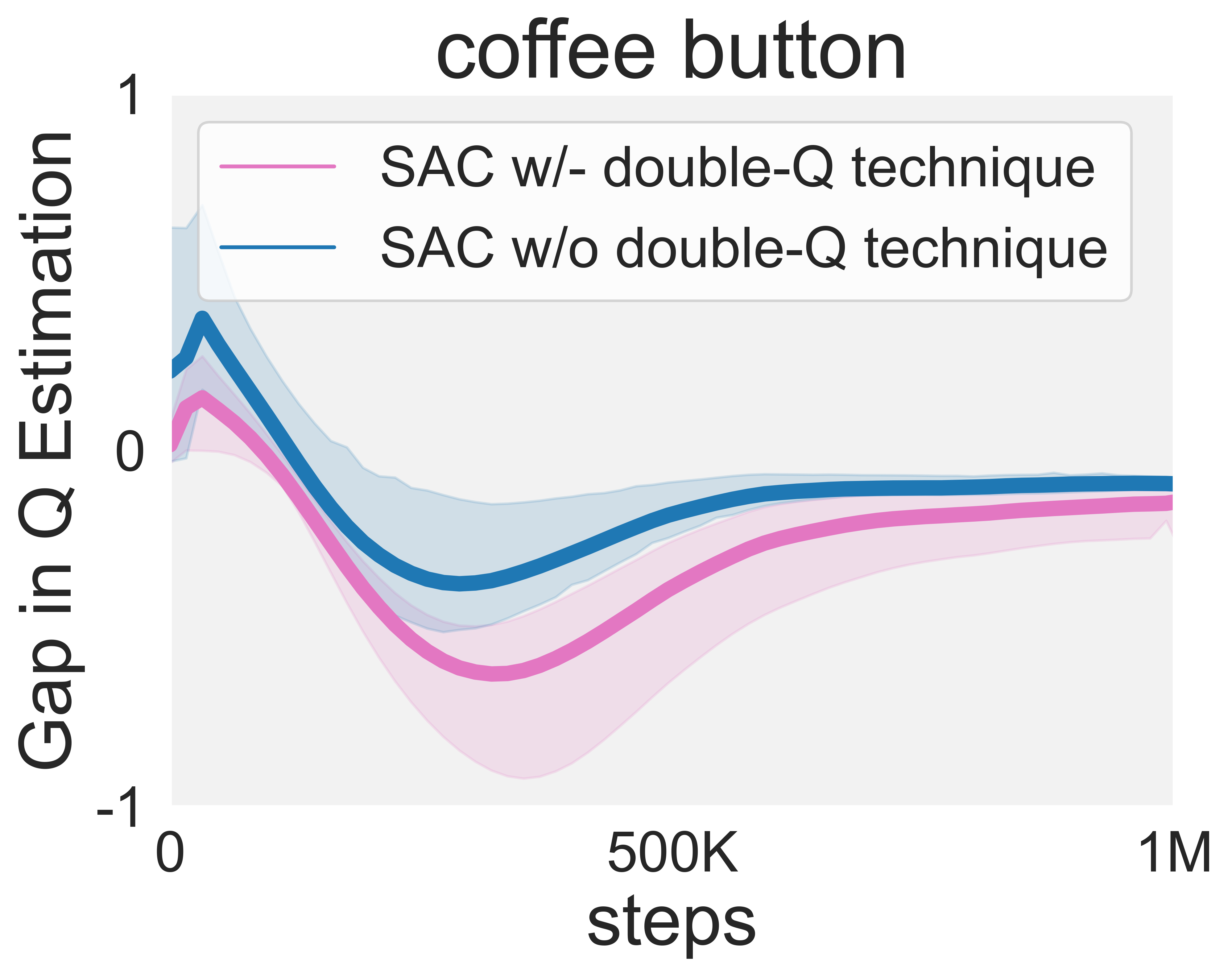}
        \includegraphics[height=2.6cm,keepaspectratio]{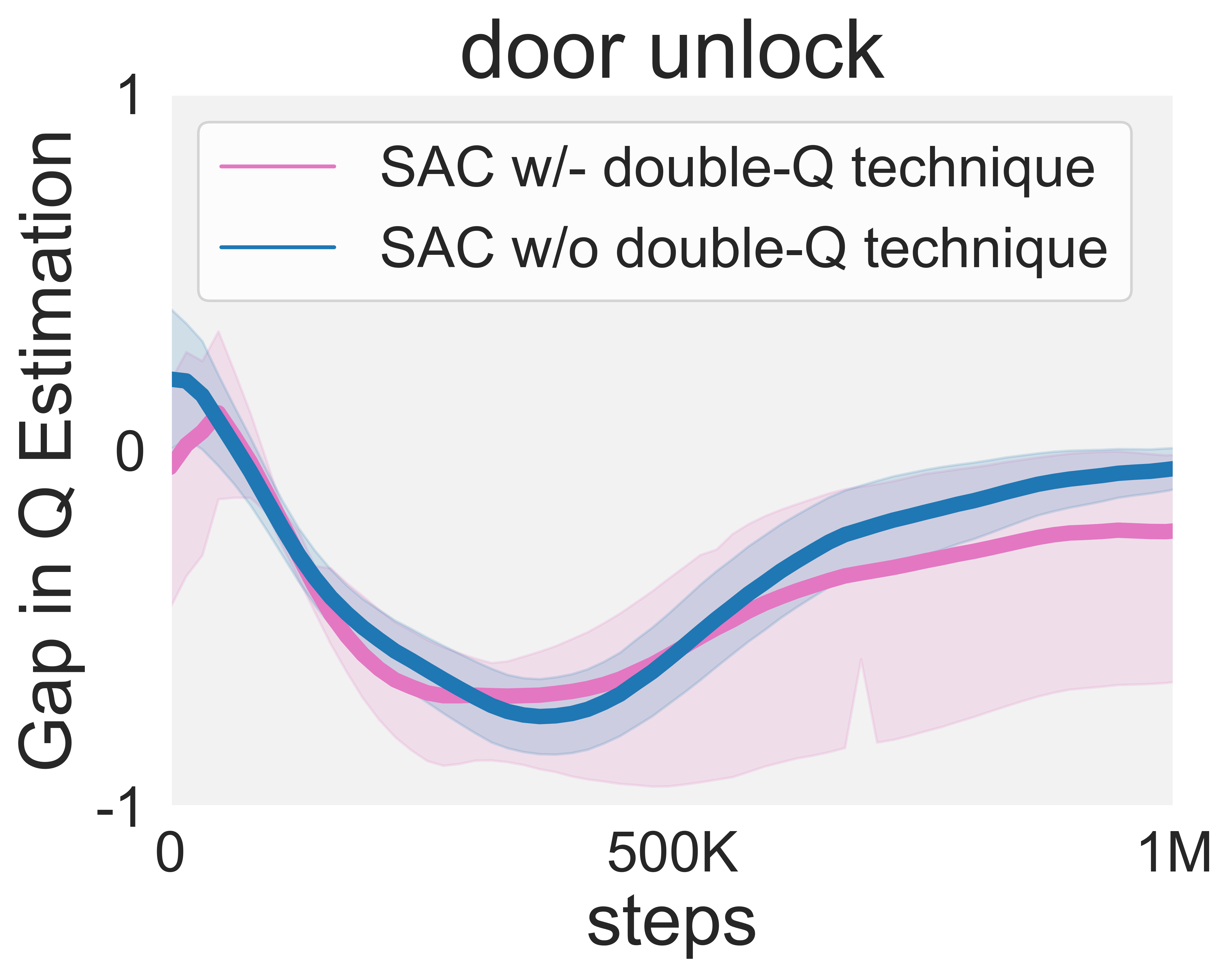}
        \includegraphics[height=2.6cm,keepaspectratio]{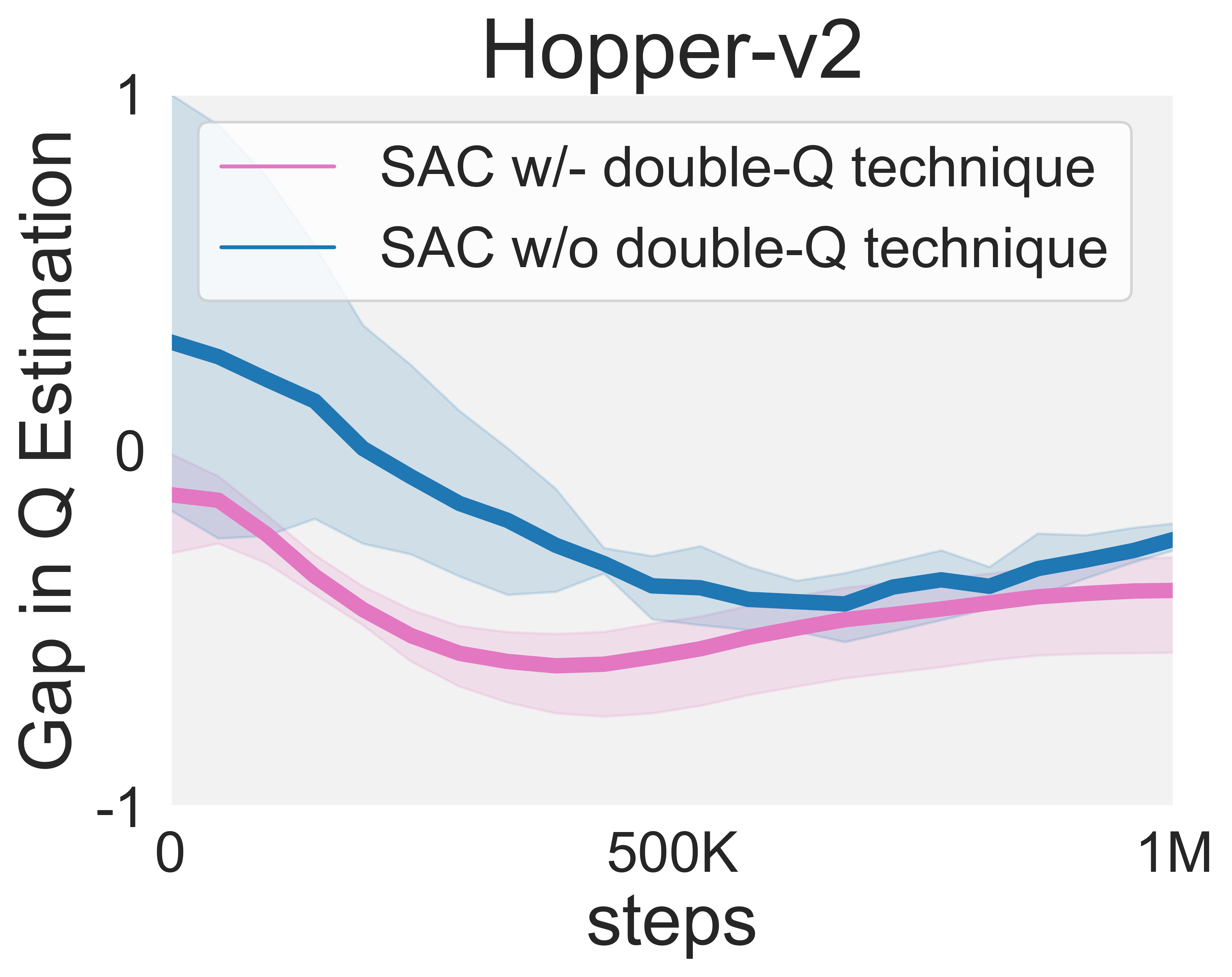}
        \\
        \includegraphics[height=2.6cm,keepaspectratio]{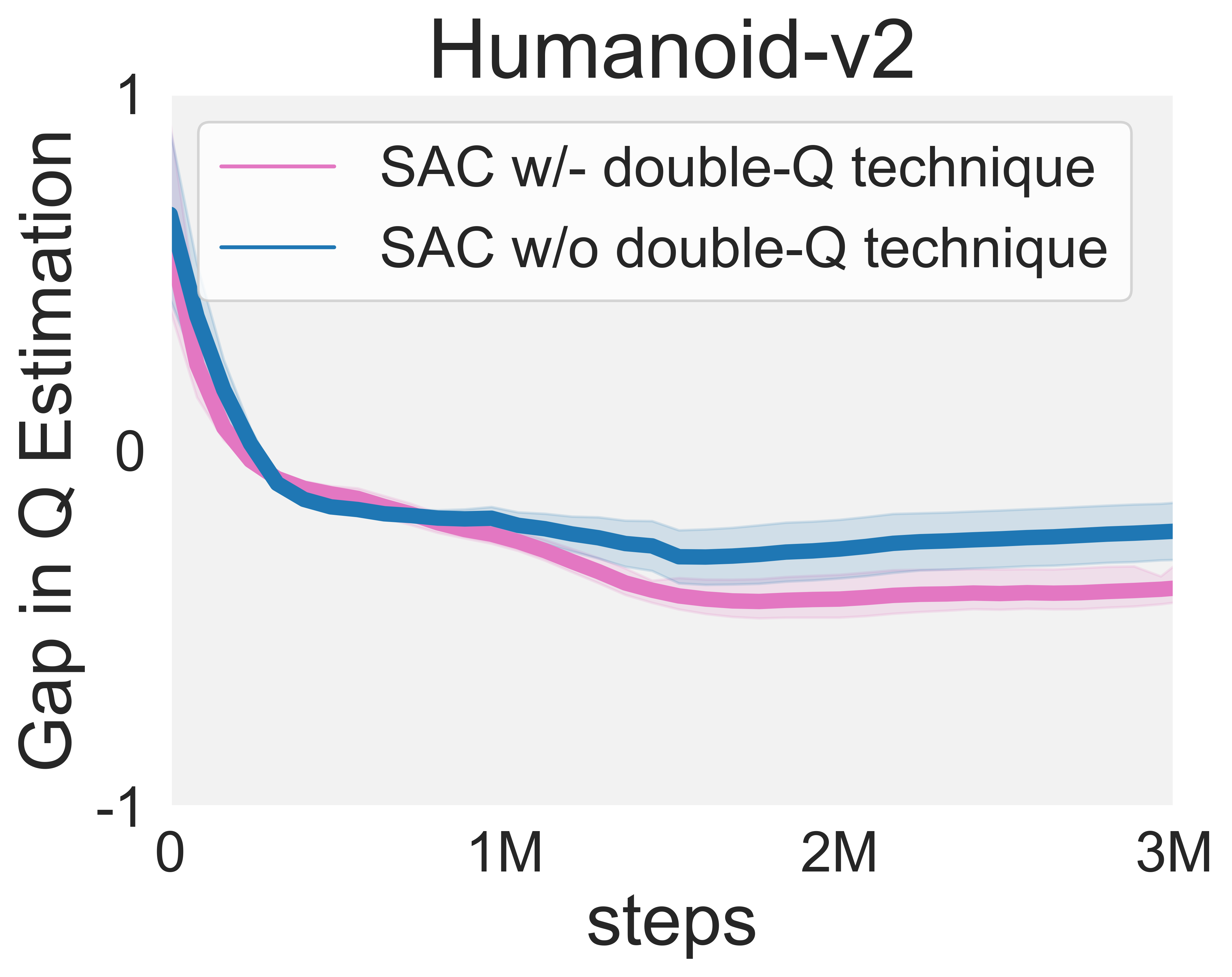}
        \includegraphics[height=2.6cm,keepaspectratio]{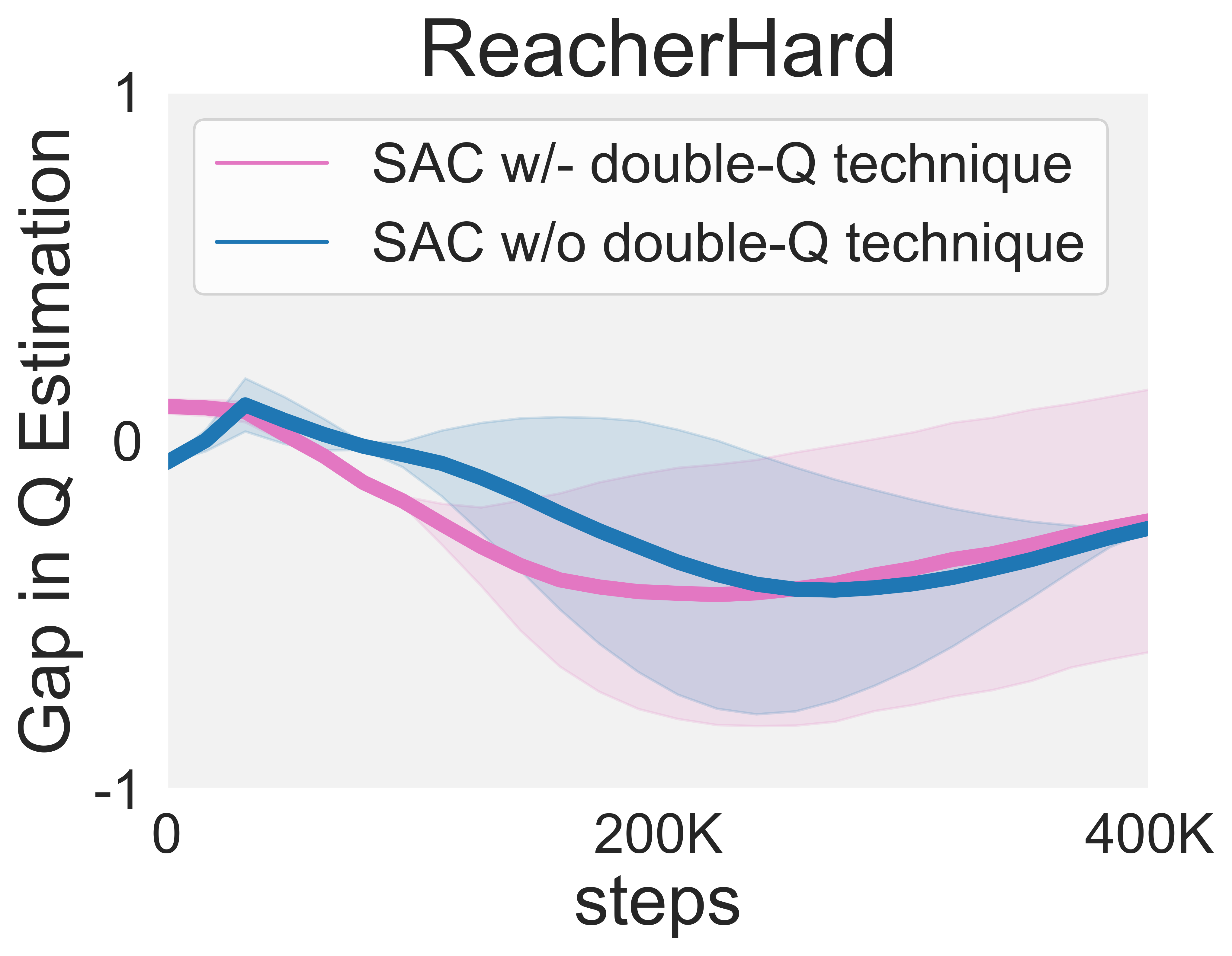}
        \includegraphics[height=2.6cm,keepaspectratio]{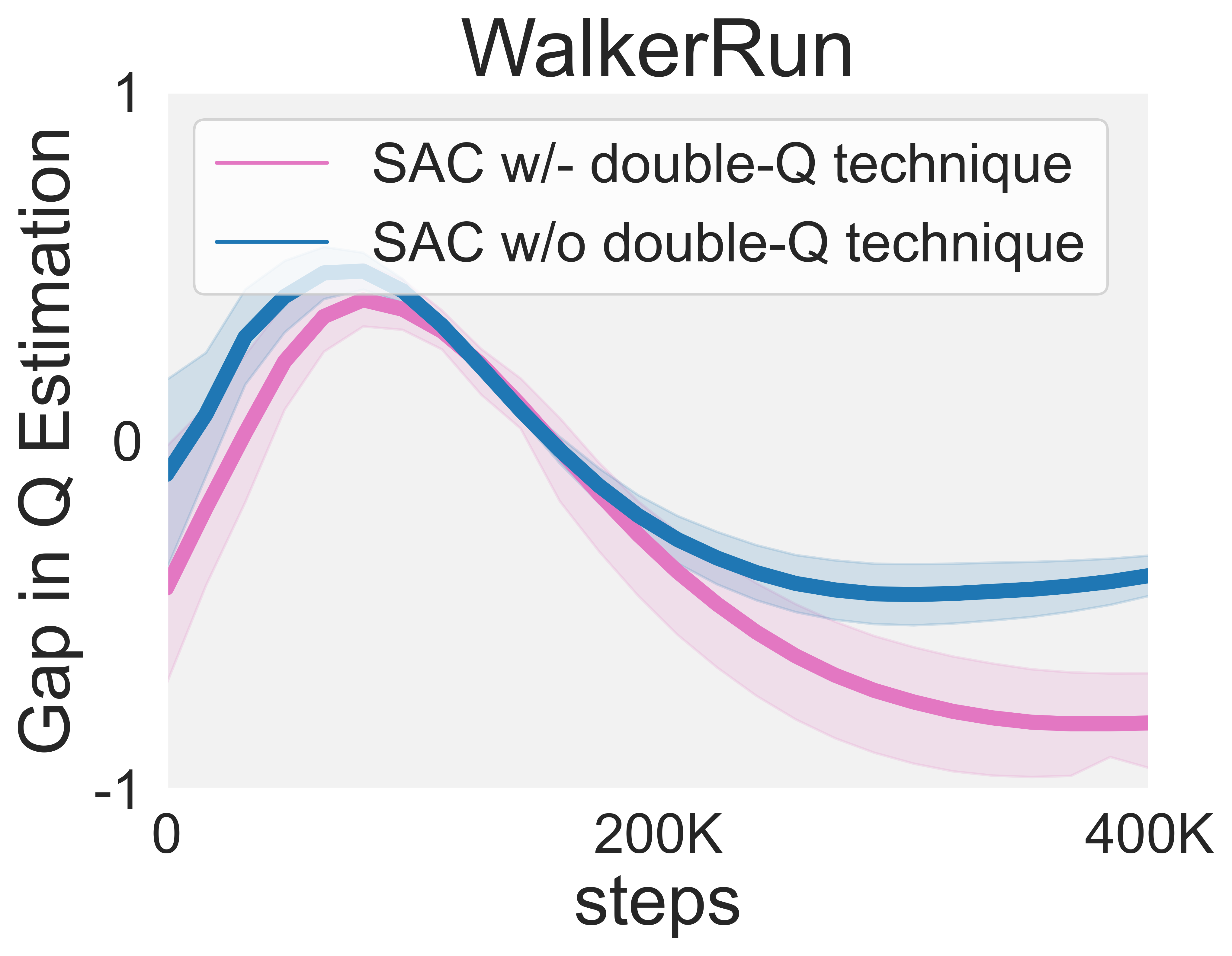}
        \caption{SAC with or without double-Q-trick faces underestimation in the latter stage of training.}
        \label{app-fig:sacwithoutdoubleQtrick}
    \end{subfigure}
    \hspace{5pt}
    \begin{subfigure}[t]{0.35\textwidth}
        \centering
        \includegraphics[height=2.6cm,keepaspectratio]{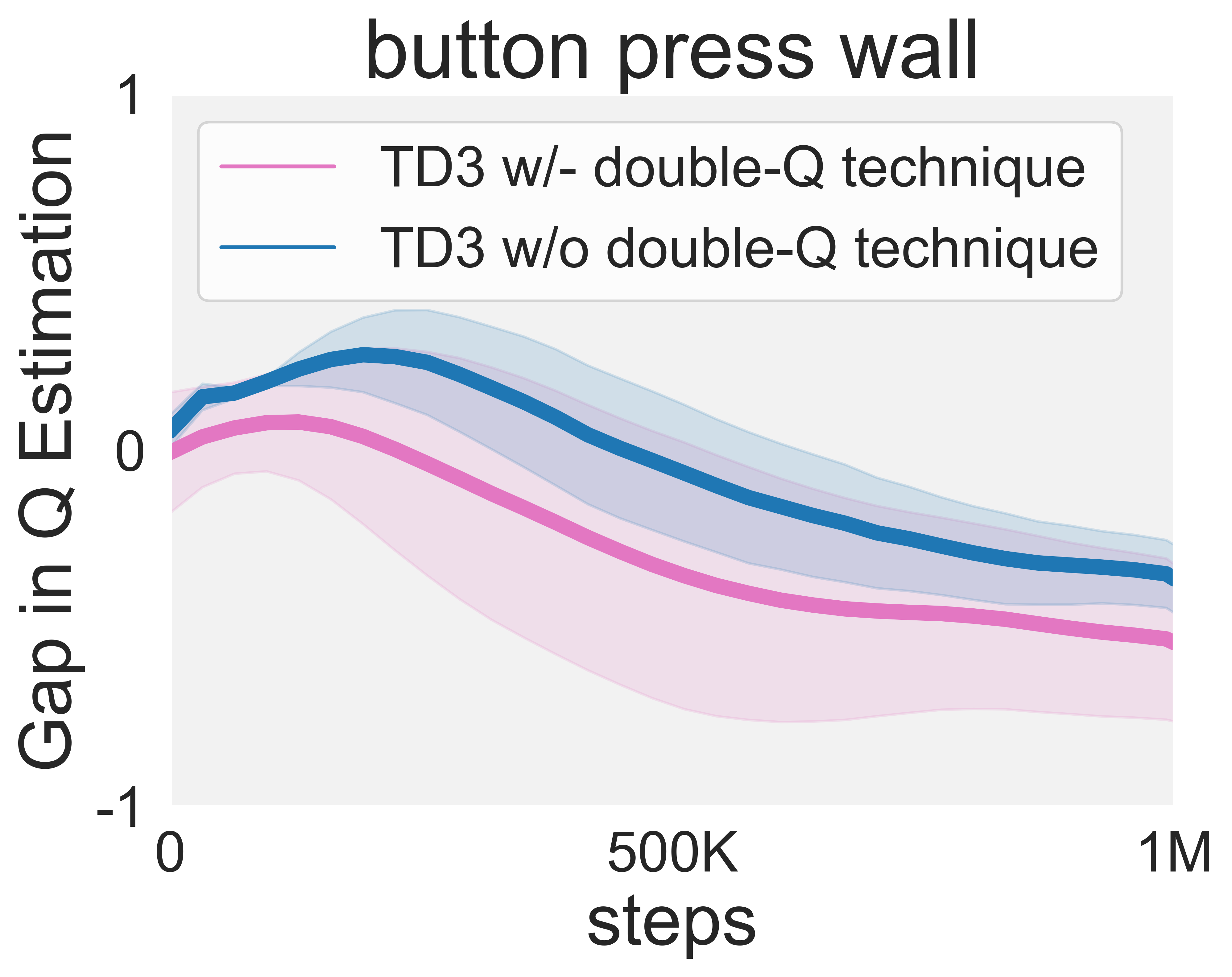}
        \includegraphics[height=2.6cm,keepaspectratio]{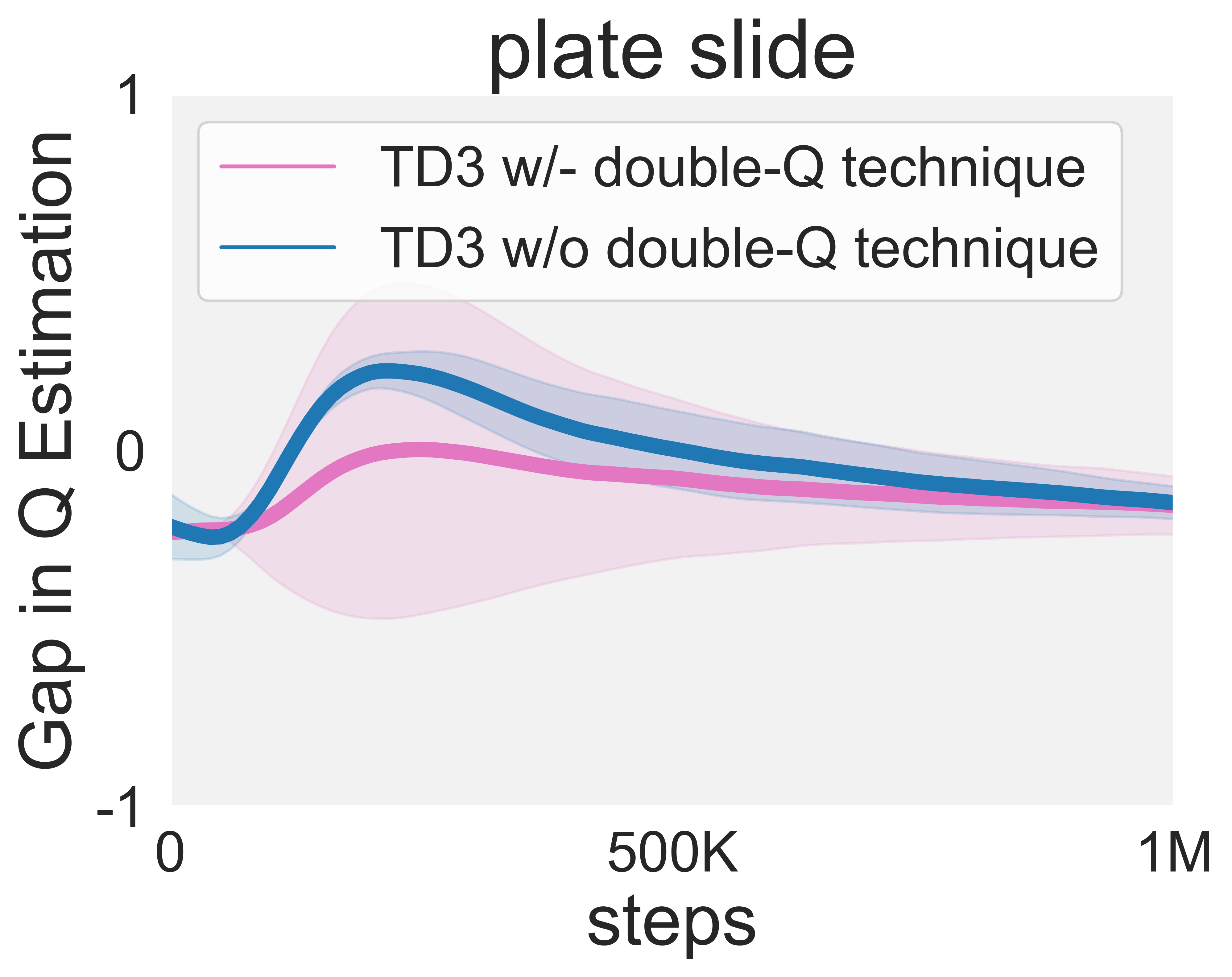}
        \caption{\Revise{TD3 with or without double-Q-trick faces underestimation in the latter stage of training.}}
        \label{app-fig:td3withoutdoubleQtrick}
    \end{subfigure}
    \caption{The gap is assessed by subtracting the $Q$-values generated by the current policy from Monte-Carlo $Q$ estimates  derived from trajectories sampled using the current policy. A value less than $0$ indicates underestimation. Off-policy Actor-Critic algorithms (SAC, TD3) with or without double-Q-trick are prone to underestimation pitfalls. }
    \label{fig:withoutdoubleQtrick}
\end{figure}

In Figure~\ref{fig:withoutdoubleQtrick}, we plot $Q$-value estimation of SAC and TD3, along with their variants by eliminating the double-Q technique. We observe that both SAC and TD3 would face the underestimation issue in various robotic tasks.

\paragraph{The optimization procedure of the AC framework can also contribute to underestimation.}
Ideally, the Bellman update needs to solve $Q(s,a)\leftarrow r(s,a)+\gamma \mathbb{E}_s[\max_a Q(s,a)]$. However, as $\max_a Q(s,a)$ operations are often impractical to calculate, so in the AC framework, we typically iteratively evaluate target Q-value as $\mathbb{E}_{\pi}[Q(s,a)]$, while implicitly conducting the max-Q operation in a separate policy improvement step to learn policy $\pi$. 

Note that the ideal $\pi=\arg\max_{a\sim \pi} Q(s,a)$ is not possible to achieve practically within only a few policy gradient updates. Hence, the actual target value used in AC Bellman update $\mathbb{E}_{s,a\sim \pi}Q(s,a)$ can have a high chance to be smaller than $\mathbb{E}_s[\max_a Q(s,a)]$, causing underestimation. In other words, the non-optimal current policy in the AC framework can also contribute to underestimation.

\textbf{In summary}, Online RL training has a relatively high variance during training, thus the agent could not  monotonically increasing its performance, then the best actions from mixture of the policies may always inferior to those from the current policy. Even if having a monotonically improvement agent, the phenomena may also exist. To give more insight, the expectation $\mathbb{E}_s[\max_{a\sim\mu_k}Q^{\mu_k}(s,a)]$ could be interpreted as the $Q$-value of the optimal policy derived from the replay buffer, not merely any arbitrary mixture policy. This concept draws from offline RL, which shows that it is possible to extract a policy from the replay buffer $\mathcal{D}_k$ that is more optimal than all preceding data-collecting policies ${\pi_0, \ldots, \pi_{k-1}}$, and thus could potentially be superior to $\pi_k$

\subsection{How to mitigate underestimation issue?}\label{section:howtomitigateunderestiamationissue}
If the policy is the true optimizer under the current $Q$-value, it would mitigate the value underestimation issue. Yet it is practically infeasible within limited policy gradient updates. 
Therefore, the immediate solution is either to improve the policy slightly under the current Q-value or to find a better way to estimate the Q-value. We experimented with increasing the number of policy update steps but found that this approach did not yield satisfactory results.

\begin{wrapfigure}[15]{r}{0.35\textwidth}
    \centering
    \includegraphics[height=3.5cm,keepaspectratio]{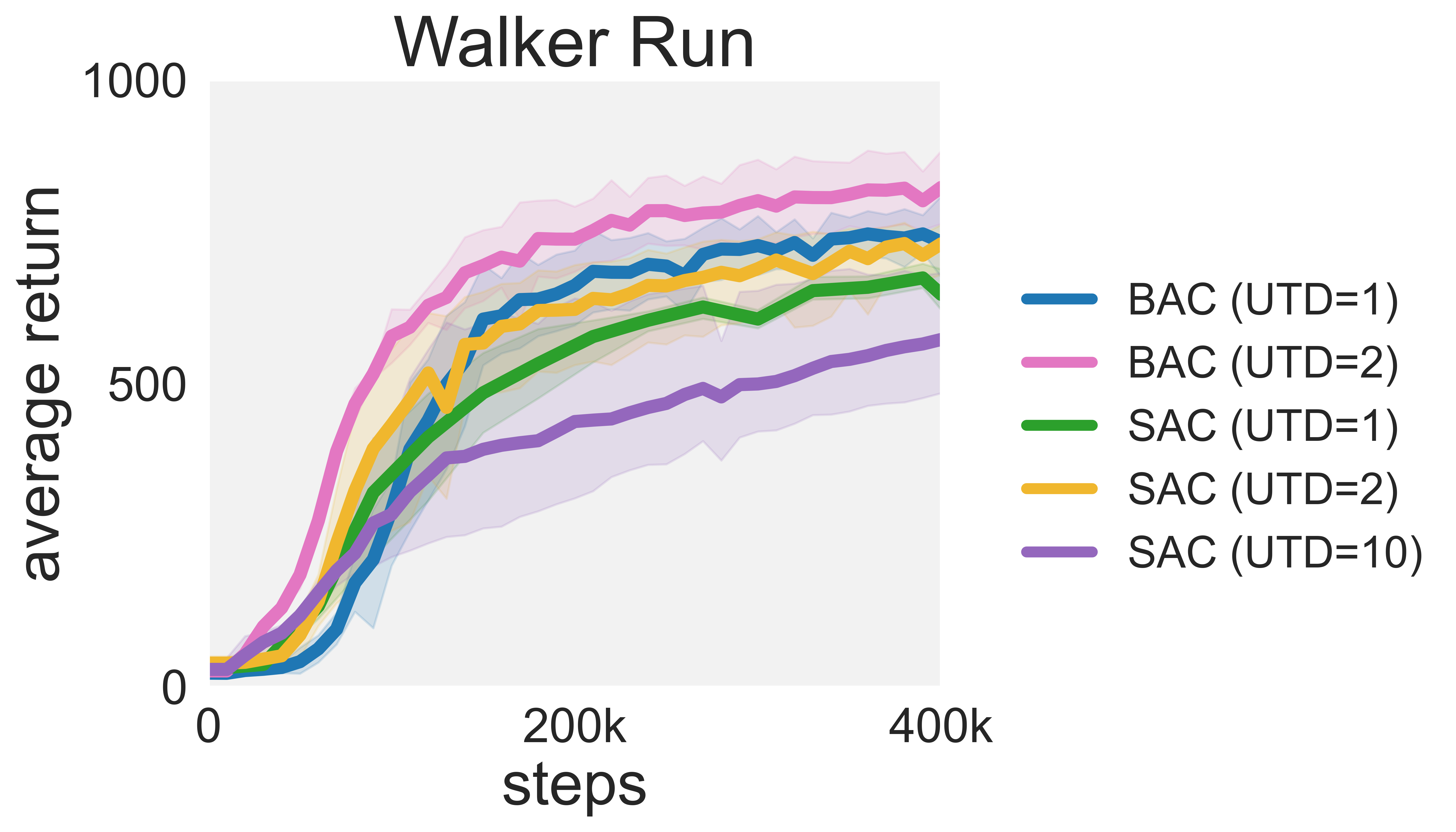} 
    \caption{\textbf{Update the policy with more steps.} While a moderate increase could enhance the performance of both SAC and BAC, excessively amplifying the number of updates might impair performance, primarily due to a diminished neural network capacity to learn and generalize}
    \label{app-fig:moreutd}
\end{wrapfigure}
\paragraph{Updating the policy with more update steps could not fully solve the issue.} 
We conducted an experiment that updated the policy with more steps, as shown in Figure~\ref{app-fig:moreutd}. While a moderate increase could enhance the performance of both SAC and BAC, excessively amplifying the number of updates might impair performance, primarily due to a diminished neural network's capacity to learn and generalize, as discussed in~\citet{d2022sample}. Moreover, updating the policy with a bit more steps might mitigate the value underestimation, yet it would lead to a prolonged training style, potentially leading to a significantly higher computational load.

\paragraph{Leveraging the more optimal actions in the replay buffer may help much.}
Many Actor-Critic algorithms commonly encounter the circumstance: the actions sampled from the current policy $\pi$ fall short of the optimal ones stored in the replay buffer $\mathcal{D}$.
The existence of more optimal actions in the replay buffer than generated by the current policy further supports the actual gap between the current policy and the ideal optimal one.

We identify that underestimation particularly occurs in the latter training stage, where we see a notable shortfall in the exploitation of the more optimal actions in the replay buffer, that is why we term it as \textbf{\textit{\textcolor{myblue}{under-exploitation}}}. Thus, exploiting the more optimal actions in the replay buffer to bootstrap $Q$ would shorten the gap to the optima, hence mitigating underestimation.

\subsection{The existence of ``under-exploitation'' stage}\label{section:existence}

In our main paper, we propose $\Delta(\mu, \pi)$ to quantify the existence of the more optimal actions in the replay buffer than those generated by the current policy. Here, we will provide more empirical evidence to show its existence.

\begin{figure}[ht]
\centering
 \includegraphics[width=0.95\linewidth]{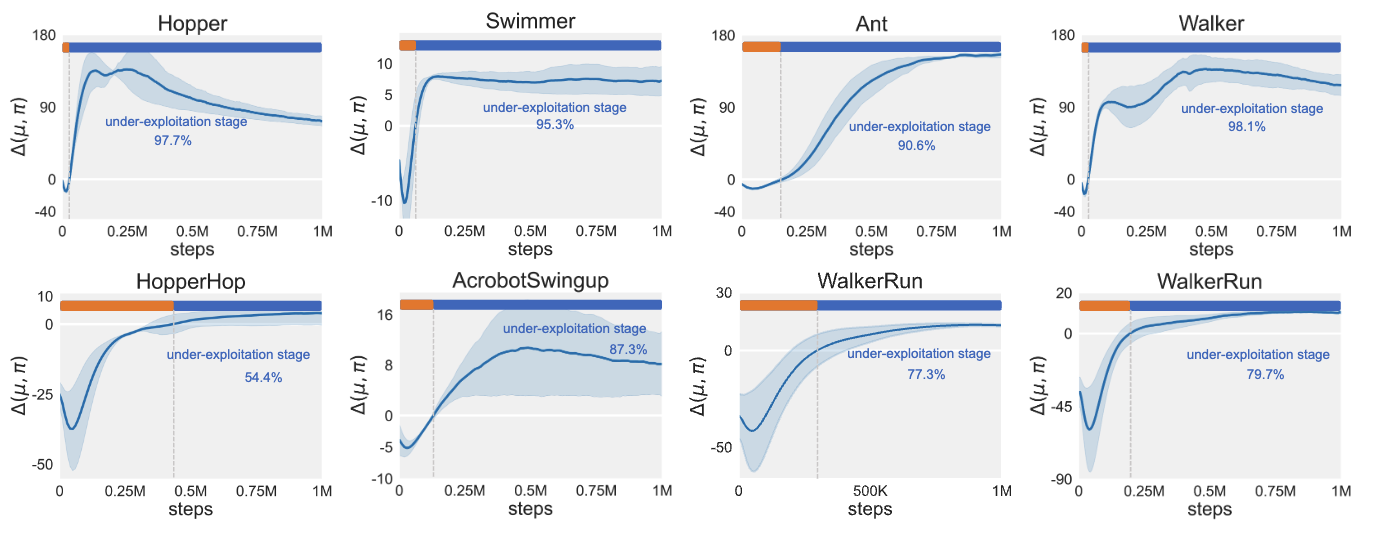}
    \caption{Visualization on under-exploitation stage on eight environments across MuJoCo and DMControl benchmark tasks, as analyzed through the lens of the SAC agent. }
    \label{fig:hard-exploitation-8}
\end{figure}

\paragraph{Existence in many scenarios.} Here we provide more results on the existence of under-exploitation stage, as shown in Figure~\ref{fig:hard-exploitation-8}, that in various scenarios, positive $\Delta(\mu_k,\pi_k)$ occupies a significantly larger portion than negative $\Delta(\mu_k,\pi_k)$, indicating that the common Bellman Exploration operator $\mathcal{T}_{explore}$ suffers from under-exploitation stages for a prolonged period of time. Our findings indicate that underutilization of the replay buffer is a common occurrence. This sheds light on the potential for significant improvements if the buffer is fully leveraged.

\begin{figure}[t]
    \centering
    \begin{minipage}[t]{0.35\textwidth}
    \centering
        \includegraphics[height=3.5cm,keepaspectratio]{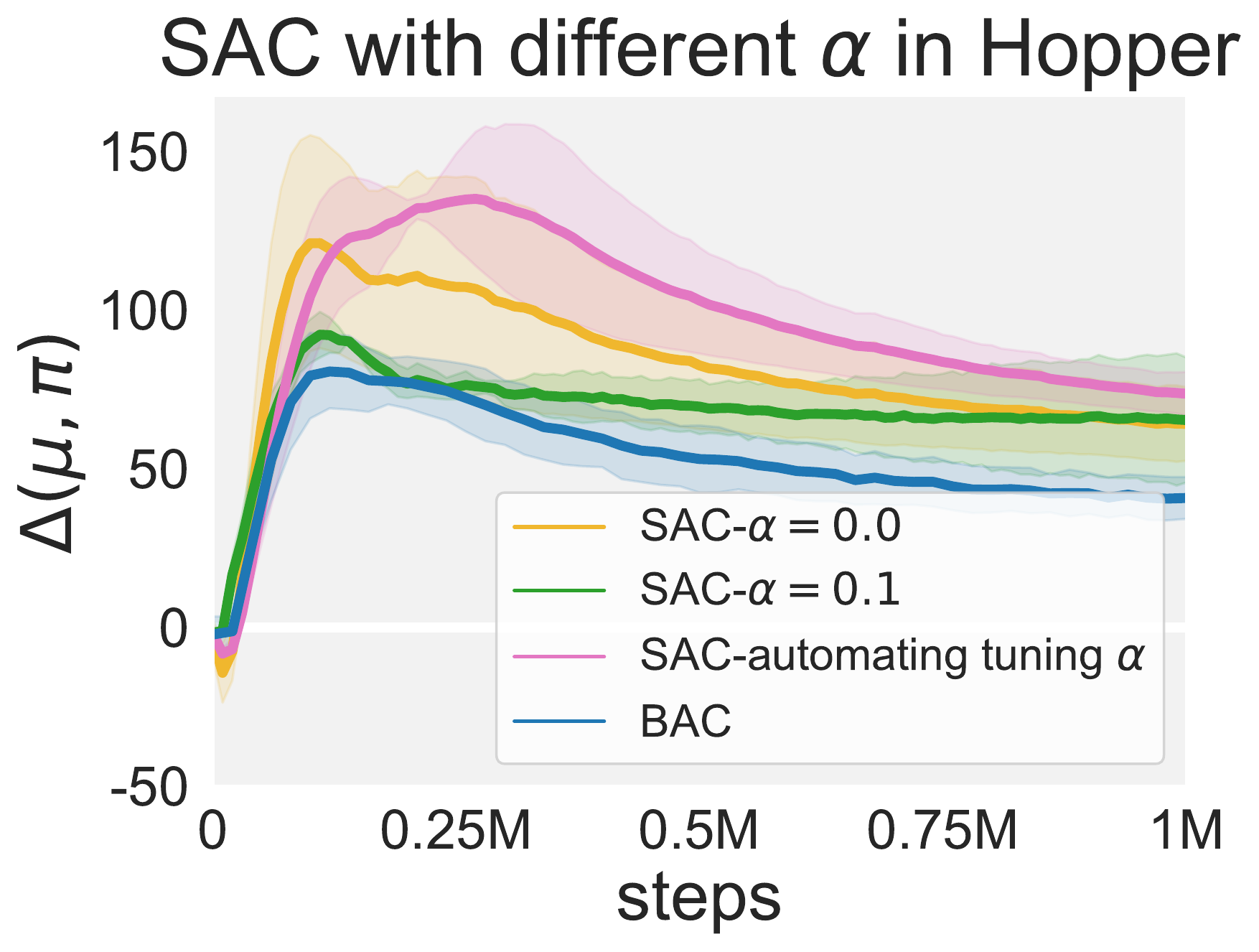}
        \caption{Under-exploitation occurs in SAC with different $\alpha$.} 
        \label{app-fig:sacwithdifferentalpha}
    \end{minipage}
    \hfill
    \begin{minipage}[t]{0.62\textwidth}
    \centering
        \includegraphics[height=3.5cm,keepaspectratio]{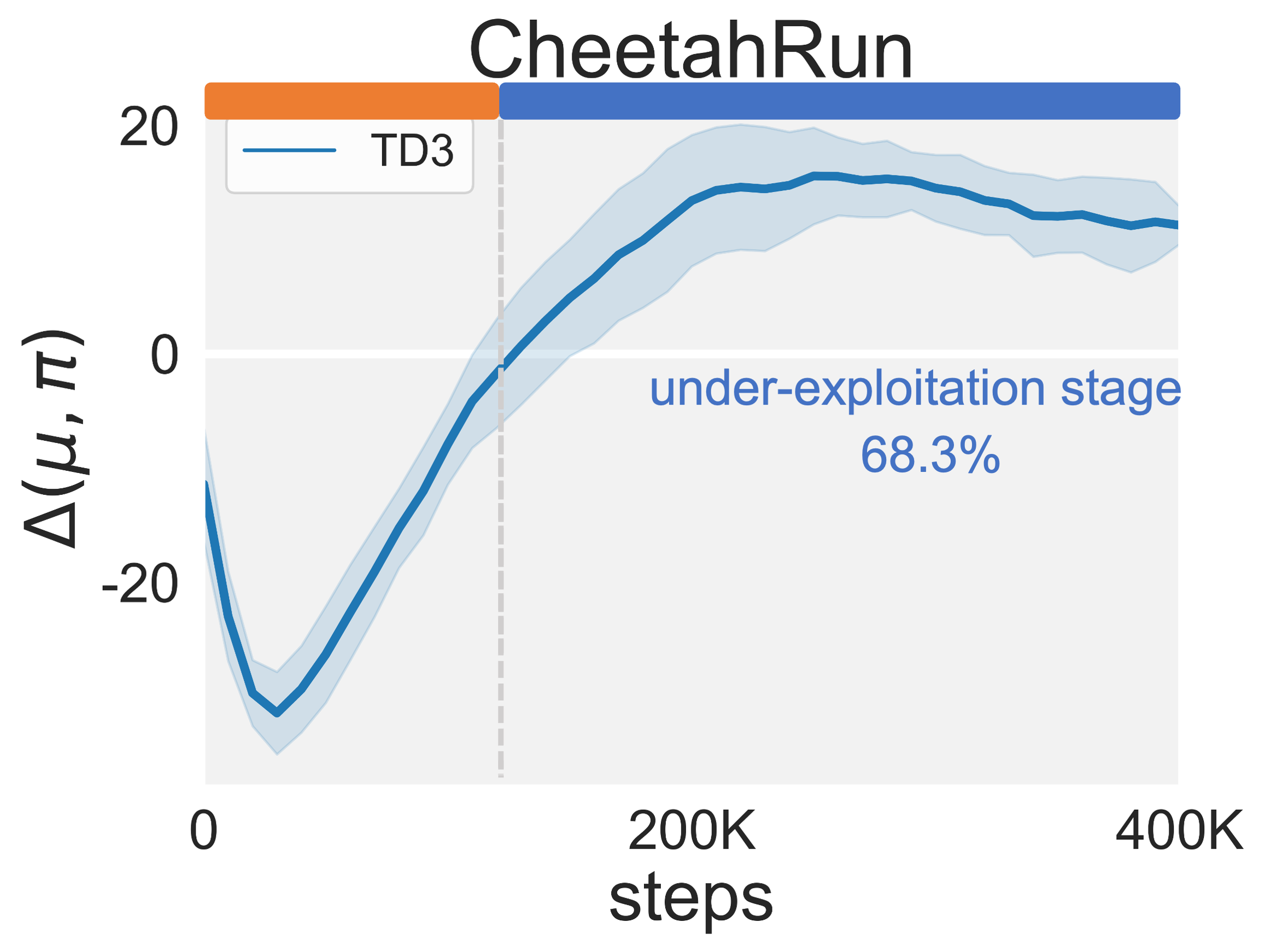}
        \includegraphics[height=3.5cm,keepaspectratio]{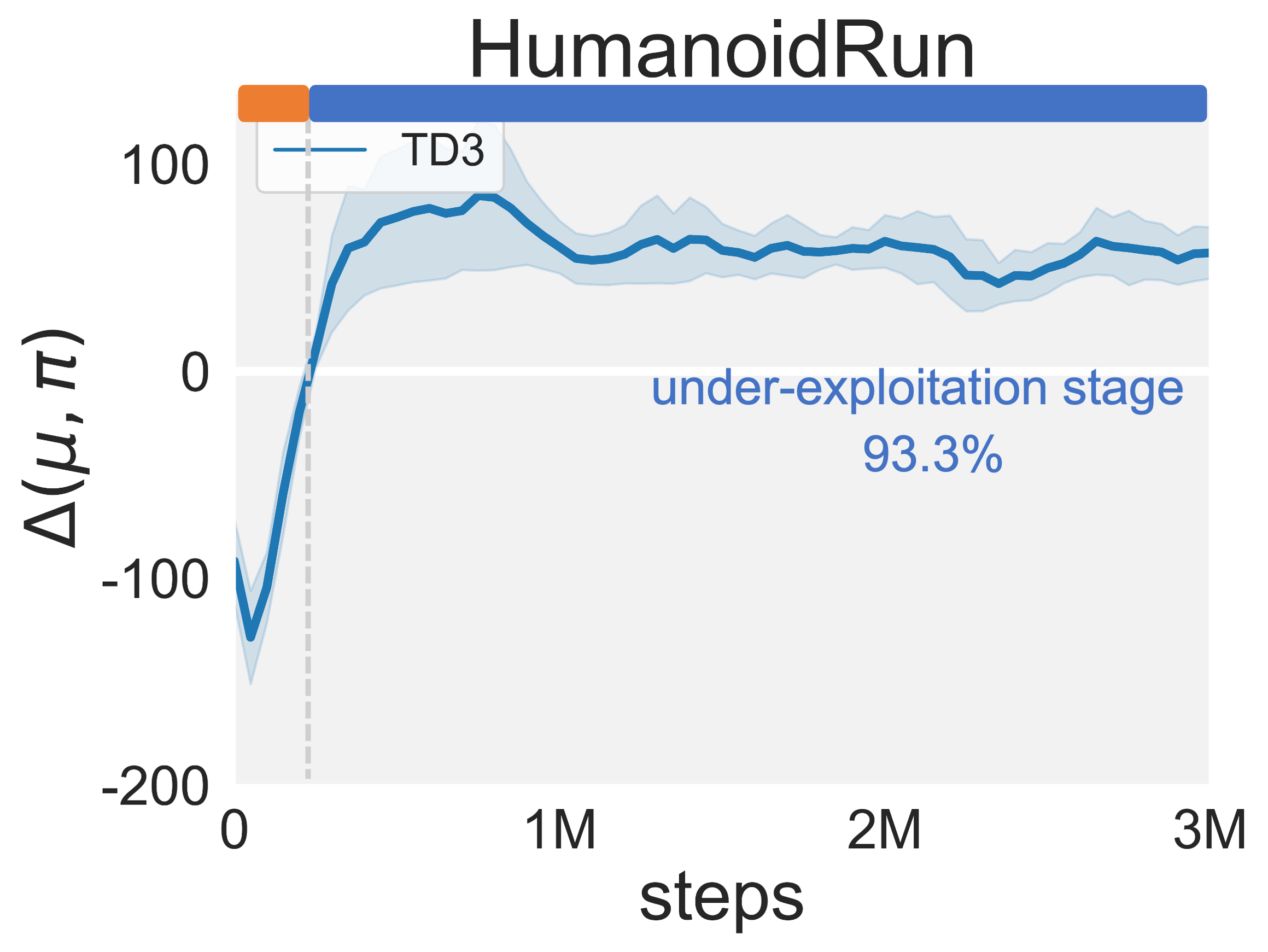}
        \caption{Visualization of $\Delta(\mu,\pi)$ on TD3 agent. Positive $\Delta(\mu,\pi)$ indicates the under-exploitation stage.}
        \label{app-fig:delta-TD3}
    \end{minipage}
\end{figure}

\paragraph{Existence without exploration bias.} 
The existence of better actions in the replay buffer stems not solely from the entropy term. It also attributes to the particulars of the optimization in AC, as obtaining an optimal policy w.r.t the current Q value is practically unattainable with a few policy gradient updates.  Many off-policy AC methods, relying solely on current policy for $Q$-value updates, may face under-exploitation issues.

Figure~\ref{app-fig:sacwithdifferentalpha} illustrates the under-exploitation that occurs in SAC with varying $\alpha$. Notably, under-exploitation is observed even in SAC instances with $\alpha=0$, indicating the presence of under-exploitation even when there is no exploration bias. BAC mitigates under-exploitation pitfalls more, even equipped with an exploration term, when compared to the SAC instance with $\alpha=0$. 

\begin{figure}[ht]
\centering
\includegraphics[height=4cm,keepaspectratio]{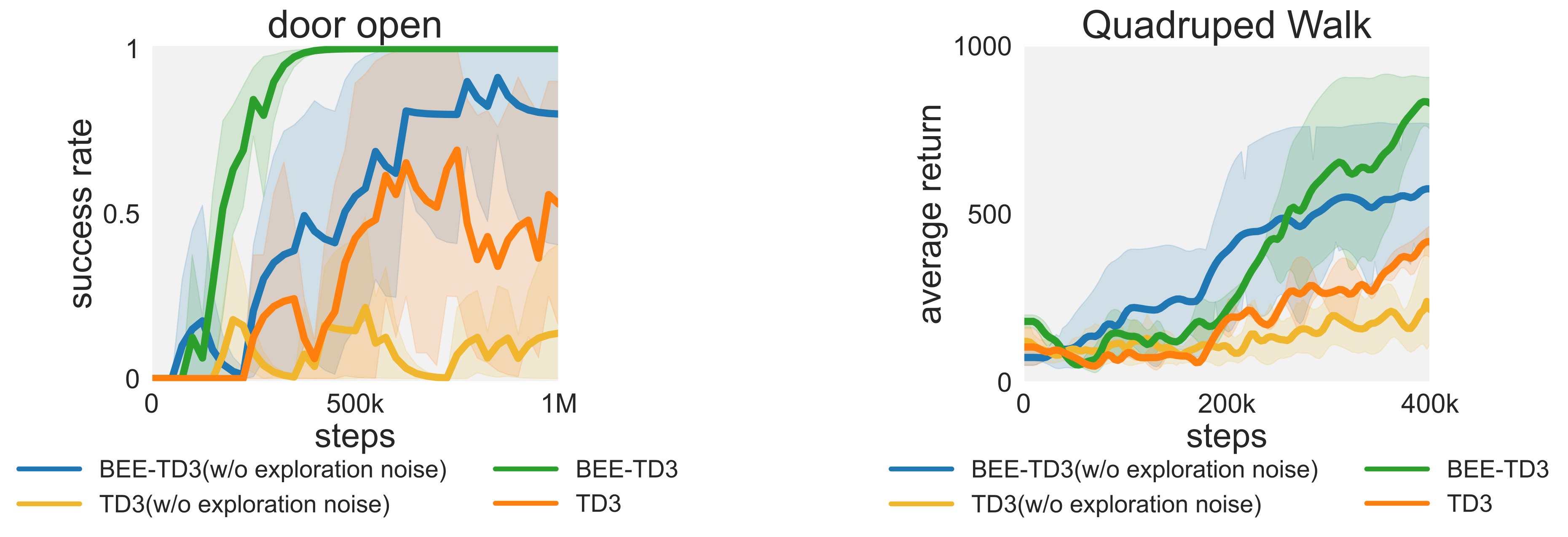}
    \caption{\textbf{Deterministic policy with/without exploration noise.} Setting exploration noise to zero would degrade performance as the algorithms would fall short in exploration. However, integrating the BEE operator with TD3 consistently improves performance by mitigating underestimation and enhancing replay buffer exploitation, regardless of exploration noise levels. }
    \label{fig:BEETD3withzeronoise}
\end{figure}

Further, we conduct experiments by applying the BEE operator to TD3 with the exploration noise setting to zero, as shown in Figure~\ref{fig:BEETD3withzeronoise}. Actually, setting exploration noise to zero would degrade performance as the algorithms would fall short in exploration. However, integrating BEE with TD3 will boost the backbone algorithm performance regardless of the exploration noise. 

\paragraph{Existence in various off-policy algorithms.} Under-exploitation exists in many off-policy algorithms, not limited to SAC. Figure~\ref{app-fig:delta-TD3} shows that TD3 also encounters under-exploitation stages during training.

\subsection{Explanations on the existence of under-exploitation circumstance}

\paragraph{Positive $\Delta(\mu,\pi)$ during later training stages.} 
From the visualization figures above, we often observe a positive $\Delta(\mu,\pi)$ during later training stages, indicating that the initial under-exploration stage is often followed by a subsequent under-exploitation stage. To give more insights,
\begin{itemize}[leftmargin=16pt]
    \item In the early training stages, the policy $\pi$ performs poorly and possibly more randomly, resulting in 1) low-reward samples in the replay buffer with corresponding low $Q$ values; 2) the exploration bonus improves the expected $Q$-value of the current policy. 
    \item As training progresses, and the agent begins to solve the task, better actions than those generated by $\pi$ may appear in the replay buffer. It is partially attributed to the iterative update nature of the Actor-Critic (AC) framework as discussed in Appendix~\ref{section:culprits}, which may lead to the existence of inferior actions after policy updates compared to the optimal ones in the replay buffer.
\end{itemize}

\paragraph{Possible causes for under-exploitation circumstance.}\label{section:culprits} Several factors contribute to this circumstance:

\begin{itemize}[leftmargin=16pt]
    \item \textbf{Exploration bias}: Exploration bias often leads to the overestimation of Q-values, promoting policy exploration of suboptimal actions.

  \item\textbf{AC framework nature}: Consideration of the iterative update nature of the Actor-Critic (AC) framework also brings two additional dimensions into play: 
     \subitem \textit{$Q$-value estimation bias}: During the training process,  either underestimation or overestimation is inevitable. In other words, the true $Q$-value of the sampled actions from the current policy might be lower than some actions in the replay buffer.
     \subitem  \textit{Suboptimal policy update}: Ideally, each new policy should be the maximizer of the current $Q$ to ensure policy improvement. However, obtaining such an optimal policy w.r.t the current $Q$ function is practically unattainable with a few policy gradient updates. 
\end{itemize}

\clearpage
\section{Superior $Q$-value Estimation using BEE Operator}\label{section:superiorQ}
\label{section:interpertation}
While being intuitively reasonable, BEE's potential benefits require further verification. 
In the following, we show that the BEE operator would facilitate the estimation of $Q$ and thus improve sample efficiency compared to the commonly used Bellman evaluation operator.

\noindent\textbf{BEE mitigates the under-exploitation pitfalls.}\quad 
The prevalent positive $\Delta(\mu,\pi)$ exposes the limitations of the Bellman Exploration operator $\mathcal{T}_{explore}$. 
The BEE operator alleviates the over-reliance on the current policy and mitigates the ``under-exploitation'' pitfalls by allowing the value of optimal actions in the replay buffer to be fully utilized in the $Q$-value update.
To be more specific, when the $\mathcal{T}_{explore}$ operator is stuck in underestimation, the BEE operator would output a higher $Q$-value, as shown by the inequality $Q_{\mathcal{B}}^{\{\mu_k,\pi_k\}}(s,a) \geq Q_{\mathcal{T}_{explore}}^{\pi_k}(s,a) + \lambda\gamma\Delta(\mu_k,\pi_k)$.
This agrees with the findings in 
Figure~\ref{fig:visualization}, the BEE operator exhibits lower underestimation bias and faster convergence of success rate, indicating its better sample efficiency.

\begin{figure}[h]
    \centering
    \begin{minipage}[t]{0.64\textwidth}
        \centering
        \includegraphics[height=3.2cm,keepaspectratio]{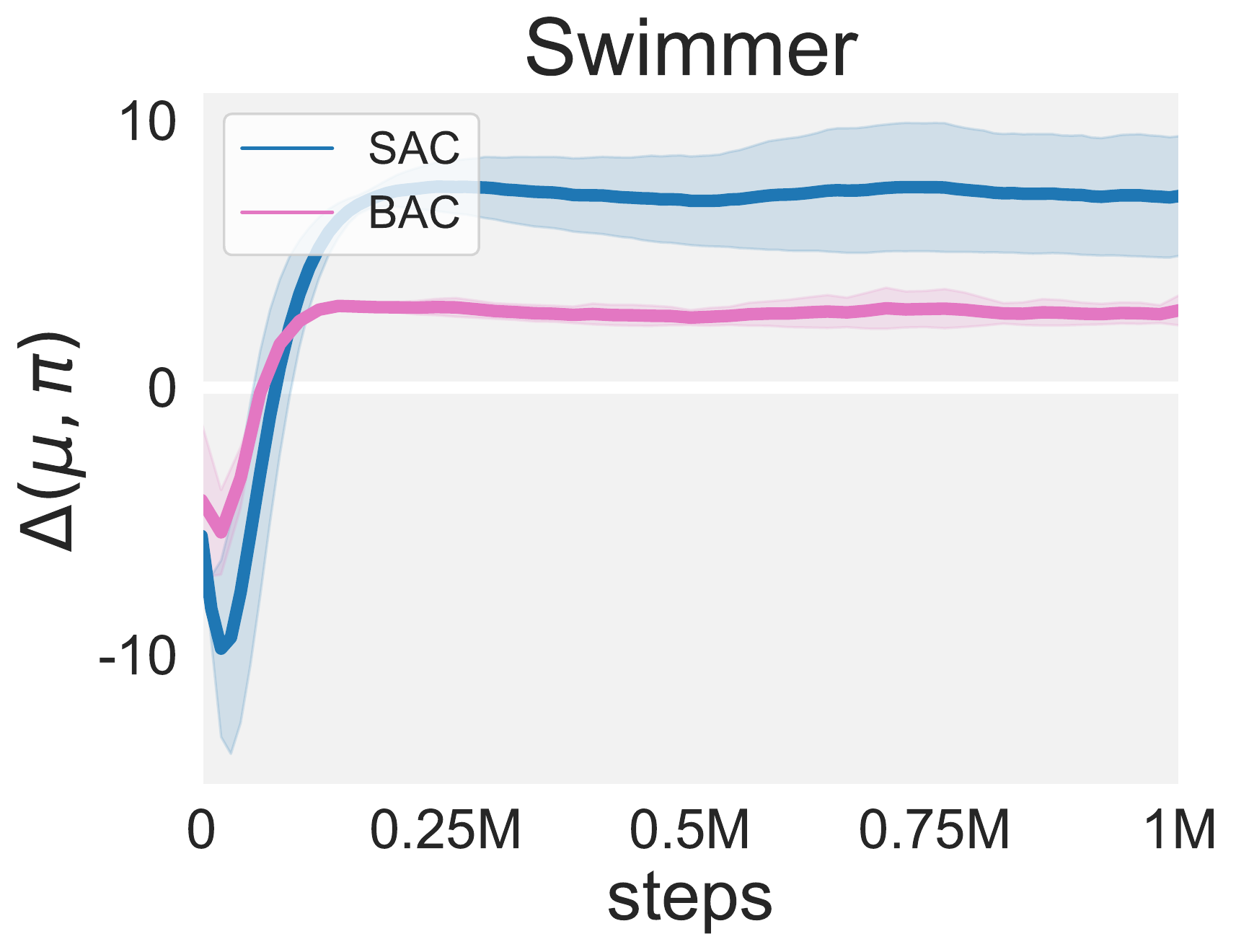} 
        \includegraphics[height=3.2cm,keepaspectratio]{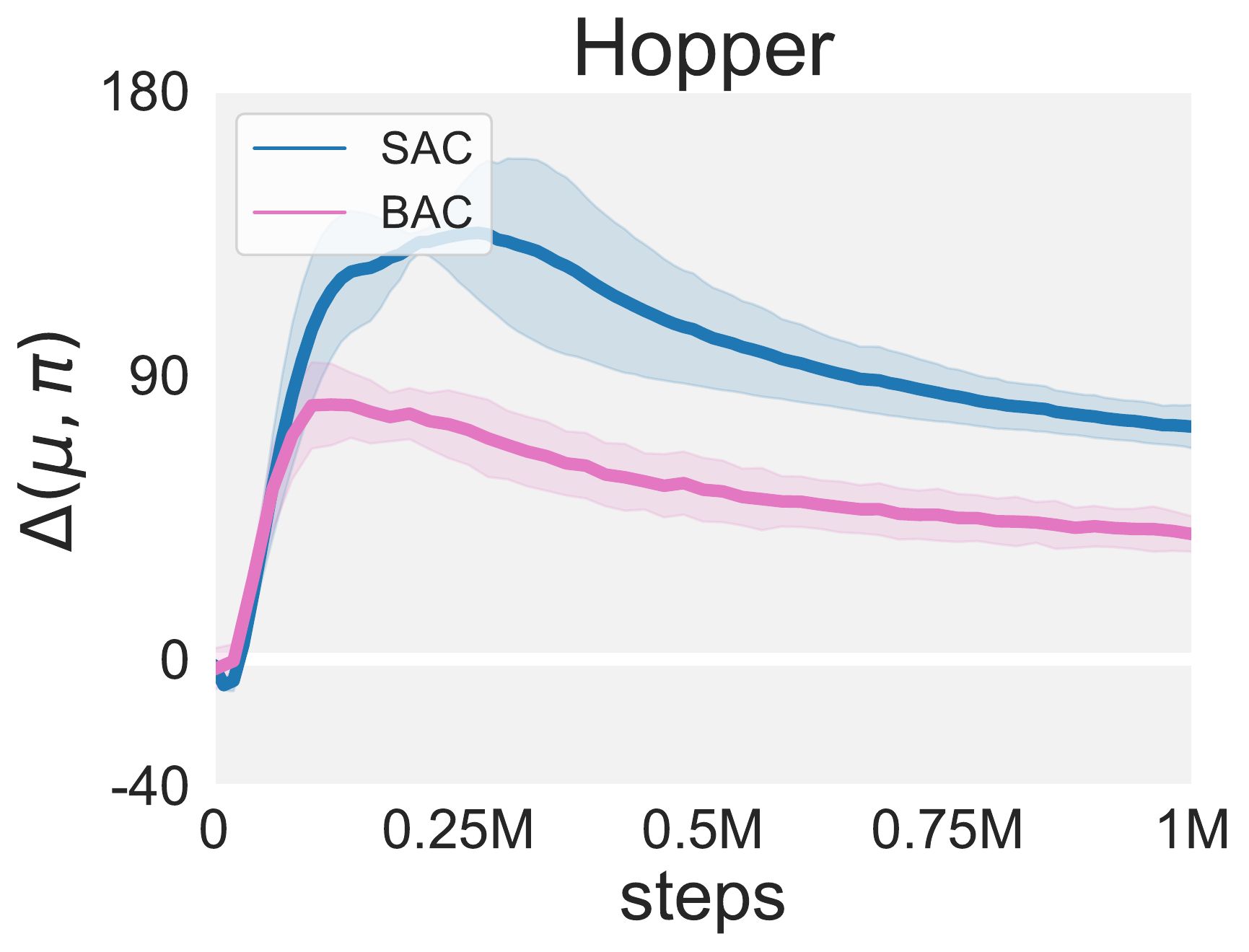} 
        \caption{Visualization of $\Delta(\mu, \pi)$ with SAC and BAC agent in Hopper and Swimmer tasks. Here, BAC improves the metric more towards 0 compared to SAC. } 
        \label{app-fig:delta-SAC-BAC}
    \end{minipage}
        \hfill
    \begin{minipage}[t]{0.32\textwidth}
        \centering
        \includegraphics[height=3.2cm,keepaspectratio]{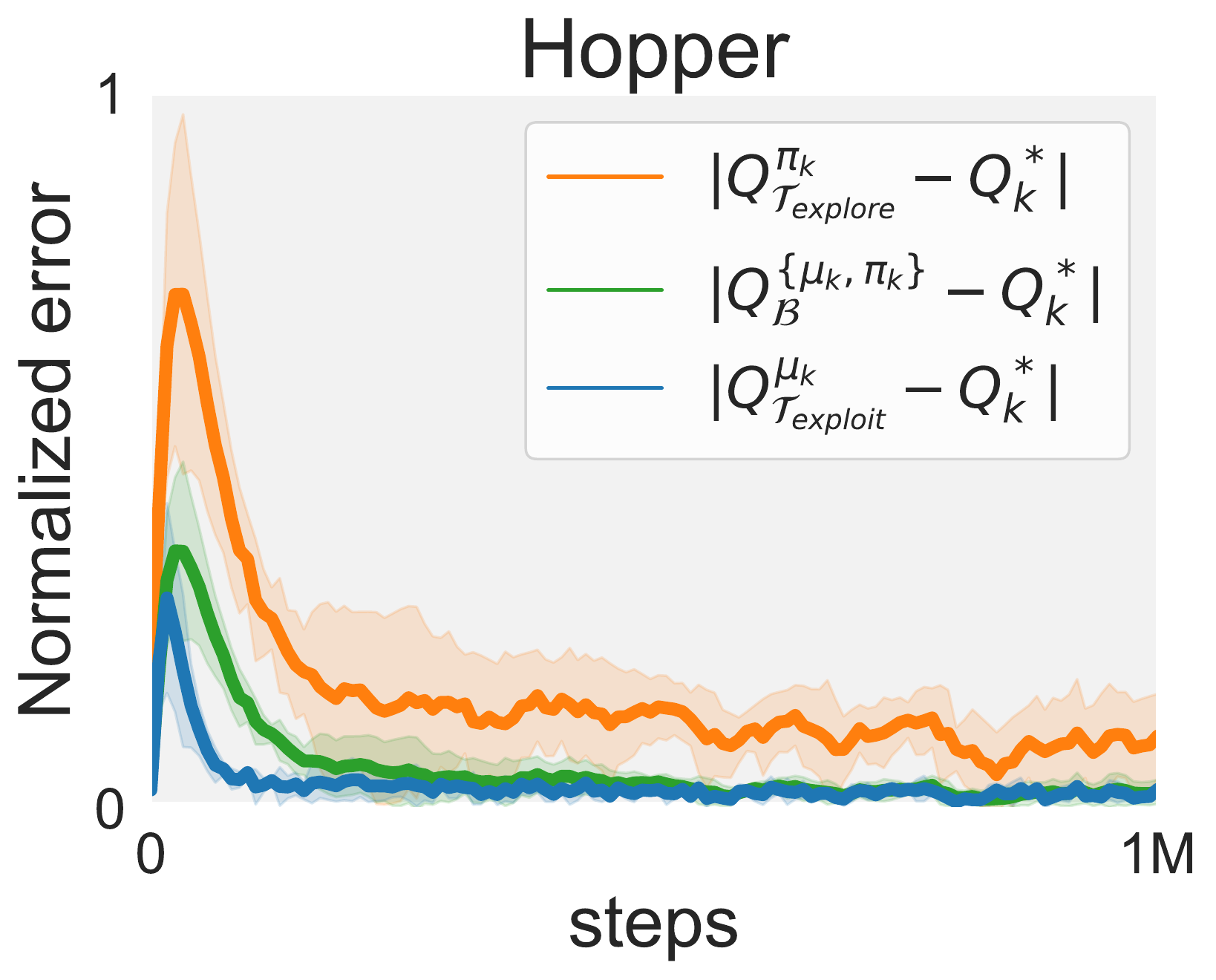} 
        \caption{$Q$-value estimation error of different operators.}
        \label{app-fig:estimationerrorofoperators}
    \end{minipage}    
\end{figure}

To further illustrate this, we visualize the $\Delta(\mu,\pi)$ metric for both SAC and BAC agents in Hopper and Swimmer environments. As shown in Figure~\ref{app-fig:delta-SAC-BAC}, BAC improves the metric $\Delta(\mu,\pi)$ more towards 0, indicating its ability to learn more accurate $Q$-values. BEE operator prevents a suboptimal current policy from diminishing the value of these actions. Thus, BAC has a higher likelihood of re-encountering these high-valued actions used for computing target $Q$-value, effectively mitigating under-exploitation pitfalls.

\begin{wrapfigure}[13]{r}{0.45\textwidth}
    \begin{minipage}{\linewidth}
            \centering           
            \includegraphics[height=3.0cm,keepaspectratio]{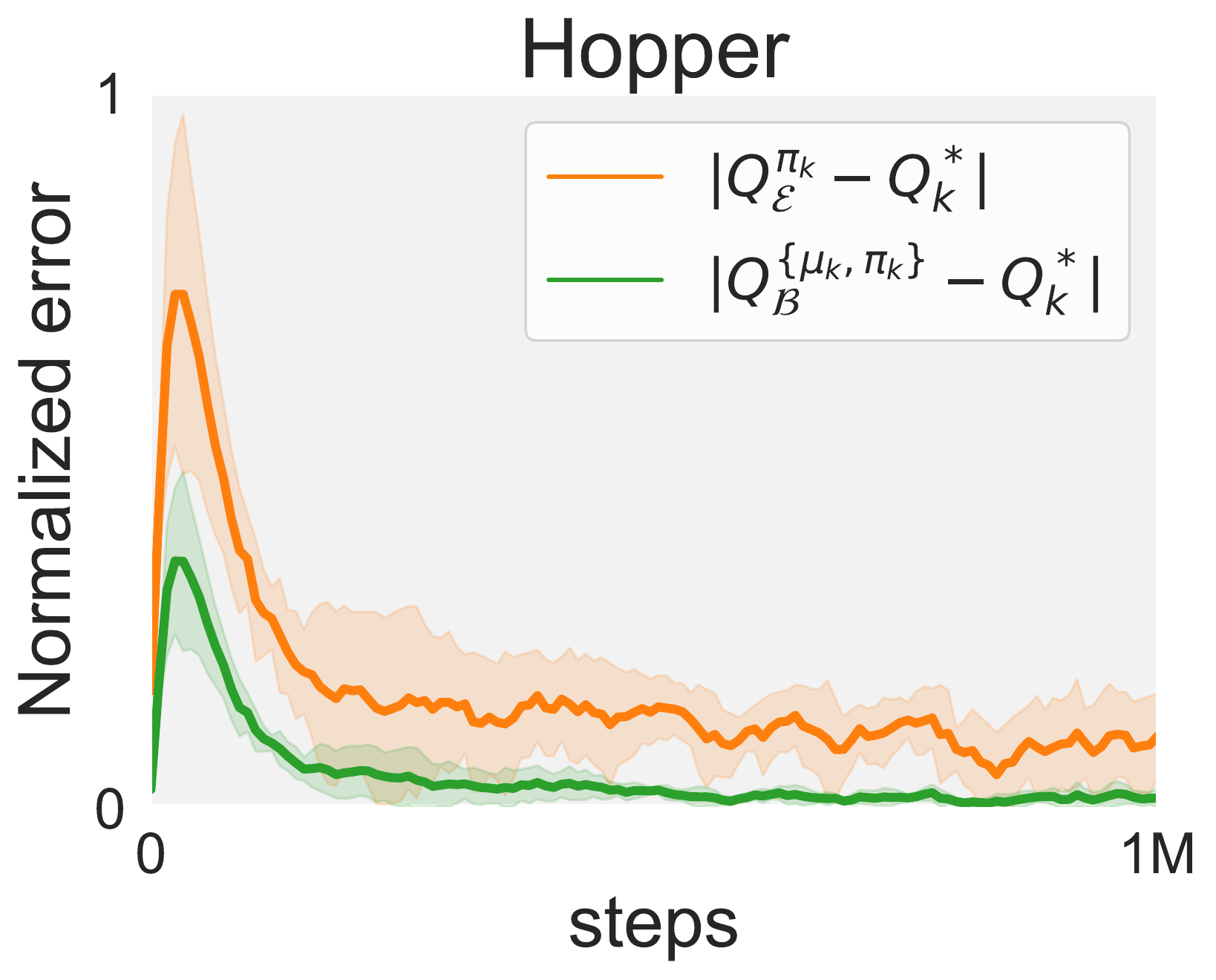}  
            \includegraphics[height=3.0cm,keepaspectratio]{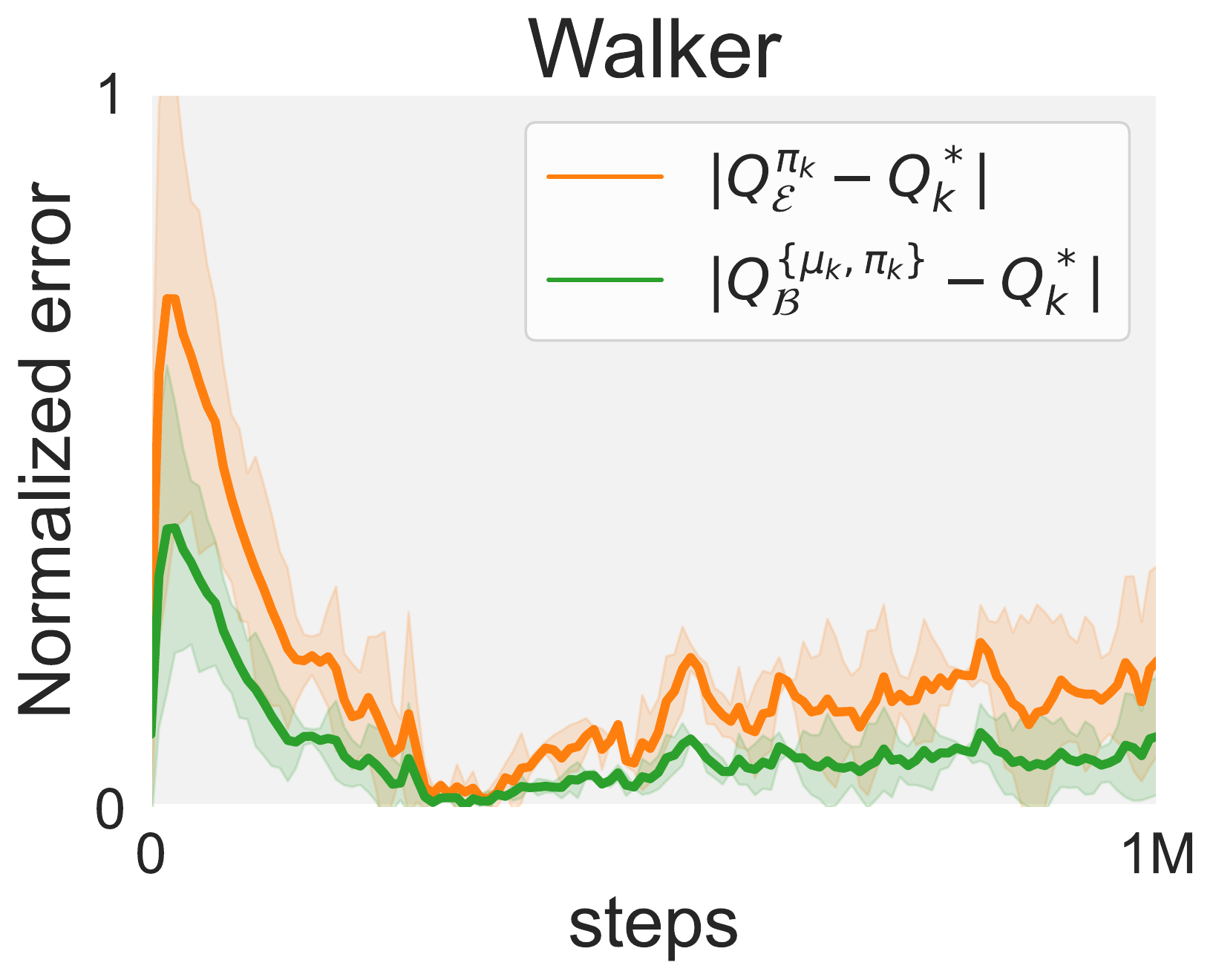}
            \vspace{-1mm}
            \caption{\small $Q$-value estimation error comparison. $\mathcal{T}_{explore}$ is referred to as $\mathcal{E}$ for brevity. And $Q^*_{k}$ is obtained practically with Monte-Carlo estimation.}
            \label{fig:estimation_bias}
    \end{minipage}
\end{wrapfigure}

\noindent\textbf{BEE exhibits no extra overestimation.}\quad
While the BEE operator seeks to alleviate underestimation, it does not incite additional overestimation. 
This is in contrast to prior techniques that excessively increase exploration bonuses or use optimistic estimation~\citep{brafman2002r,kim2019emi,pathak2019self}, which may
distort the $Q$-value estimates and potentially cause severe overestimation~\citep{oac}. The Bellman Exploitation operator, $\mathcal{T}_{exploit}$ does not introduce artificial bonus items and instead relies solely on the policy mixture induced by the replay buffer to calculate the maximum $Q$-value. Consequently, $\mathcal{T}_{exploit}$ is grounded in real experiences.

As illustrated in Figure~\ref{fig:estimation_bias}, the $Q$-value function induced by the BEE operator enjoys a lower level of overestimation and underestimation.  
Further, as empirically shown in Figure~\ref{fig:visualization} and~\ref{grid-world-visualization}, with enhanced exploitation, the BEE operator enables faster and more accurate $Q$-value learning, thereby reducing the chains of ineffective exploration on some inferior samples, and leading to improved sample efficiency. Moreover, we consider an extreme situation, $\lambda=1$. We plot Q-estimation-error under $\lambda=1$ in Figure~\ref{app-fig:estimationerrorofoperators}, and find that it does not cause overestimation.

Actually, $\mathcal{T}_{exploit}^{\mu}$, the reduced form  BEE operator when $\lambda=1$, relies on real experience and may lead to conservative estimation.
To give more insights, online learning's dynamic replay memory could be treated as a static dataset at a specific time step. Then, in practice,  the  Bellman exploitation operator $\mathcal{T}_{exploit}^{\mu}$  could be obtained by several effective techniques from offline RL. The pessimistic treatments in offline RL penalize overestimation heavily.  Thus a pure exploitation operator practically even might help to reduce overestimation.

\clearpage
\section{Effectiveness in Failure-prone Scenarios}\label{section:failure-prone}
In our main paper, we have shown the effectiveness of the BEE operator in terms of \textbf{ability to seize serendipity} and  \textbf{more stable $Q$-value in practice}. Here, we investigate the superiority of the BEE operator in terms of \textbf{ability to counteract failure}, \textbf{effectiveness in noisy environments}, and \textbf{effectiveness in sparse reward environments}.

\subsection{The ability to counteract failure} 
The BEE operator can not only grasp success but also \textbf{counteract failure}.  Here, we conduct some extreme experiments to show it. We simultaneously train SAC and BAC, and at 100k steps, both have reached a certain level of performance. This suggests that there already exists several high-value~(successful) samples in the replay buffer. At this point, we abruptly apply a massive perturbation to the policy and value networks~(\emph{i.e.}, \textcolor{myred}{at 100k steps, we substitute the current policy with a random one and reinitialize the value networks}). Keep other components the same, we continue the training. This setup is a magnification of a situation often seen in failure-prone scenarios: the agent is prone to performance drop, which consequently disrupts the $Q$ value estimate and necessitates additional sampling for recovery, thus forming a stark gap in the learning curve.  

\begin{figure}[h!]
    \centering
    \includegraphics[height=3.5cm,keepaspectratio]{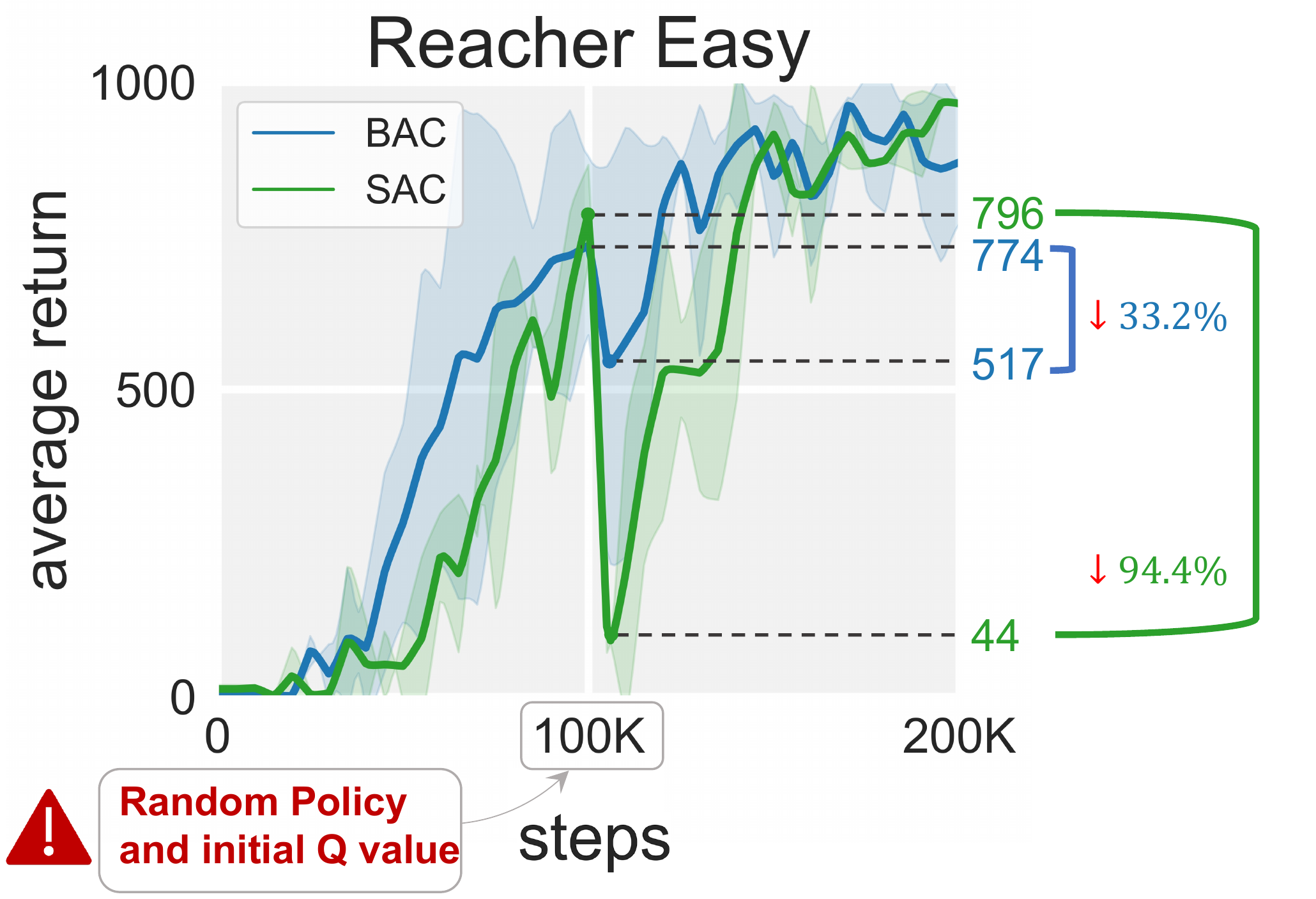}
    \hspace{5pt}
    \includegraphics[height=3.5cm,keepaspectratio]{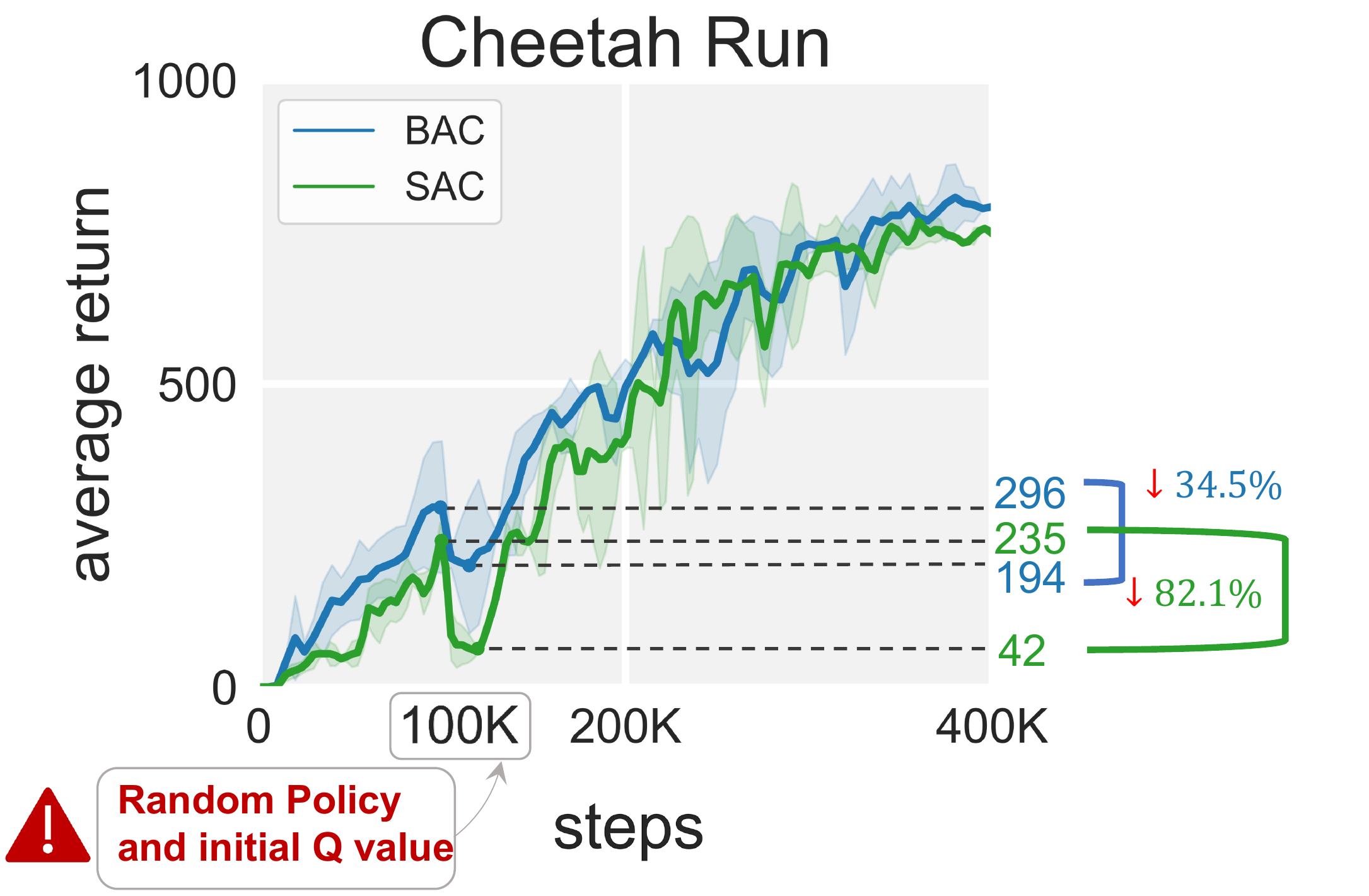}
    \caption{Comparison of the ability to counteract failure. \ourshort\ exhibits less performance drop (33.2\% in ReachEasy and 34.5\% in CheetahRun) and faster recovery. }
    \label{fig:counteredfailure}
\end{figure}

As shown in Figure~\ref{fig:counteredfailure}, we can observe that the degree of performance drop in \ourshort\ after 100k steps is significantly less than that in SAC, coupled with a faster recovery speed, demonstrating its better resilience against failure.
This capability possibly stems from the fact that the learned $Q$ value by the BEE operator is less influenced by the optimal level of the current policy.

\subsection{Effectiveness in noisy environments} 
We conduct experiments in noisy environments to investigate the robustness of the BEE operator. 
Noisy environments are created by adding Gaussian noise to the agent’s action at each step. Specifically, the environment executes $a' = a+ \rm{WN}(\sigma^2)$ as the action, where $a$ is the agent's original action, and $\rm{WN}(\sigma^2)$ is an unobserved Gaussian white noise with a standard deviation of $\sigma$.

\begin{figure}[h!]
    \centering
    \begin{subfigure}[t]{0.49\textwidth}
        \includegraphics[width=0.49\linewidth]{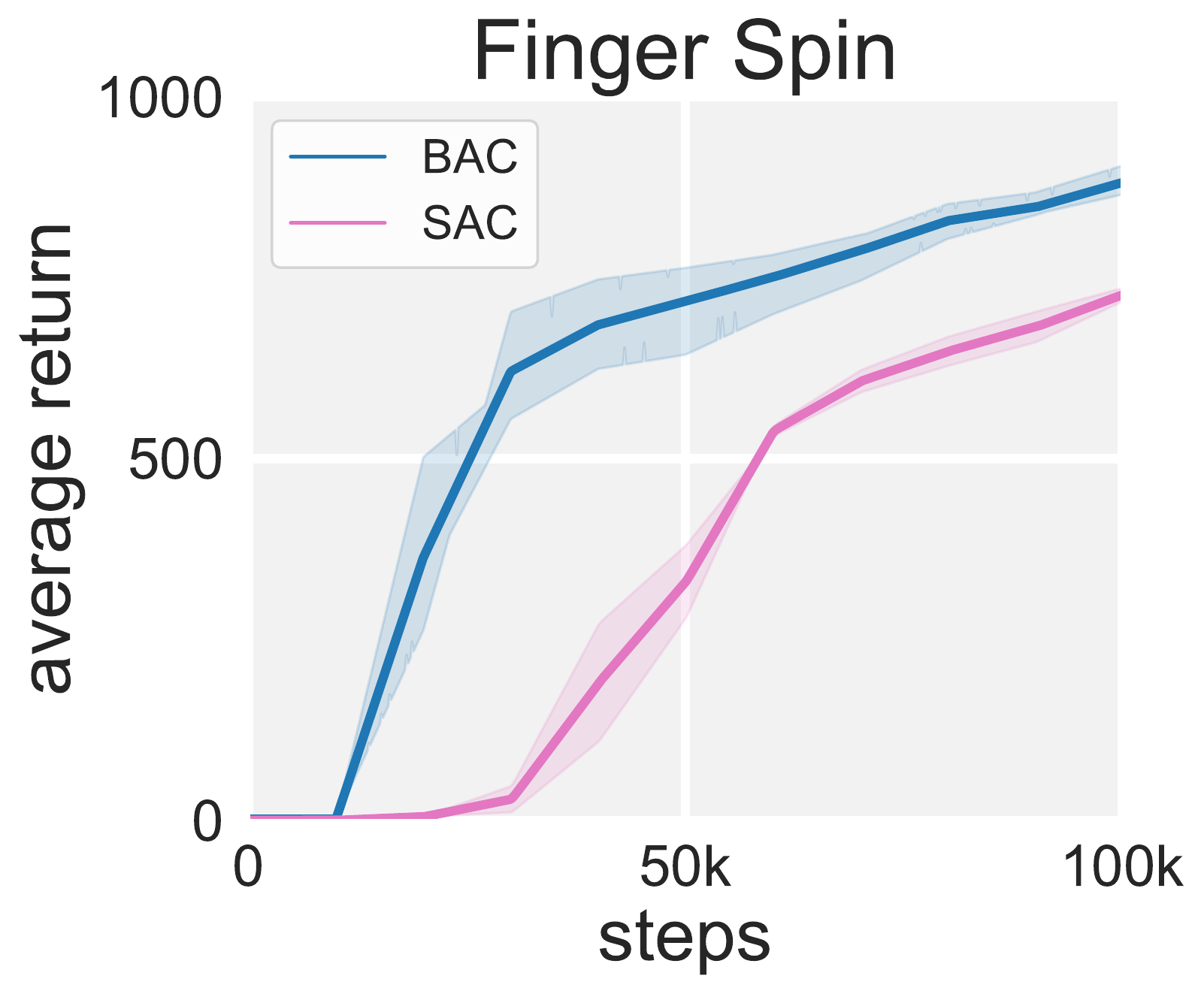}
        \includegraphics[width=0.49\linewidth]{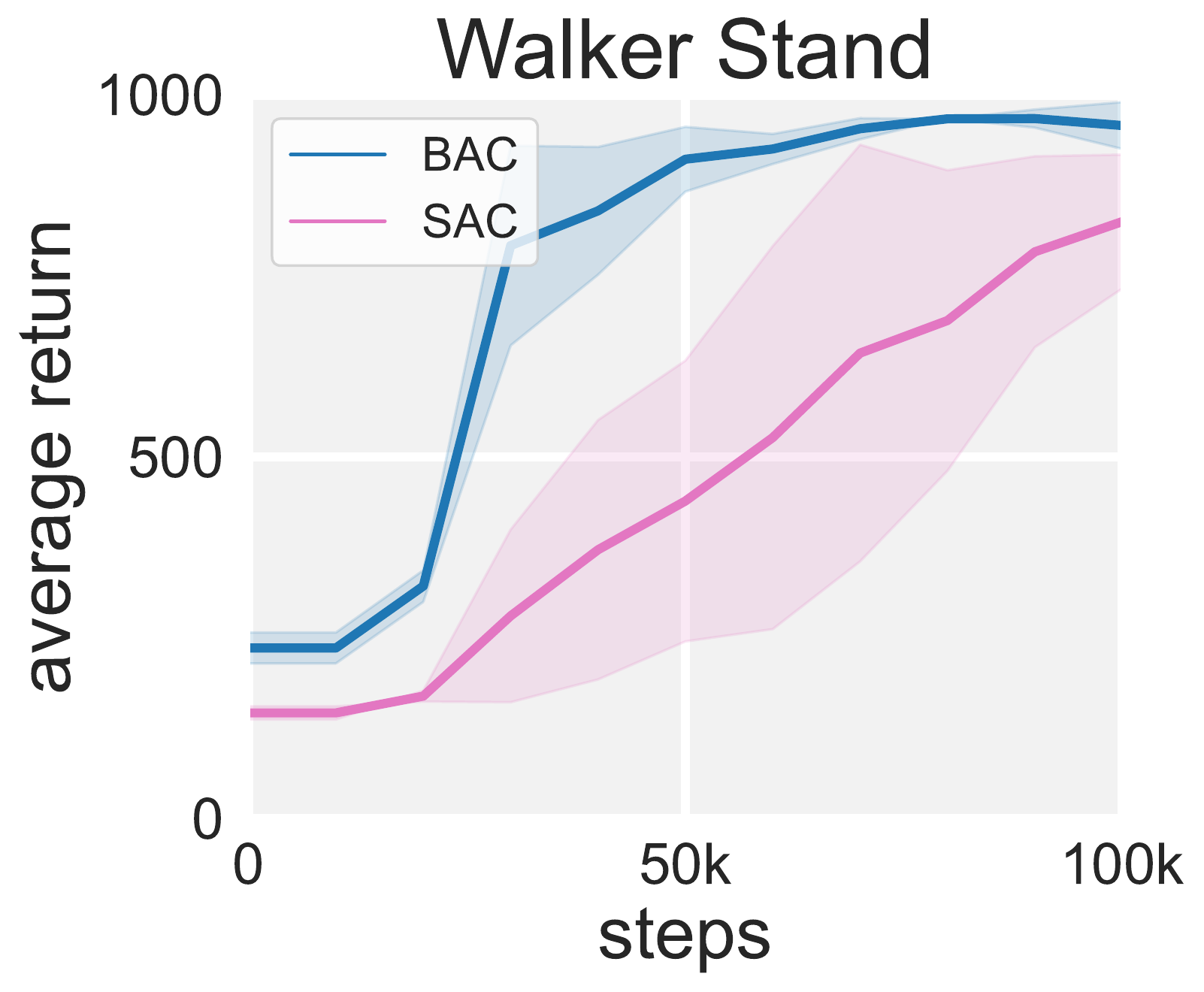}
        \caption{$\sigma=0.1$}
        \label{fig:noisy_1}
    \end{subfigure}
    \begin{subfigure}[t]{0.49\textwidth}
        \includegraphics[width=0.49\linewidth]{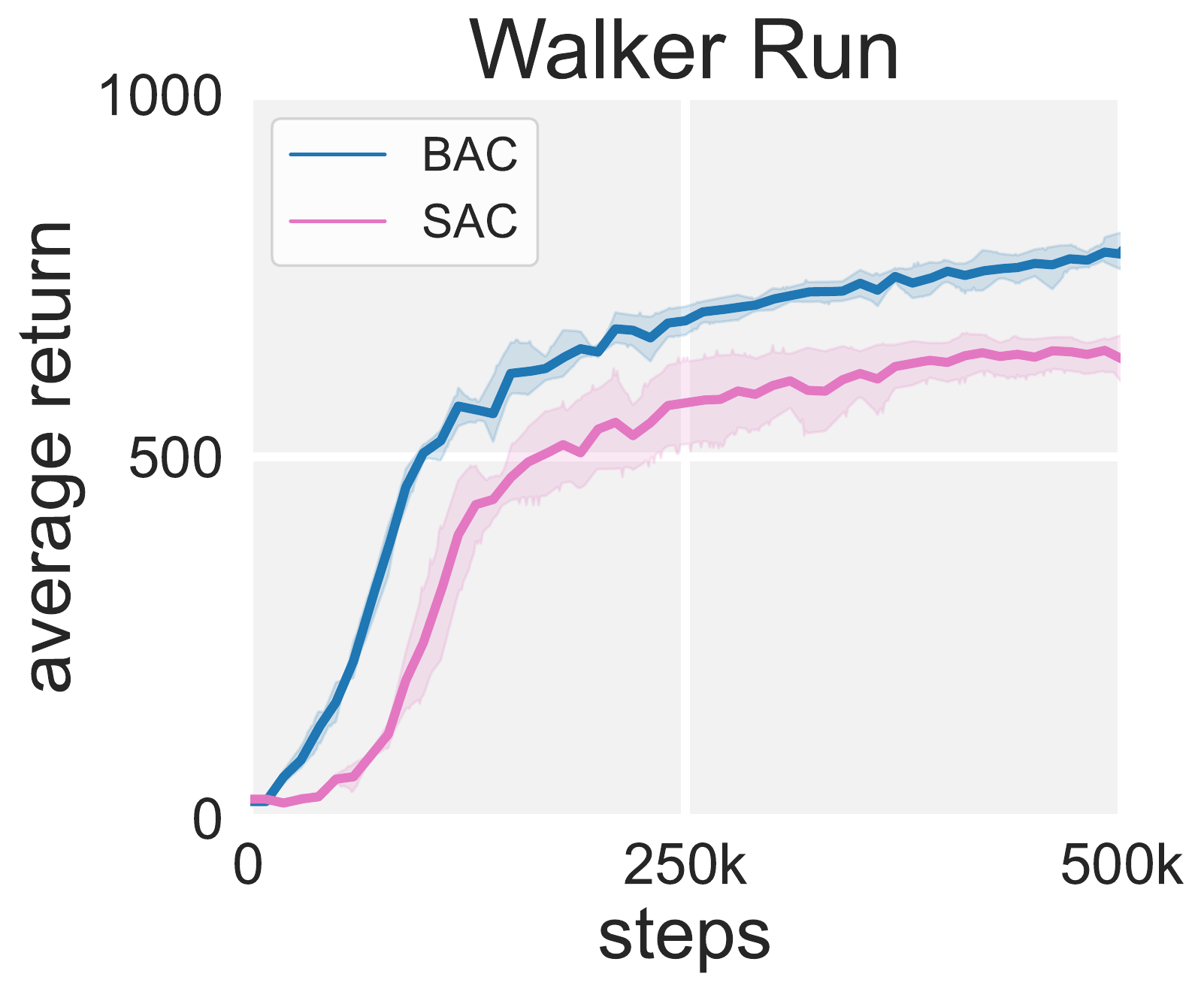}
        \includegraphics[width=0.49\linewidth]{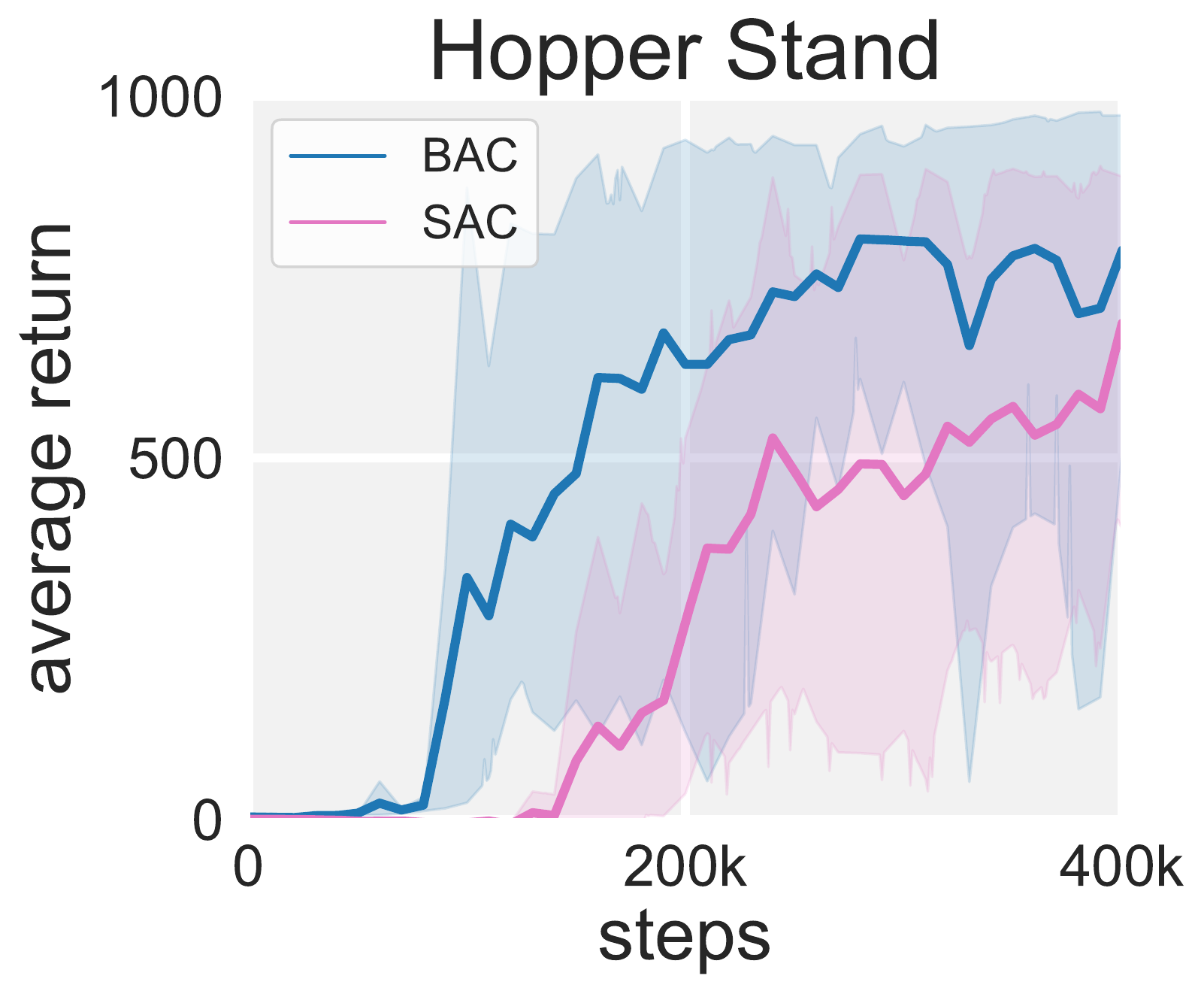}
        \caption{$\sigma=0.2$}
        \label{fig:noisy_2}
    \end{subfigure}
    \caption{Results in noisy environments: (a) in noisy FingerSpin and WalkerStand tasks with $\sigma=0.1$; (b) in noisy WalkerRun and HopperStand tasks with a server noise $\sigma=0.2$.}
    \label{fig:noisy}
\end{figure}

Despite the noisy settings that can destabilize $Q$ values and impede sample efficiency, as shown in Figure~\ref{fig:noisy}, \ourshort\ demonstrates desirable performance, even outperforming SAC more significantly than in noise-free environments.

\Revise{\subsection{\Revise{Illustrative example on the failure-prone scenario.}}}

\Revise{We provide a typical failure-prone scenario to illustrate the effectiveness of our operator.}

\begin{figure}[h!]
    \centering
    \centering
    \includegraphics[width=0.5\linewidth]{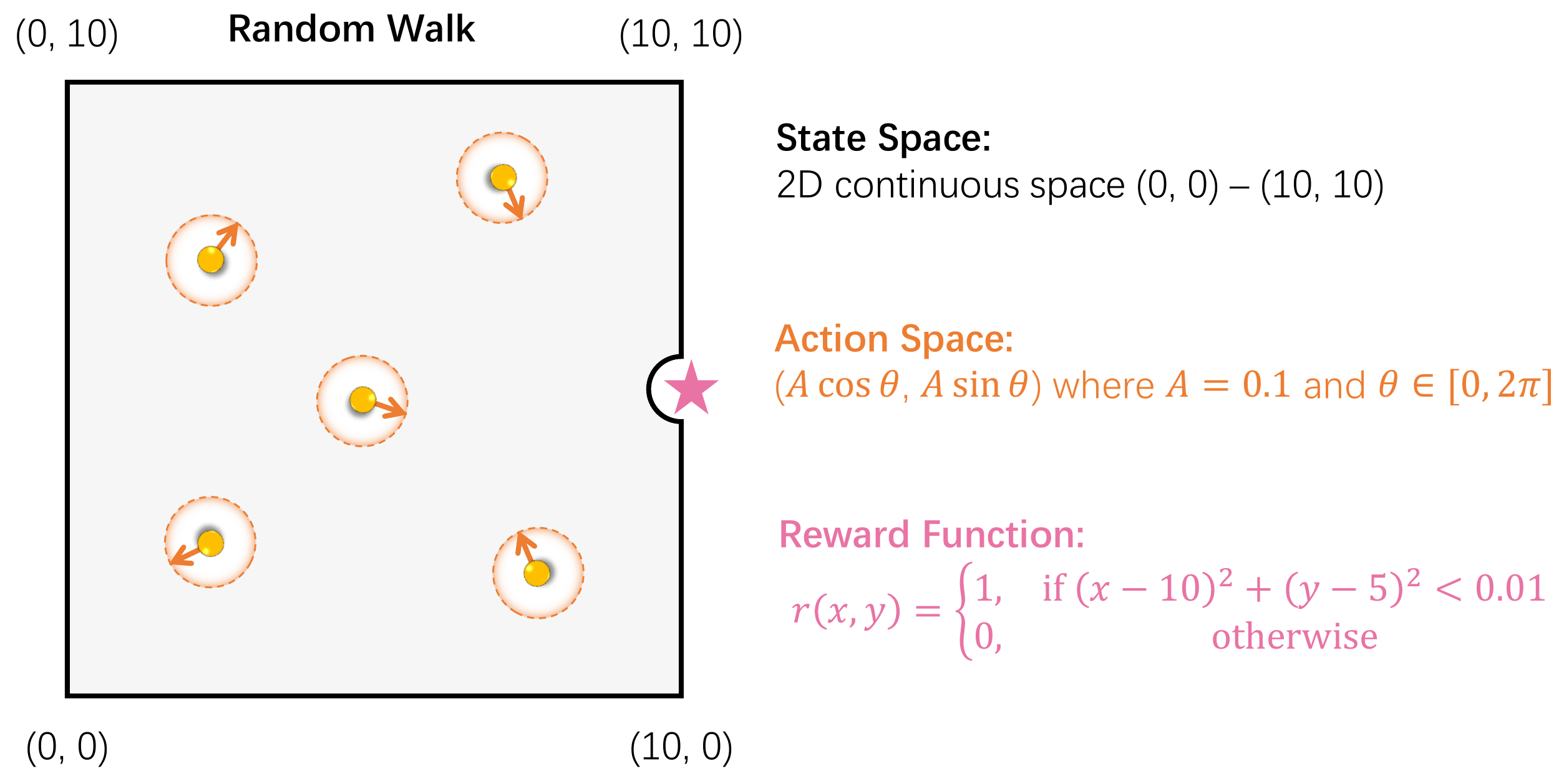}
    \caption{\Revise{We construct a failure-prone scenario: Random Walk. The yellow particle has to explore the 2D space, and the target is to reach the small hole around $(10, 5)$ (pink star).}}
    \label{fig:failure-prone-scenario}
\end{figure}

\Revise{\textbf{Task description.}}
\Revise{As shown in Figure \ref{fig:failure-prone-scenario}, a small particle spawns in a 2D continuous space of $[0, 10] \times [0, 10]$.
The particle could take any random moves inside the space with a length of 0.1.
The objective is to let the particle hit the small hole of radius 0.1 at $(10, 5)$.
In other words, the particle receives a non-zero reward if and only if it is in the hole.
Starting from a random policy, the particle has to explore the space and find the hole.}

\begin{figure}[h]
    \centering
    \includegraphics[width=0.9\linewidth]{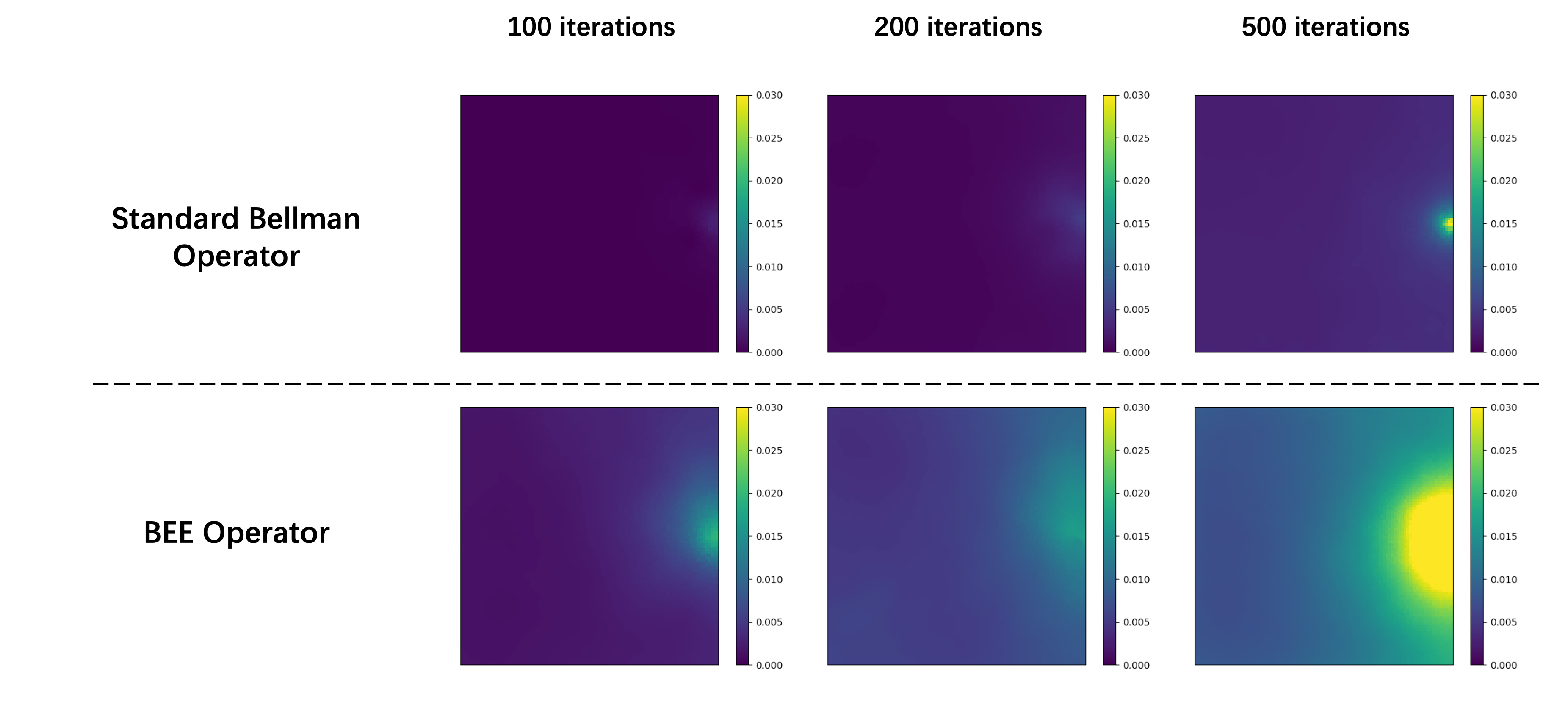}
    \caption{\Revise{$Q$-value heatmaps with standard Bellman operator and the BEE operator after 100, 200, and 500 iterations.}}
    \label{fig:failure-prone-scenario-iterations}
\end{figure}

\Revise{
\textbf{$Q$-value comparison.} Only 10 of 100000 samples have reached the hole in the replay buffer.
Figure \ref{fig:failure-prone-scenario-iterations} shows the $Q$-value heatmaps with the standard Bellman operator and our proposed BEE operator after 100, 200, and 500 $Q$-learning iterations. $Q$-values learned by the BEE operator are much closer to the expected ones in limited iterations.}

\Revise{
Let's dive deeper. Given $\bar{s}$  is one of the successor of a tuple $(s,a)$,  the target update  $r+ \gamma \mathbb{E}_{a’\sim \pi}Q(s',a')$ for Standard Bellman Operator, only focuses on actions $a'$ from the current policy $\pi$, ignoring a more optimal one $\bar{a}'$.  Thus   $Q(s,a)$ which should be valued higher is underestimated.  Then next policy derived from the current misleading $Q$ may prefer not to sample $(s,a)$ again as it does not have a high value.  Thus, algorithms based on the standard Bellman operator might take a substantially longer time to re-encounter these serendipities for training with decreased sample efficiency.
In contrast, the BEE operator extracts the best actions in the replay buffer to construct referenced value to the $Q$-value function thus mitigates such underestimation issue. 
}

\clearpage
\Revise{\subsection{\Revise{Effectiveness in sparse-reward tasks}}}
\Revise{We conduct experiments in sparse reward tasks to further demonstrate the generalizability of the BEE operator. 
We evaluate in both robot locomotion and manipulation tasks, based on the sparse reward version of benchmark tasks from \textbf{Meta-World}~\citep{yu2019meta}, \textbf{panda-gym}~\citep{panda-gym}, \textbf{ROBEL}~\citep{dkitty}. Here is the task description:}

\Revise{Meta-World manipulation tasks are based on a Sawyer robot with end-effector displacement control.
\begin{itemize}[leftmargin=16pt]
    \item coffee button: Push a button on the coffee machine whose position is randomized. 
    \item hand insert: Insert the gripper into a hole.
    \item door open: Open a door with a revolving joint. Randomize door positions.
\end{itemize}}

\begin{figure}[h!]
    \centering
    \begin{minipage}{0.8\textwidth}
        \begin{subfigure}[t]{1.0\textwidth}
            \begin{subfigure}[t]{0.32\textwidth}
                \includegraphics[height=3.5cm,keepaspectratio]{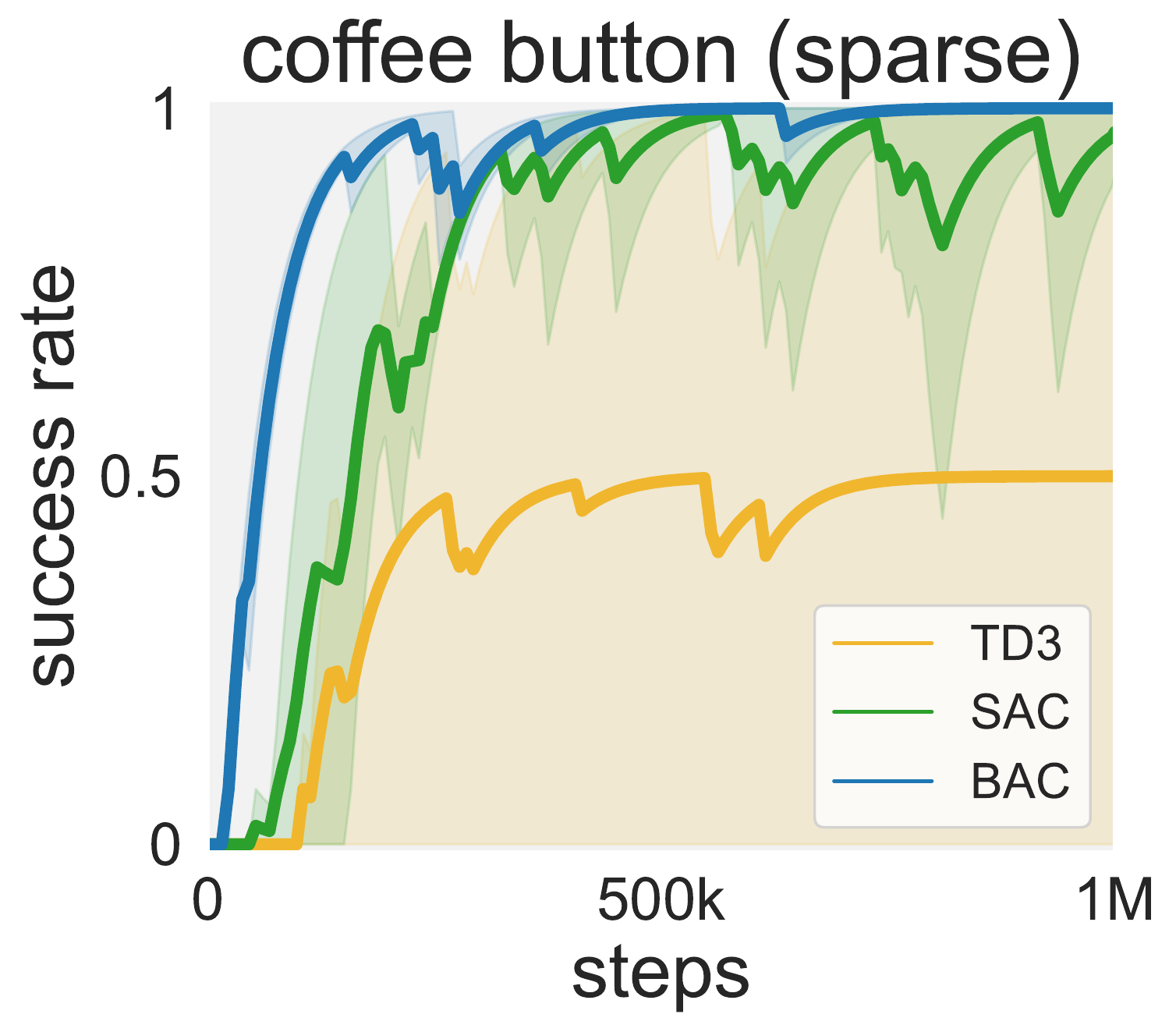}
            \end{subfigure}
            \hfill
            \begin{subfigure}[t]{0.64\textwidth}
                \includegraphics[height=3.5cm,keepaspectratio]{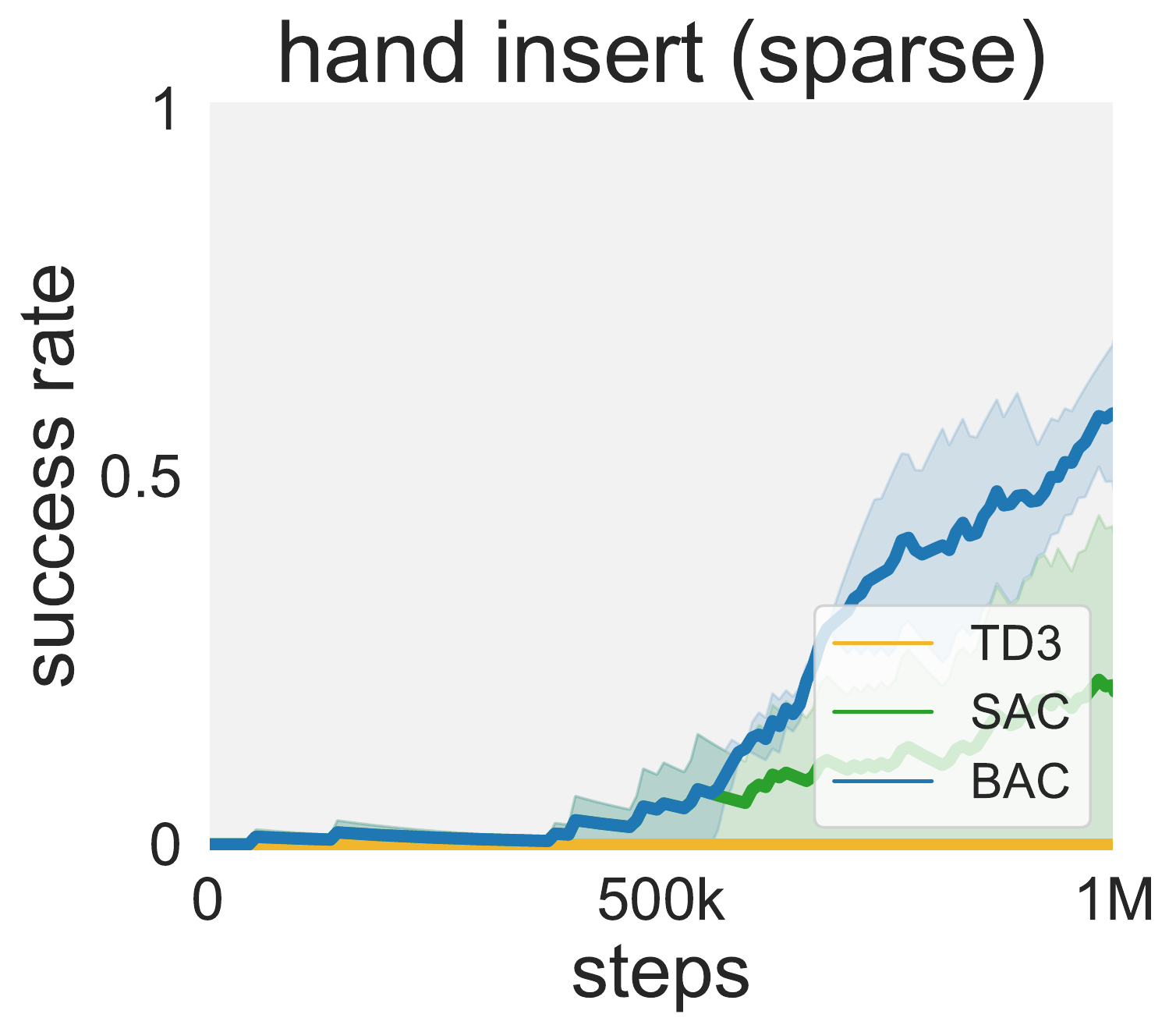}
                \includegraphics[height=3.5cm,keepaspectratio]{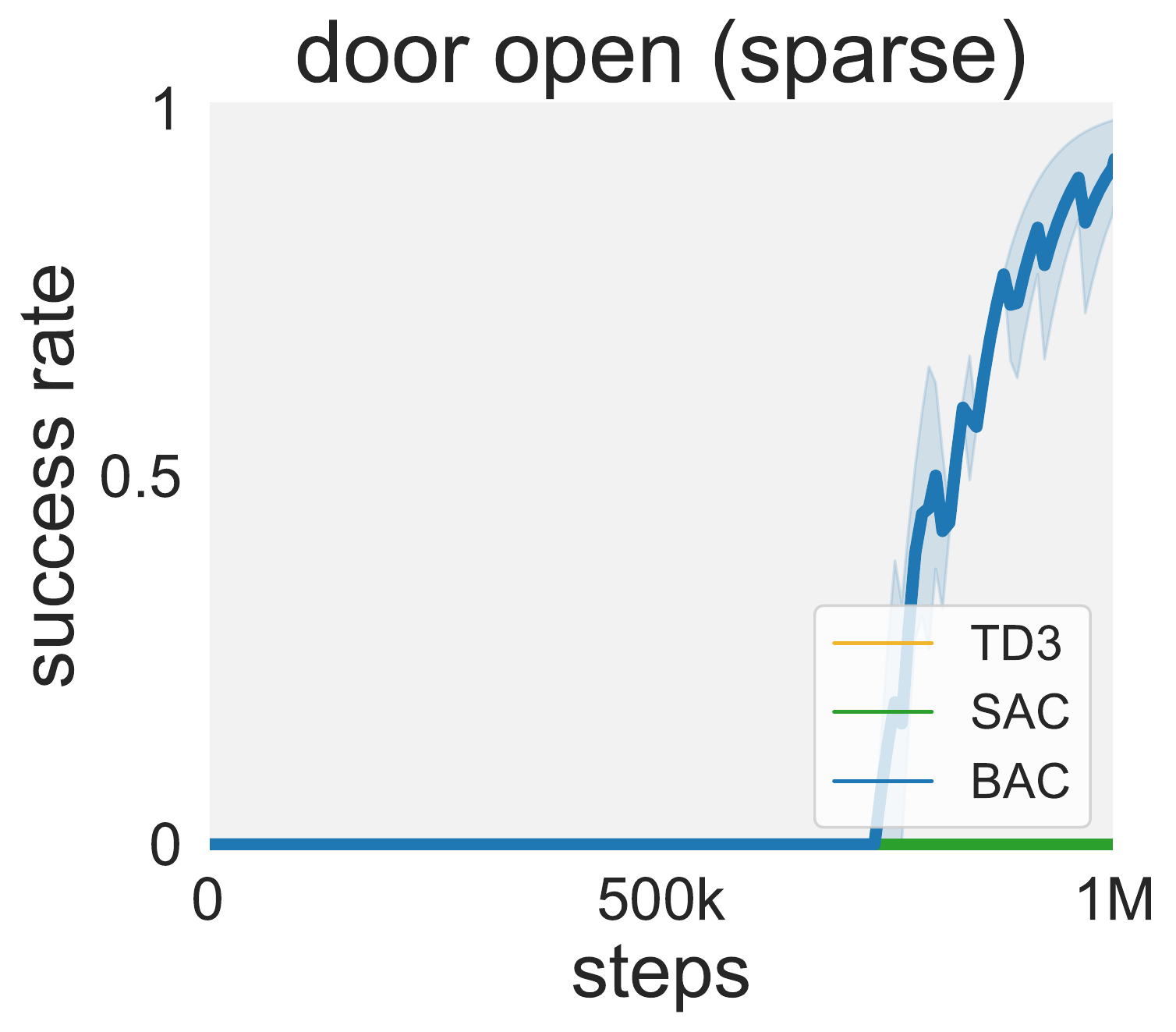}
            \end{subfigure}
            \caption{\Revise{\texttt{Meta-World} manipulation tasks on end-effector displacement control, using a Sawyer robot.}}
            \vspace{5mm}
        \end{subfigure}
        \begin{subfigure}[t]{1.0\textwidth}
            \begin{subfigure}[t]{0.32\textwidth}
            \includegraphics[height=3.5cm,keepaspectratio]{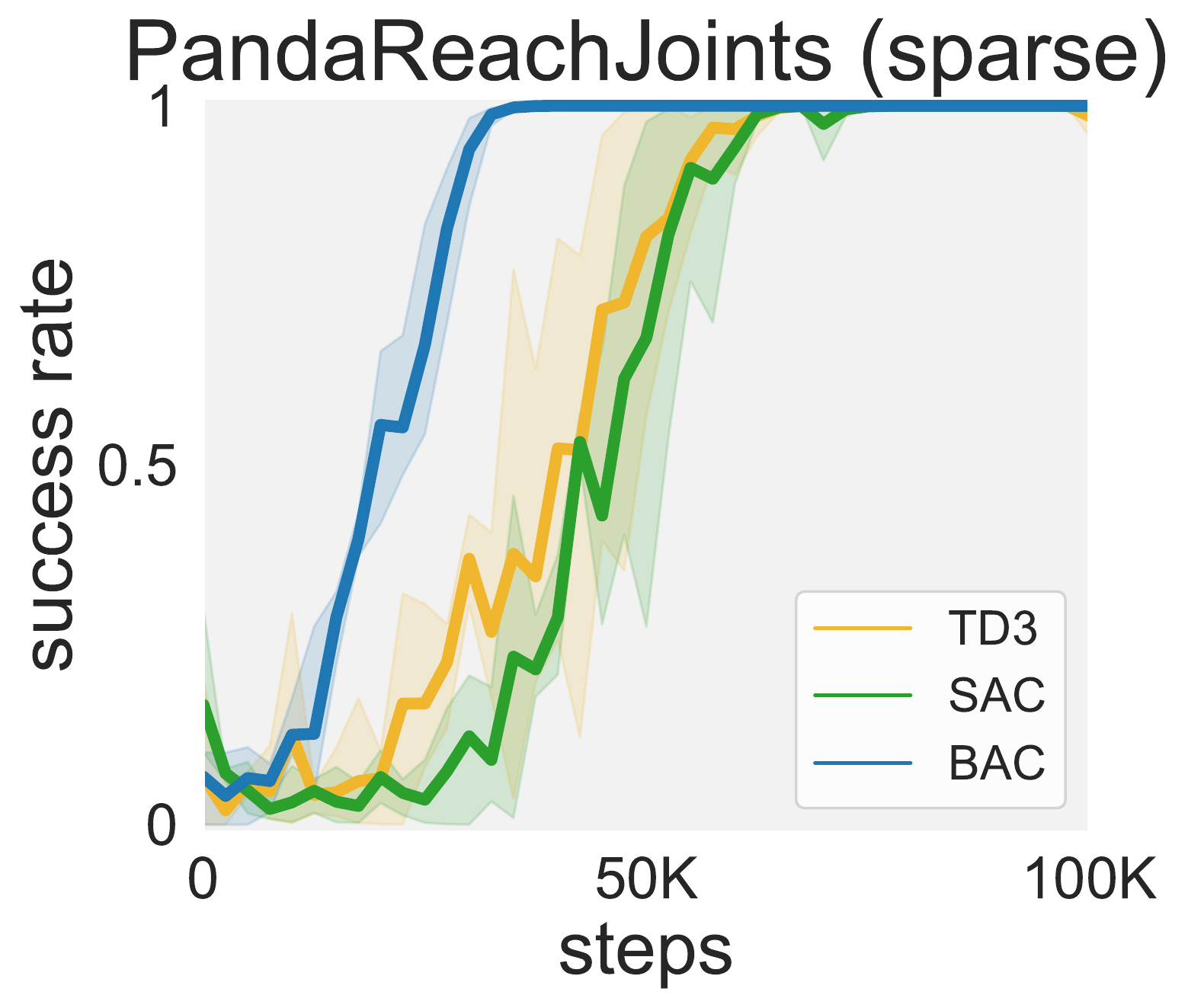}
                \caption{\Revise{\texttt{Panda-gym} manipulation tasks on joint angles control, using a Franka Emika Panda robot.}}
            \end{subfigure}
            \hfill
            \begin{subfigure}[t]{0.64\textwidth}
                \includegraphics[height=3.5cm,keepaspectratio]{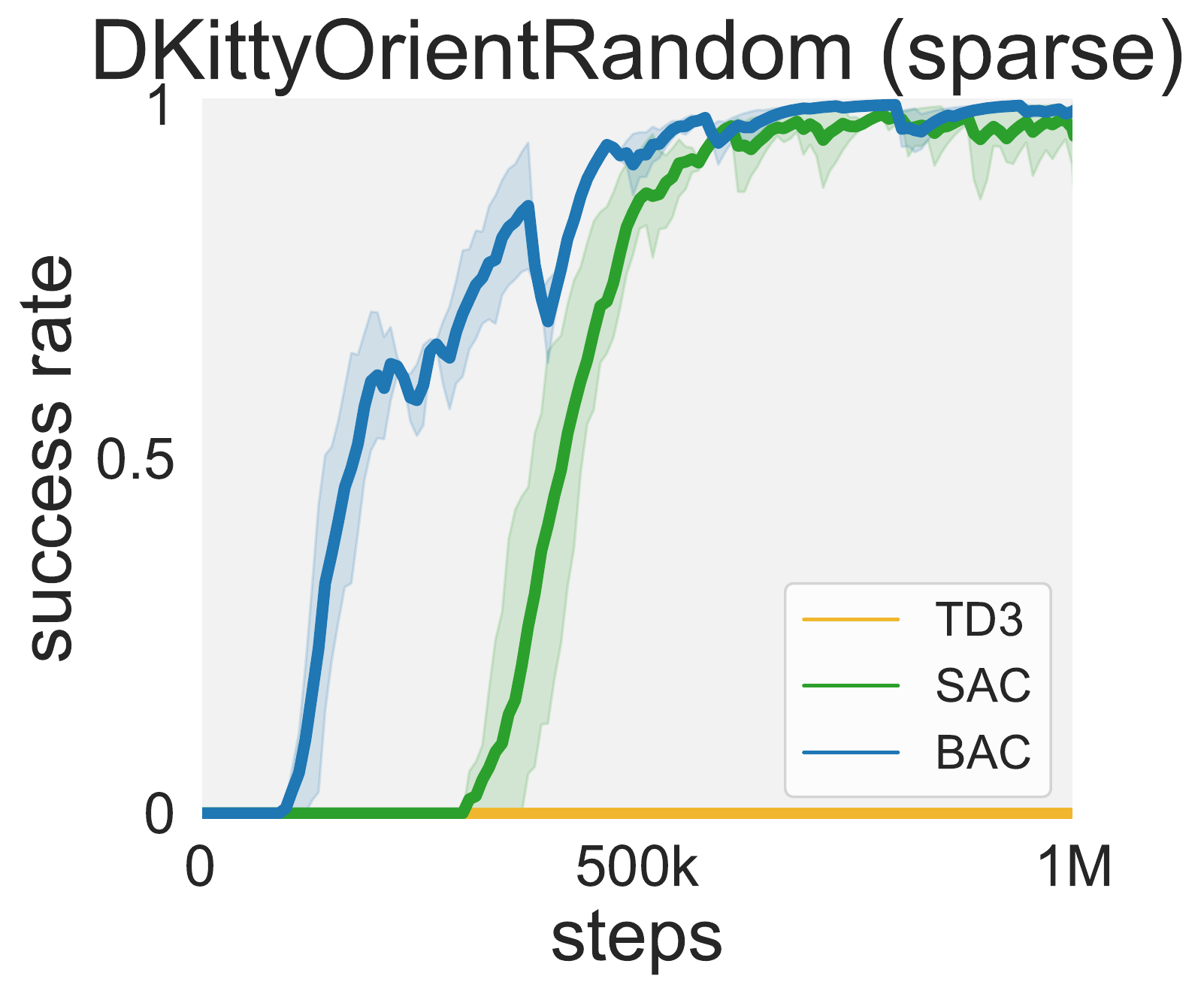}
                \includegraphics[height=3.5cm,keepaspectratio]{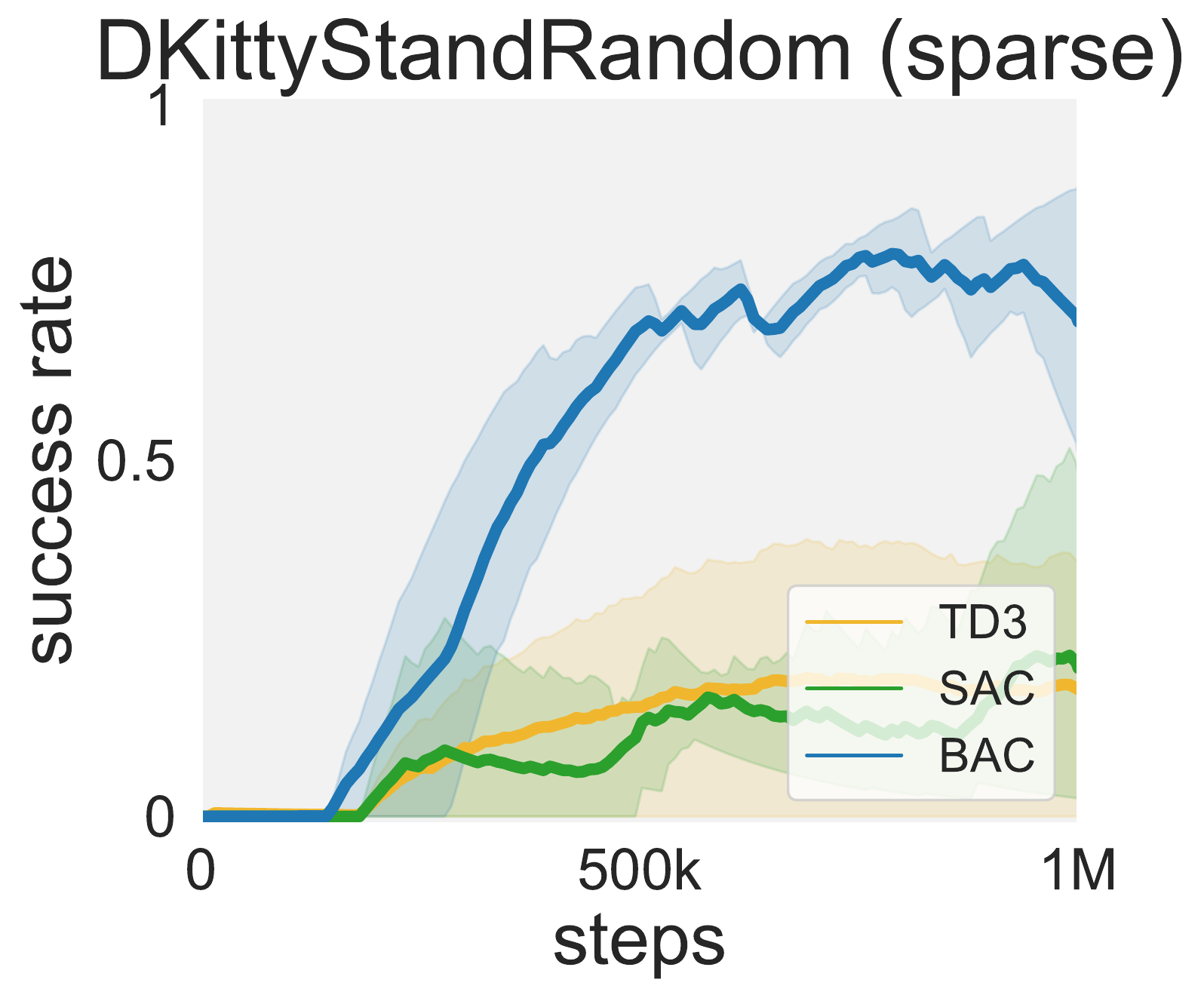}
            \caption{\Revise{\texttt{ROBEL} quadruped locomotion tasks, on 12 joint positions control, using a ROBEL D'Kitty robot.}}
            \end{subfigure}
        \end{subfigure}
    \end{minipage}
    \caption{\Revise{\textbf{Sparse reward tasks.} BAC outperforms the baselines on six sparse reward tasks across various control types and robotic platforms, including manipulation and locomotion tasks.}}
    \label{fig:sparsereward}
\end{figure}

\Revise{Panda-gym manipulation tasks are based on a Franka Emika Panda robot with joint angle control.
\begin{itemize}[leftmargin=16pt]
    \item PandaReachJoints: A target position must be reached with the gripper. This target position is randomly generated in a volume of $30cm\times 30cm\times 30 cm$.
\end{itemize}}
\Revise{
ROBEL quadruped locomotion tasks are based on a D'Kitty robot with 12 joint positions control.
\begin{itemize}[leftmargin=16pt]
   \item DKittyStandRandom:  The D'Kitty robot needs to reach a pose while being upright from a random initial conﬁguration. A successful strategy requires maintaining the stability of the torso via the ground reaction forces. 
   \item DKittyOrientRandom:  The D'Kitty robot needs to change its orientation from an initial facing direction to a  random target orientation.   A successful strategy requires maneuvering the torso via the ground reaction forces while maintaining balance.
\end{itemize}}
\Revise{
As shown in Figure~\ref{fig:sparsereward}, our BAC surpasses the baselines by a large margin. }

\clearpage
\subsection{Task visualizations in failure-prone scenarios}
\begin{wrapfigure}[15]{r}{0.40\textwidth}
    \centering
    \includegraphics[width=0.8\linewidth]{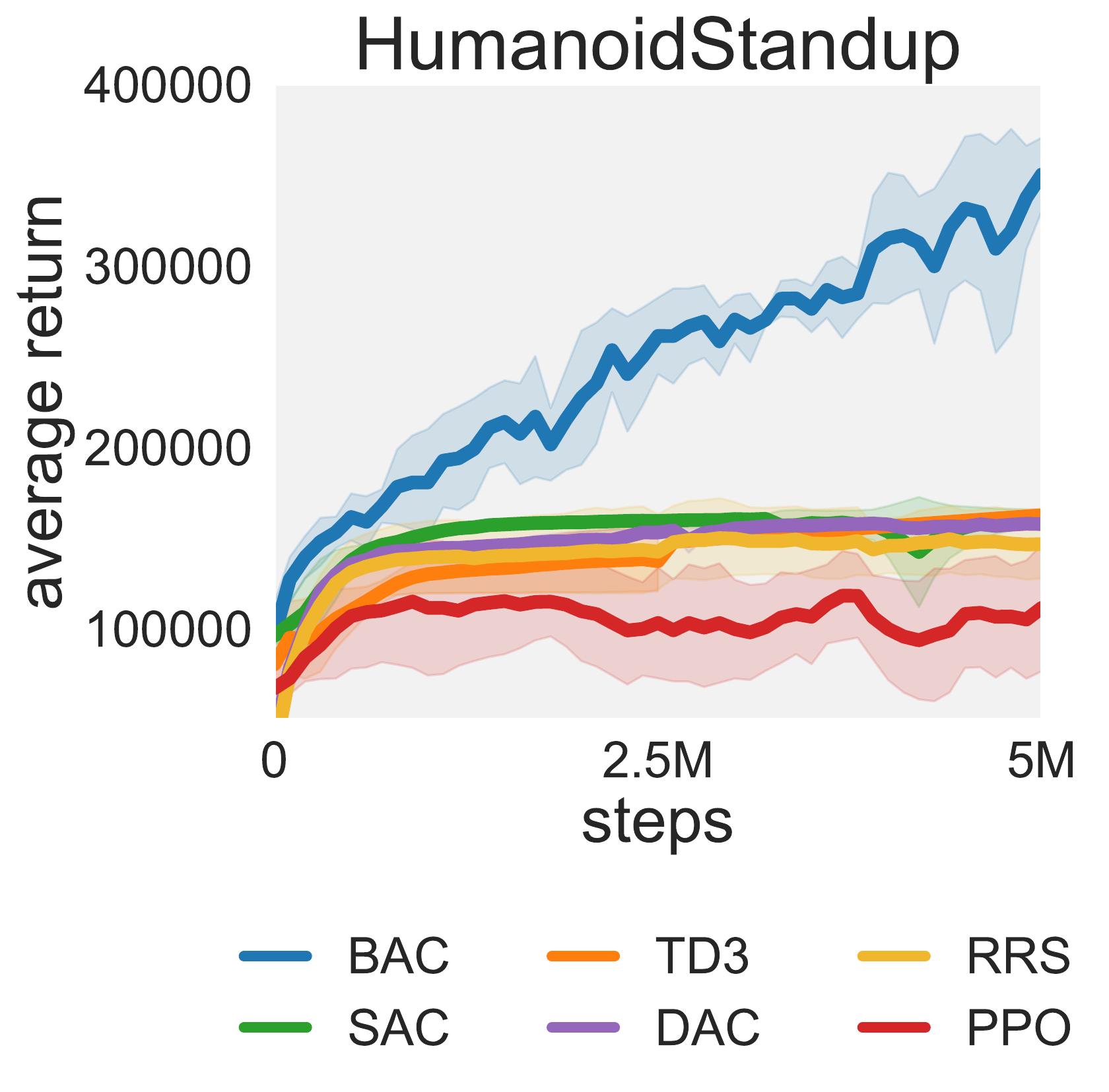}
    \caption{Learning curves of \ourshort\ and other baselines on HumanoidStandup task.}
    \label{fig:standup}
    \vspace{0.5cm}
\end{wrapfigure}
\paragraph{HumanoidStandup.}

HumanoidStandup, provided by MuJoCo~\citep{Mujoco}, is a challenging locomotion task. The environment begins with the humanoid lying on the ground, and the goal is to enable the humanoid to stand up and then keep it standing by applying torques on the various hinges. The agent takes a 17-element vector for actions.

In the HumanoidStandup task, BAC demonstrates a significantly superior performance than all other algorithms. With average returns reaching approximately 280,000 at 2.5 million steps and 36,000 at 5 million steps, BAC surpasses other algorithms whose asymptotic performance peaks at around 170,000, as illustrated in Figure~\ref{fig:standup}.

\begin{figure}[h!]
    \resizebox{1.0\textwidth}{!}{
    \begin{tabular}{cc}
        & time $\longrightarrow$\\ 
        \rotatebox{90}{{\textbf{\textcolor{myorange}{DAC}}}} & \includegraphics[width=0.95\linewidth,valign=c]{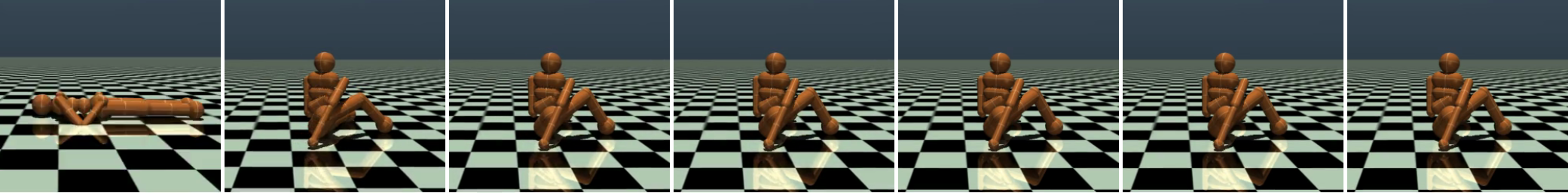} \\
        & \\
        \rotatebox{90}{{\textbf{\textcolor{myyellow}{RRS}}}} & \includegraphics[width=0.95\linewidth,valign=c]{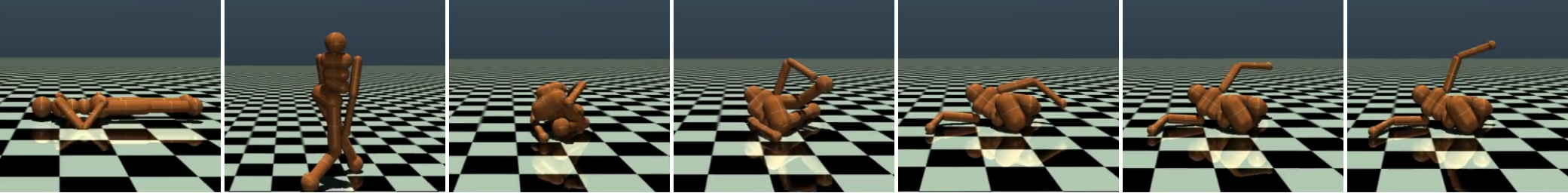} \\
        & \\
        \rotatebox{90}{\textbf{\textcolor{mygreen}{SAC}}}  & \includegraphics[width=0.95\linewidth,valign=c]{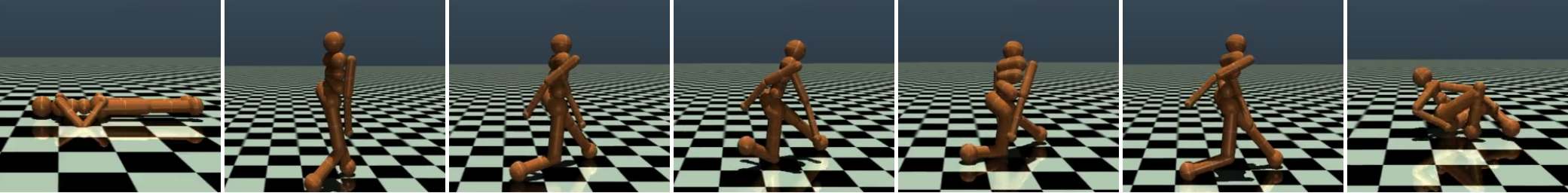} \\
        & \\
        \rotatebox{90}{\textbf{\textcolor{myblue}{BAC$^1$}}} & \includegraphics[width=0.95\linewidth,valign=c]{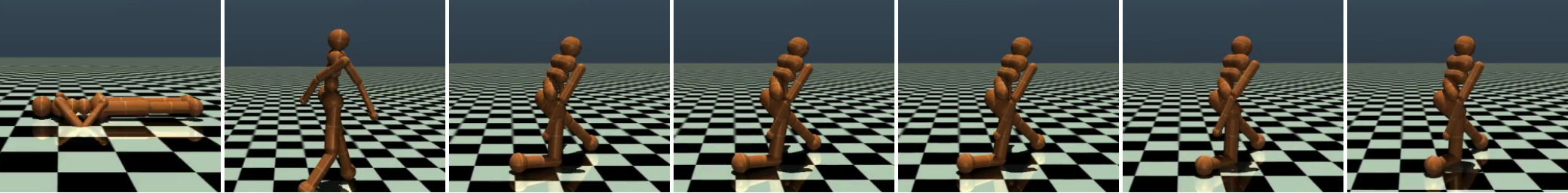} \\
        & \\
        \rotatebox{90}{\textbf{\textcolor{myblue}{BAC$^2$}}} & \includegraphics[width=0.95\linewidth,valign=c]{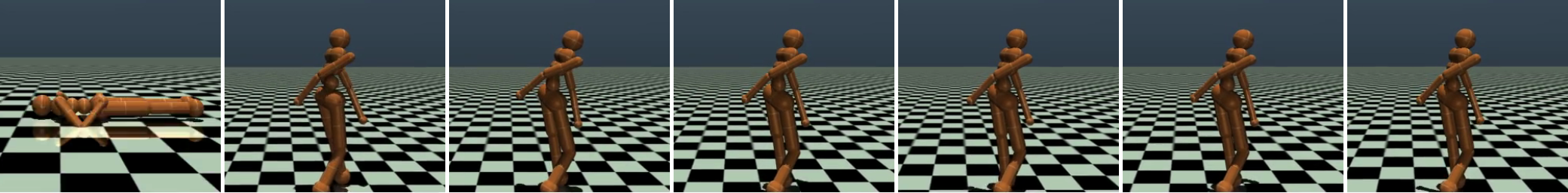} \\
    \end{tabular}
    }
    \caption{Visualization on HumanoidStandup task. BAC$^1$ is the visualization using the learned policy at 2.5M steps, and BAC$^2$ reveals the behaviors learned at 5M steps. For DAC, RRS, and SAC, we visualize the learned policy at 5M steps.}
    \label{fig:standup_visualization}
\end{figure}

Visualization in Figure~\ref{fig:standup_visualization} depicts that the BAC agent can quickly achieve a stable standing pose. In contrast, the SAC agent ends up in an unstable, swaying kneeling position, DAC ends up sitting on the ground, and the RRS agent, regrettably, is seen rolling around.

\vspace{1cm}

\clearpage
\paragraph{DogRun.}
\begin{wrapfigure}[15]{r}{0.40\textwidth}
    \centering
    \includegraphics[width=0.8\linewidth]{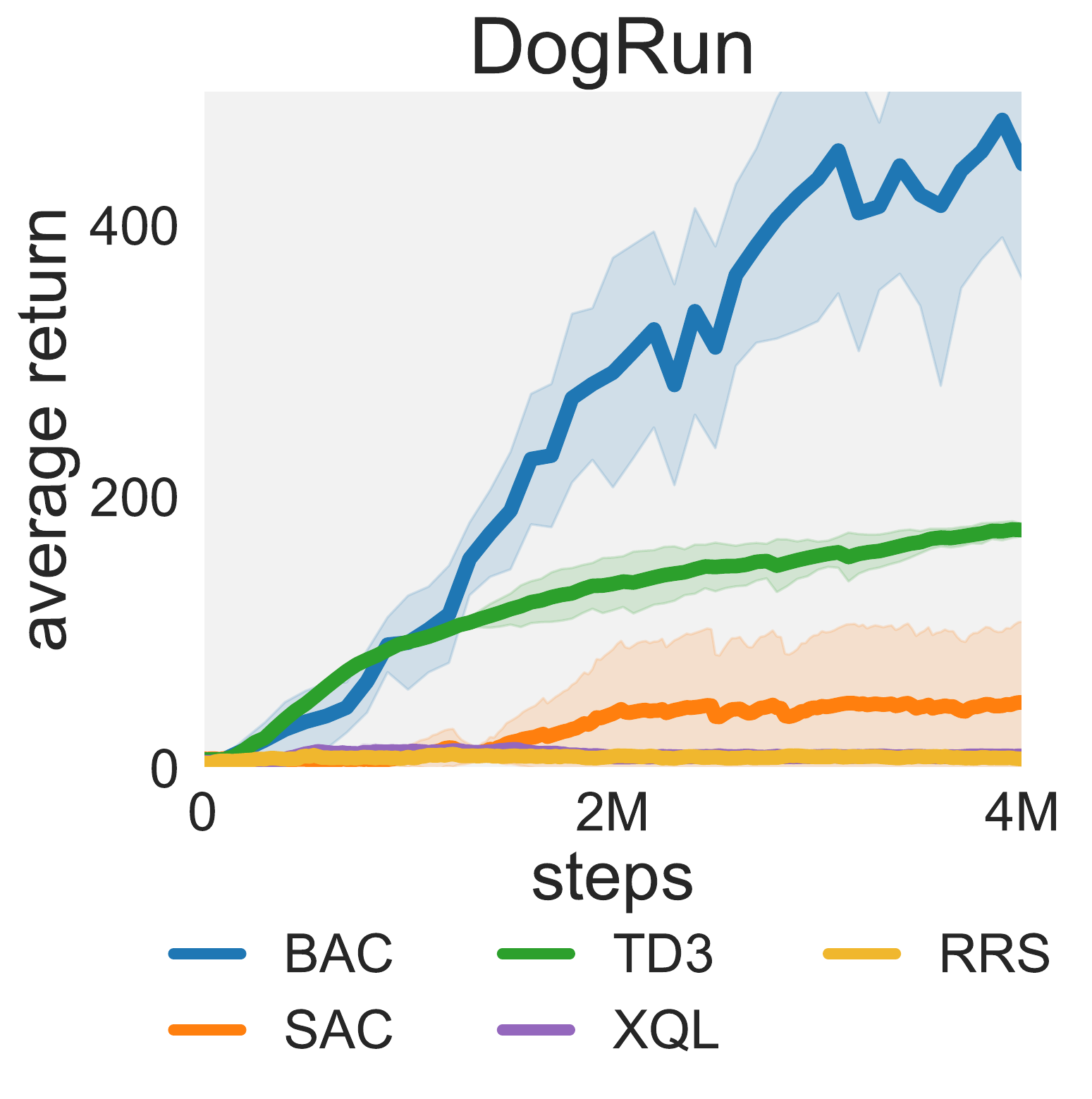}
    \vspace{-5mm}
    \caption{Learning curves \ourshort\ and other baselines on DogRun task.}
    \label{fig:dogrun_performance}
\end{wrapfigure}
DogRun, provided by the DMControl~\citep{dmcontrol}, is a challenging task with a high-dimensional action space ($\mathcal{A}\in \mathbb{R}^{38}$). The task is based on a sophisticated model of a Pharaoh Dog, including intricate kinematics, skinning weights, collision geometry, as well as muscle and tendon attachment points. This complexity makes the DogRun task extremely difficult for algorithms to learn and control effectively.

We conducted extensive experiments in the DogRun task to compare the performance of \ourshort\ against other state-of-the-art algorithms. 
Here, we include Extreme Q-Learning~(XQL)~\citep{garg2023extreme} as our baseline, which falls into the MaxEntropy RL framework but directly models the maximal $Q$ value.
The results, depicted in Figure~\ref{fig:dogrun_performance}, reveal that \ourshort\ significantly surpasses its counterparts, attaining higher average returns in fewer interactions.  It demonstrates a remarkable capability of learning to control the high-dimensional, complex robot, such as facilitating the dog's run. To the best of our knowledge, \textbf{\textit{\textcolor{myred}{it is the first documented result of model-free methods effectively tackling the challenging DogRun task.}}}

\begin{figure}[h!]
    \resizebox{1.0\textwidth}{!}{
    \begin{tabular}{cc}
        & time $\longrightarrow$\\ 
        \rotatebox{90}{{\textbf{\textcolor{mypurple}{XQL}}}} & \includegraphics[width=0.95\linewidth,valign=c]{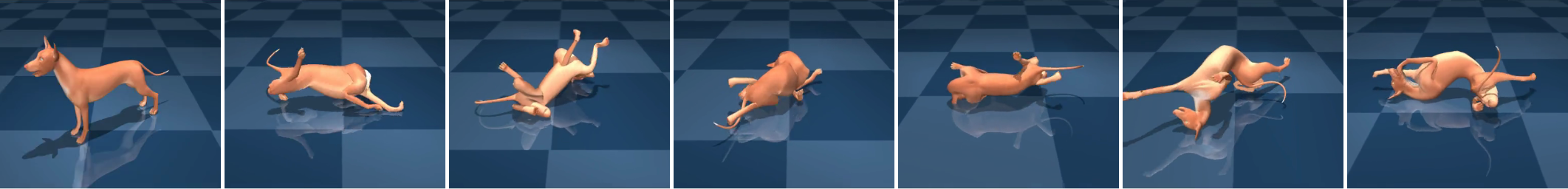} \\
        & \\
        \rotatebox{90}{{\textbf{\textcolor{myyellow}{RRS}}}} & \includegraphics[width=0.95\linewidth,valign=c]{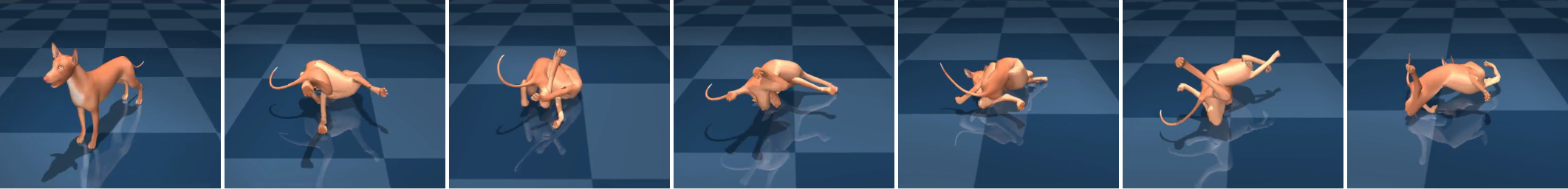} \\
        & \\
        \rotatebox{90}{\textbf{\textcolor{mygreen}{SAC}}}  & \includegraphics[width=0.95\linewidth,valign=c]{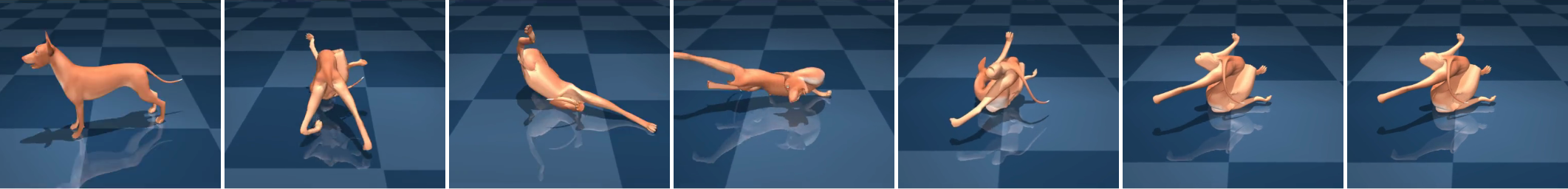} \\
        & \\
        \rotatebox{90}{\textbf{\textcolor{myorange}{TD3}}} & \includegraphics[width=0.95\linewidth,valign=c]{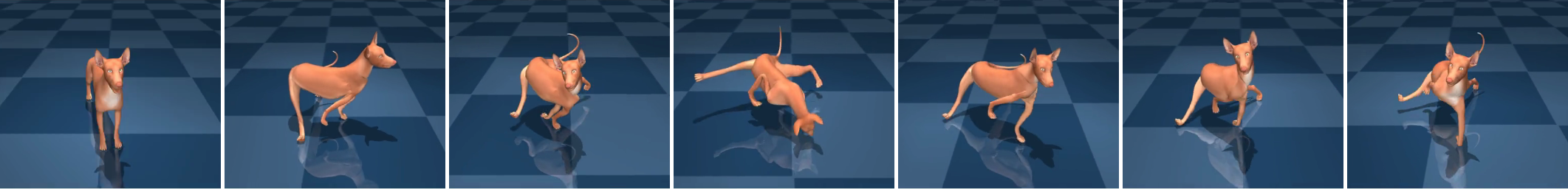} \\
        & \\
        \rotatebox{90}{\textbf{\textcolor{myblue}{BAC}}} & \includegraphics[width=0.95\linewidth,valign=c]{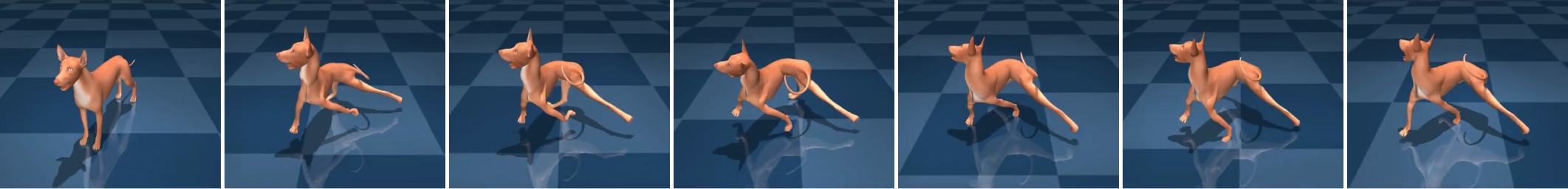} \\
    \end{tabular}
    }
    \caption{Visualization on DogRun task. We visualize the keyframes of the trajectories induced by the learned policy of each algorithm at 4M steps. }
    \label{fig:dogrun_trajectory}
\end{figure}

In addition to the quantitative results, we also offer a visualization of keyframes in the trajectory in Figure~\ref{fig:dogrun_trajectory}. Here, the superior performance of \ourshort\ becomes even more apparent. While competing algorithms struggle to prevent the dog from falling, \ourshort\ successfully achieves a running motion. This aptitude for handling complex, high-dimensional tasks further reaffirms the efficacy and robustness of \ourshort\ when dealing with failure-prone scenarios.

\vspace{0.3cm}

\clearpage

\section{Additional Baselines Comparison}
Our paper spotted the underestimation phenomenon, and other options may fix it. 

\begin{figure}[h!]
    \centering
    \begin{subfigure}[t]{0.45\textwidth}
        \includegraphics[width=1.0\linewidth]{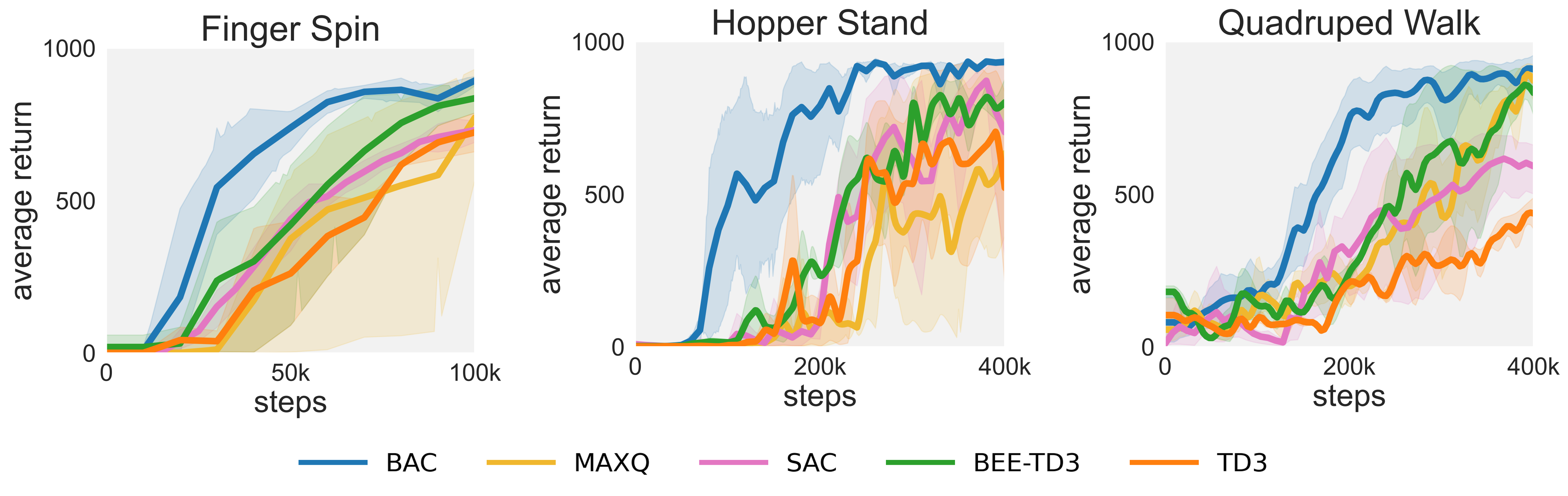}
        \caption{\textbf{Comparison with MAXQ.} Training curves of BAC, MAXQ, BEE-TD3, SAC, TD3 in DMControl benchmark tasks. Solid curves depict the mean of ten trials and shaded regions correspond to the one standard deviation.}
        \label{fig:maxQ}
    \end{subfigure}
    \hspace{10mm}
    \begin{subfigure}[t]{0.45\textwidth}
        \includegraphics[width=1.0\linewidth]{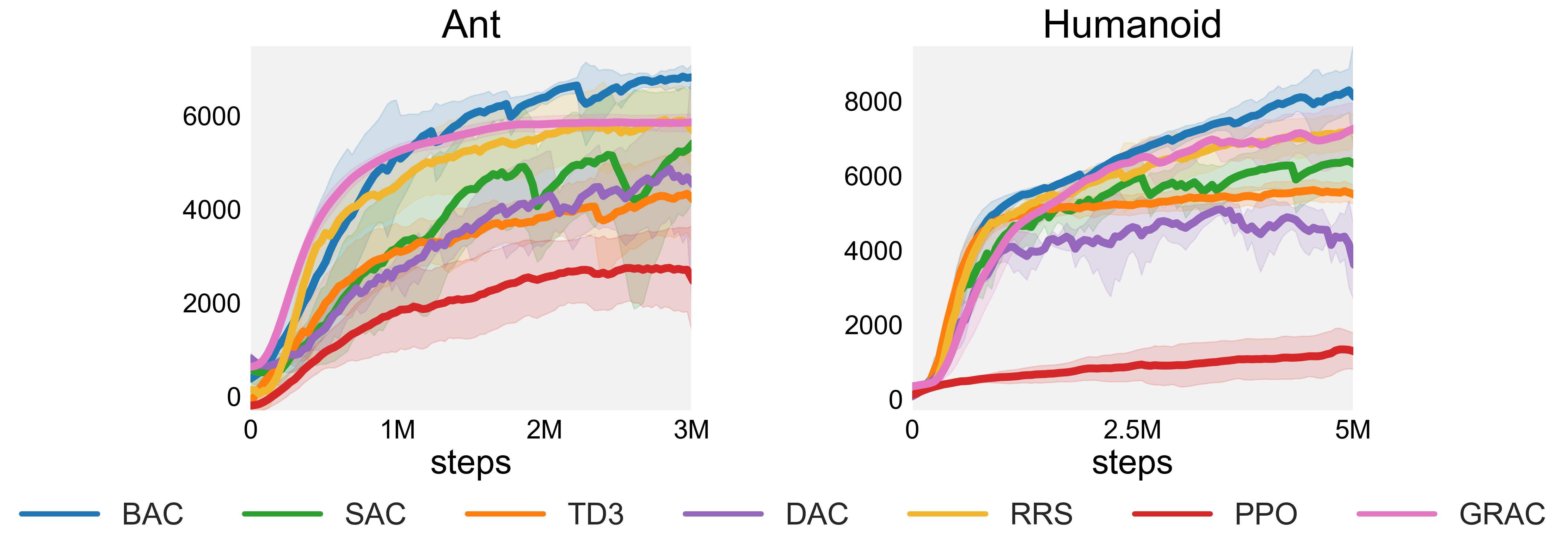}
        \caption{\\textbf{Comparison with GRAC.} Training curves of BAC, GRAC, and other standard AC algorithms(SAC, TD3, DAC, RRS, and PPO) in MuJoCo benchmark tasks. Solid curves depict the mean of ten trials and shaded regions correspond to the one standard deviation. GRAC does outperform standard AC algorithms, underscoring the advantage of addressing underestimation to boost performance.}
        \label{fig:grac}
    \end{subfigure}
    \caption{Performance comparison to two further baselines.}
    \label{fig:gracandmaxq}
\end{figure}

Here, we compare two baselines, $\max(Q(s',\pi(s')), Q(s',a'))$ and GRAC~\citep{shao2022grac}, as shown in Figure~\ref{fig:gracandmaxq}. However, both the MAXQ and GRAC lag behind BAC for a crucial reason: while they address underestimation, they inadvertently introduce extra overestimation, which can hinder performance. Interestingly, GRAC does outperform standard AC algorithms, underscoring the advantage of addressing underestimation to boost performance—one of the key contributions of our paper. Here's a deeper dive into these comparisons:

\begin{itemize}[leftmargin=16pt]
    \item Comparison to MAXQ: The MAXQ method, although simple, faces notable instability issues. When $Q(s’,a’)$ gets erroneously overestimated, a common issue in early training, it leads to overly optimistic outcomes from $\max(Q(s',\pi(s')),Q(s',a'))$, potentially leading to more severe overestimation issues.

    \item Comparison to GRAC: GRAC, by selecting the maximum Q-values around the policy $\pi$, does introduce a degree of optimism. However, this approach suffers from considerable overestimation (extrapolation error), especially when it encounters actions that are out of the distribution during training~\citep{BCQ}. Such overestimation can negatively affect performance. Also, the additional computational load required for action sampling could be a drawback in resource-constrained environments.
\end{itemize}

BEE does not handle it one-sidedly. While the BEE seeks to alleviate underestimation, it might not incite extra overestimation. As the exploitation operator only relies on real experience and may lead to a more stable estimation, which would help to reduce overestimation practically.

\clearpage
\section{More Benchmark Results}\label{section:more-benchmark-results}
Given that MuJoCo benchmark tasks have been solved well by popular baselines, we conduct experiments on the more complex locomotion and manipulation tasks from \textbf{DMControl}~\citep{dmcontrol}, \textbf{Meta-World}~\citep{yu2019meta}, \textbf{Adroit}~\citep{adroit}, \textbf{ManiSkill2}~\citep{gu2023maniskill2}, \textbf{Shadow Dexterous Hand}~\citep{dexshadowhand} and \textbf{MyoSuite}~\citep{MyoSuite2022} for further evaluation of BAC and the baselines. Currently, several tasks in these benchmarks pose a formidable challenge that stumps most model-free methods.  Notably, \ourshort\ has demonstrated its effectiveness by successfully solving many of these challenging tasks.

\subsection{Evaluation on DMControl benchmark tasks}\label{section:dmc_benchmark}
We tested BAC and its variant, BEE-TD3, on 15 continuous control tasks from DMControl. BAC successfully solves many challenging tasks like HumanoidRun, DogWalk, and DogRun, where both SAC and TD3 fail. Also, BEE-TD3 boosts TD3's performance by a large margin, demonstrating the generalizability of the BEE operator.

\begin{figure}[h!]
    \resizebox{1.0\textwidth}{!}{
        \begin{tabular}{cc}
            & time $\longrightarrow$\\ 
            \rotatebox{90}{\parbox[c]{2cm}{\centering  Quadruped \\ Walk}} &
            \includegraphics[width=0.95\linewidth,valign=b]{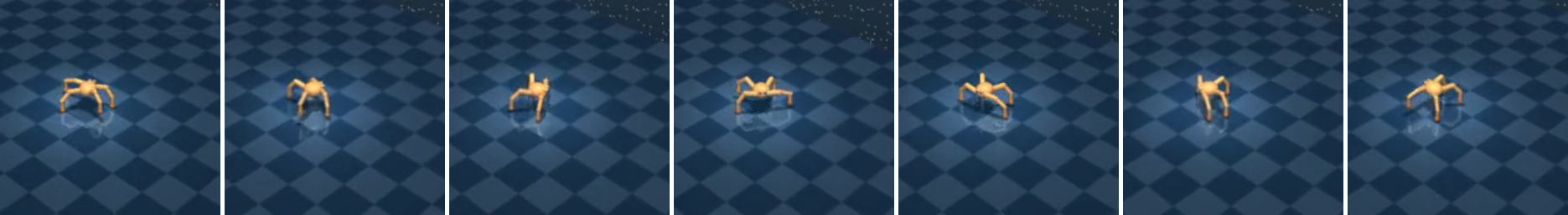} 
            \\[-0.6ex]
            \rotatebox{90}{\parbox[c]{2cm}{\centering  Dog \\ Walk}} &
            \includegraphics[width=0.95\linewidth,valign=b]{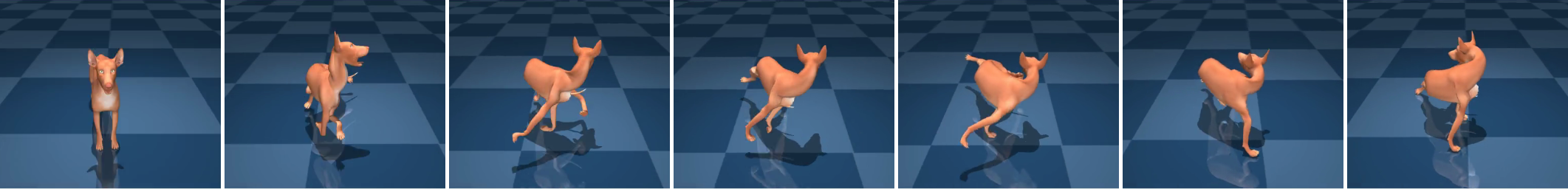} 
            \\[-0.6ex]
            \rotatebox{90}{\parbox[c]{2cm}{\centering  Dog \\ Run}} &
            \includegraphics[width=0.95\linewidth,valign=b]{Figures/554.png} 
            \\[-0.6ex]
            \rotatebox{90}{\parbox[c]{2cm}{\centering  Humanoid \\ Walk}} &
            \includegraphics[width=0.95\linewidth,valign=b]{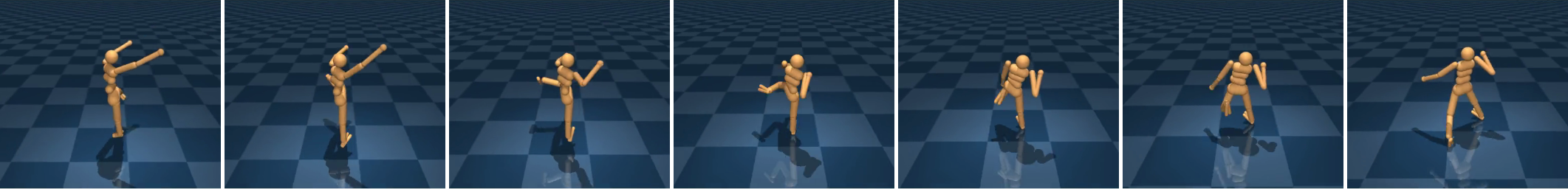}
            \\[-0.6ex]
            \rotatebox{90}{\parbox[c]{2cm}{\centering  Humanoid \\ Run}} &
            \includegraphics[width=0.95\linewidth,valign=b]{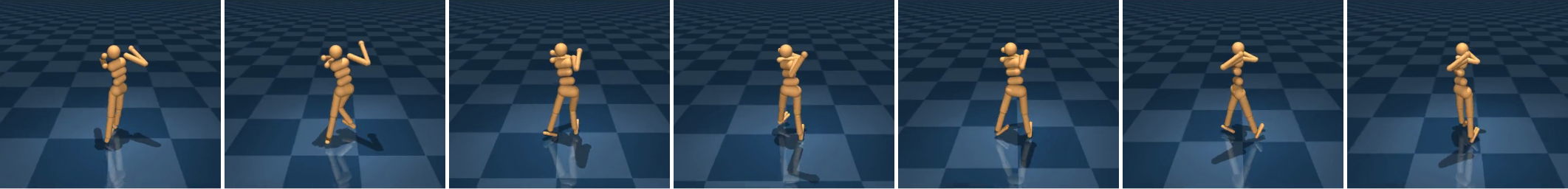}
        \end{tabular}
    }
    \caption{\textbf{Trajectory Visualizations.} Visualizations of the learned policy of \ourshort\ on five DMControl benchmark tasks.}
    \label{fig:dmc_trajectoryvisualization}
\end{figure}

\paragraph{Trajectory Visualizations.}
Figure~\ref{fig:dmc_trajectoryvisualization}  provides visualizations of trajectories generated by BAC on five tasks from DMControl. For each trajectory, we display seven keyframes.

\begin{figure}[h!]
    \centering
        \includegraphics[width=1.0\linewidth]{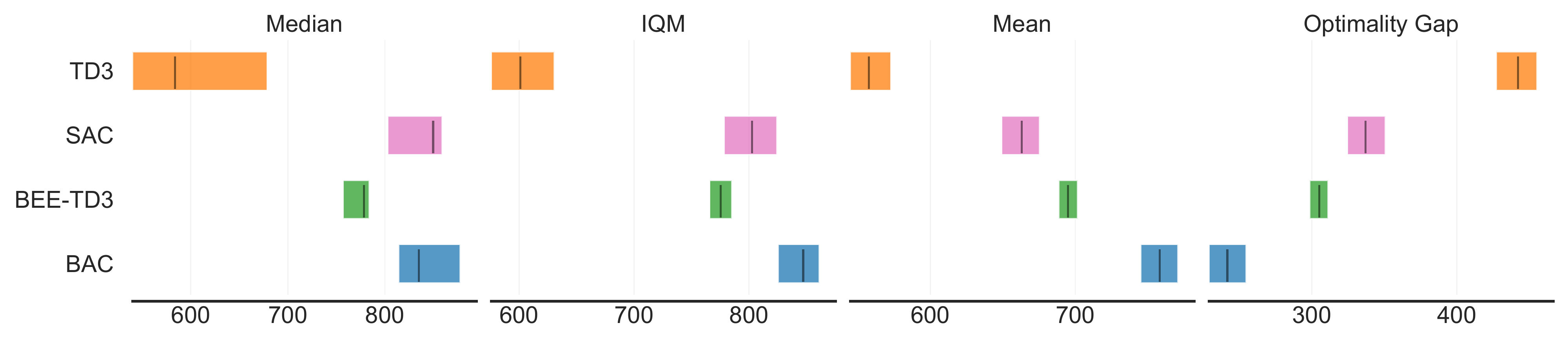}
    \caption{\textbf{Reliable metrics on DMControl tasks.} Median, IQM, Mean~(higher values are better), and Optimality Gap~(lower values are better) performance of \ourshort\ , BEE-TD3 and baselines (SAC, TD3) on the 15 DMControl tasks. }
    \label{fig:rliable}
\end{figure}

\paragraph{Rliable metrics.} 
We report additional~(aggregate) performance metrics of \ourshort\ and SAC  on the set of 15 DMControl tasks using the \texttt{rliable} toolkit~\citep{agarwal2021deep}. As shown in Figure~\ref{fig:rliable}, \ourshort\ outperforms SAC in terms of Median, interquantile mean~(IQM), Mean, and Optimality Gap.

\begin{figure}[h!]
    \centering
    \includegraphics[width=0.24\linewidth]{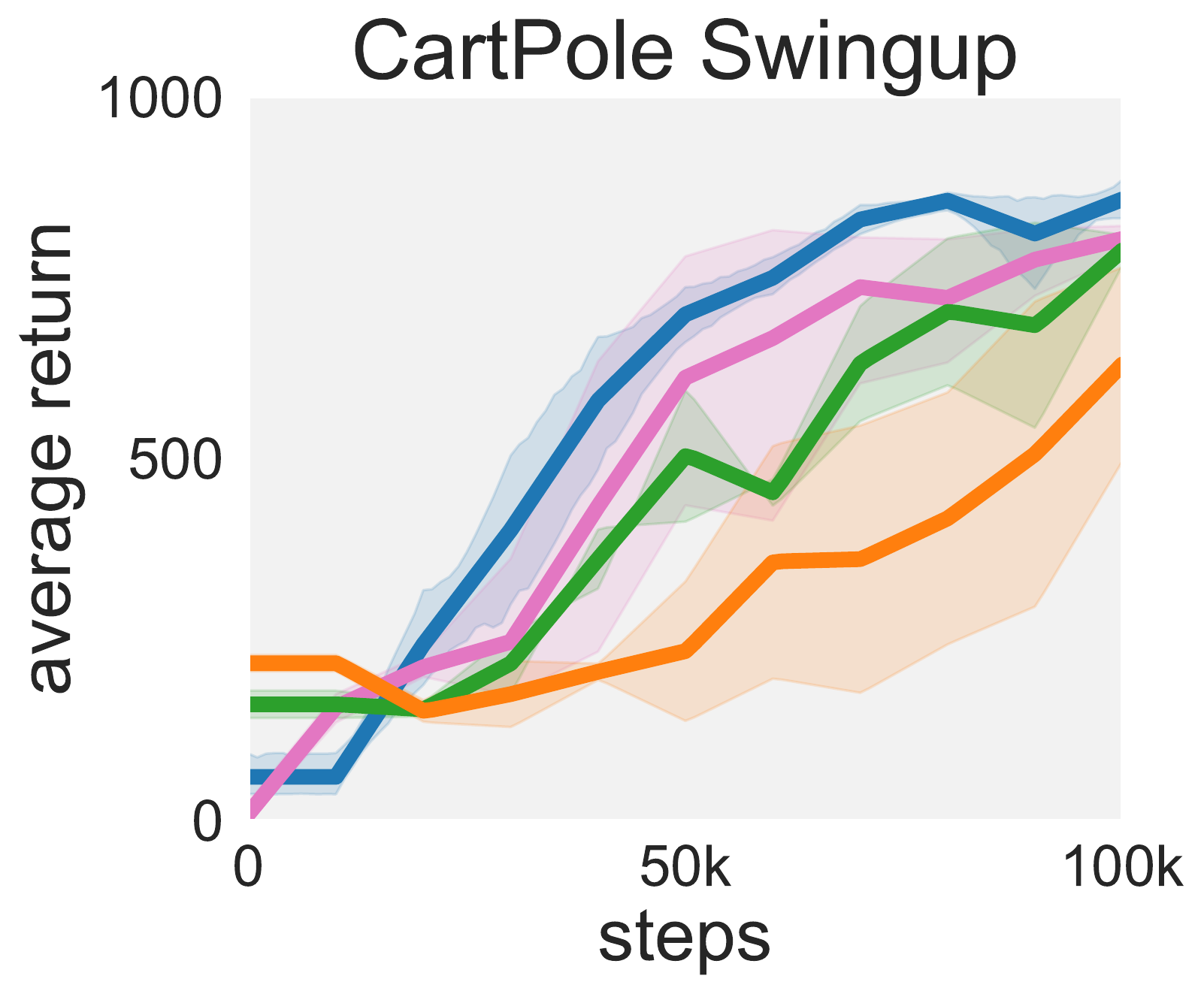} 
    \includegraphics[width=0.24\linewidth]{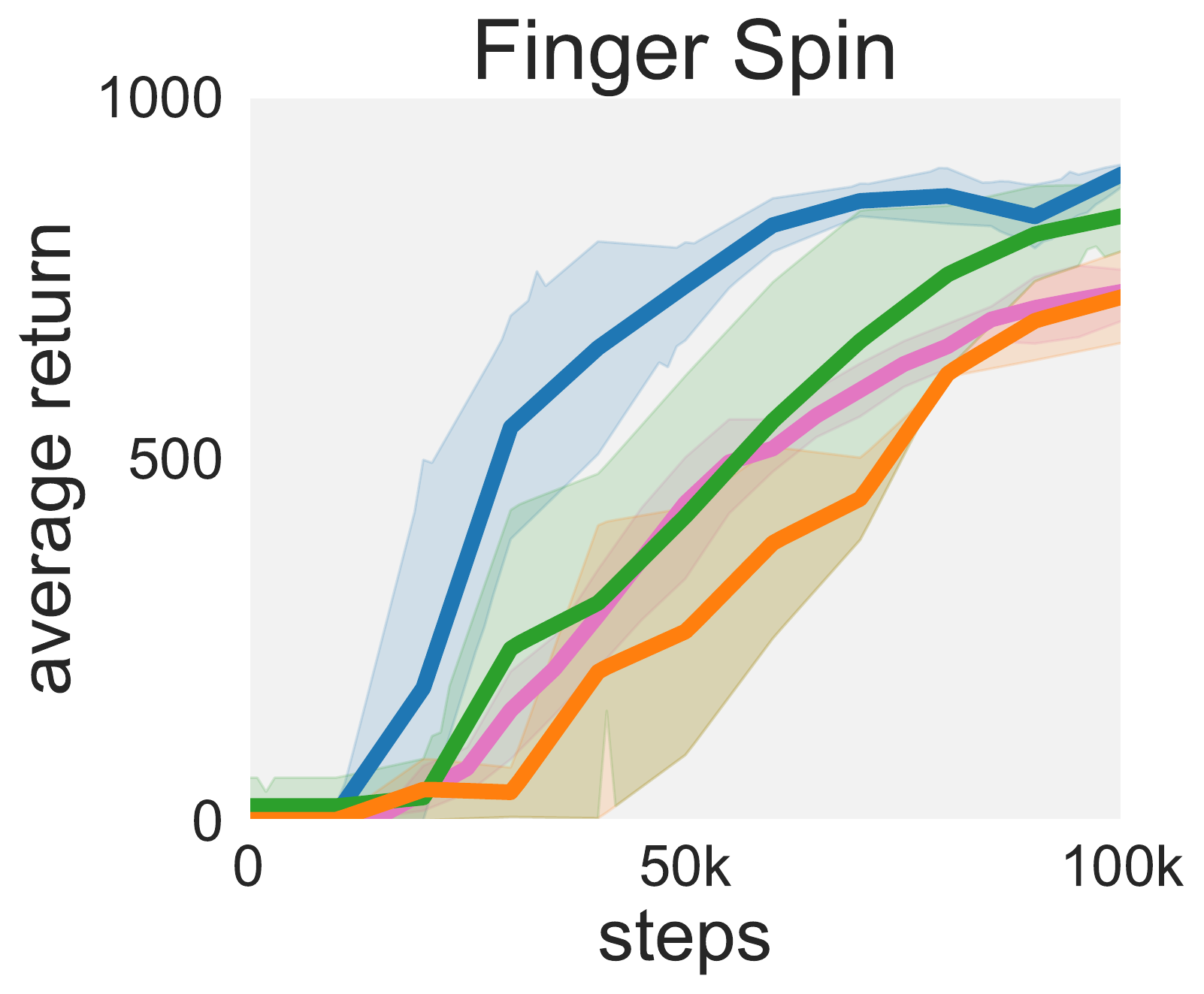} 
    \includegraphics[width=0.24\linewidth]{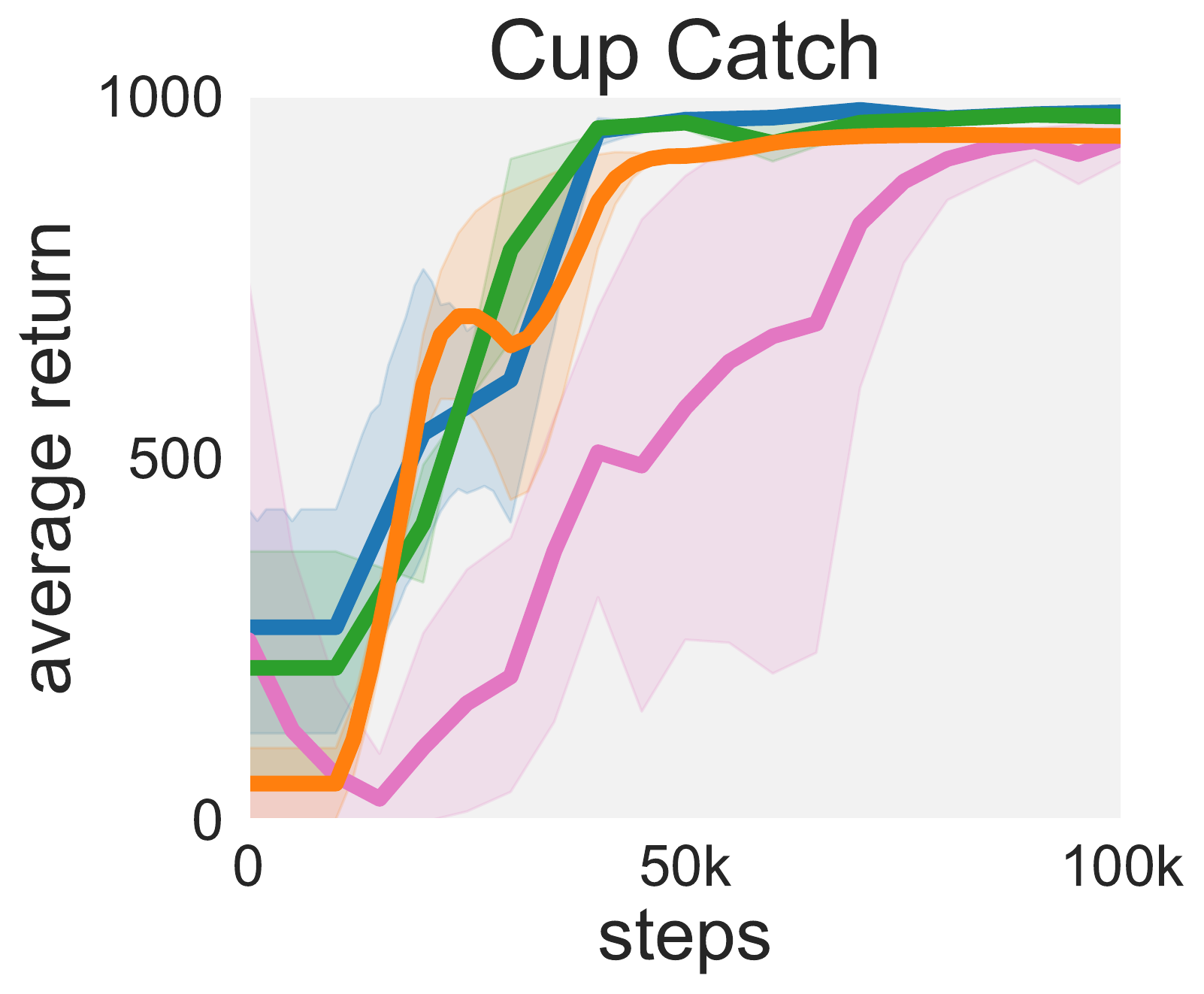}
    \includegraphics[width=0.24\linewidth]{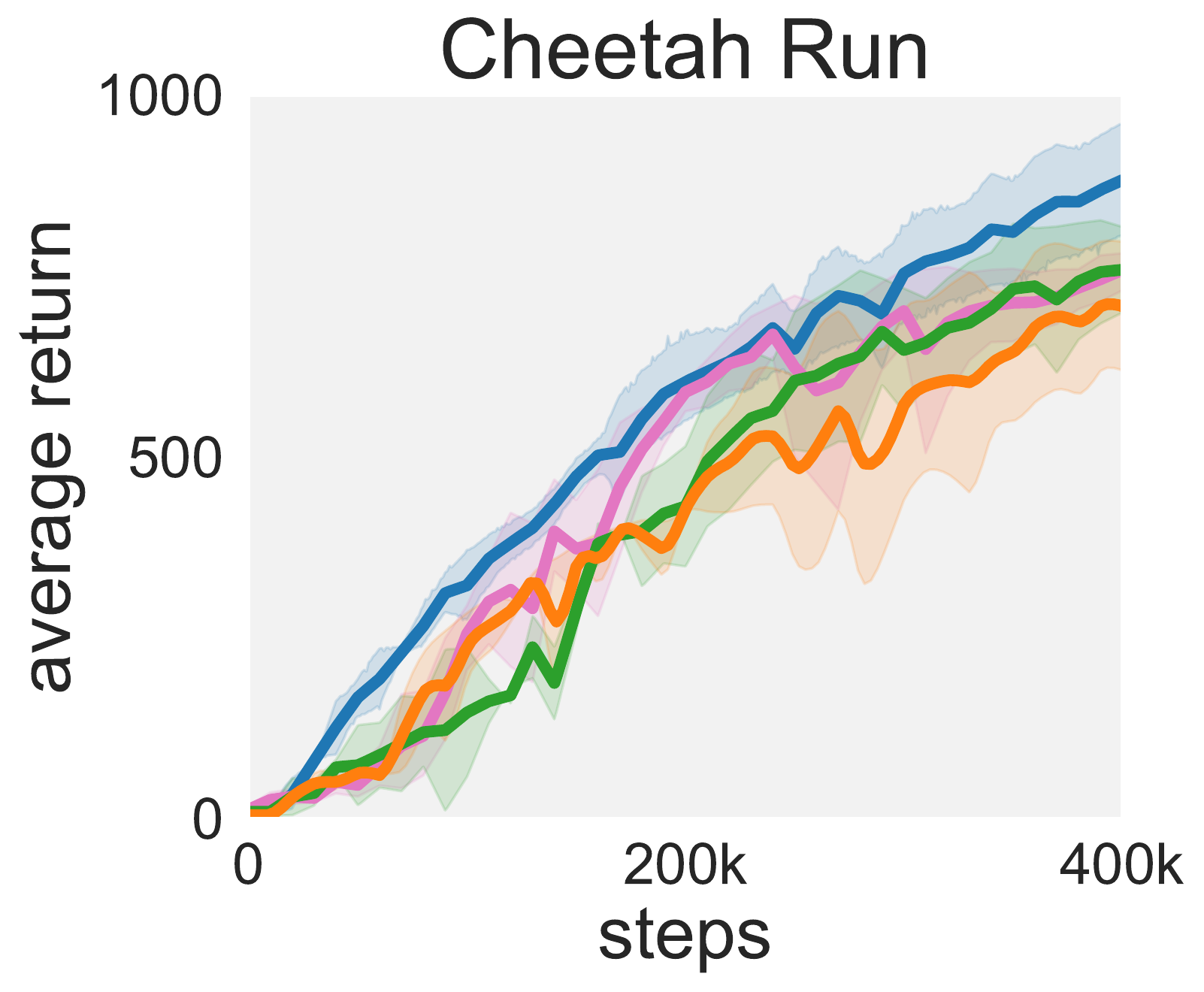}
    \includegraphics[width=0.24\linewidth]{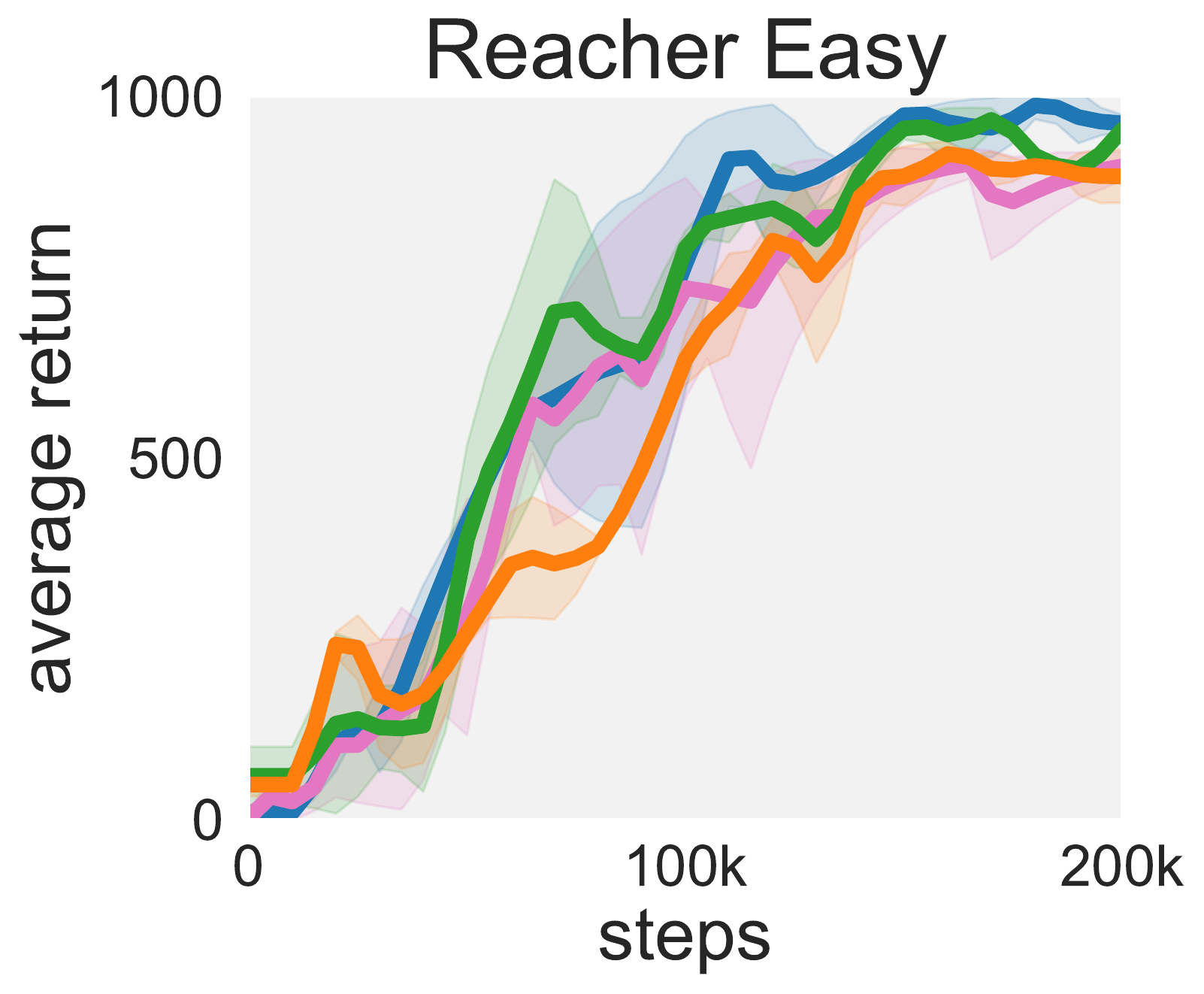} 
    \includegraphics[width=0.24\linewidth]{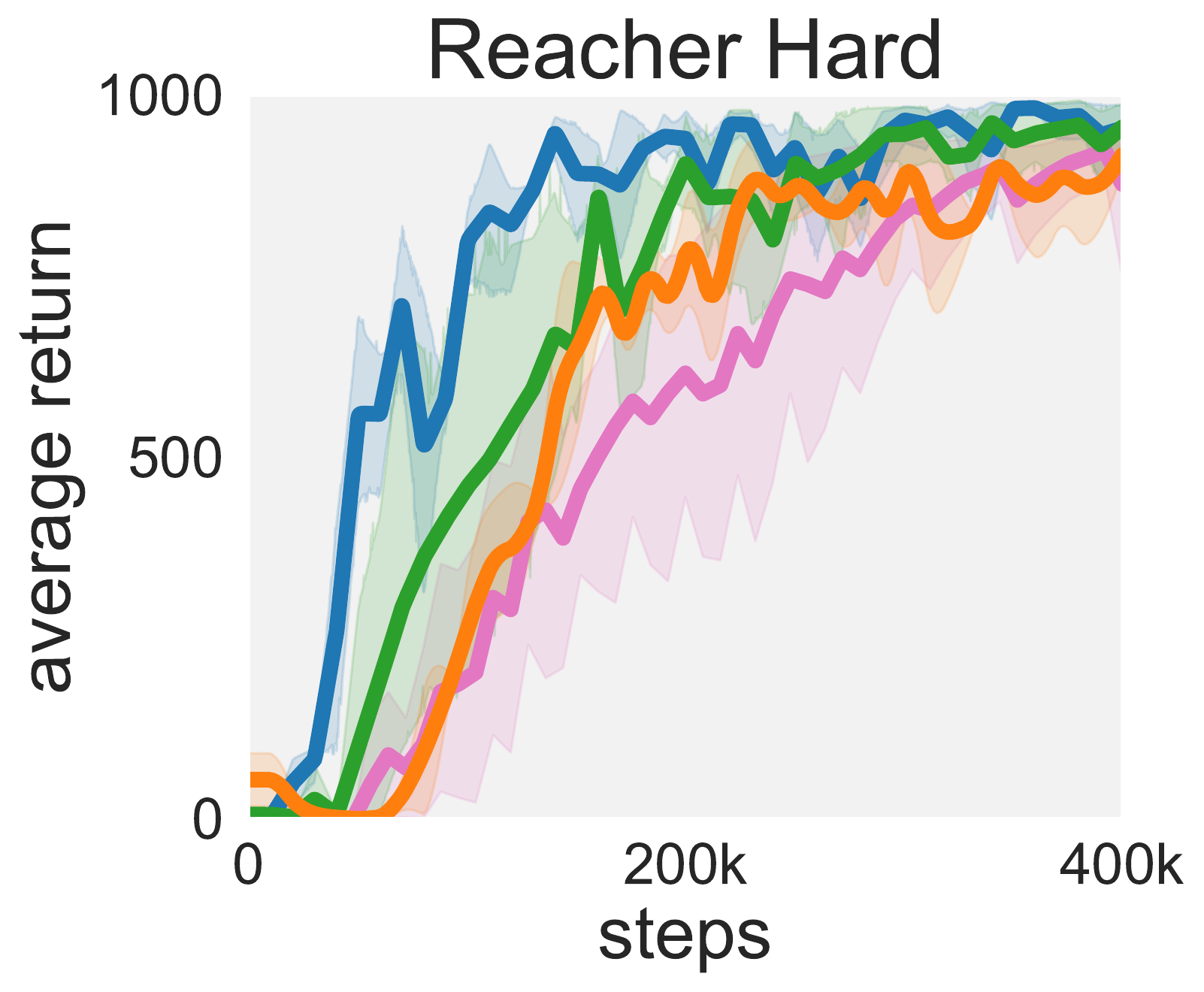} 
    \includegraphics[width=0.24\linewidth]{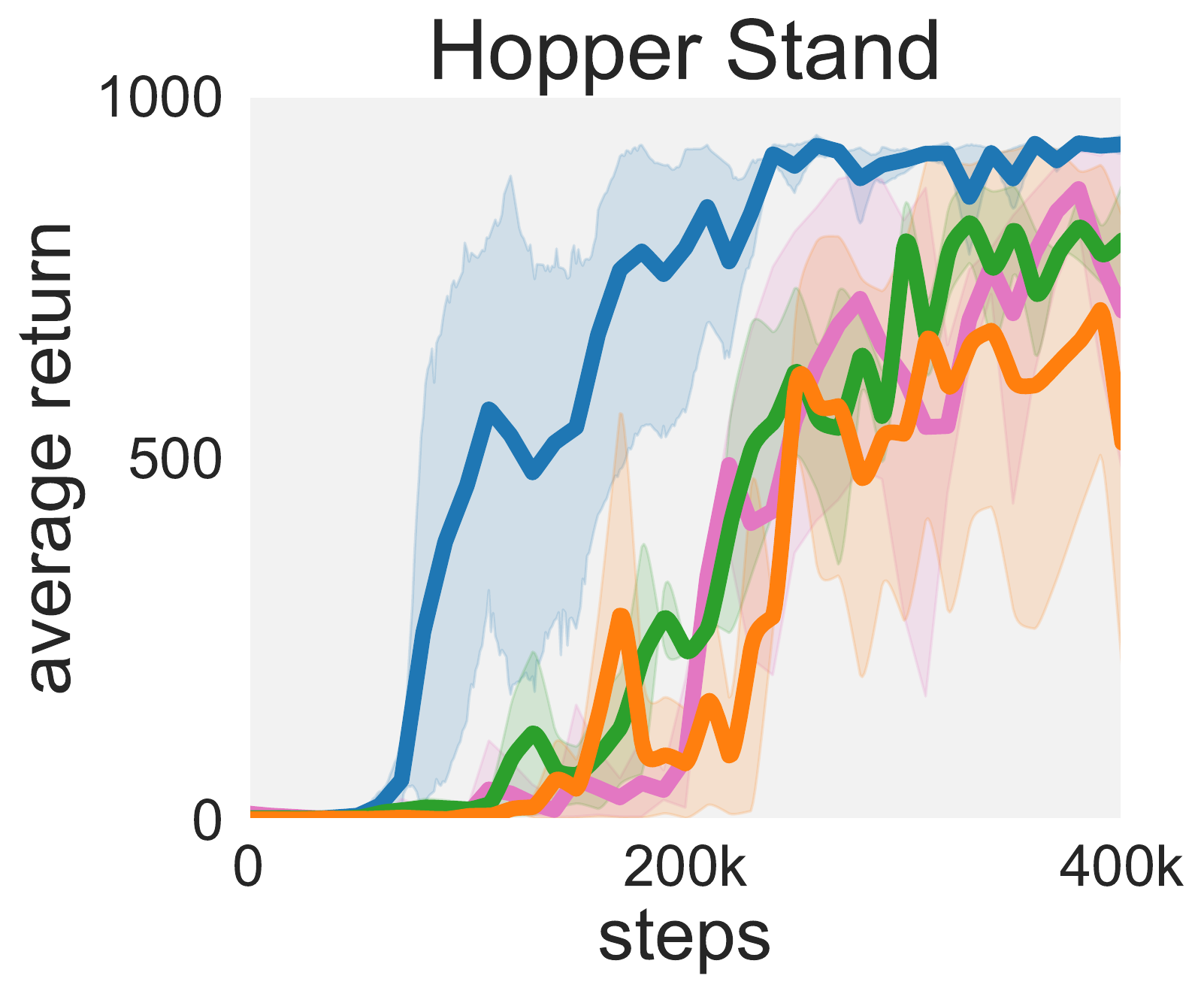} 
    \includegraphics[width=0.24\linewidth]{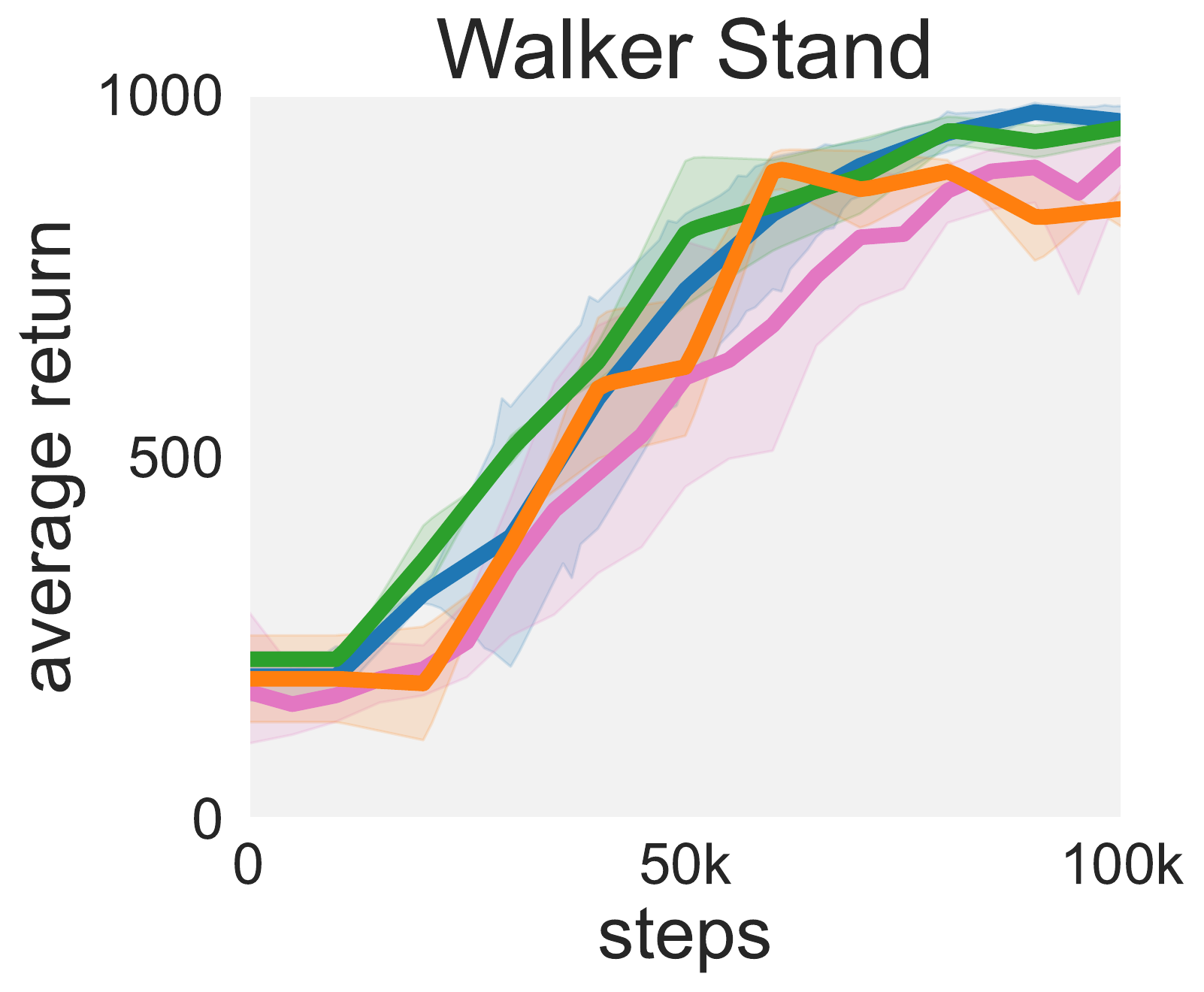} 
    \includegraphics[width=0.24\linewidth]{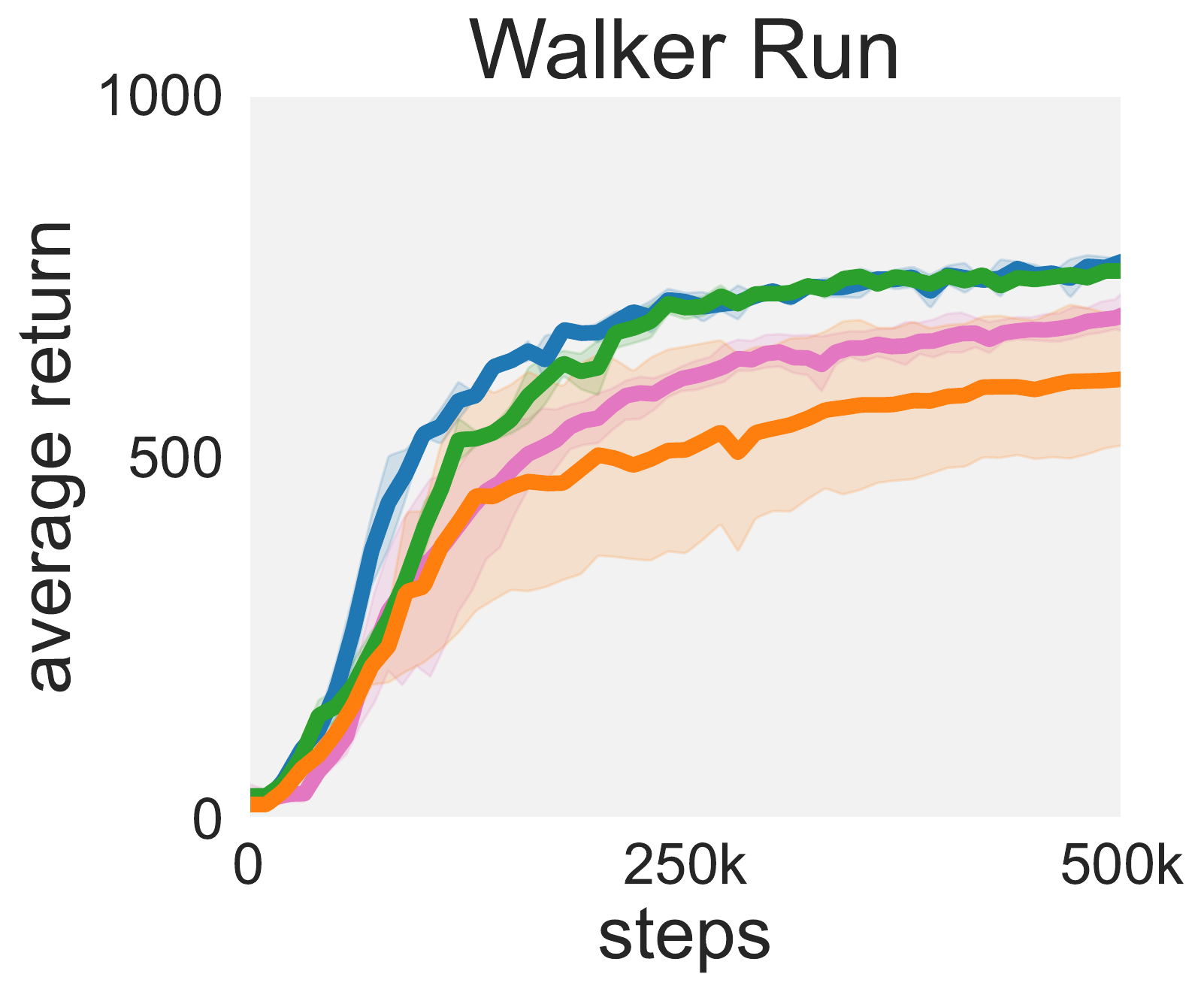} 
    \includegraphics[width=0.24\linewidth]{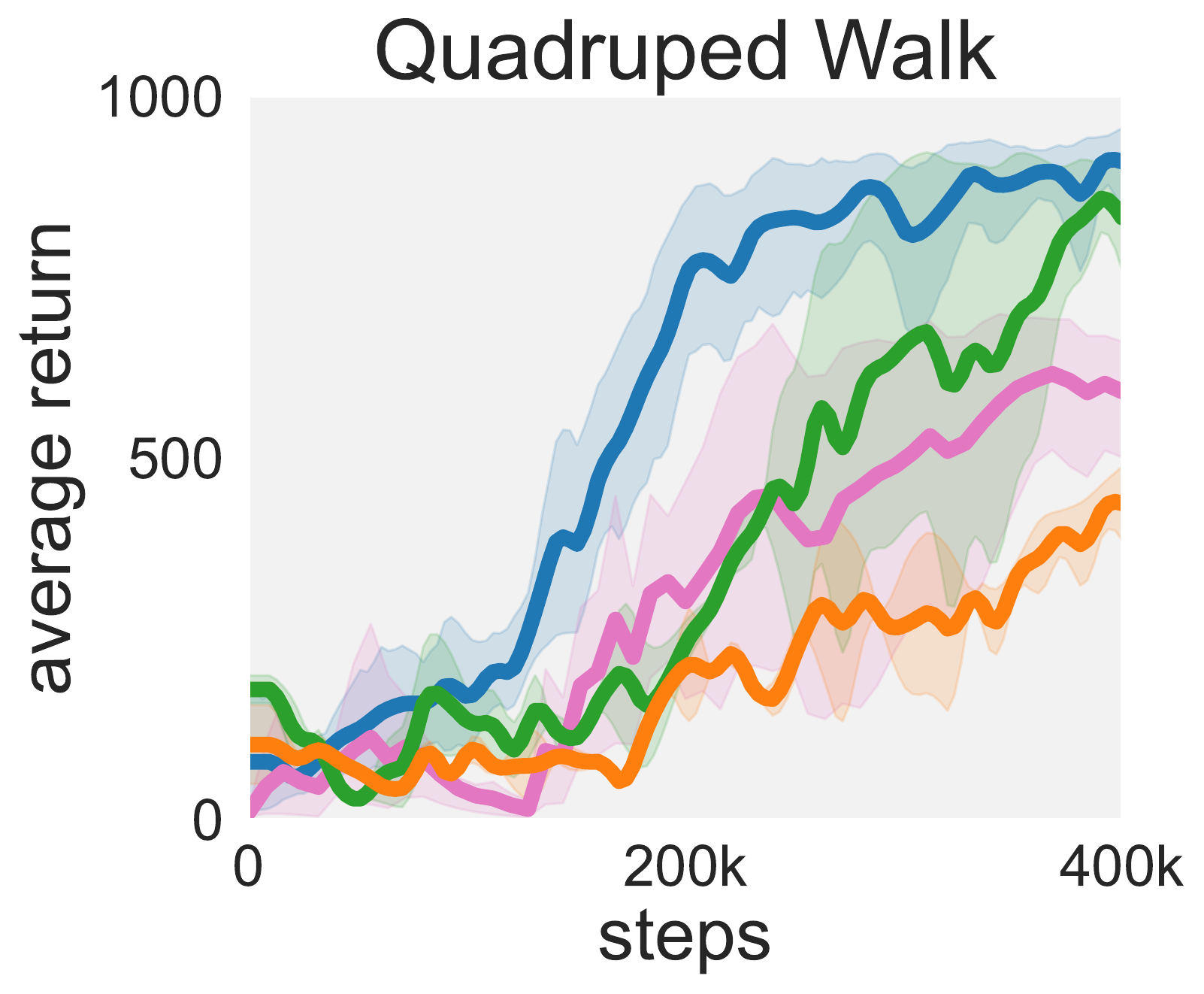} 
    \includegraphics[width=0.24\linewidth]{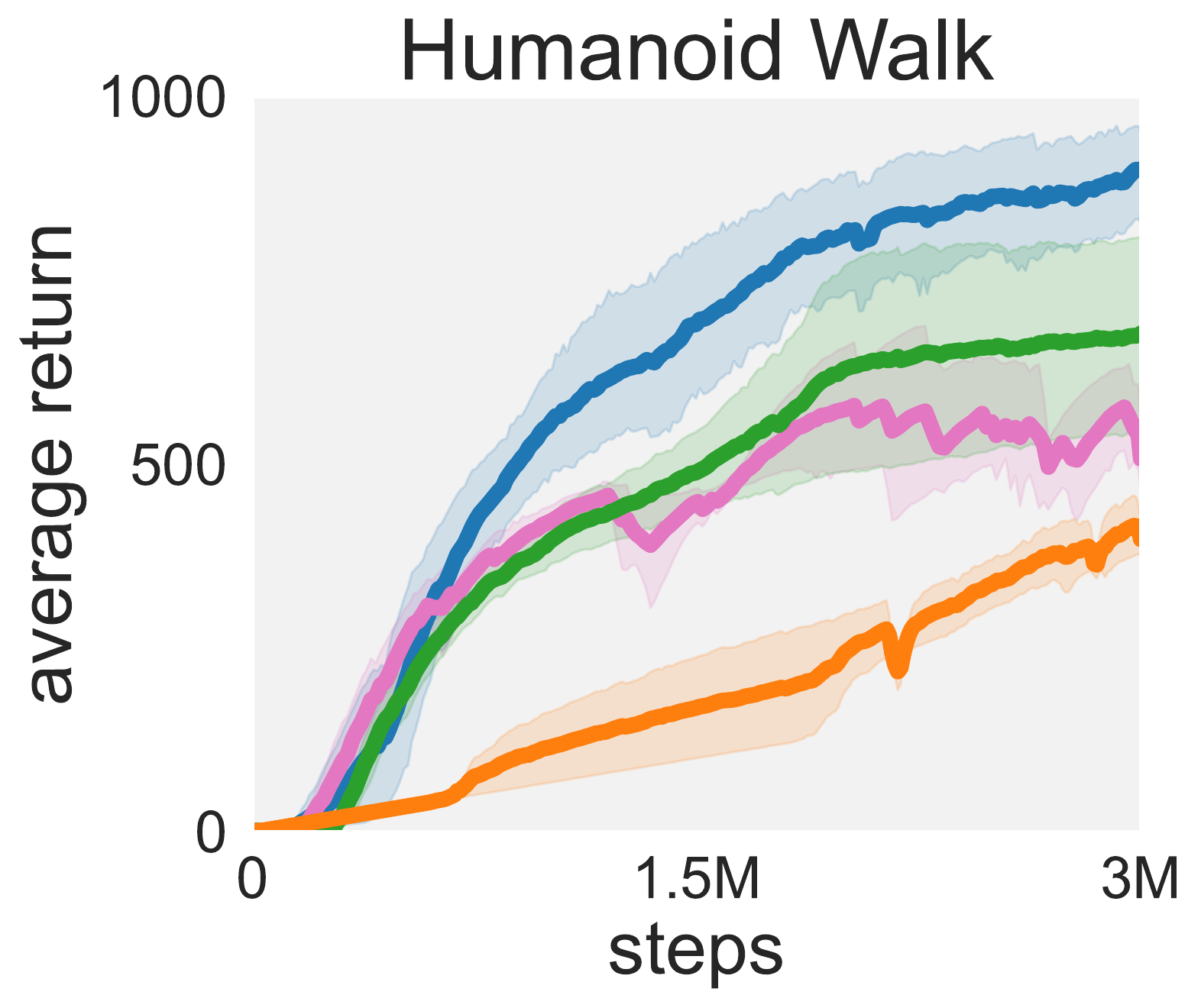} 
    \includegraphics[width=0.24\linewidth]{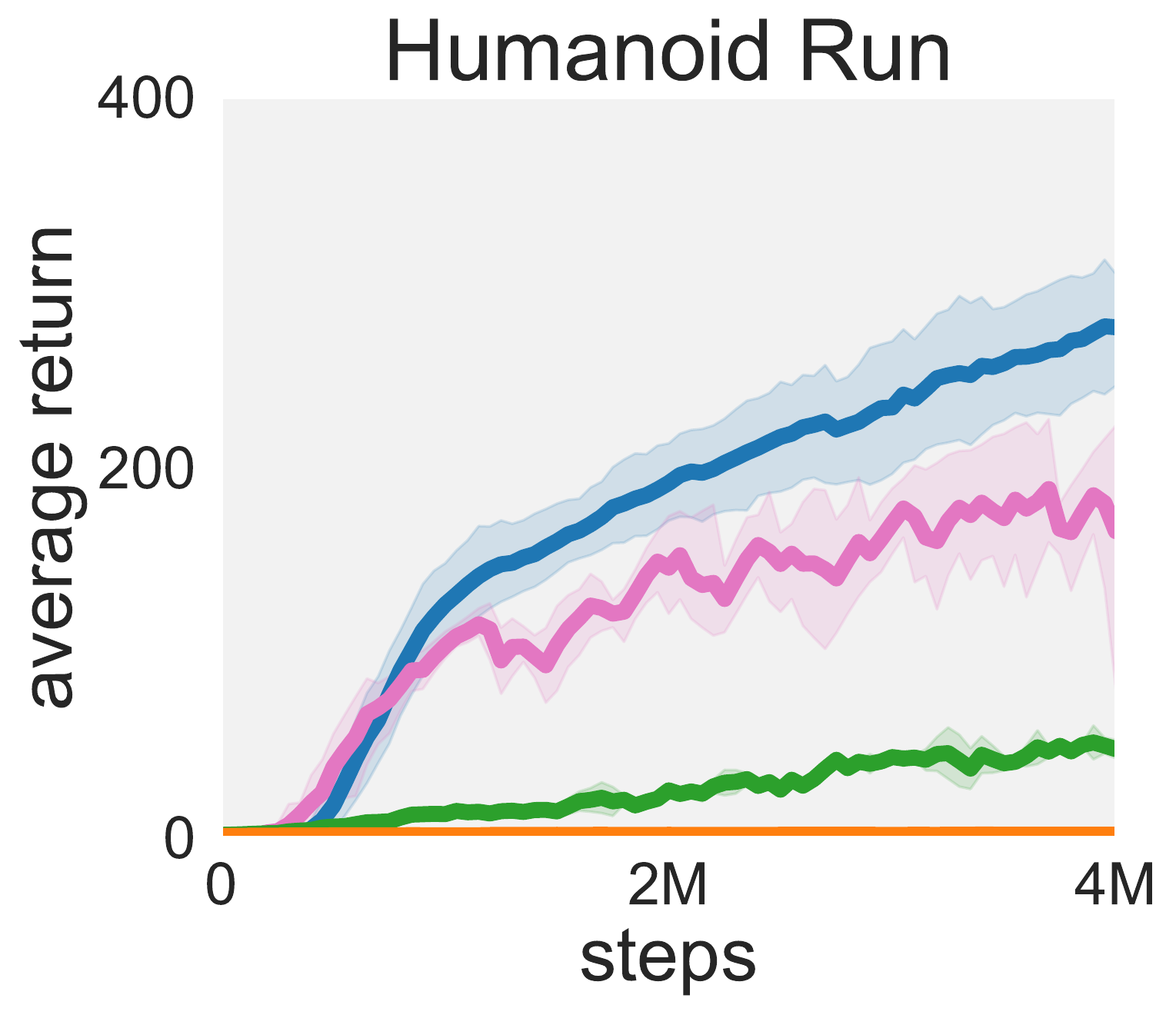} 
    \includegraphics[width=0.24\linewidth]{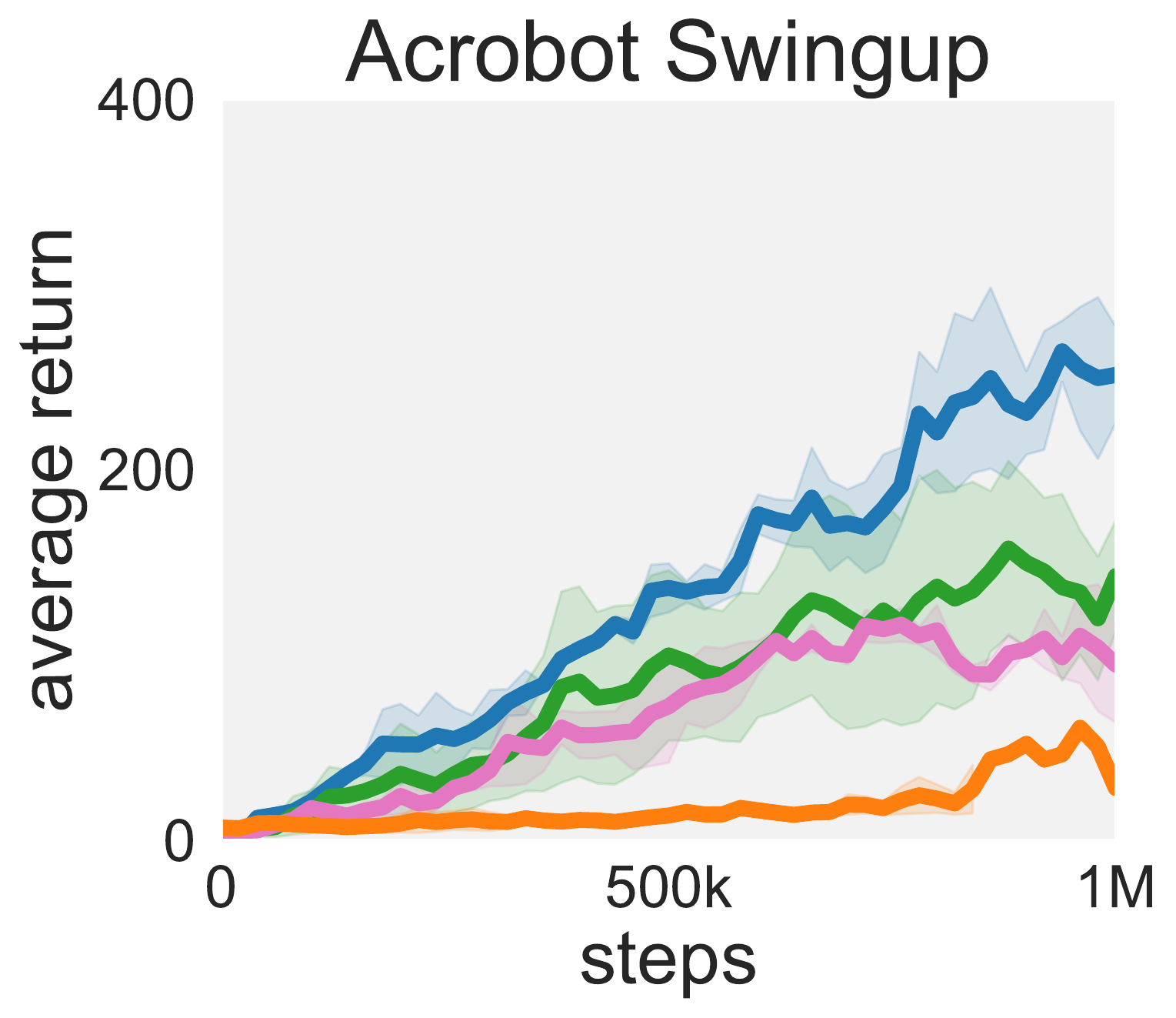} 
    \includegraphics[width=0.24\linewidth]{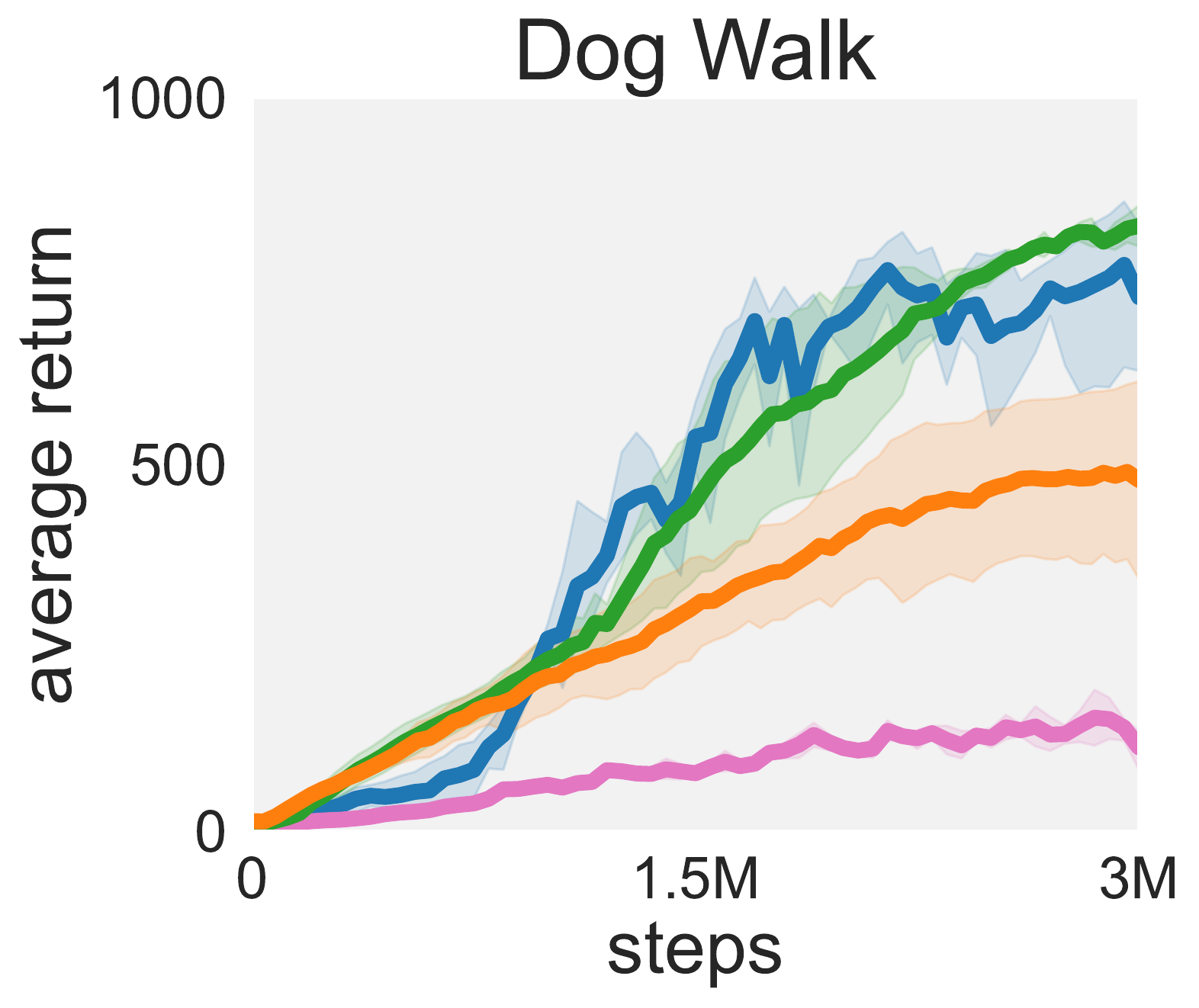} 
    \includegraphics[width=0.24\linewidth]{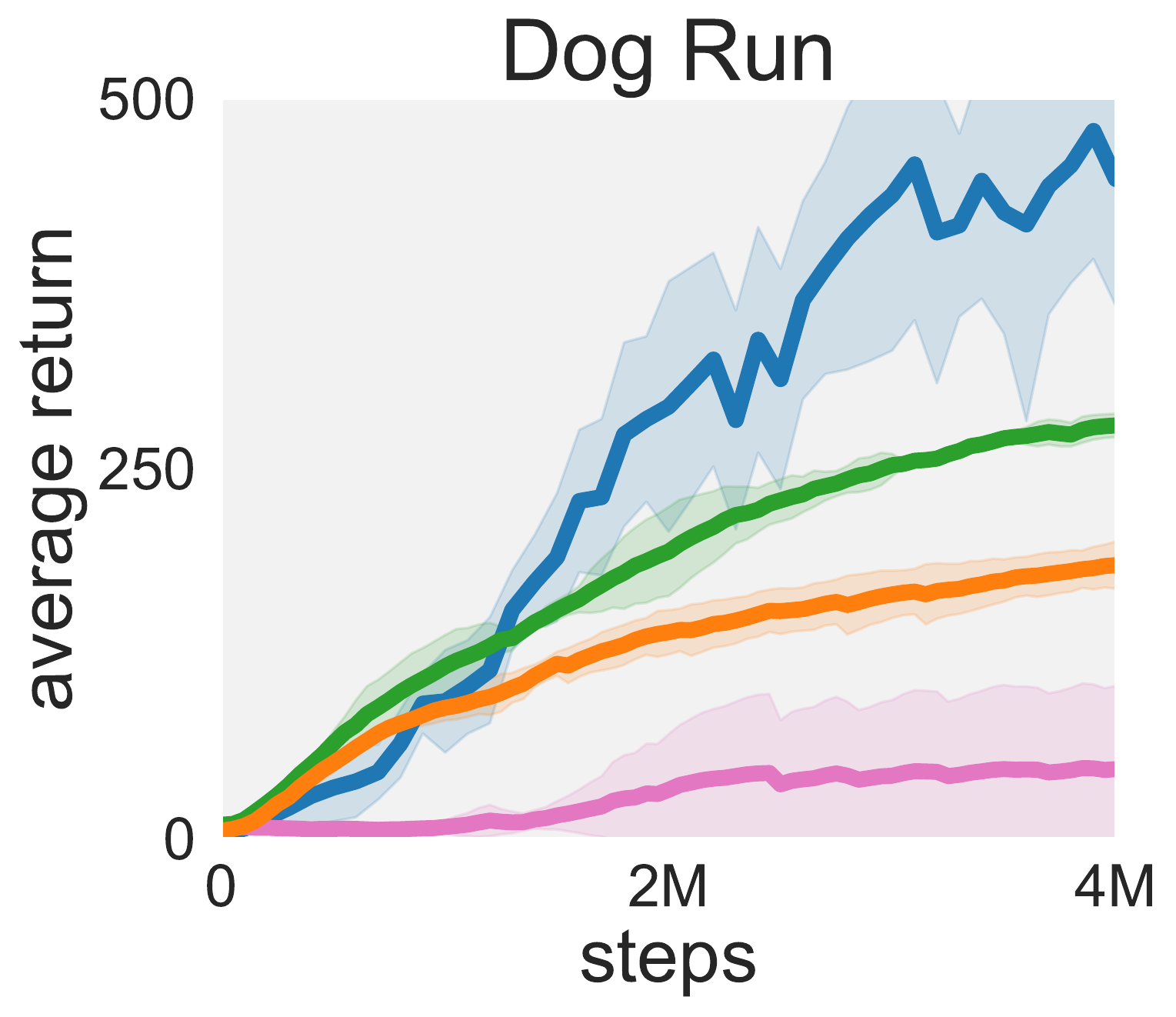}
    \includegraphics[width=0.24\linewidth]{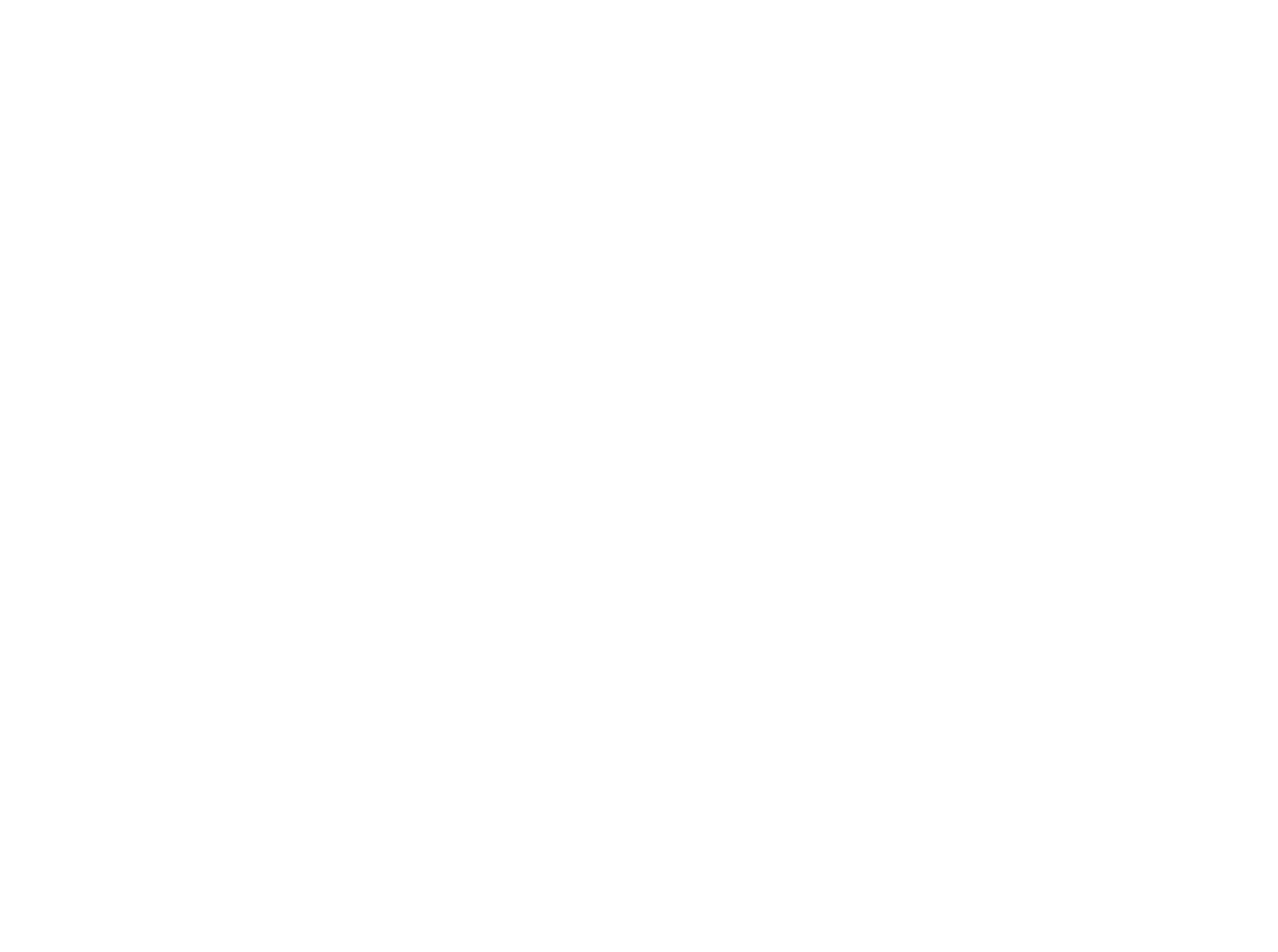}
    \includegraphics[width=1.0\linewidth]{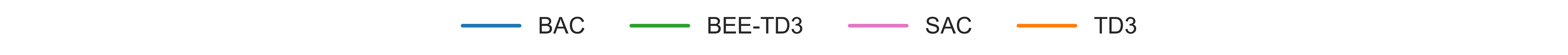}
    \caption{\textbf{DMControl tasks.} Training curves of \ourshort\ , BEE-TD3, SAC, TD3 in DMControl benchmark tasks. Solid curves depict the mean of ten trials and shaded regions correspond to the one standard deviation. }
    \label{fig:dmcontrol}
\end{figure}

\paragraph{Performance comparison. }
Training curves for 15 DMControl tasks are shown in Figure~\ref{fig:dmcontrol}. 
For simple locomotion/manipulation tasks~(\emph{e.g.}, HopperStand, WalkerStand, CupCatch), we generally find that while SAC's eventual performance is competitive with BAC, BAC shows better sample efficiency. In the more complex, failure-prone tasks~(\emph{e.g.}, HumanoidWalk, HumanoidRun, DogWalk, and DogRun), BAC significantly surpasses SAC. 
As shown in the visualizations\footnote{Please refer to \textcolor{mydarkblue}{\url{https://jity16.github.io/BEE/}} for videos or \textcolor{mydarkblue}{Section D.4} for key frames.}, SAC struggles to learn meaningful behaviors in Dog Run, whereas the BAC agent yields superior performance. 

\clearpage
\subsection{Evaluation on Meta-World benchmark tasks}
\begin{figure}[ht]
    \centering
    \begin{minipage}{0.9\textwidth}
        \begin{subfigure}[t]{0.48\textwidth} \centering
            \includegraphics[height=3.0cm,keepaspectratio]{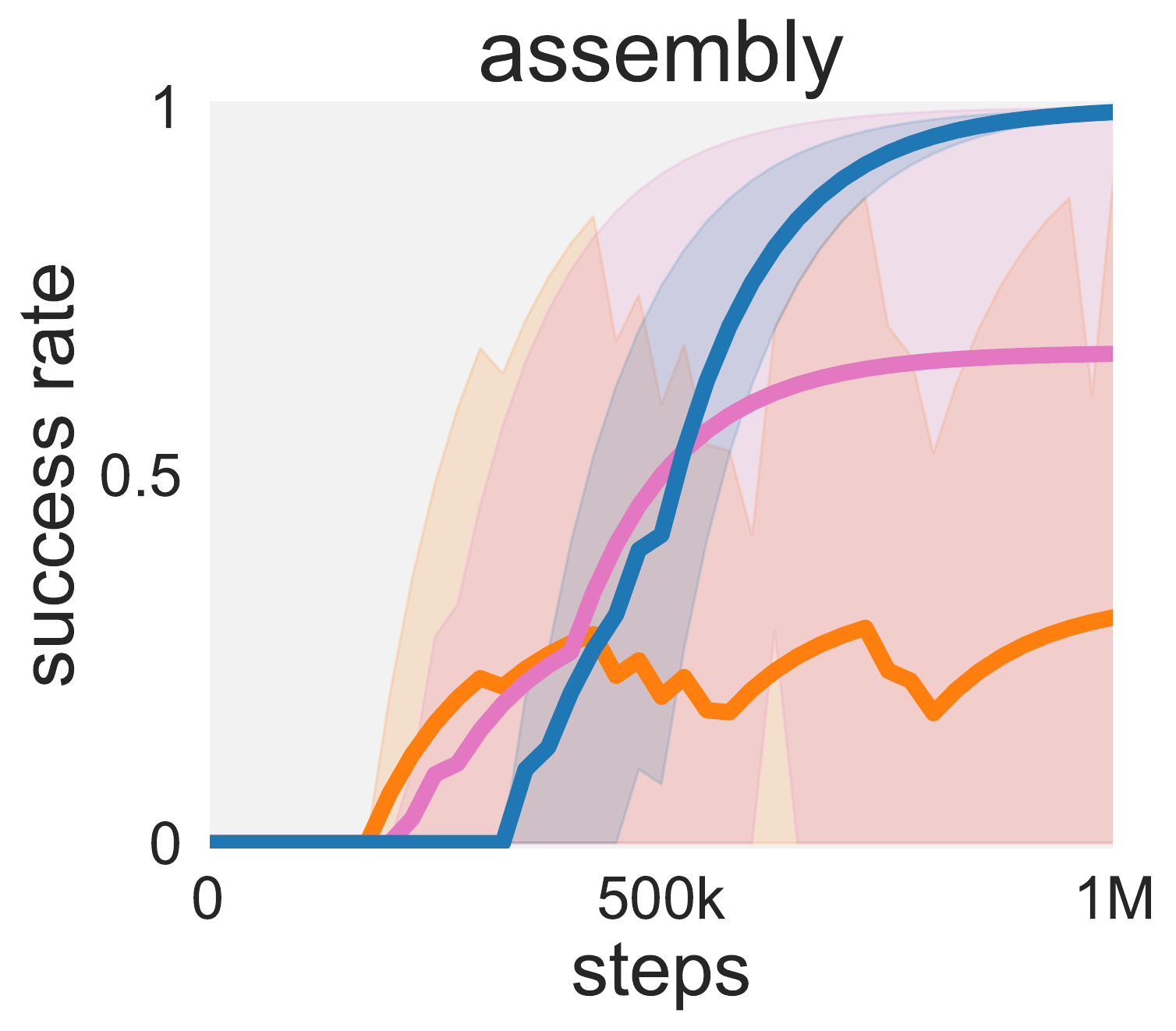}
            \includegraphics[height=3.0cm,keepaspectratio]{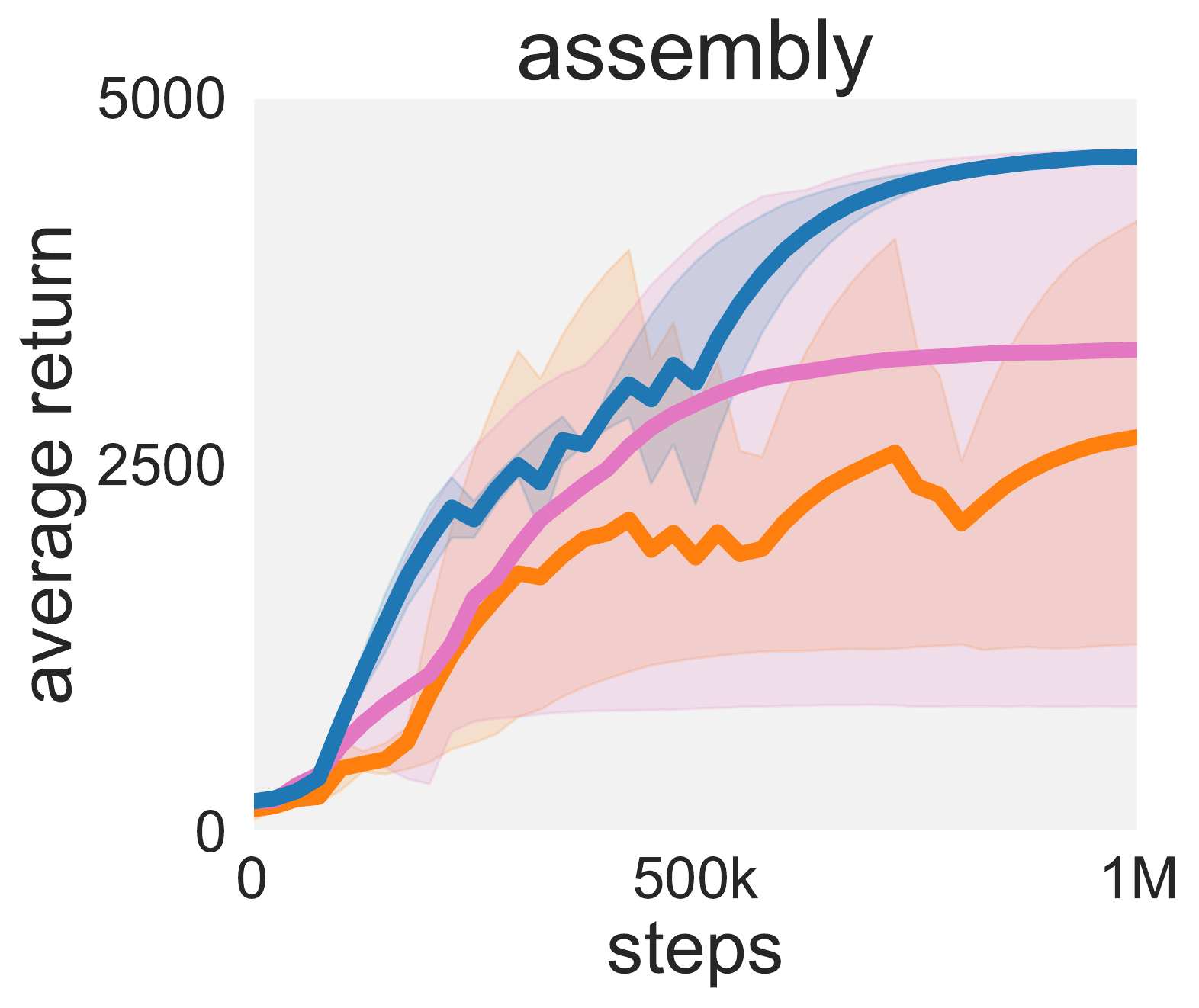}
        \end{subfigure}%
            \hfill    \begin{subfigure}[t]{0.48\textwidth} \centering
            \includegraphics[height=3.0cm,keepaspectratio]{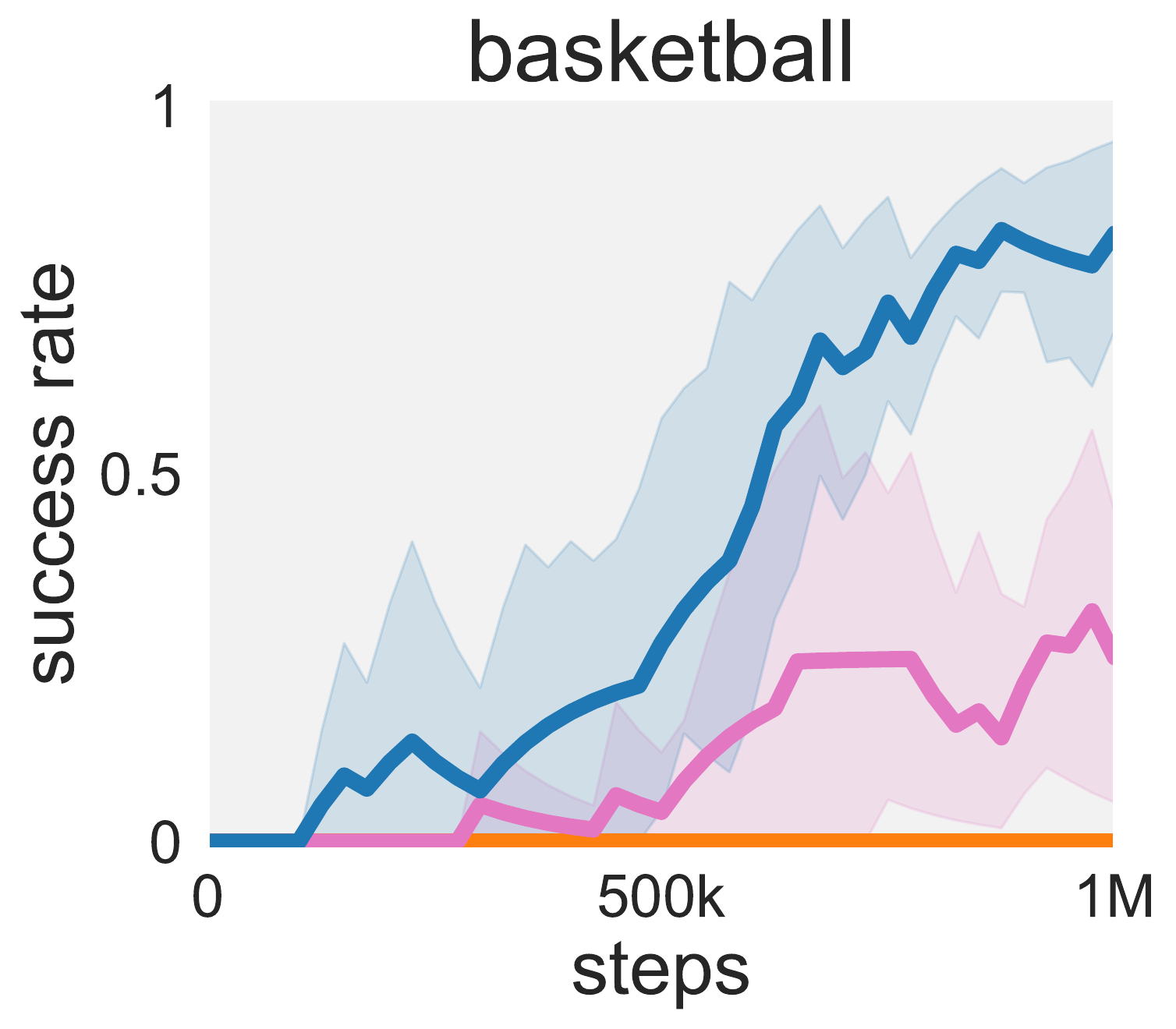}
            \includegraphics[height=3.0cm,keepaspectratio]{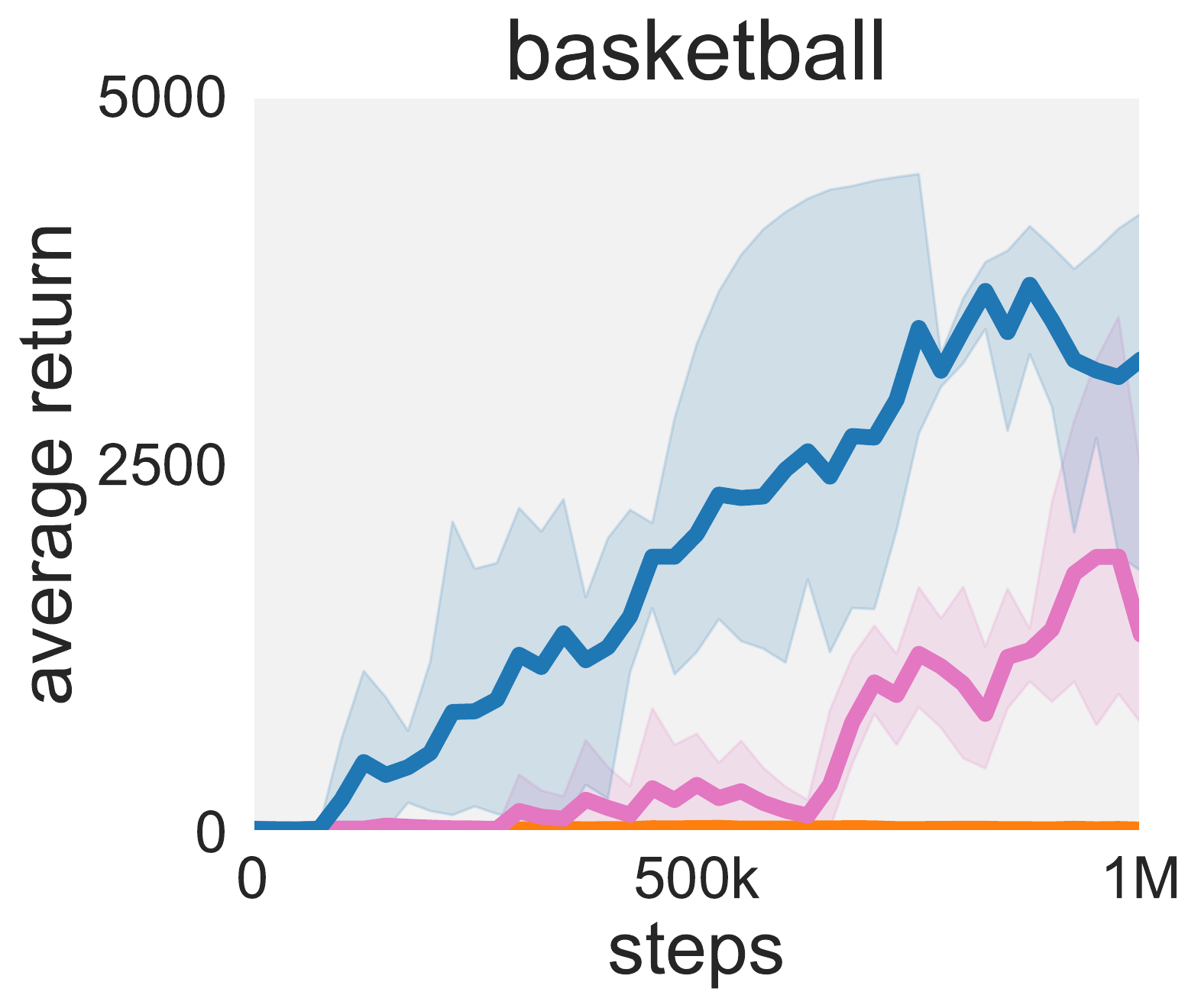}
        \end{subfigure}%
        \\ 
        \begin{subfigure}[t]{0.48\textwidth} \centering
            \includegraphics[height=3.0cm,keepaspectratio]{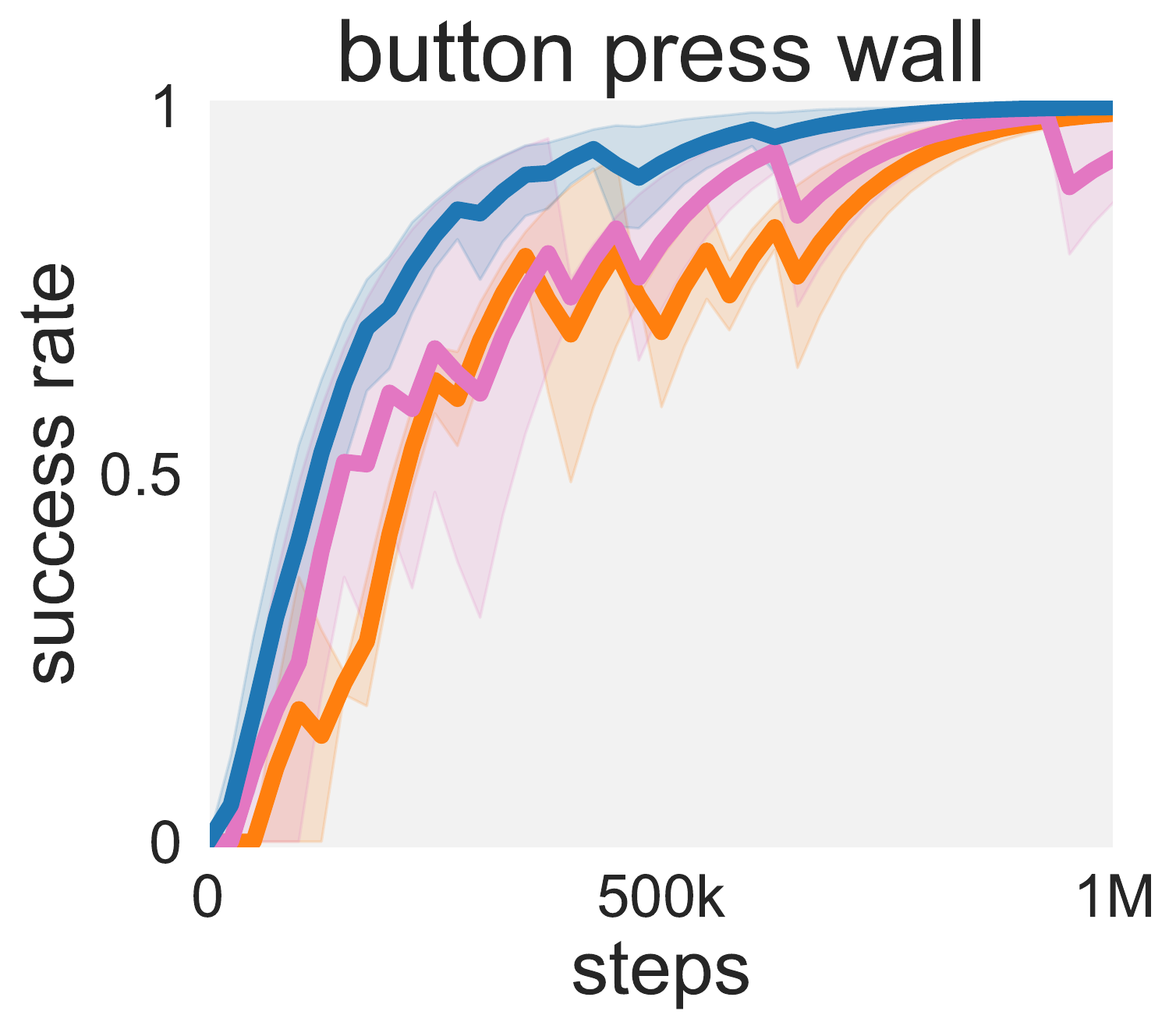}
            \includegraphics[height=3.0cm,keepaspectratio]{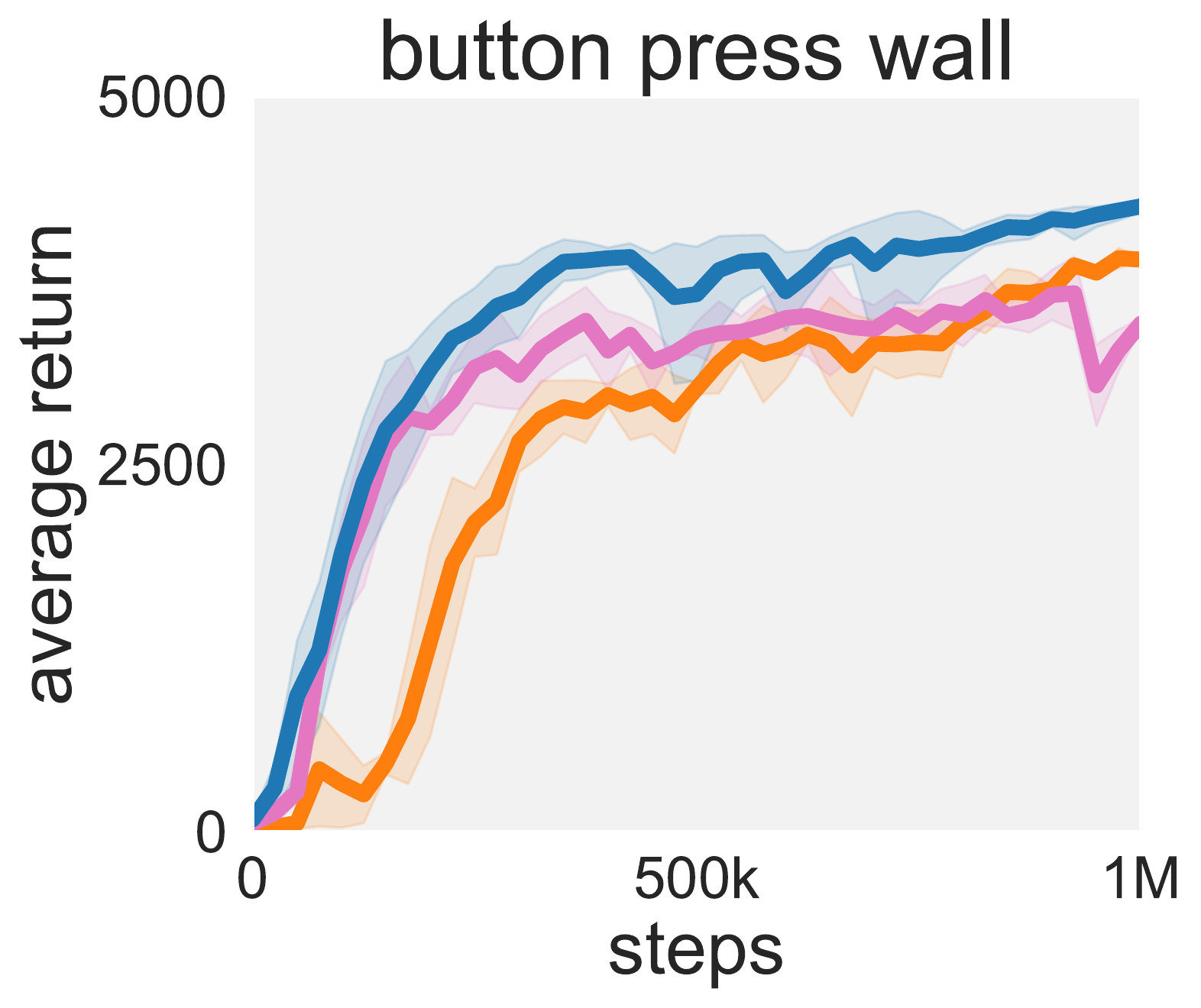}
        \end{subfigure}%
            \hfill    \begin{subfigure}[t]{0.48\textwidth} \centering
            \includegraphics[height=3.0cm,keepaspectratio]{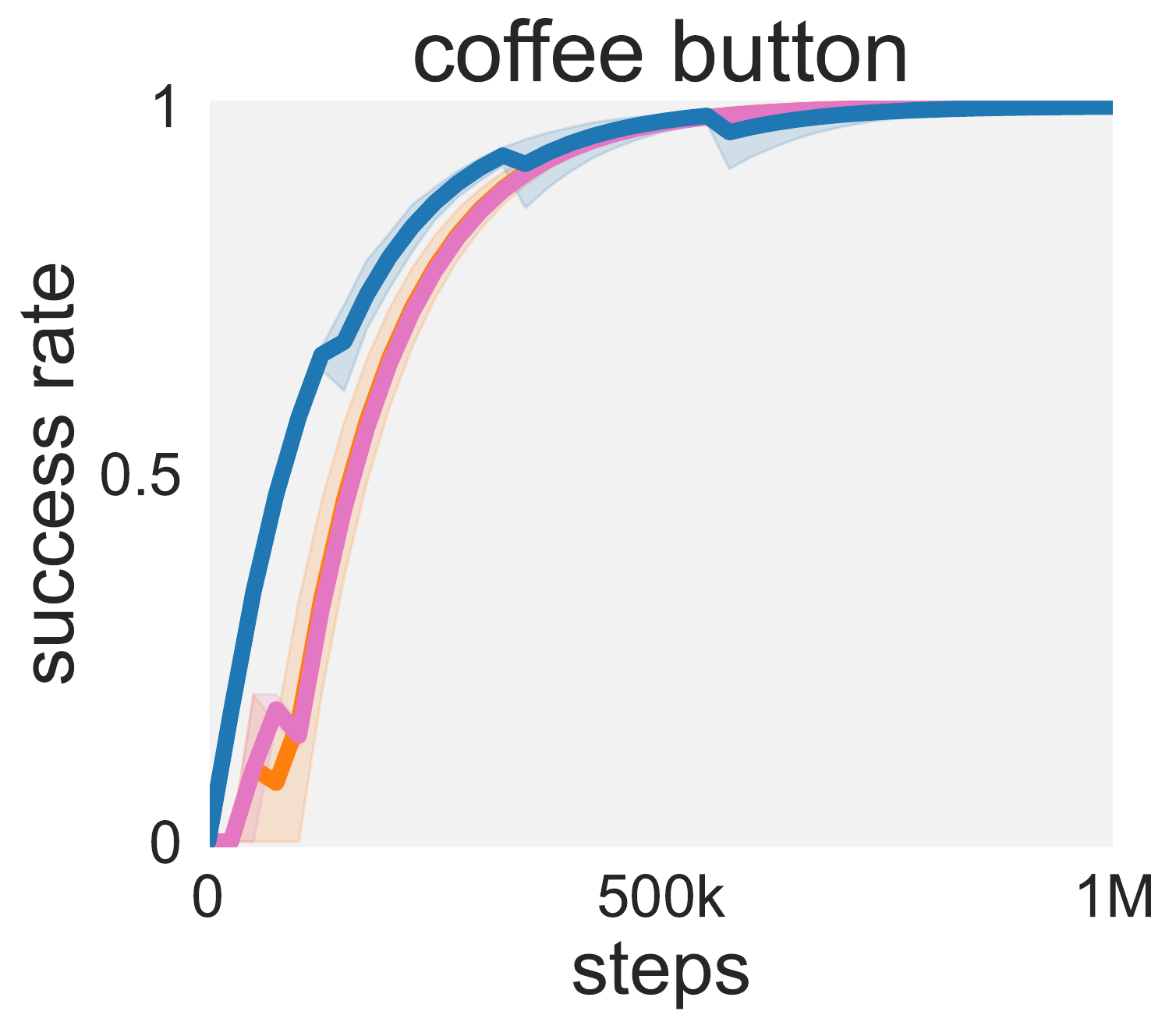}
            \includegraphics[height=3.0cm,keepaspectratio]{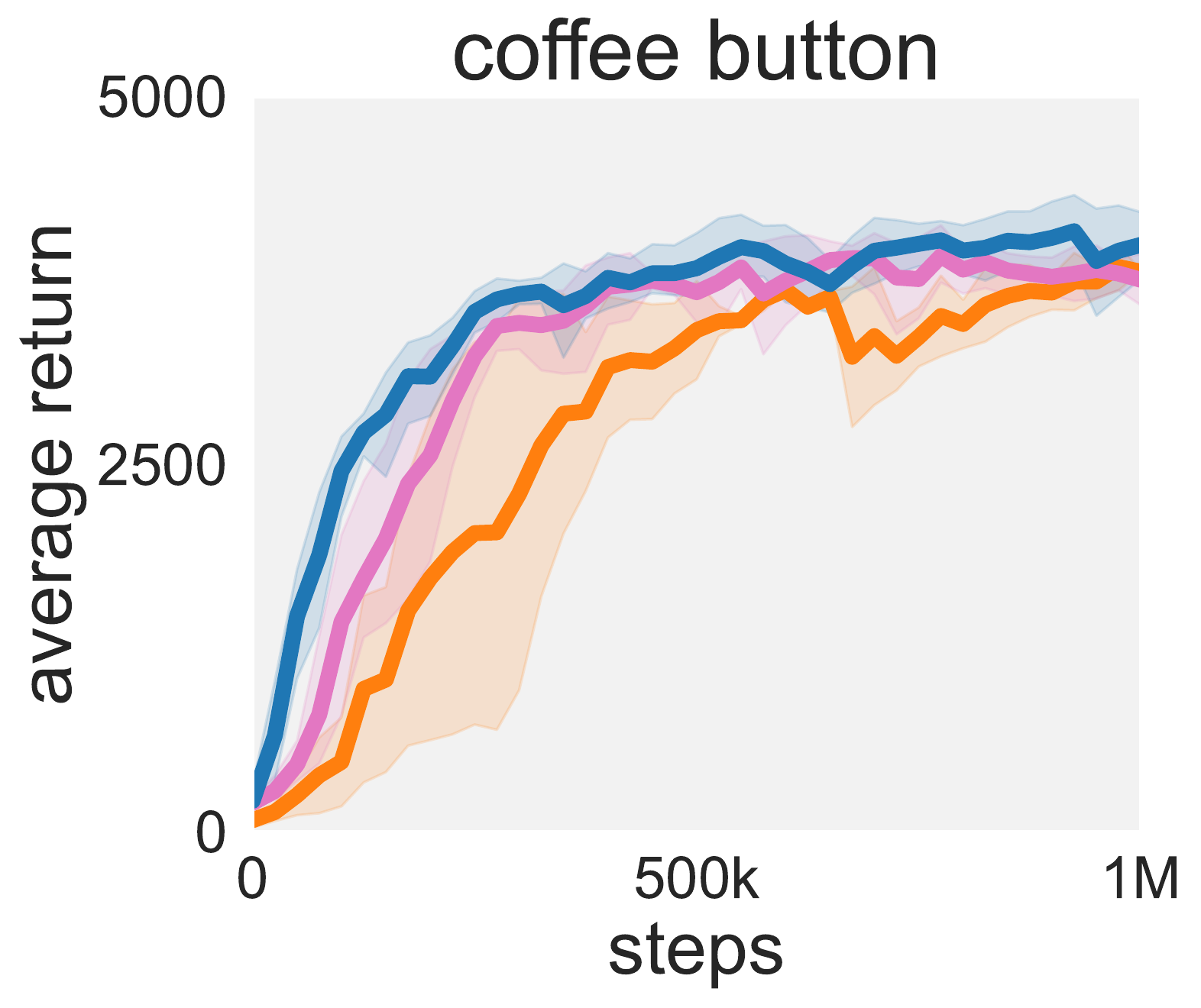}
        \end{subfigure}%
        \\ 
        \begin{subfigure}[t]{0.48\textwidth} \centering
            \includegraphics[height=3.0cm,keepaspectratio]{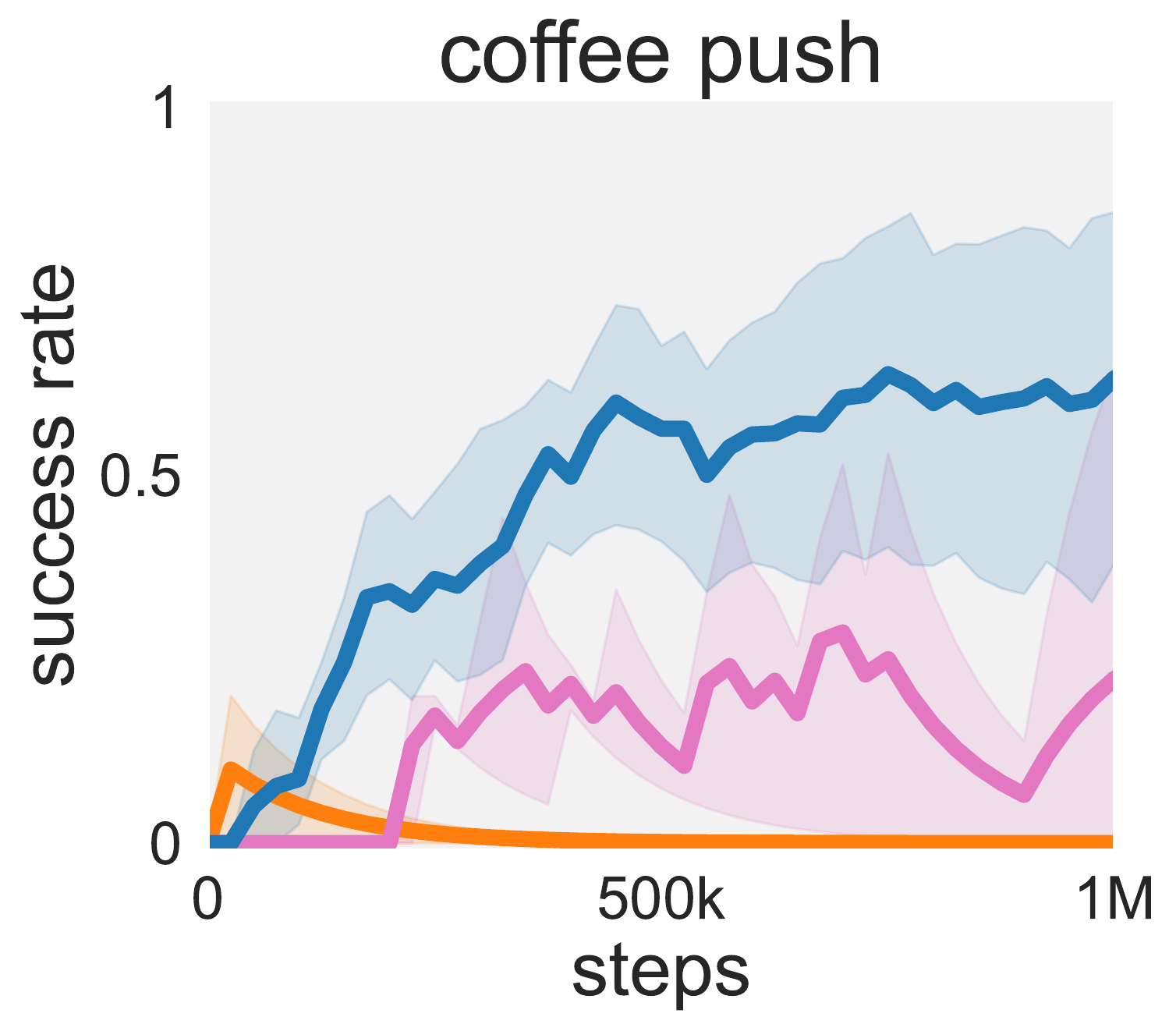}
            \includegraphics[height=3.0cm,keepaspectratio]{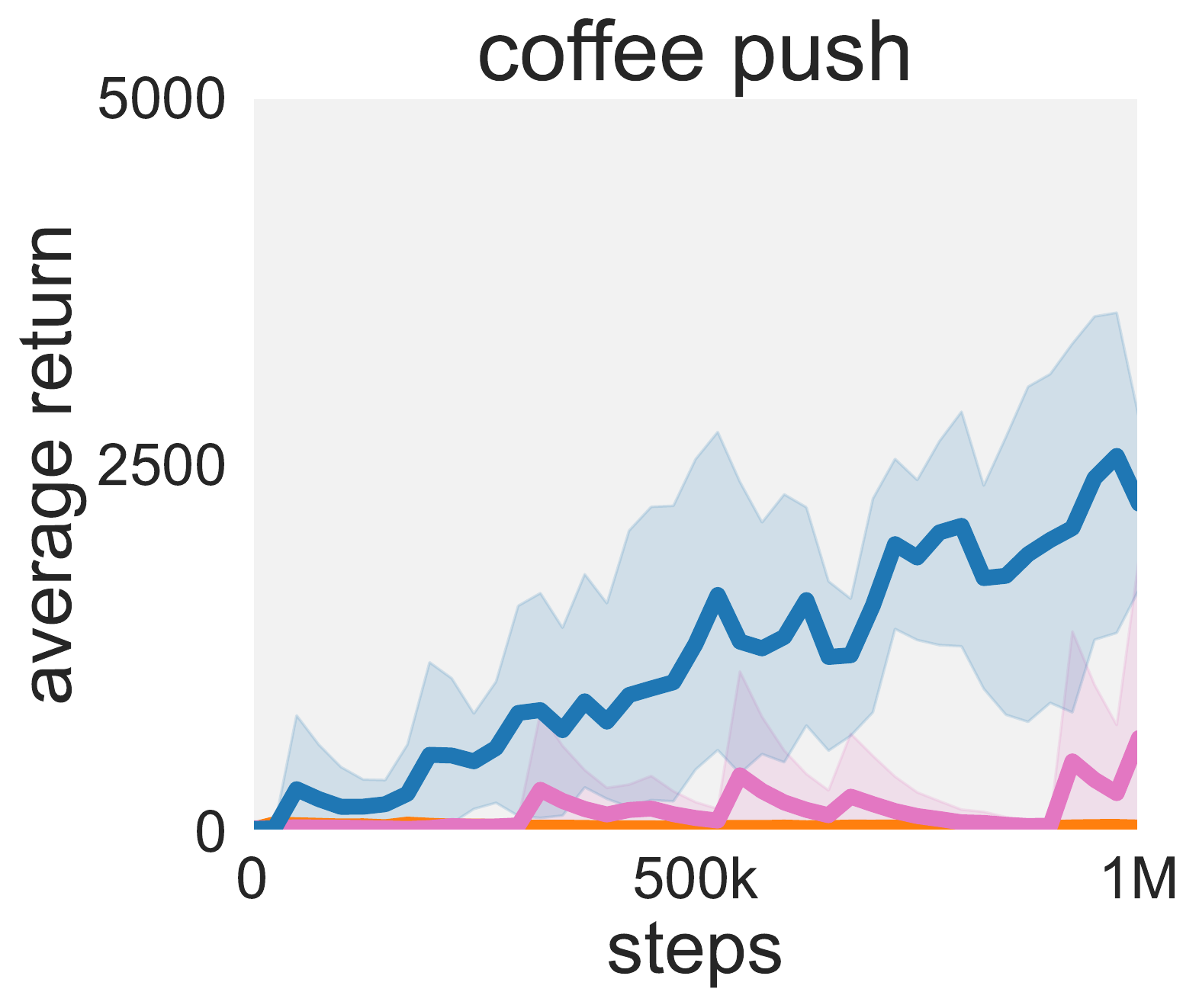}
        \end{subfigure}%
            \hfill    
            \begin{subfigure}[t]{0.48\textwidth} \centering
            \includegraphics[height=3.0cm,keepaspectratio]{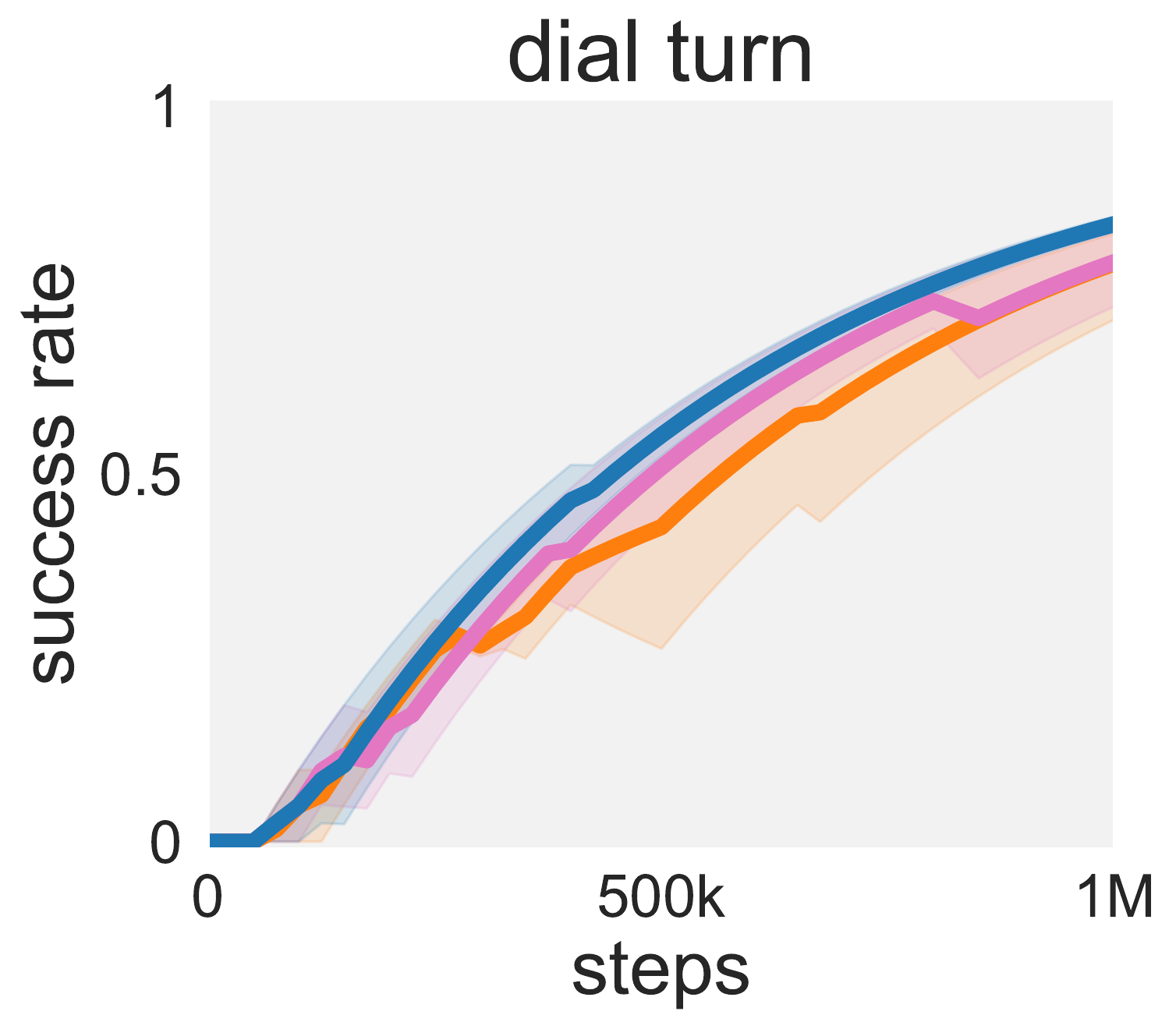}
            \includegraphics[height=3.0cm,keepaspectratio]{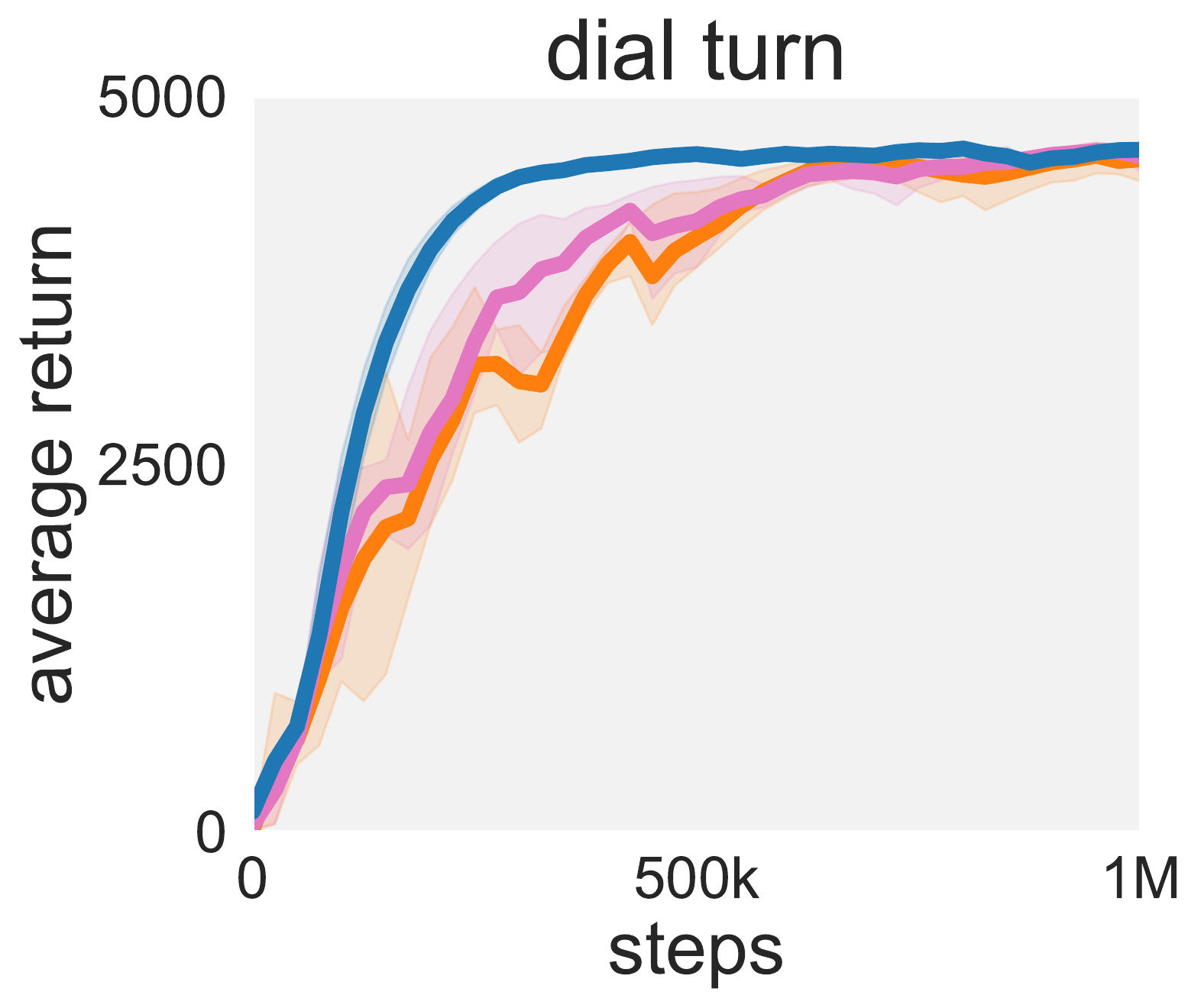}
        \end{subfigure}%
        \\ 
        \begin{subfigure}[t]{0.48\textwidth} \centering
            \includegraphics[height=3.0cm,keepaspectratio]{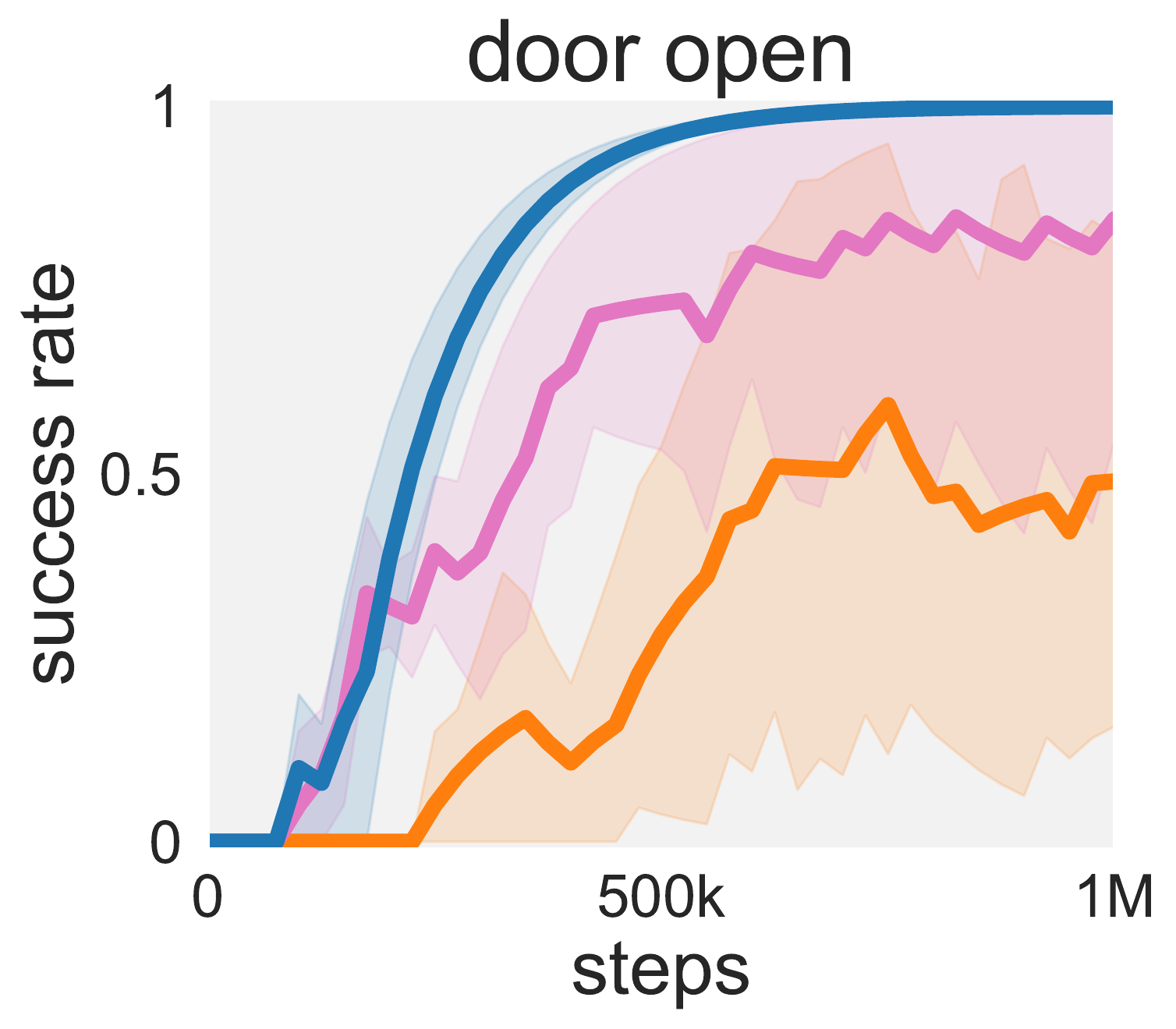}
            \includegraphics[height=3.0cm,keepaspectratio]{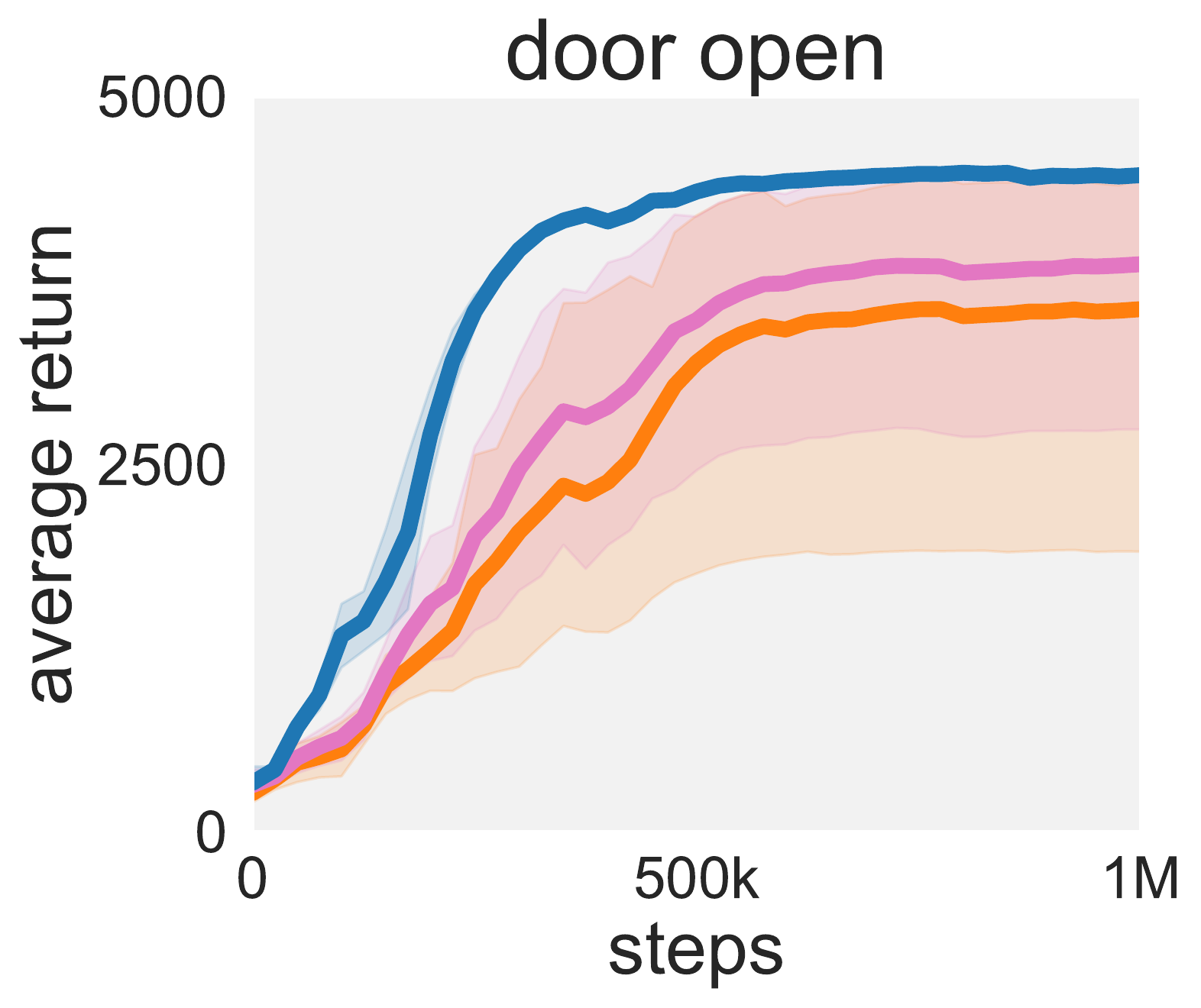}
        \end{subfigure}%
            \hfill    \begin{subfigure}[t]{0.48\textwidth} \centering
            \includegraphics[height=3.0cm,keepaspectratio]{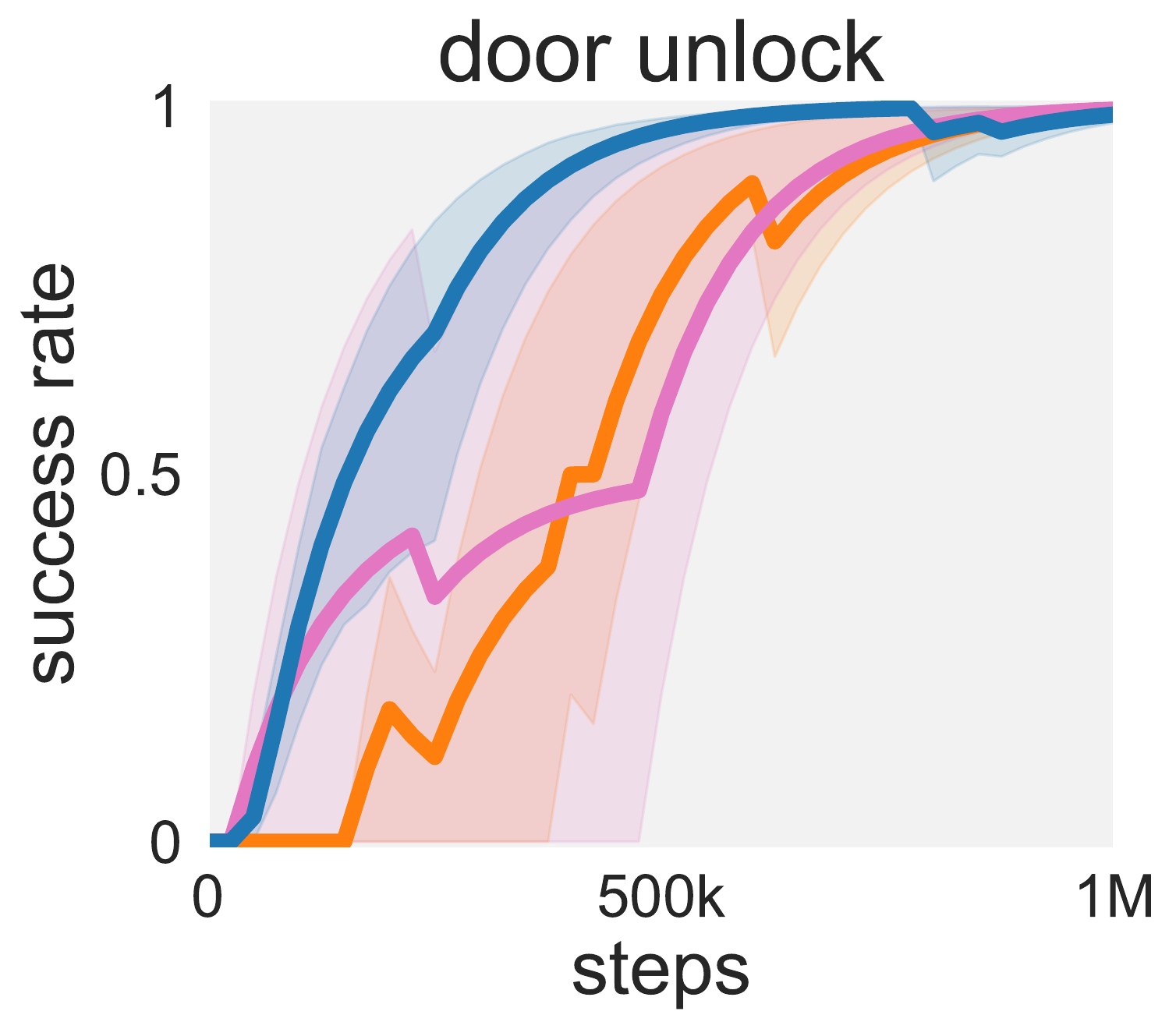}
            \includegraphics[height=3.0cm,keepaspectratio]{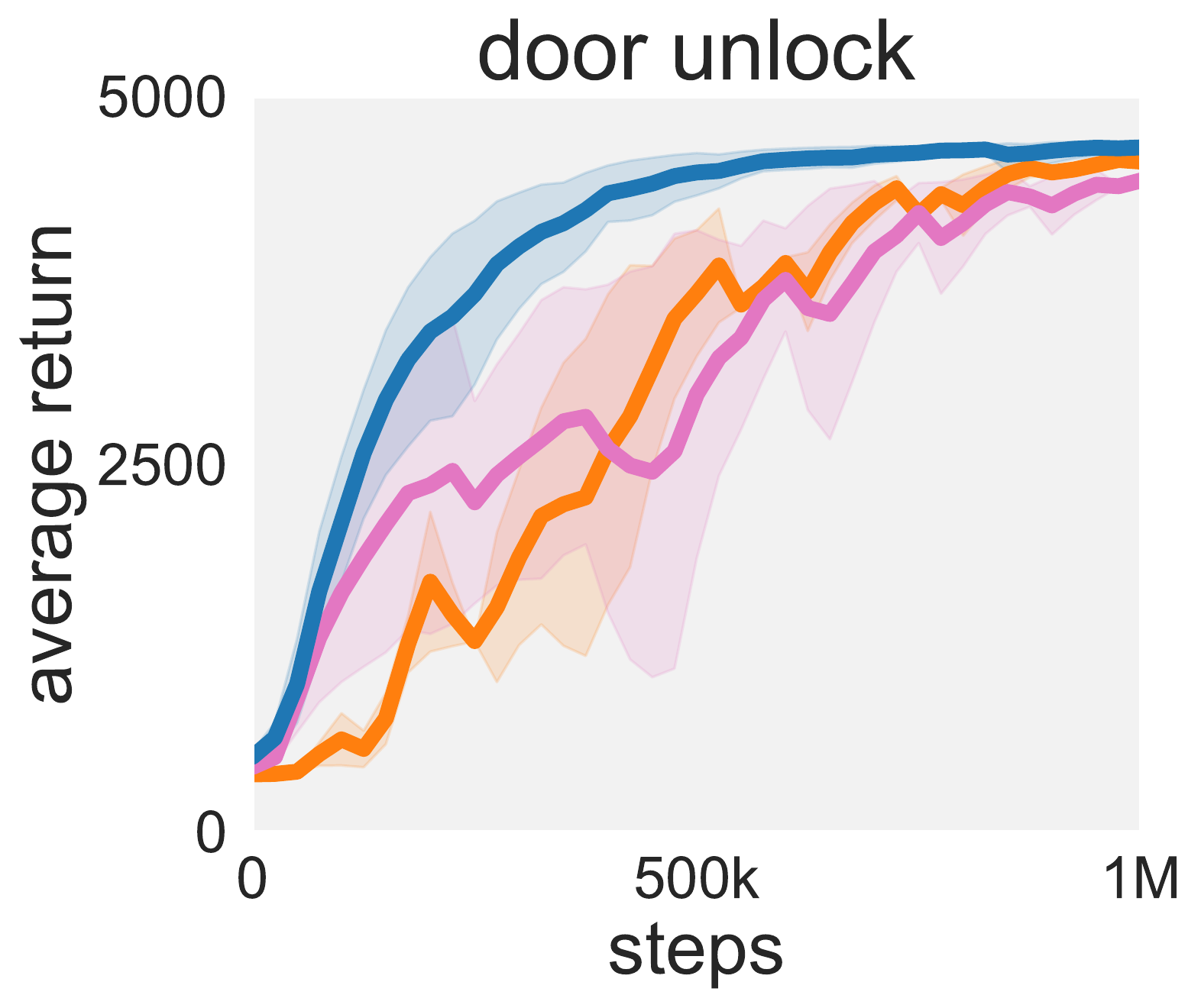}
        \end{subfigure}%
        \\ 
        \begin{subfigure}[t]{0.48\textwidth} \centering
            \includegraphics[height=3.0cm,keepaspectratio]{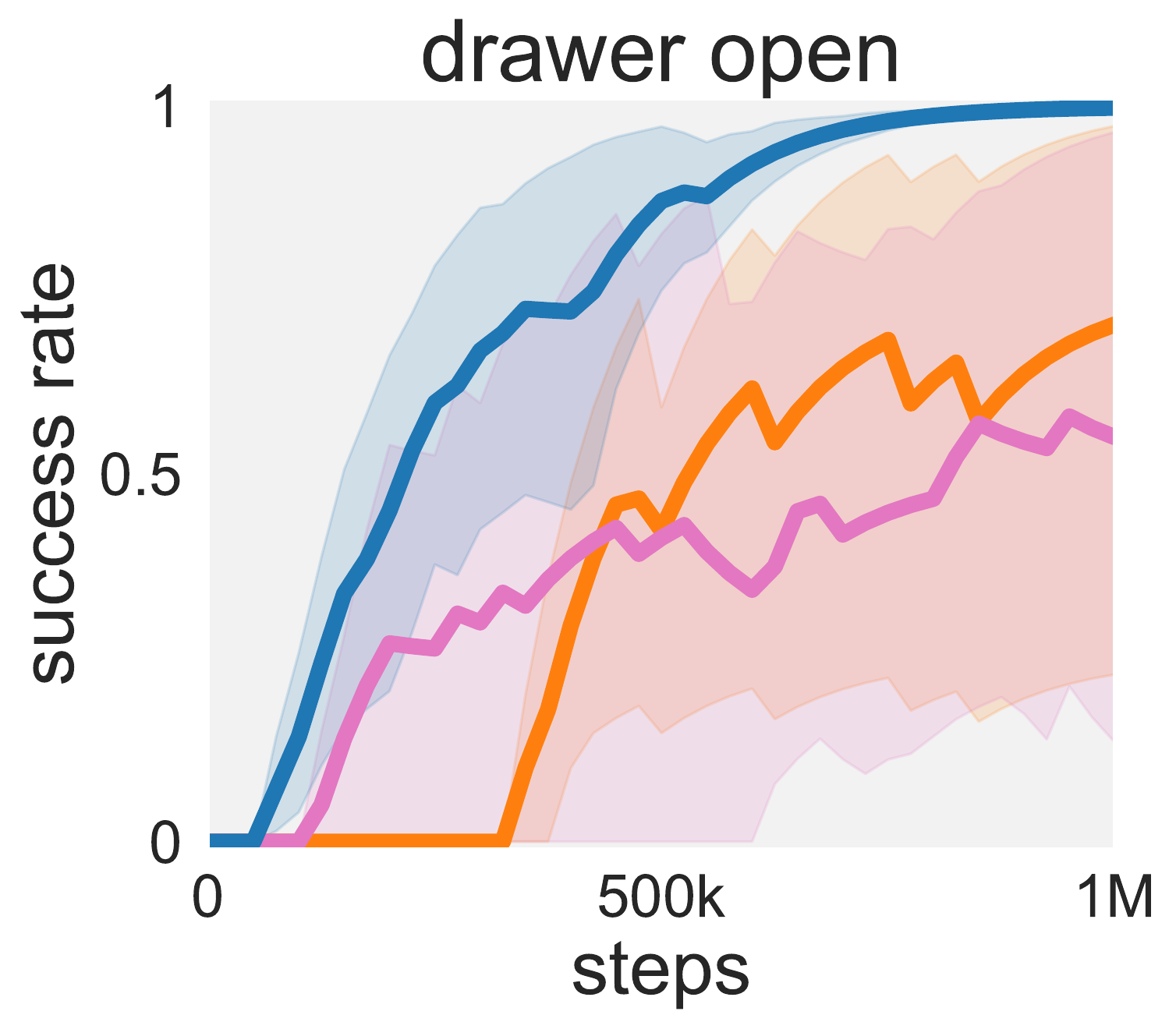}
            \includegraphics[height=3.0cm,keepaspectratio]{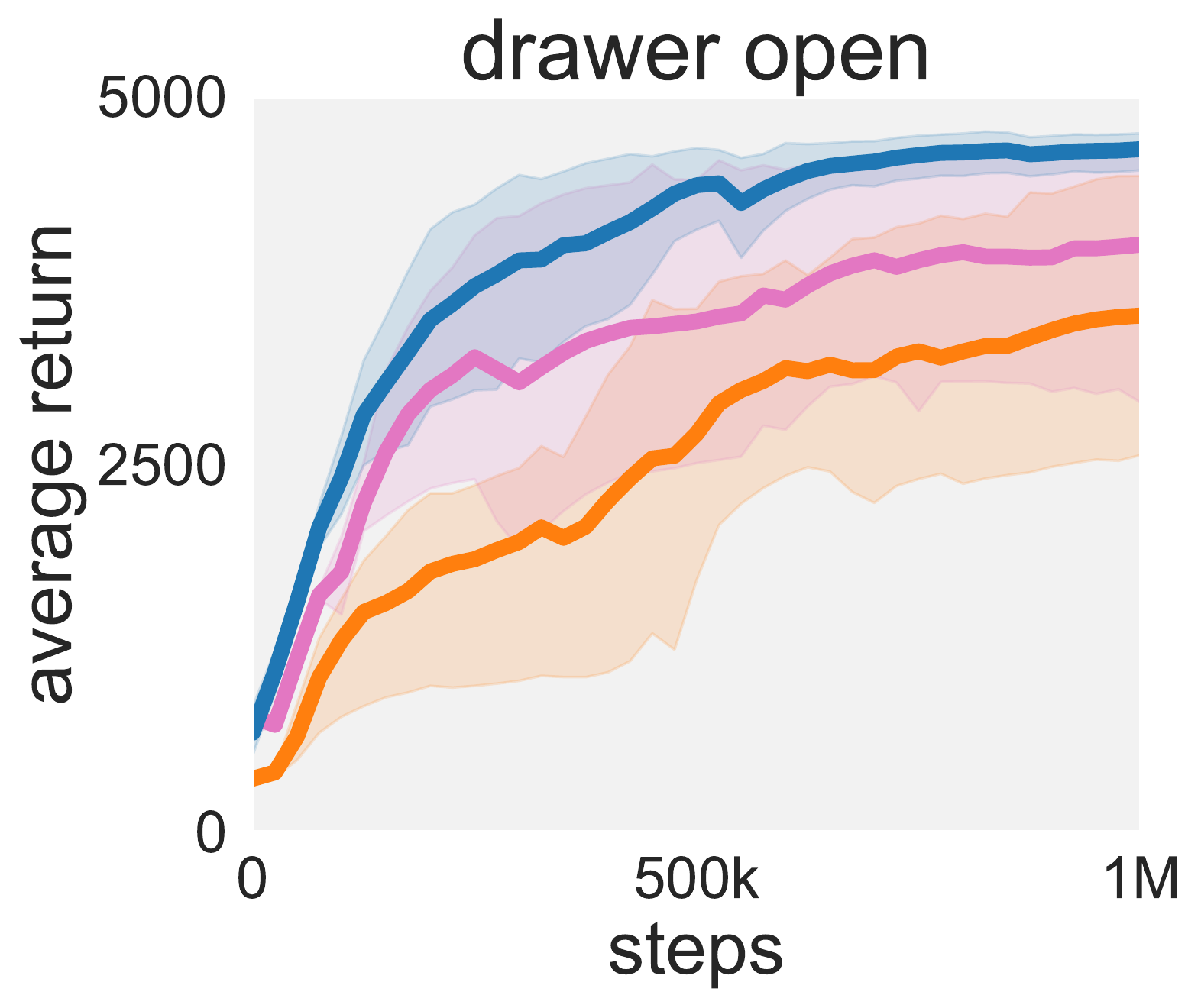}
        \end{subfigure}%
            \hfill    \begin{subfigure}[t]{0.48\textwidth} \centering
            \includegraphics[height=3.0cm,keepaspectratio]{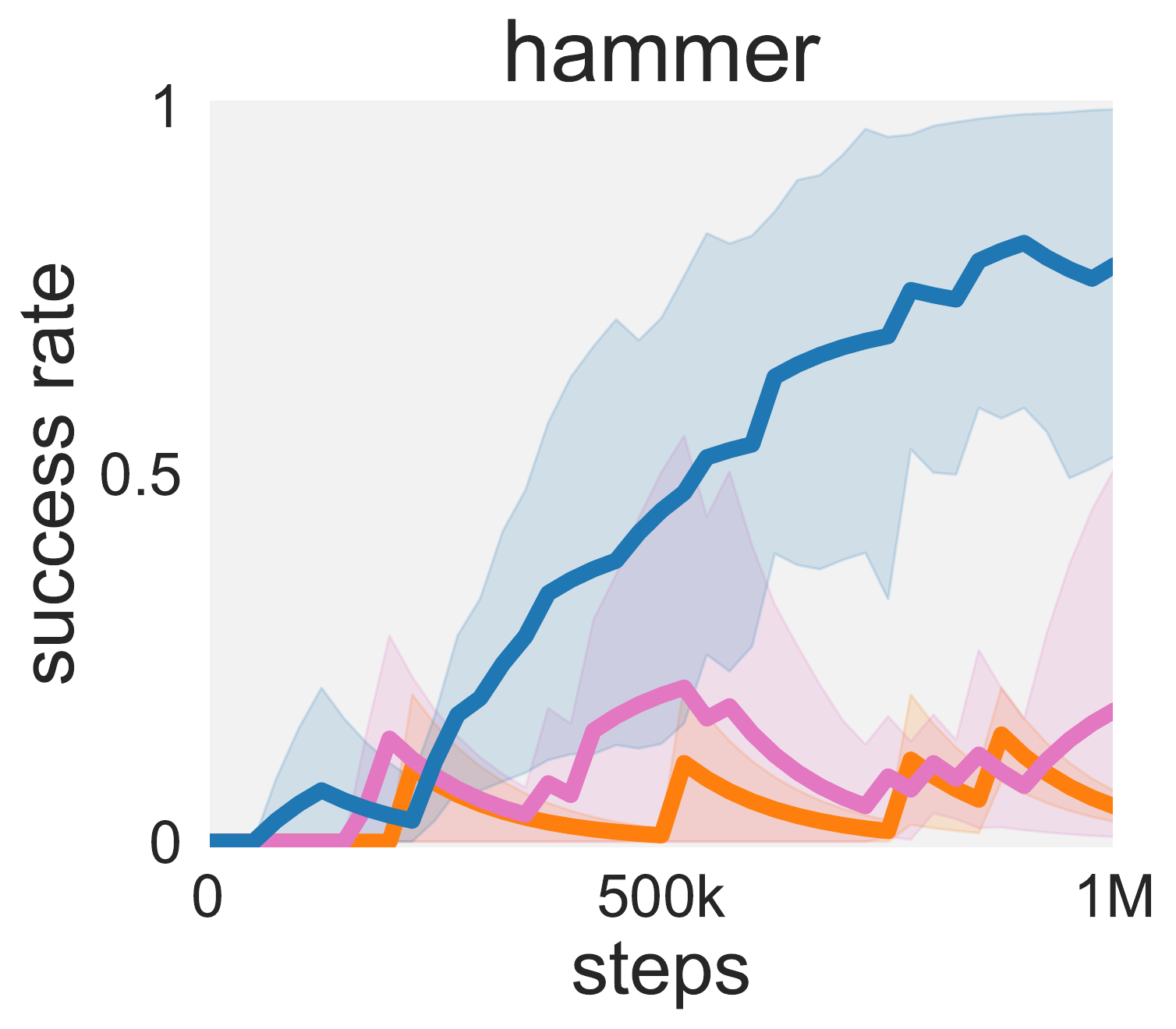}
            \includegraphics[height=3.0cm,keepaspectratio]{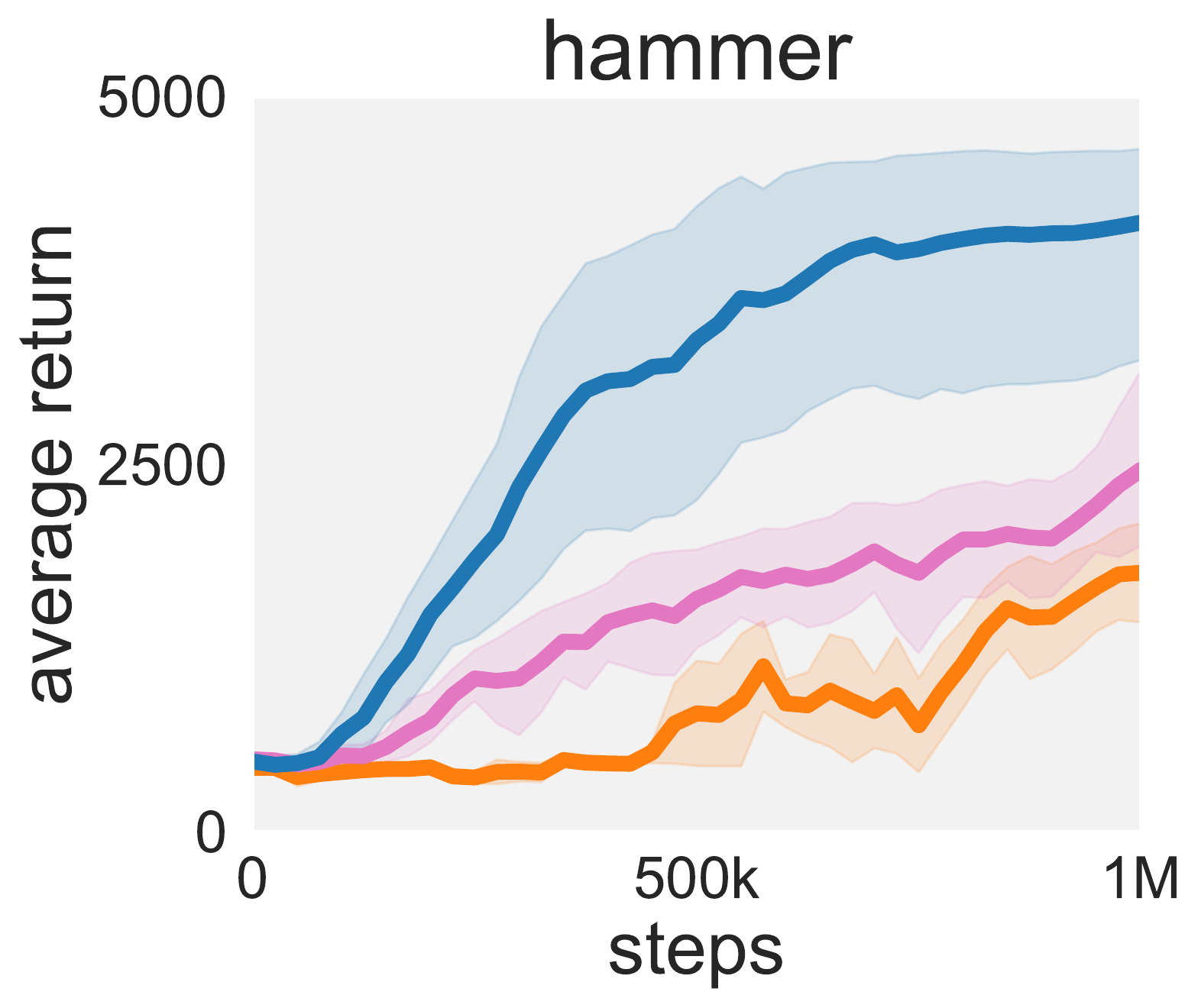}
        \end{subfigure}%
        \\ 
        \begin{subfigure}[t]{0.48\textwidth} \centering
            \includegraphics[height=3.0cm,keepaspectratio]{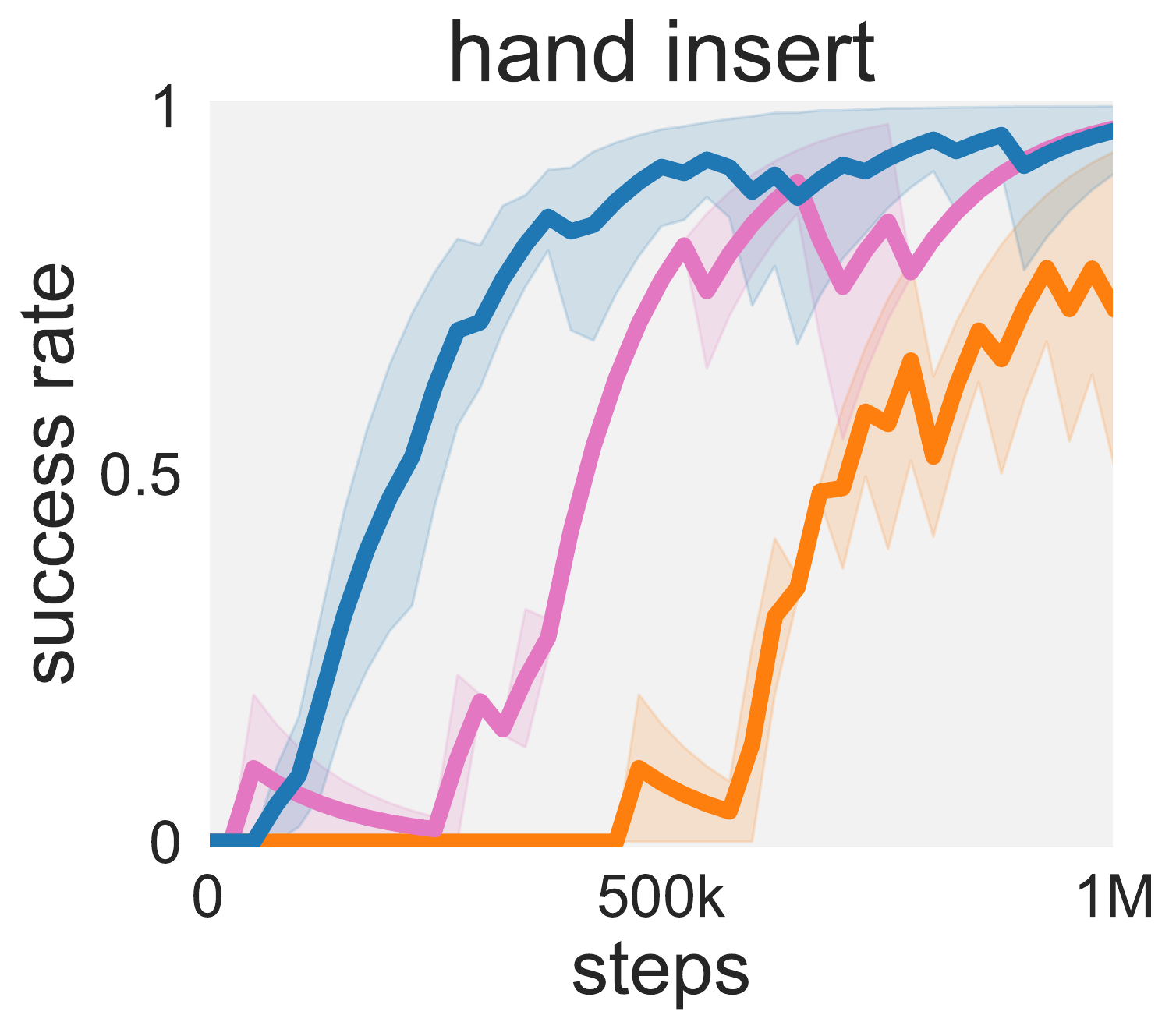}
            \includegraphics[height=3.0cm,keepaspectratio]{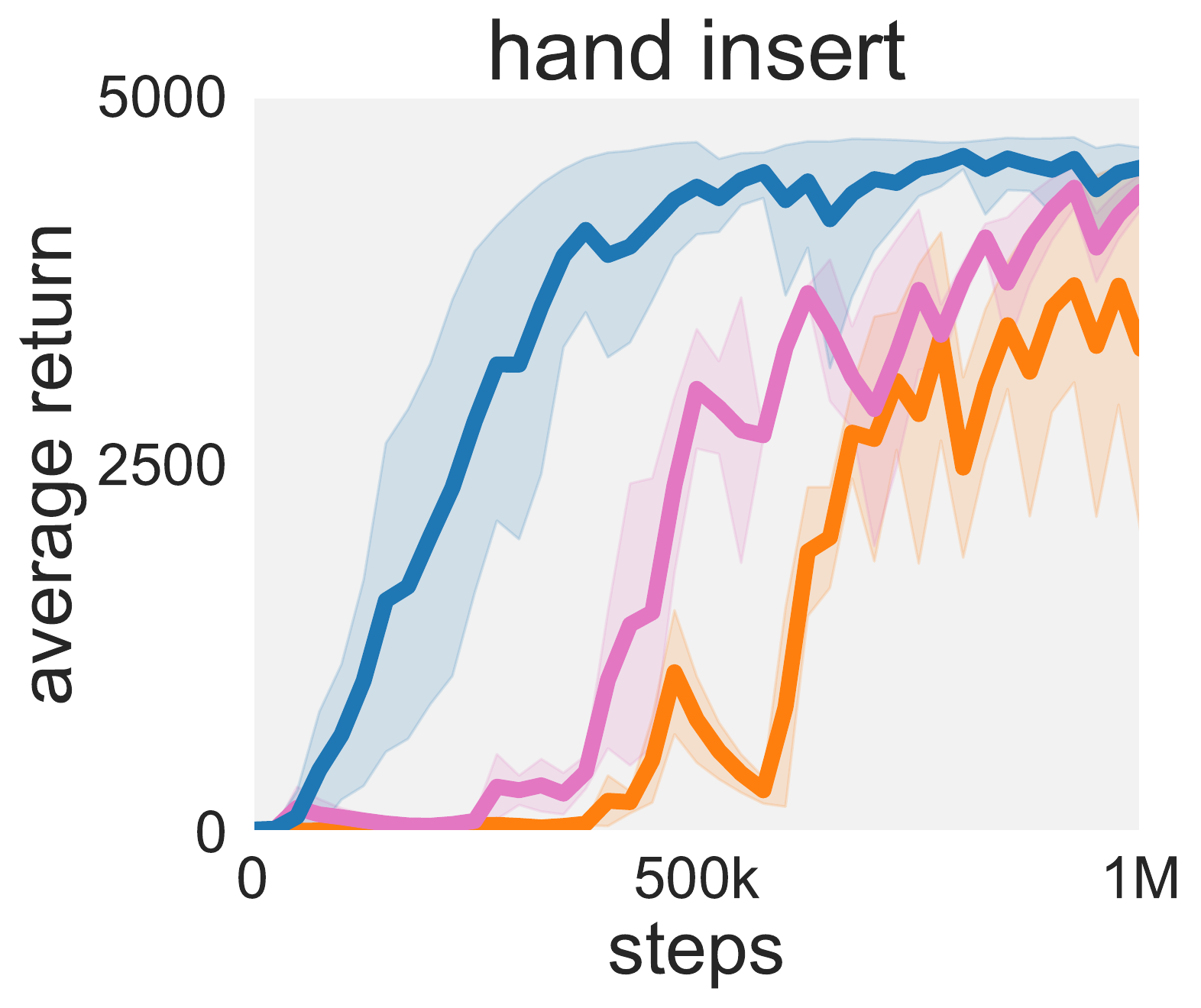}
        \end{subfigure}%
            \hfill    \begin{subfigure}[t]{0.48\textwidth} \centering
            \includegraphics[height=3.0cm,keepaspectratio]{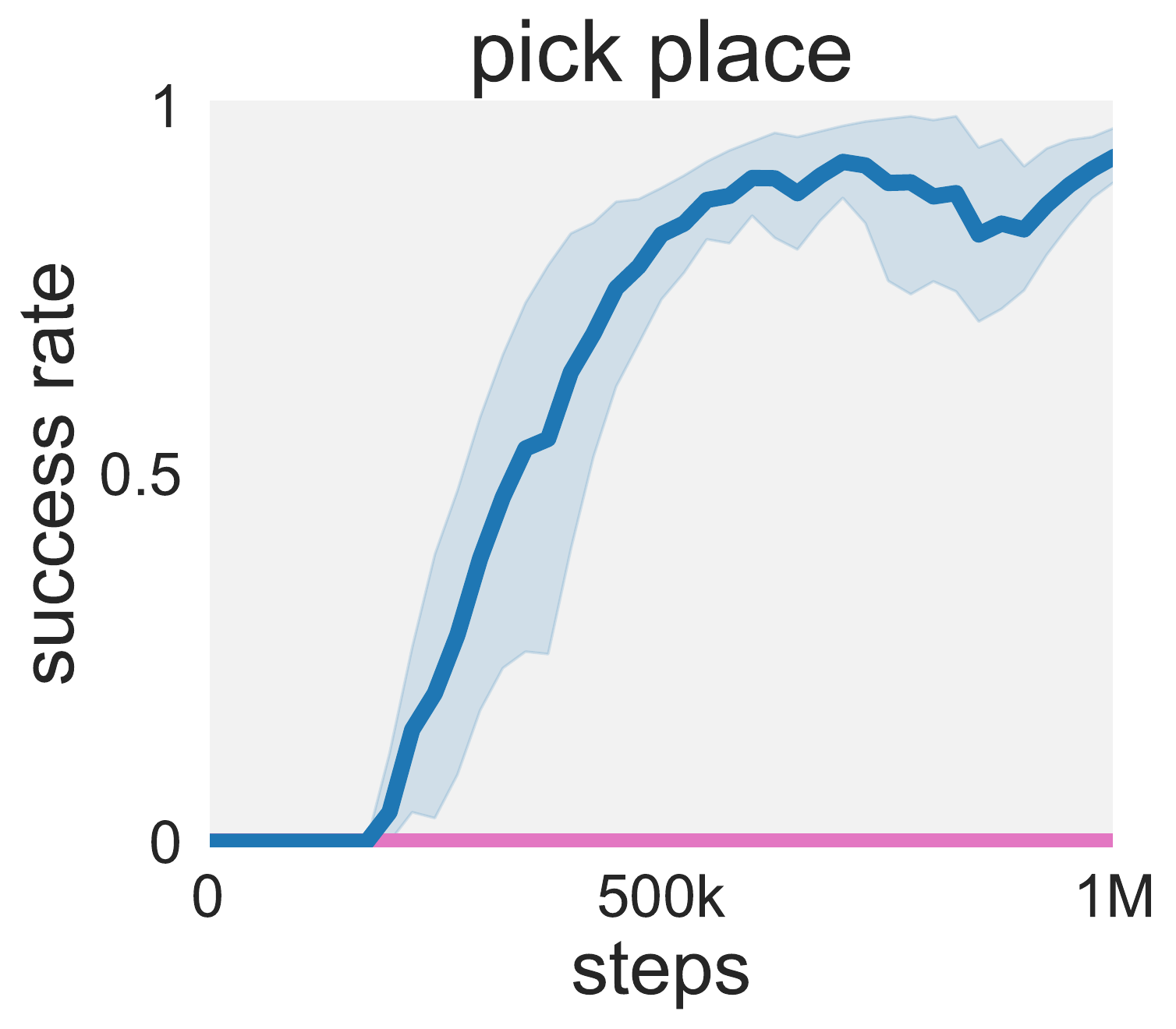}
            \includegraphics[height=3.0cm,keepaspectratio]{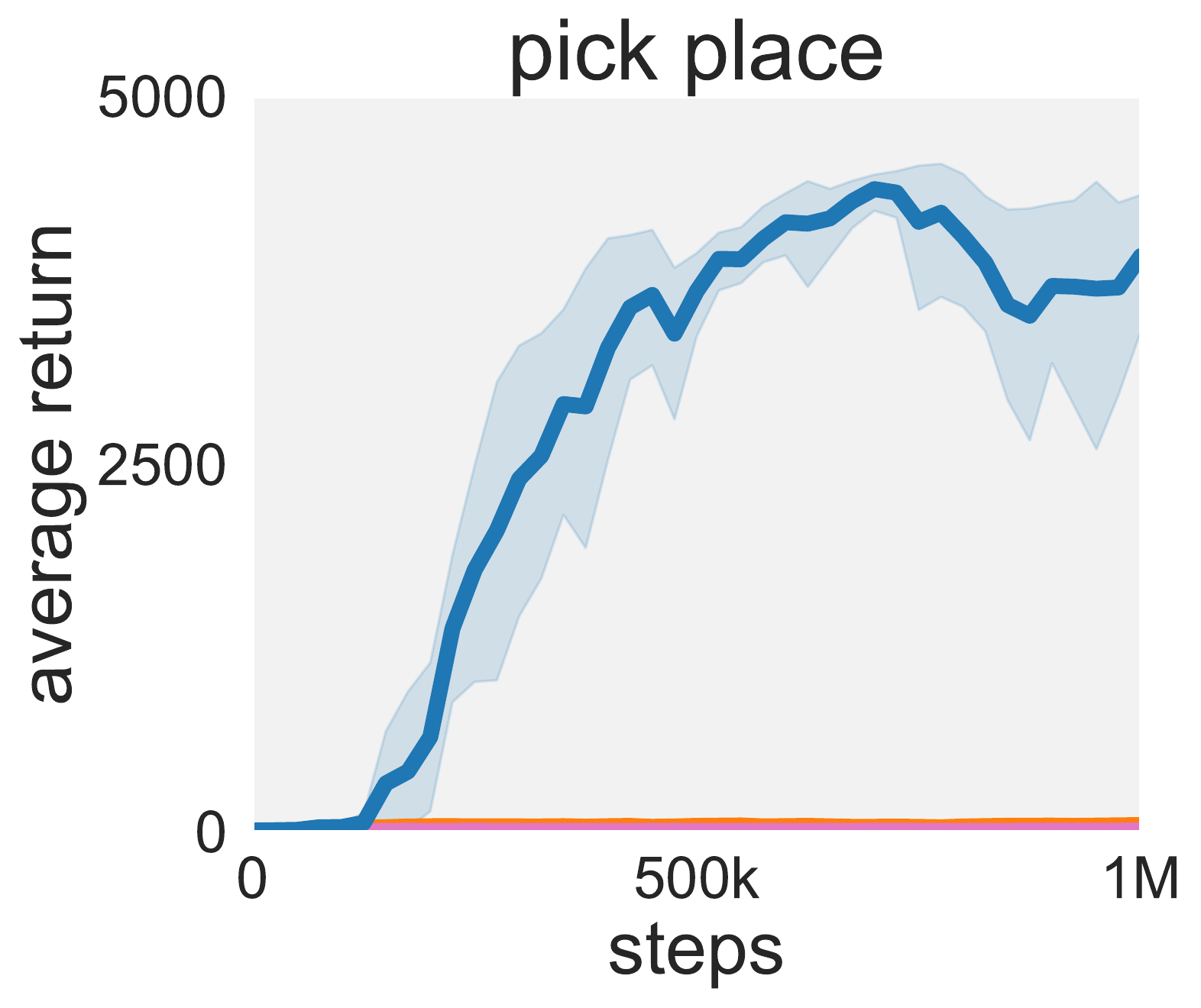}
        \end{subfigure}%
        \\ 
        \begin{subfigure}[t]{0.48\textwidth} \centering
            \includegraphics[height=3.0cm,keepaspectratio]{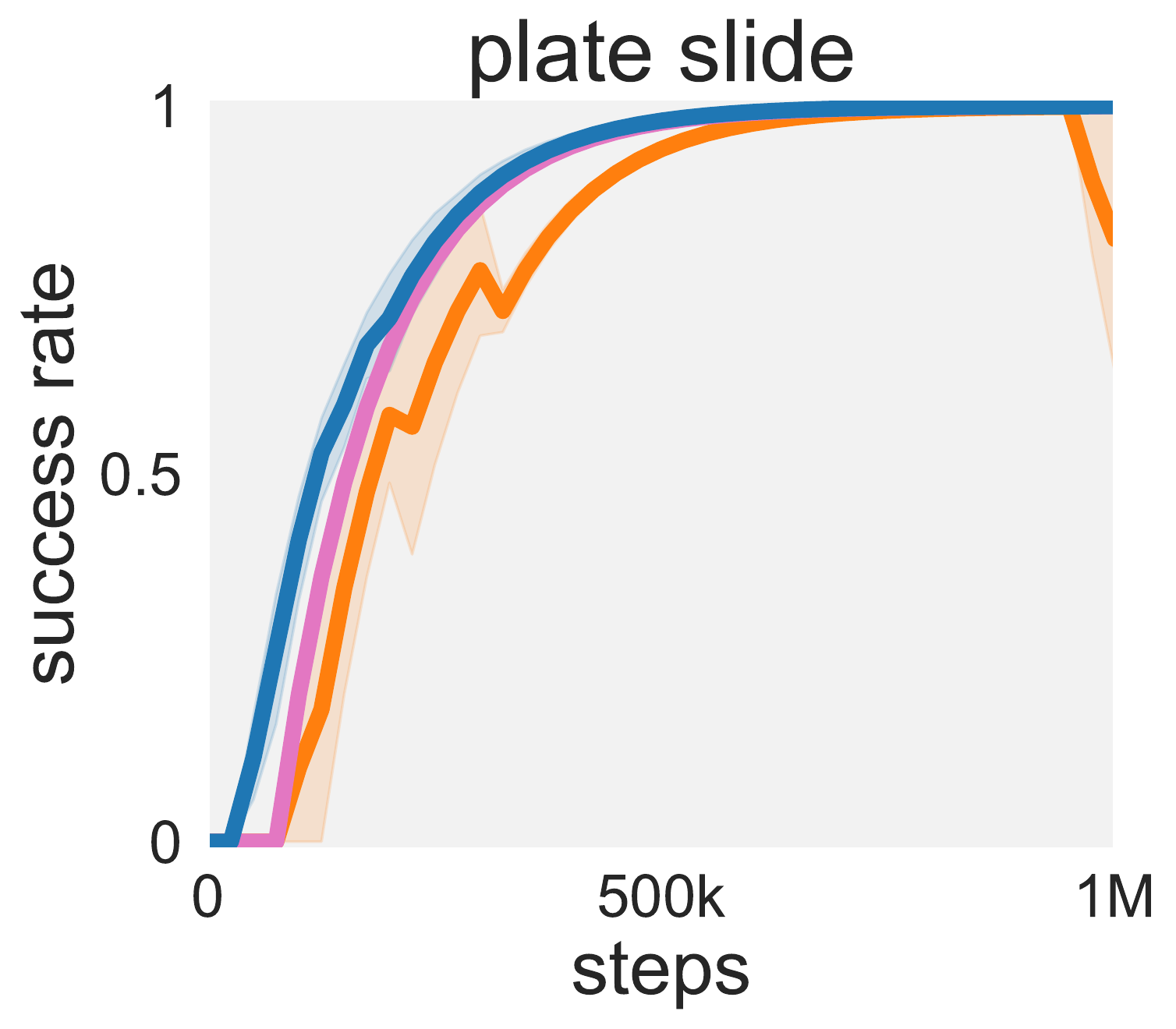}
            \includegraphics[height=3.0cm,keepaspectratio]{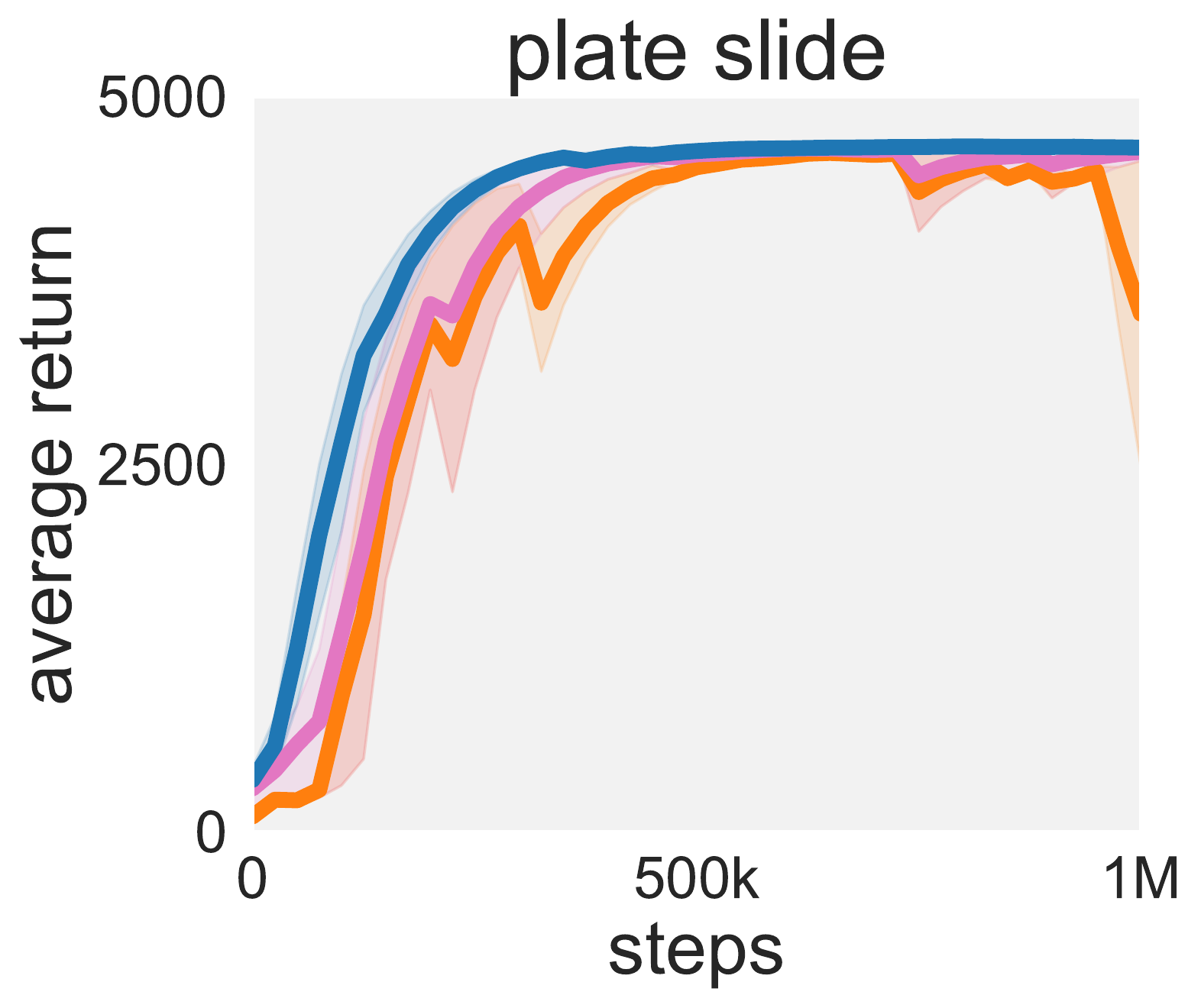}
        \end{subfigure}%
            \hfill    
            \begin{subfigure}[t]{0.48\textwidth} \centering
            \includegraphics[height=3.0cm,keepaspectratio]{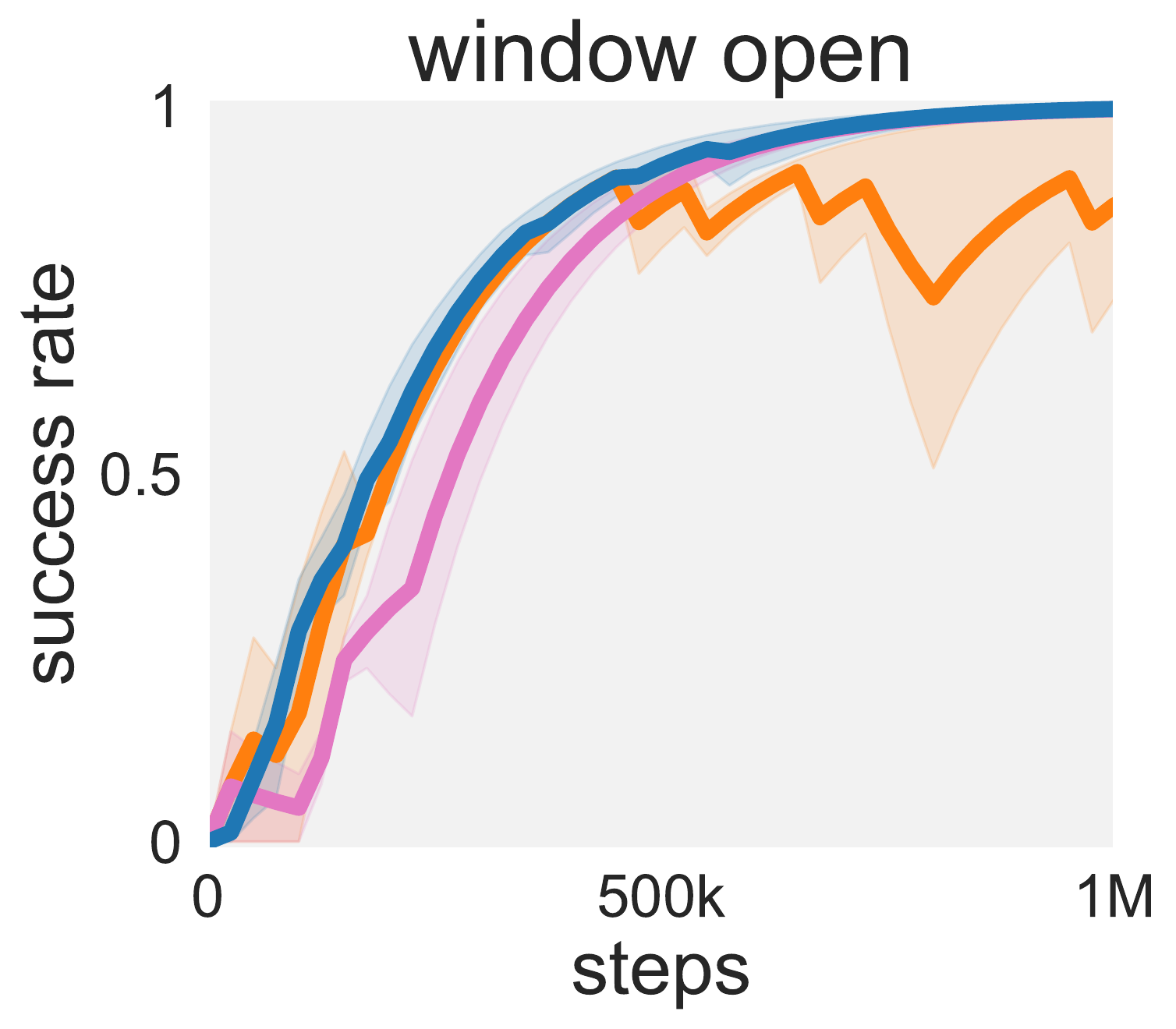}
            \includegraphics[height=3.0cm,keepaspectratio]{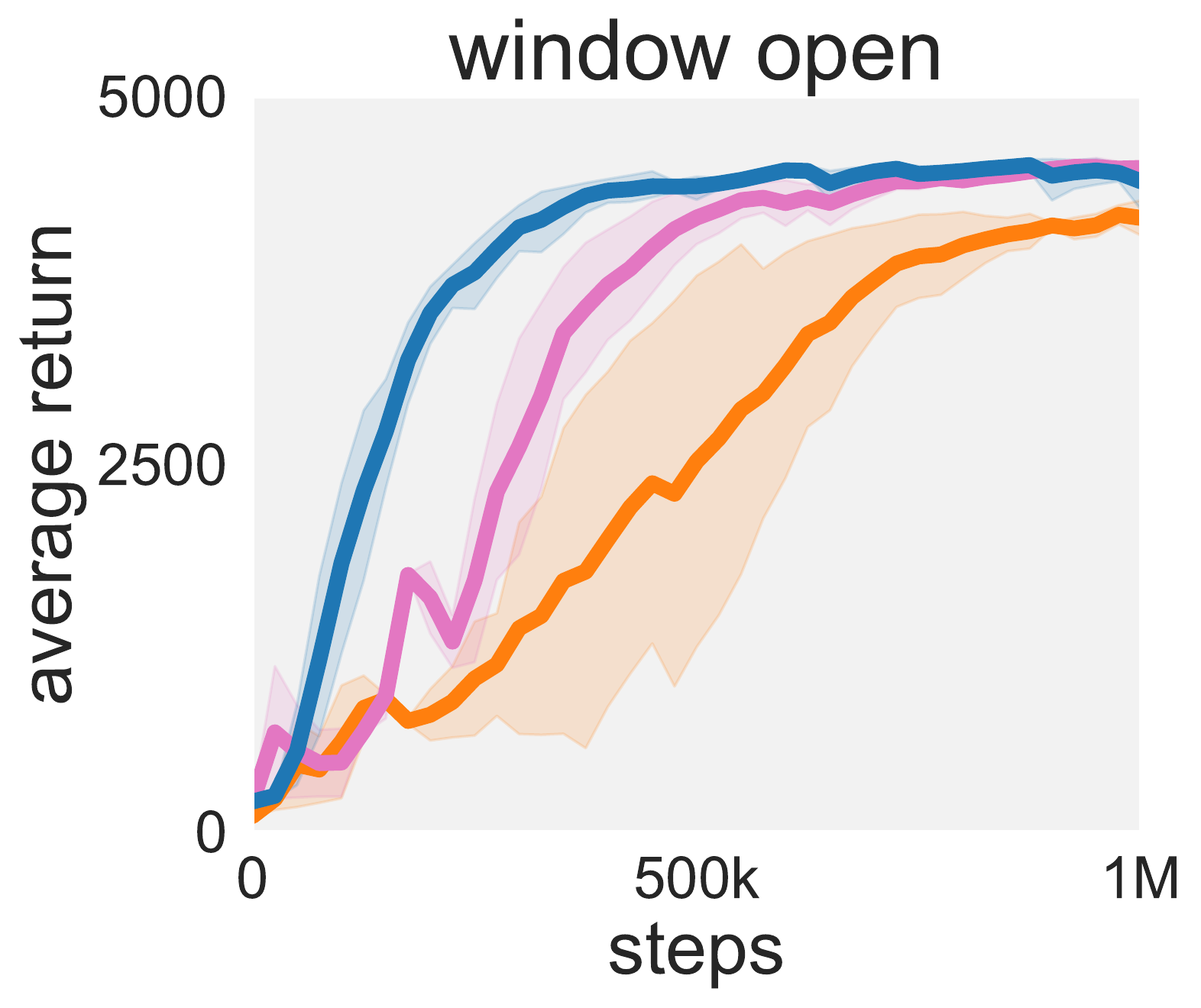}
        \end{subfigure}%
        \\
        \begin{subfigure}[t]{1.0\textwidth}
        \centering
            \includegraphics[width=0.3\textwidth]{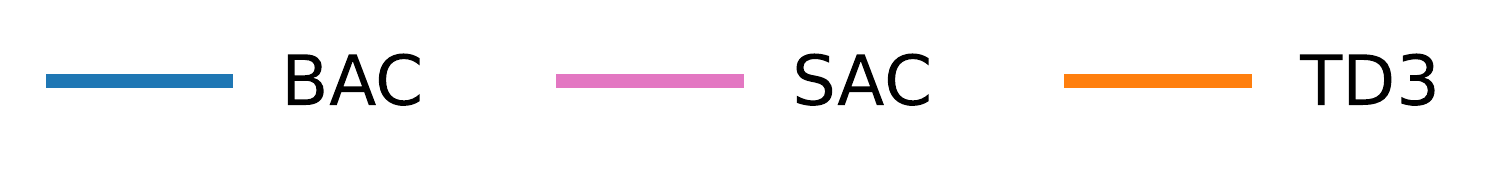}
        \end{subfigure}
        \caption{\textbf{Individual Meta-World tasks.} Success rate and average return of \ourshort\ , SAC, TD3 on twelve manipulation tasks from MetaWorld (sorted alphabetically). Solid curves depict the mean of ten trials, and shaded regions correspond to the one standard deviation. }
        \label{fig:metaworld-results}
    \end{minipage}
\end{figure}

\paragraph{Performance comparison.} In Figure~\ref{fig:metaworld-results}, we present learning curves of both the success rate and average return for twelve individual Meta-World tasks. Note that we conduct experiments on the goal-conditioned versions of the tasks from Meta-World-v2, which are considered harder than the single-goal variant. 

In tasks typically categorized as simple, where both SAC and TD3 succeed within 1M steps, it is noteworthy that BAC still outperforms in terms of sample efficiency.

In tasks involving complex manipulation, such as pick place, basketball, hand insert, coffee push and hammer, BAC exhibits strong performance. Consider the hammer task, while SAC and TD3 occasionally achieve serendipitous successes before reaching 500K steps, their $Q$-value estimations are susceptible to misguidance by the inferior follow-up actions that occur frequently, resulting in a sustained low success rate. In contrast, BAC efficiently exploits the value of success and mitigates the impact of inferior samples on the $Q$-value, leading to a significant performance improvement beyond 500K steps, and finally surpasses SAC and TD3 by a large margin in terms of success rate.

These results highlight the promising potential of BAC in manipulation tasks.

\begin{figure}[ht]
    \resizebox{1.0\textwidth}{!}{
        \begin{tabular}{cc}
            & time $\longrightarrow$\\ 
            \rotatebox{90}{\parbox[c]{2cm}{\centering  Drawer \\ Open}} &
            \includegraphics[width=0.95\linewidth,valign=b]{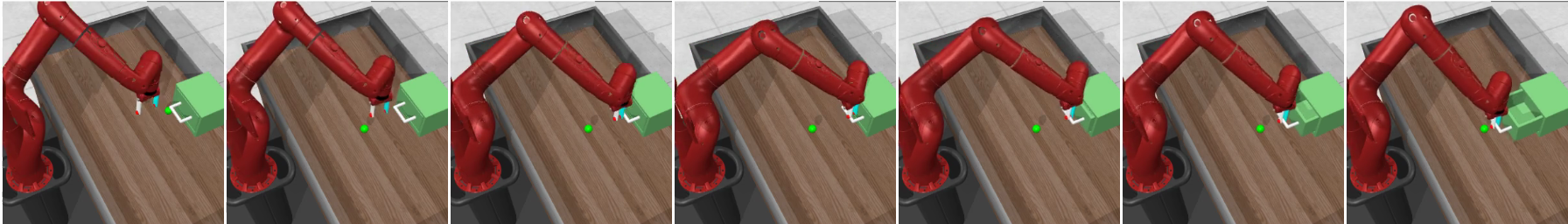} 
            \\[-0.6ex]
            \rotatebox{90}{\parbox[c]{2cm}{\centering  Basketball}} &
            \includegraphics[width=0.95\linewidth,valign=b]{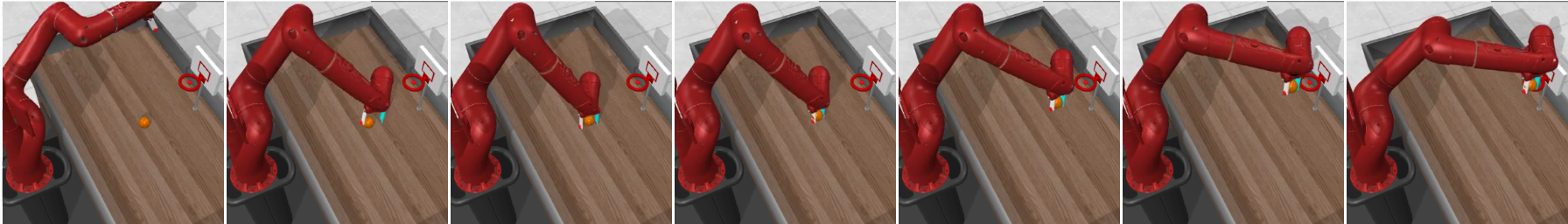} 
            \\[-0.6ex]
            \rotatebox{90}{\parbox[c]{2cm}{\centering  Hammer}} &
            \includegraphics[width=0.95\linewidth,valign=b]{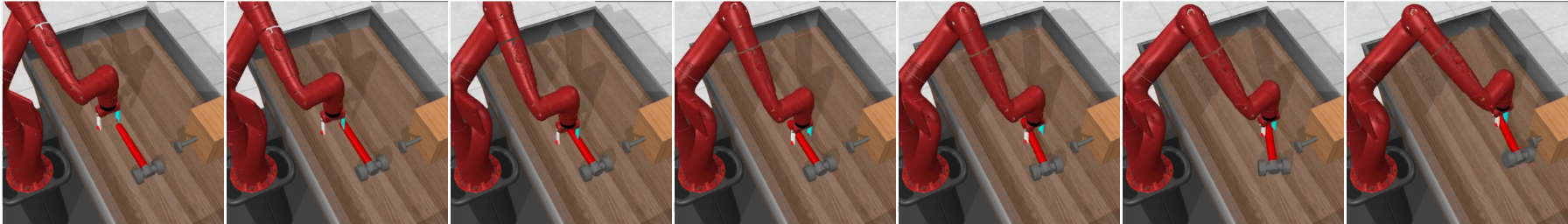} 
            \\[-0.6ex]
            \rotatebox{90}{\parbox[c]{2cm}{\centering  Coffee \\ Push}} &
            \includegraphics[width=0.95\linewidth,valign=b]{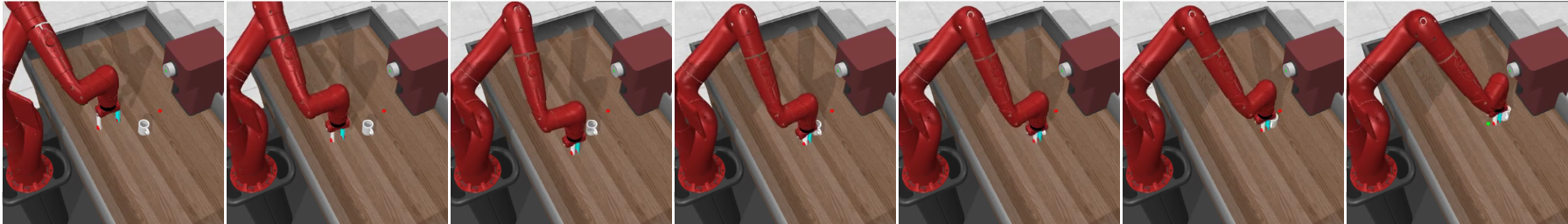}
            \\[-0.6ex]
            \rotatebox{90}{\parbox[c]{2cm}{\centering  Hand \\ Insert}} &
            \includegraphics[width=0.95\linewidth,valign=b]{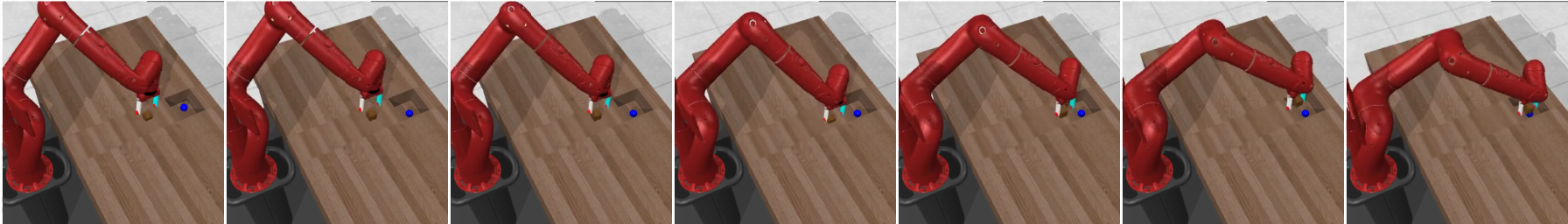}
            \\[-0.6ex]
            \rotatebox{90}{\parbox[c]{2cm}{\centering  Pick \\ Place}} &
            \includegraphics[width=0.95\linewidth,valign=b]{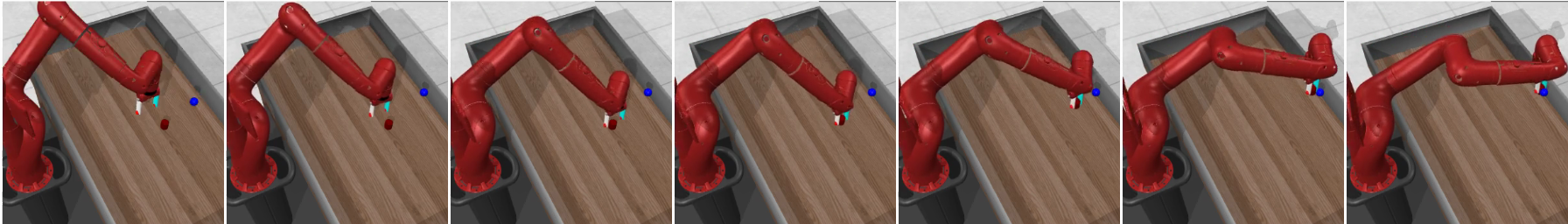}
        \end{tabular}
    }
    \caption{\textbf{Trajectory Visualizations.} We visualize trajectories generated by \ourshort\ on six Meta-World tasks. Our method (\ourshort) is capable of solving each of these tasks within 1M  steps.}
    \label{fig:metaworld_visualization}
\end{figure}

\paragraph{Trajectory Visualizations.} Successful trajectories for one simple task and five aforementioned complex tasks are visualized in Figure~\ref{fig:metaworld_visualization}. For each trajectory, we display seven keyframes. 

\clearpage
\subsection{Evaluation on Adroit benchmark tasks}\label{section:adroit_benchmark}

\begin{figure}[h]
    \centering
    \includegraphics[width=3.8cm,keepaspectratio]{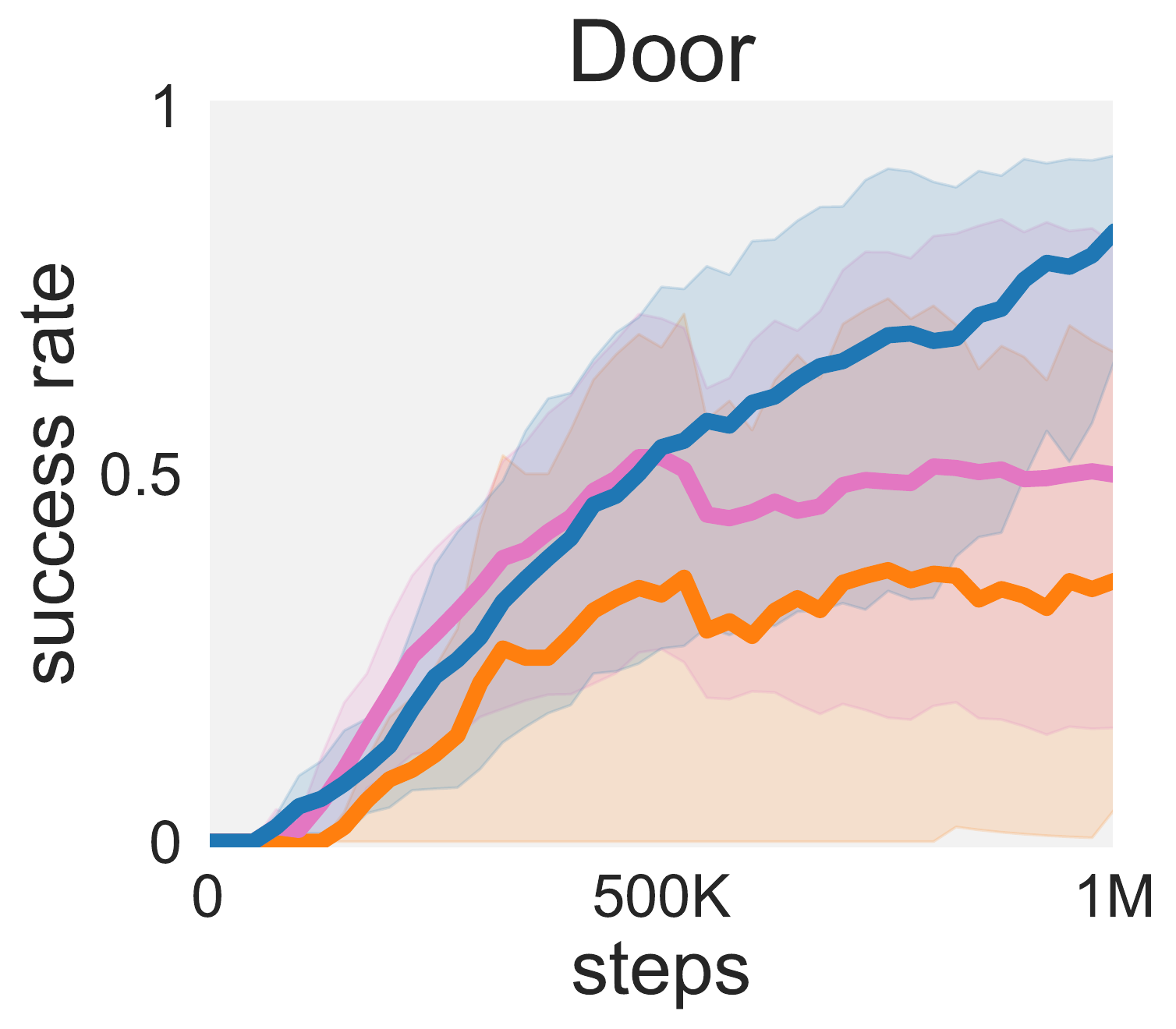}
    \includegraphics[width=3.8cm,keepaspectratio]{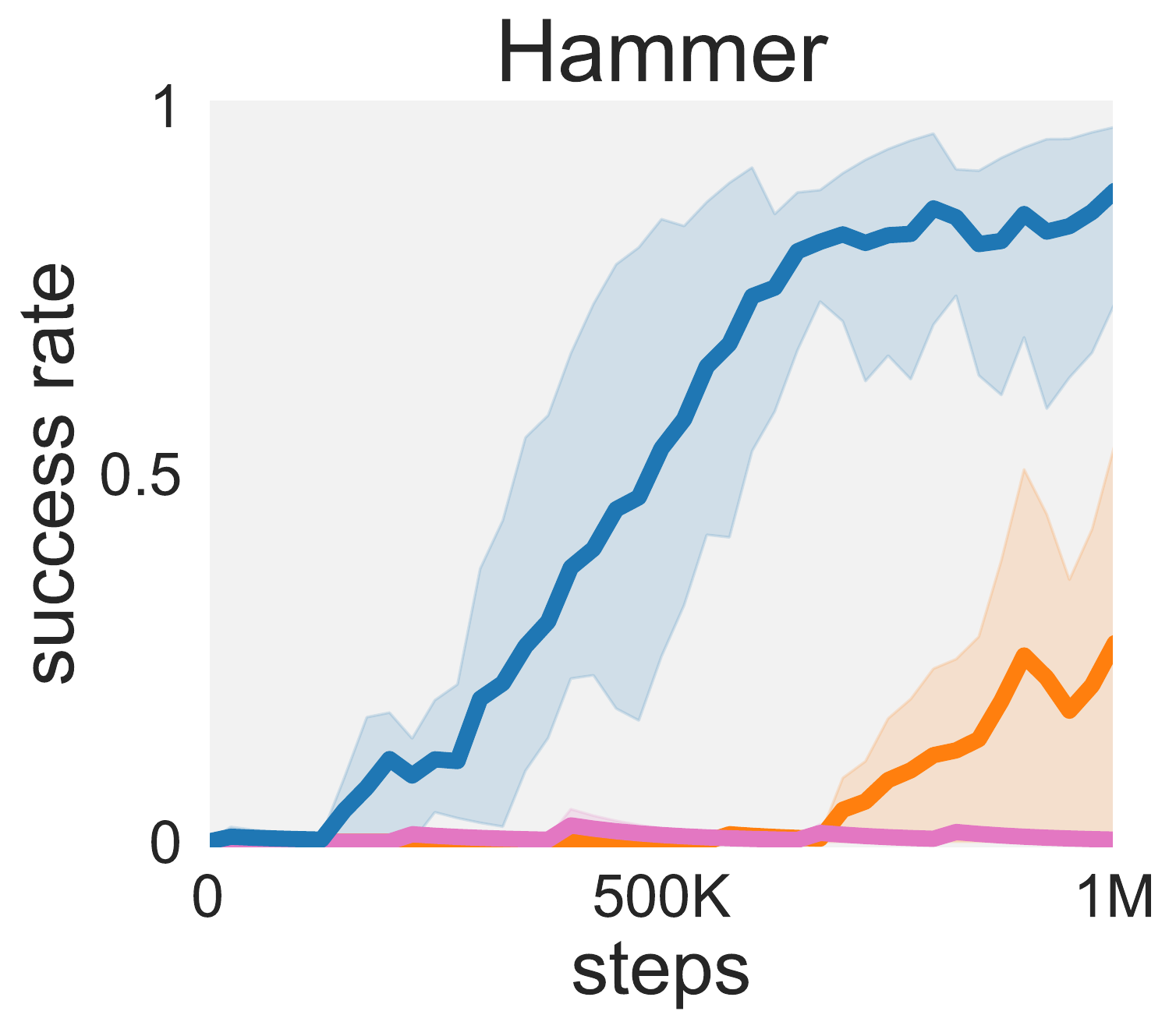}
    \includegraphics[width=3.8cm,keepaspectratio]{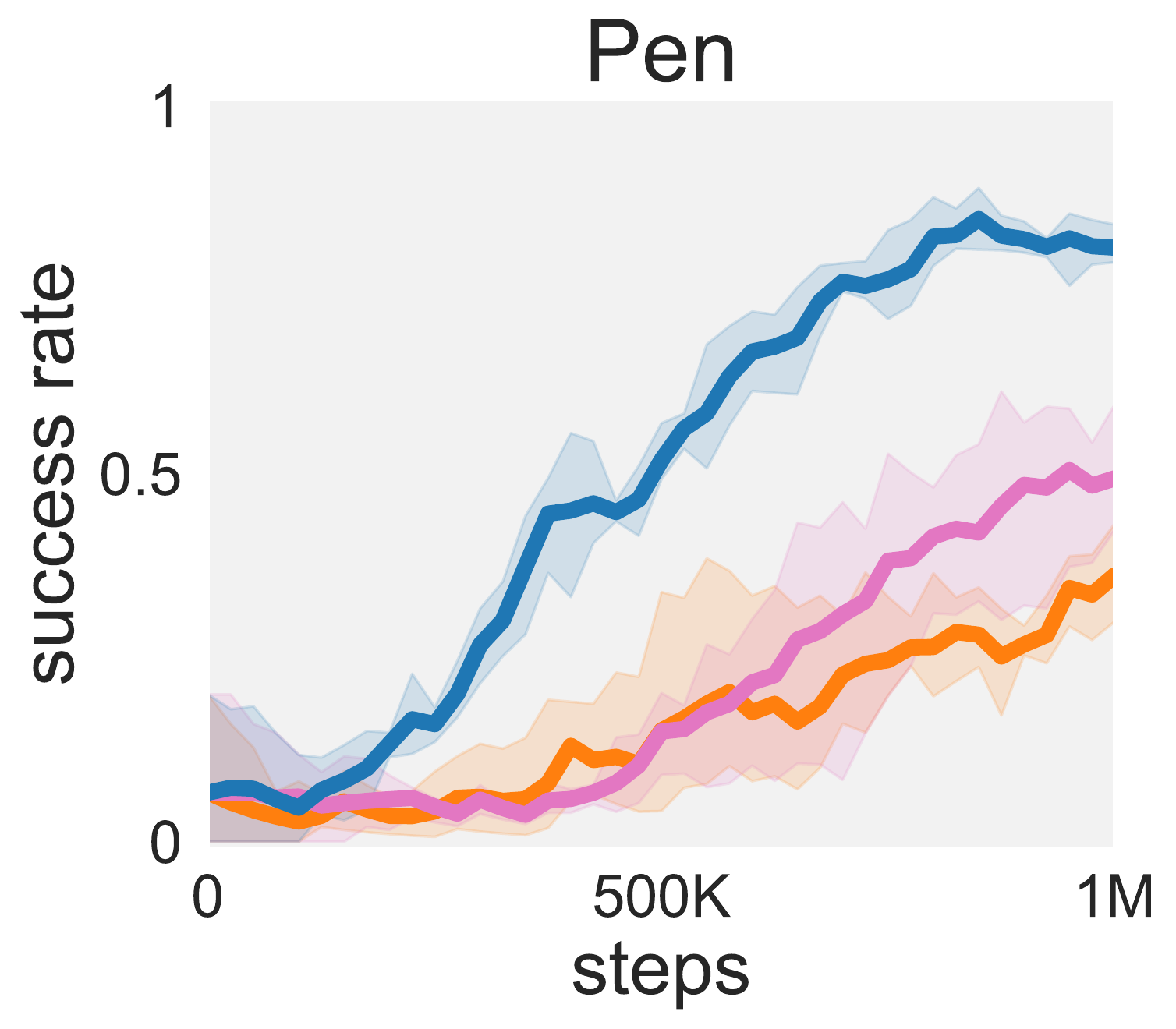}
    \\
    \includegraphics[width=3.8cm,keepaspectratio]{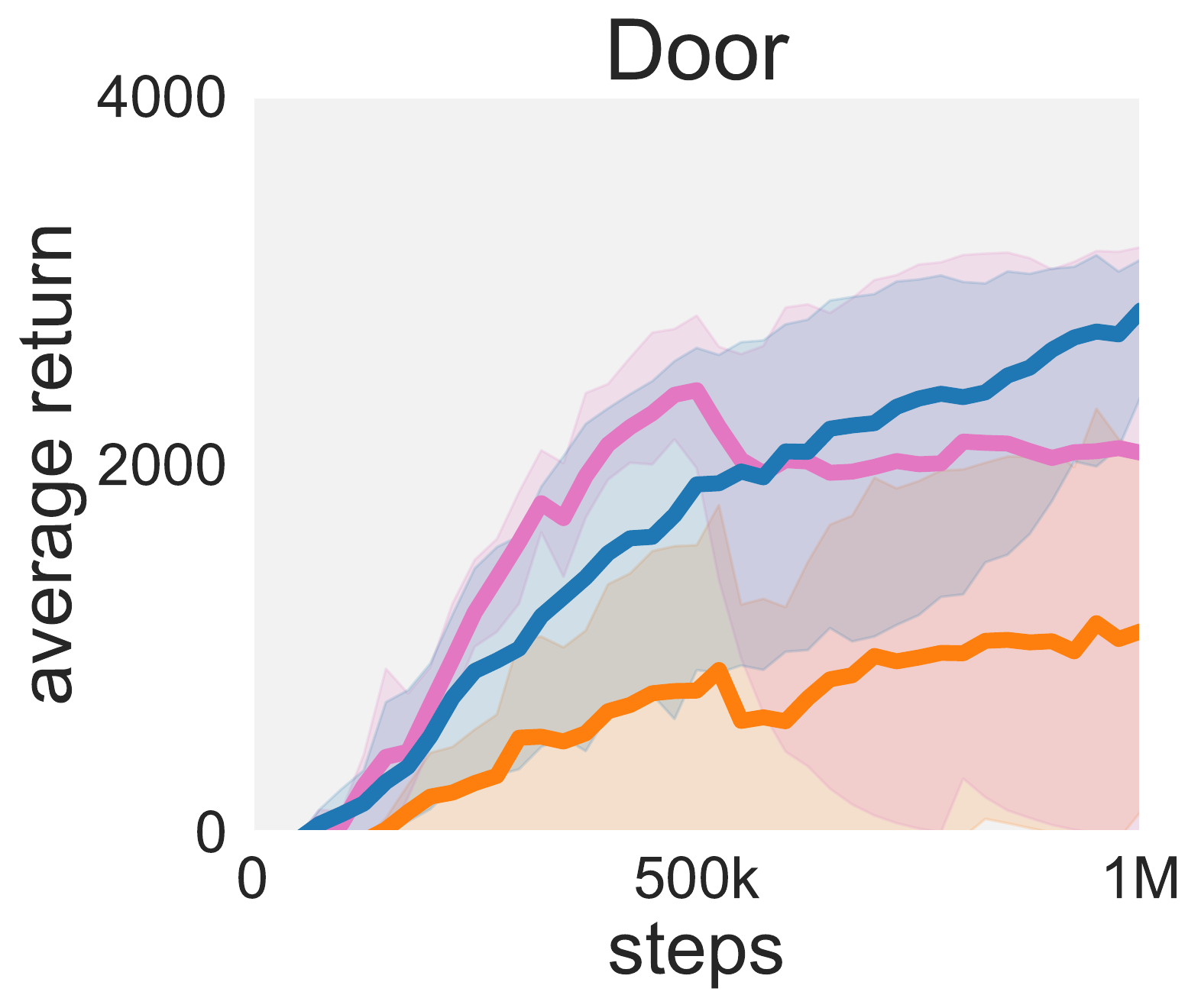}
    \includegraphics[width=3.8cm,keepaspectratio]{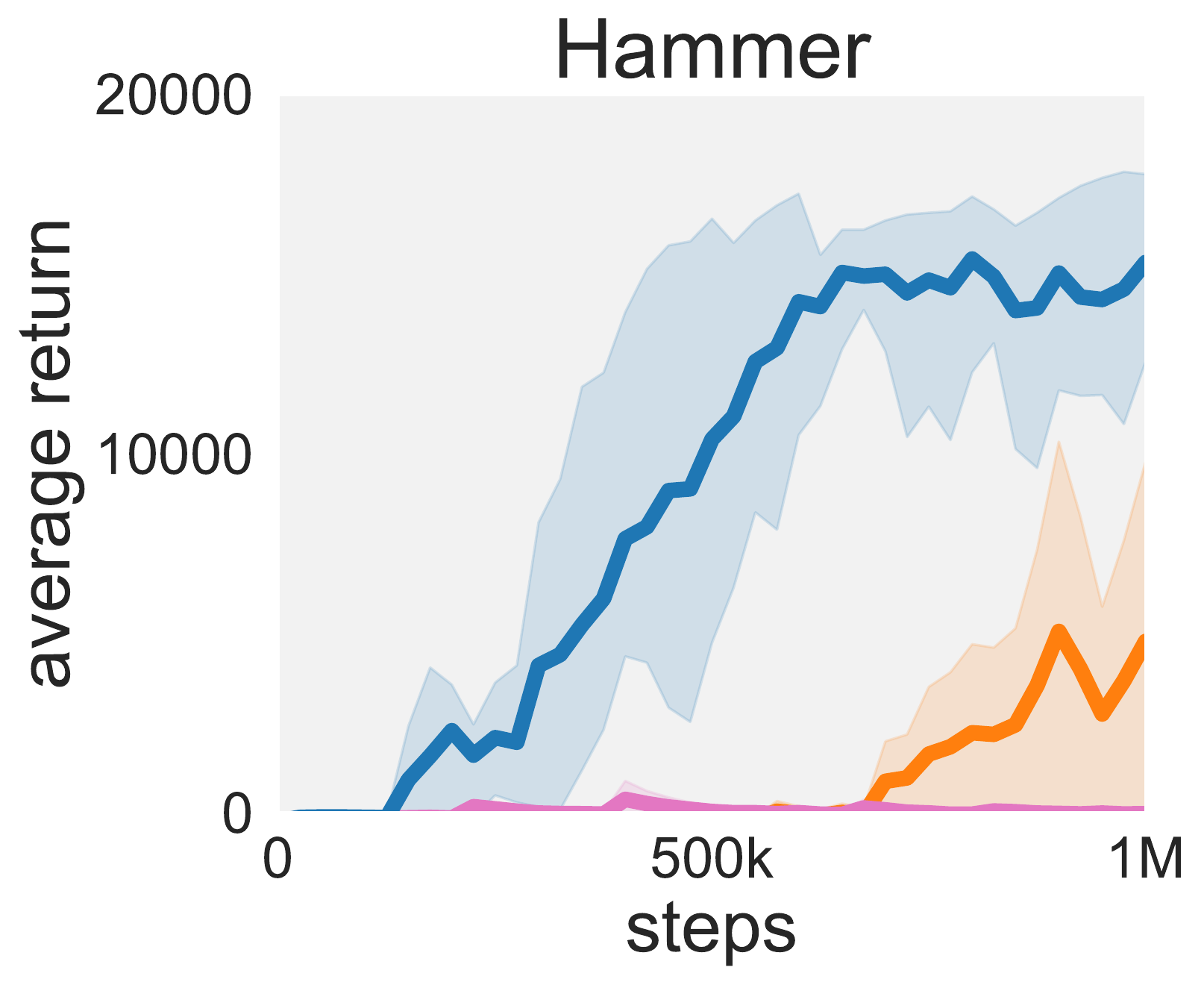}
    \includegraphics[width=3.8cm,keepaspectratio]{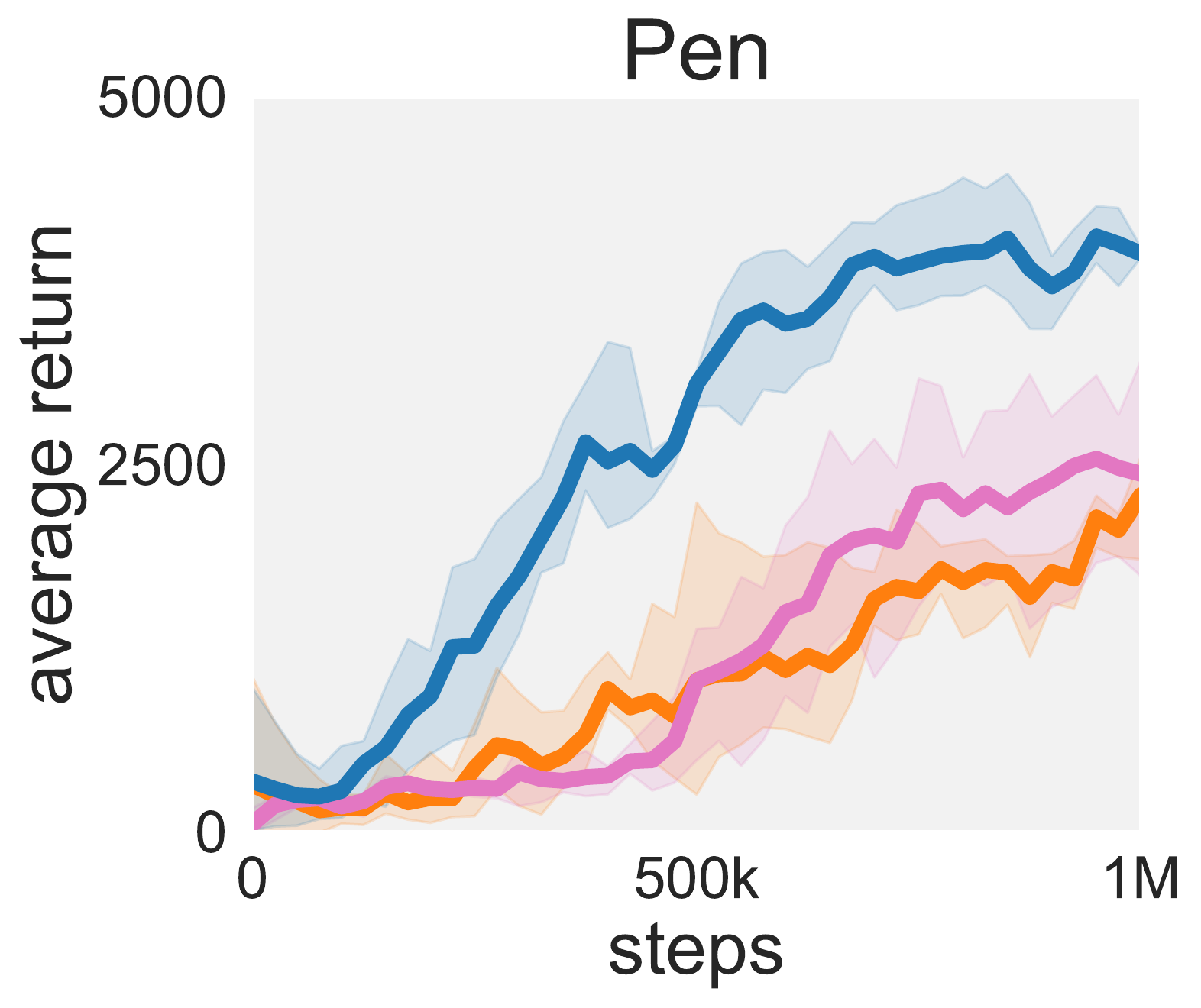}
    \\
    \includegraphics[width=0.3\textwidth]{Figures/3-legend_only.pdf}
    \caption{\textbf{Adroit tasks.} Success rate and average return of \ourshort\ , SAC, TD3 on Adroit benchmark tasks. Solid curves depict the mean of ten trials, and shaded regions correspond to the one standard deviation. }
    \label{fig:adroit-res}
\end{figure}

\subsection{Evaluation on MyoSuite benchmark tasks}\label{section:myosuite_benchmark}
\begin{figure}[H]
    \centering
    \includegraphics[width=3.8cm,keepaspectratio]{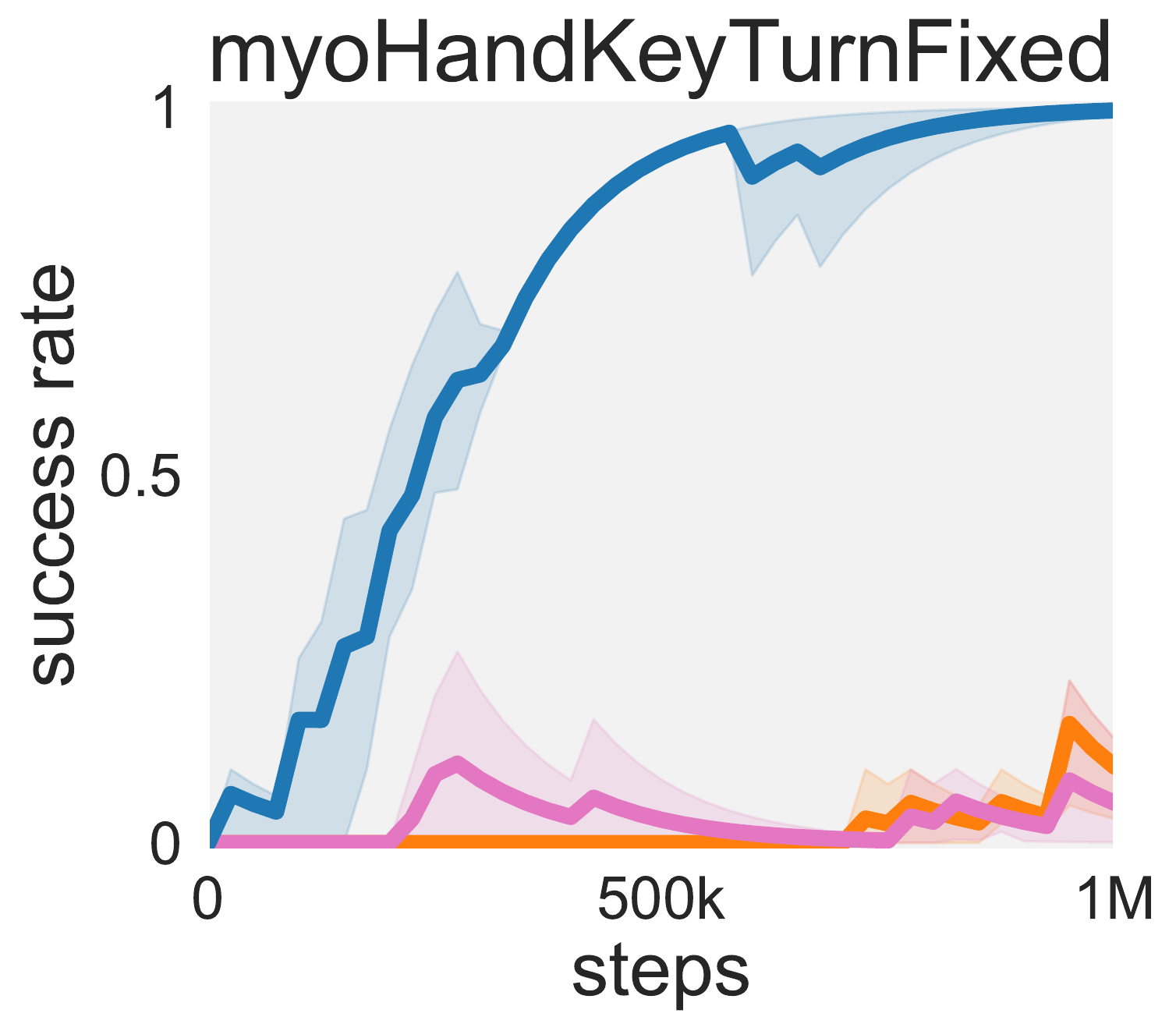}
    \includegraphics[width=3.8cm,keepaspectratio]{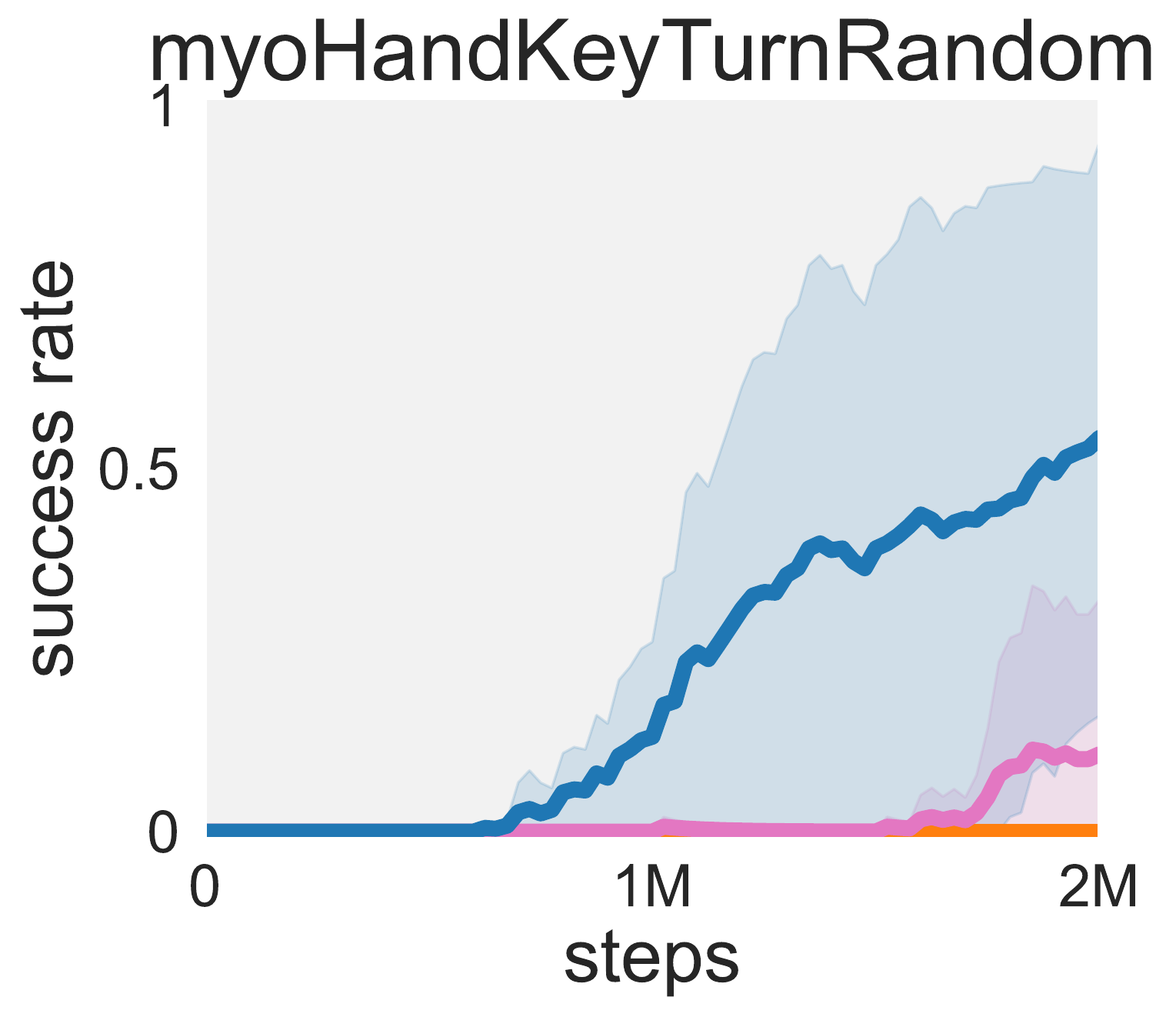}
    \includegraphics[width=3.8cm,keepaspectratio]{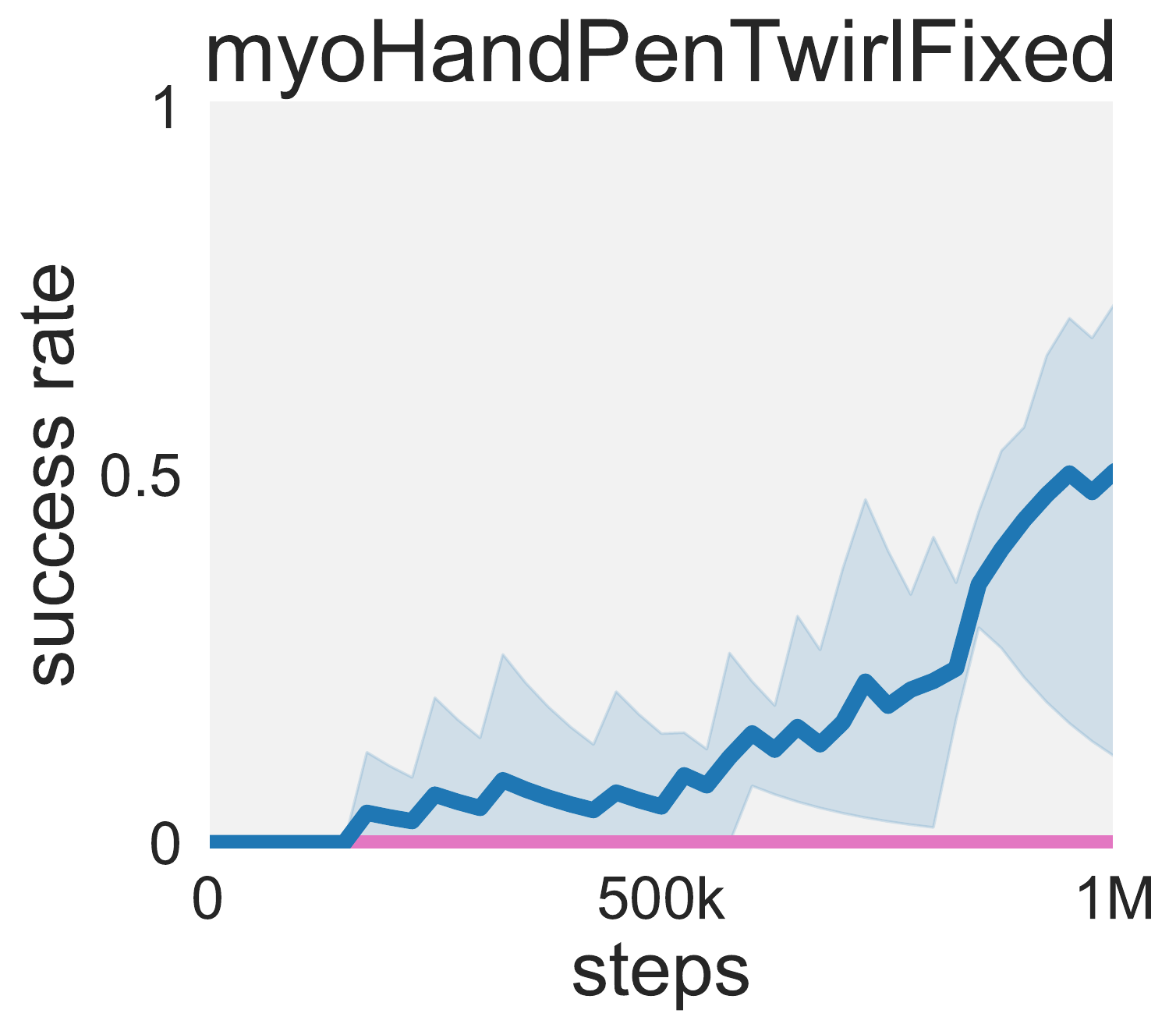}
    \\
    \includegraphics[width=3.8cm,keepaspectratio]{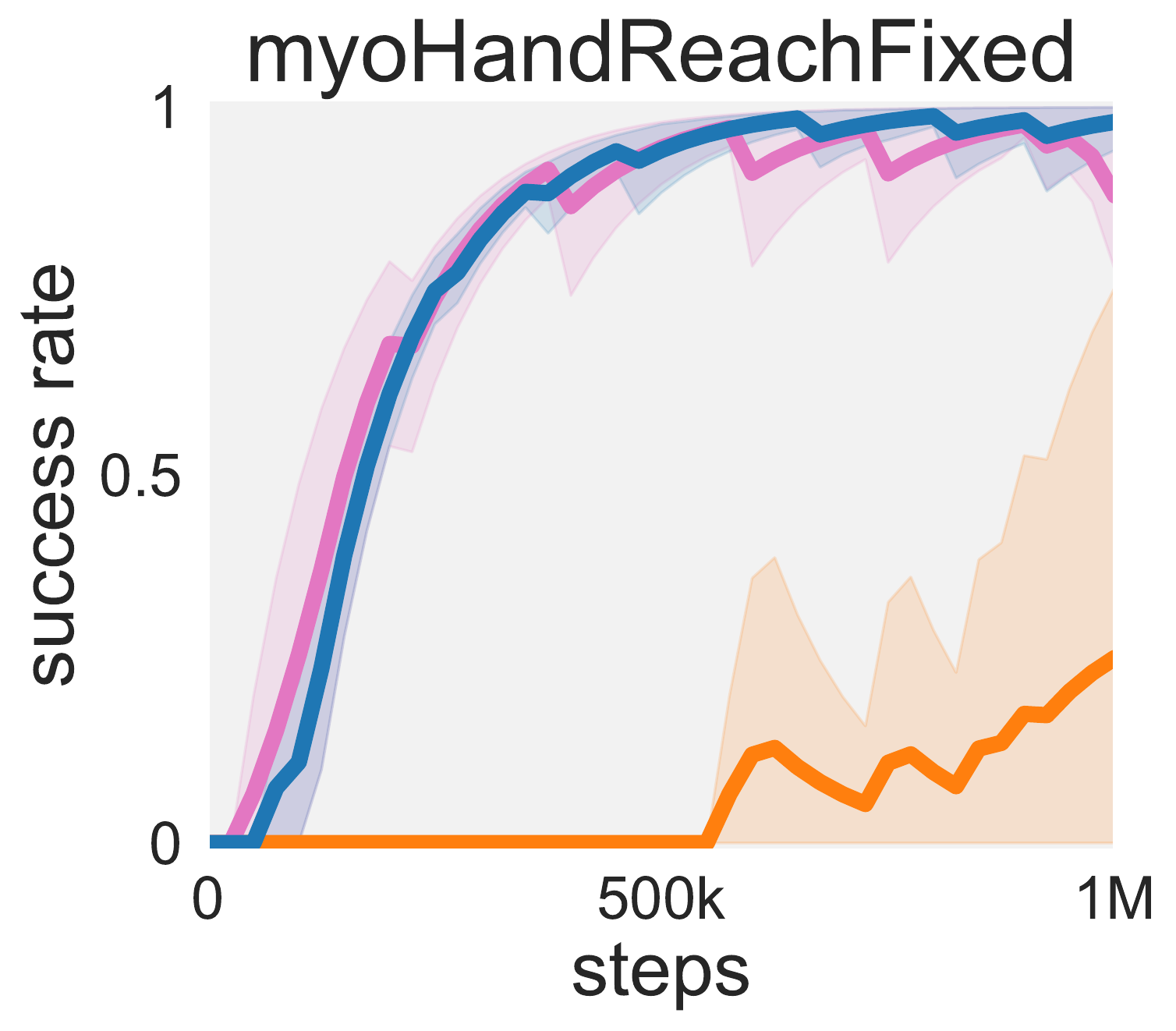}
    \includegraphics[width=3.8cm,keepaspectratio]{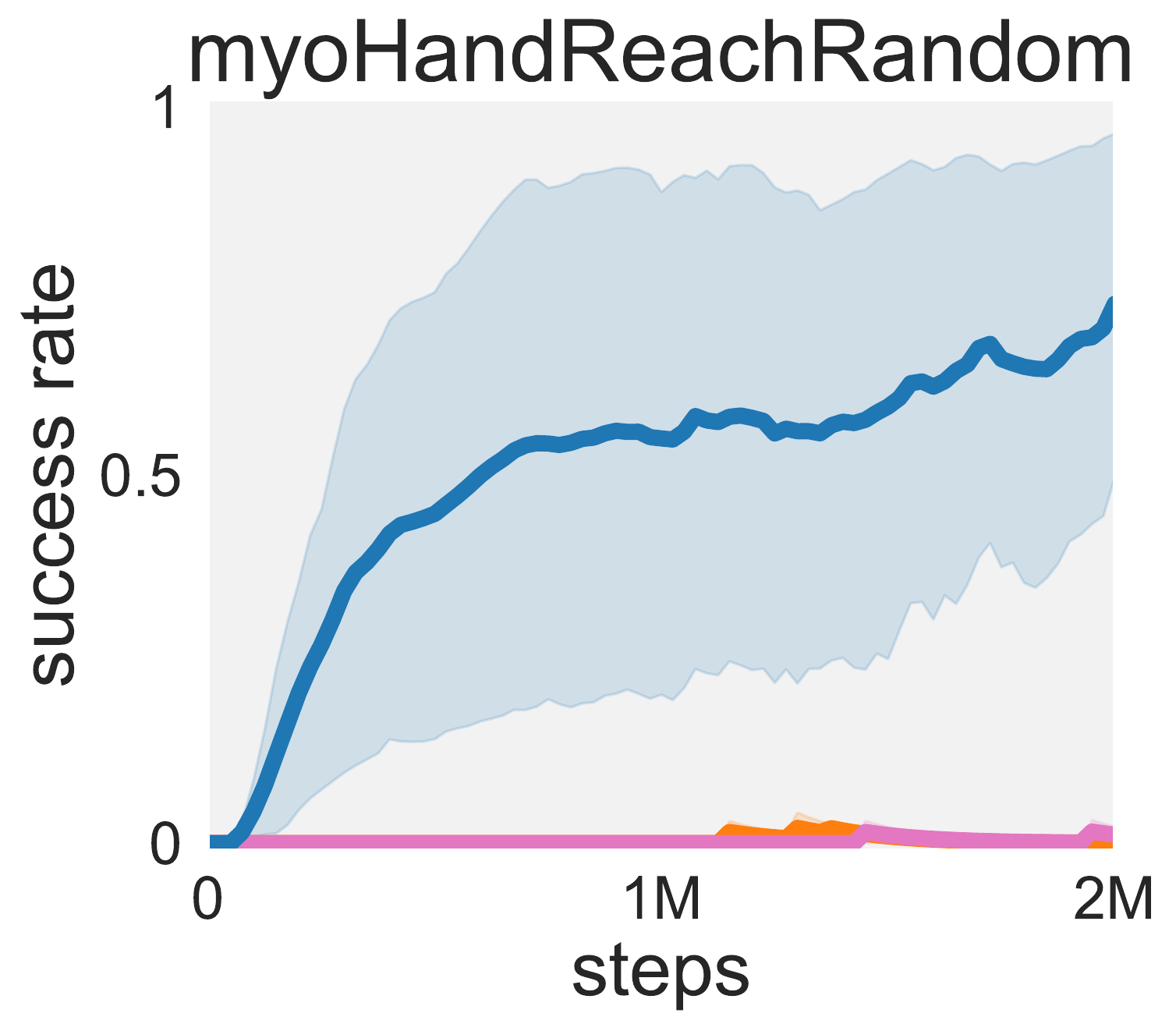}
    \\
    \includegraphics[width=0.3\textwidth]{Figures/3-legend_only.pdf}
    \caption{\textbf{MyoSuite tasks.} Success rate of \ourshort\ , SAC, TD3 on MyoSuite benchmark tasks. Solid curves depict the mean of ten trials, and shaded regions correspond to the one standard deviation. }
    \label{fig:myosuite-res}
\end{figure}

\subsection{Evaluation on ManiSkill2 benchmark tasks}\label{section:maniskill_benchmark}
\begin{figure}[H]
    \centering
    \includegraphics[width=3.6cm,keepaspectratio]{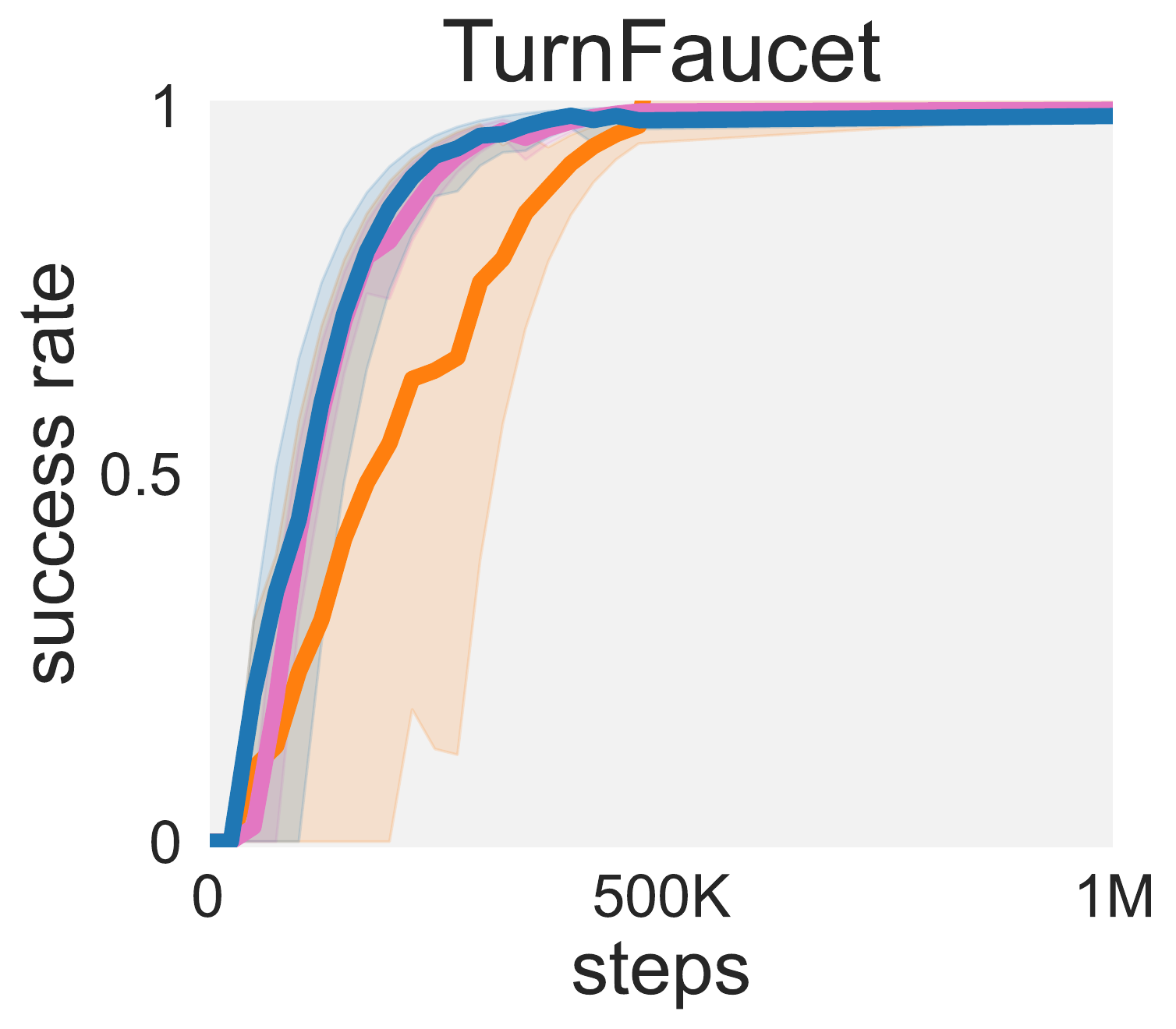}
    \includegraphics[width=3.6cm,keepaspectratio]{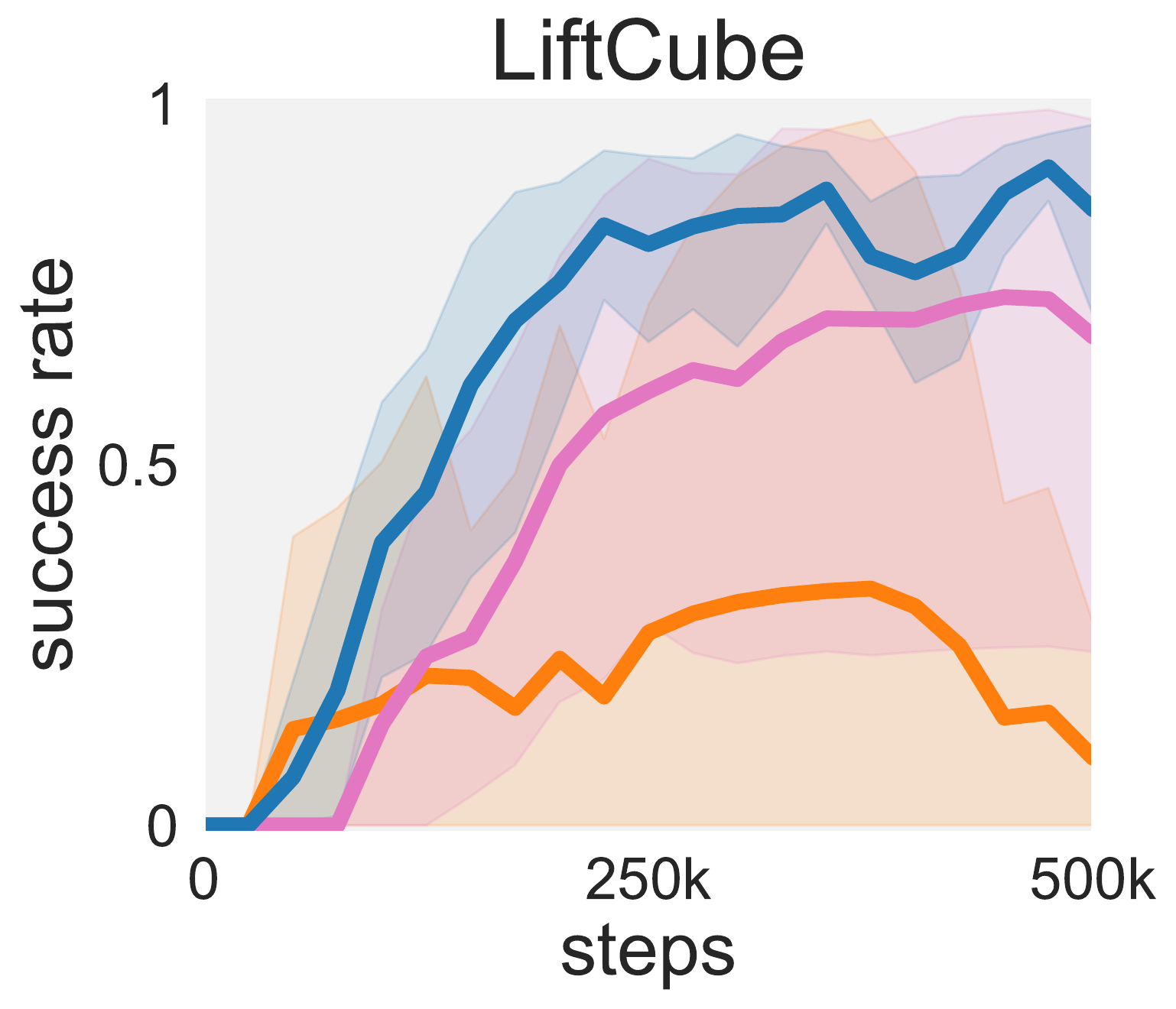}
    \includegraphics[width=3.6cm,keepaspectratio]{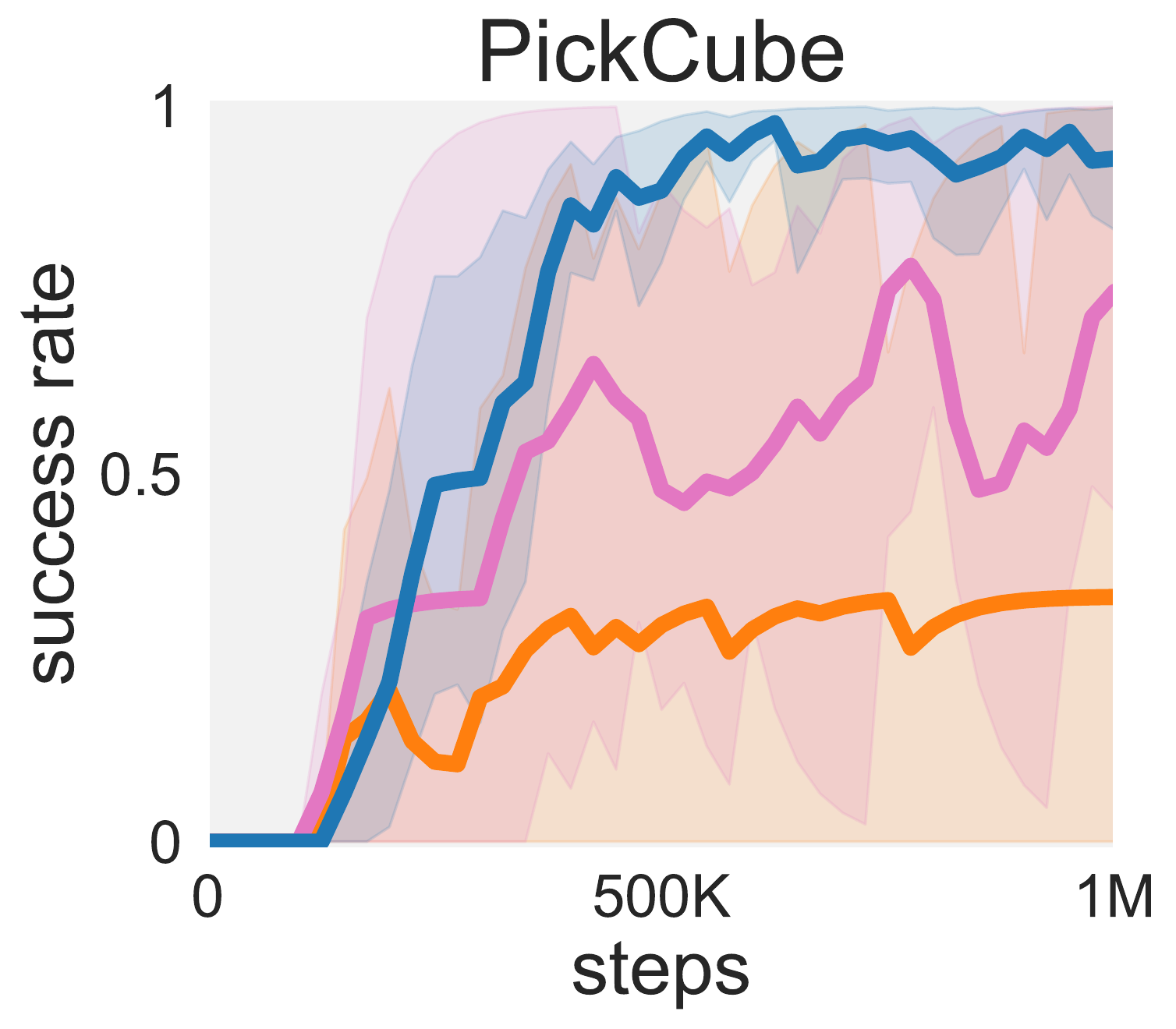}
    \\
    \includegraphics[width=3.6cm,keepaspectratio]{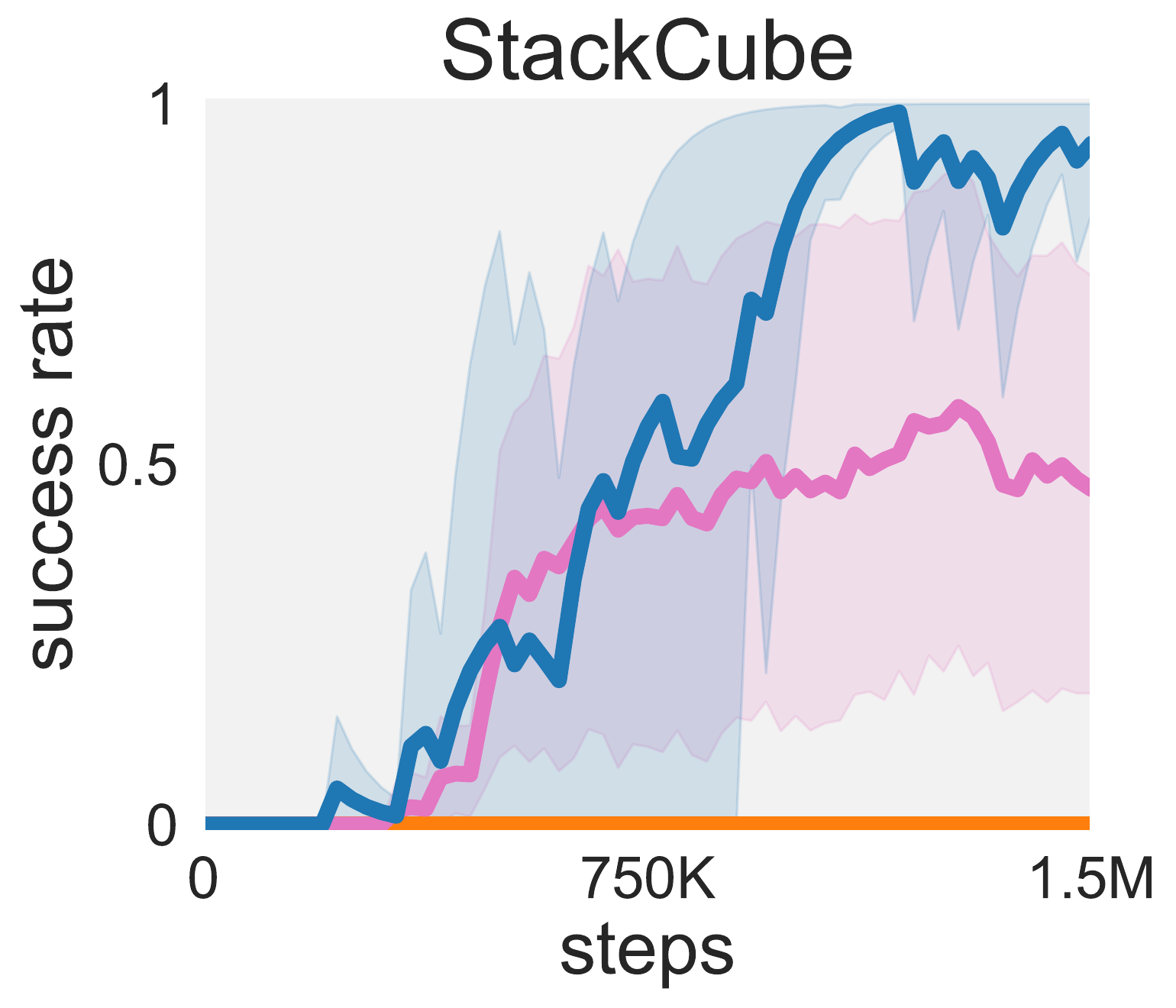}
    \includegraphics[width=3.6cm,keepaspectratio]{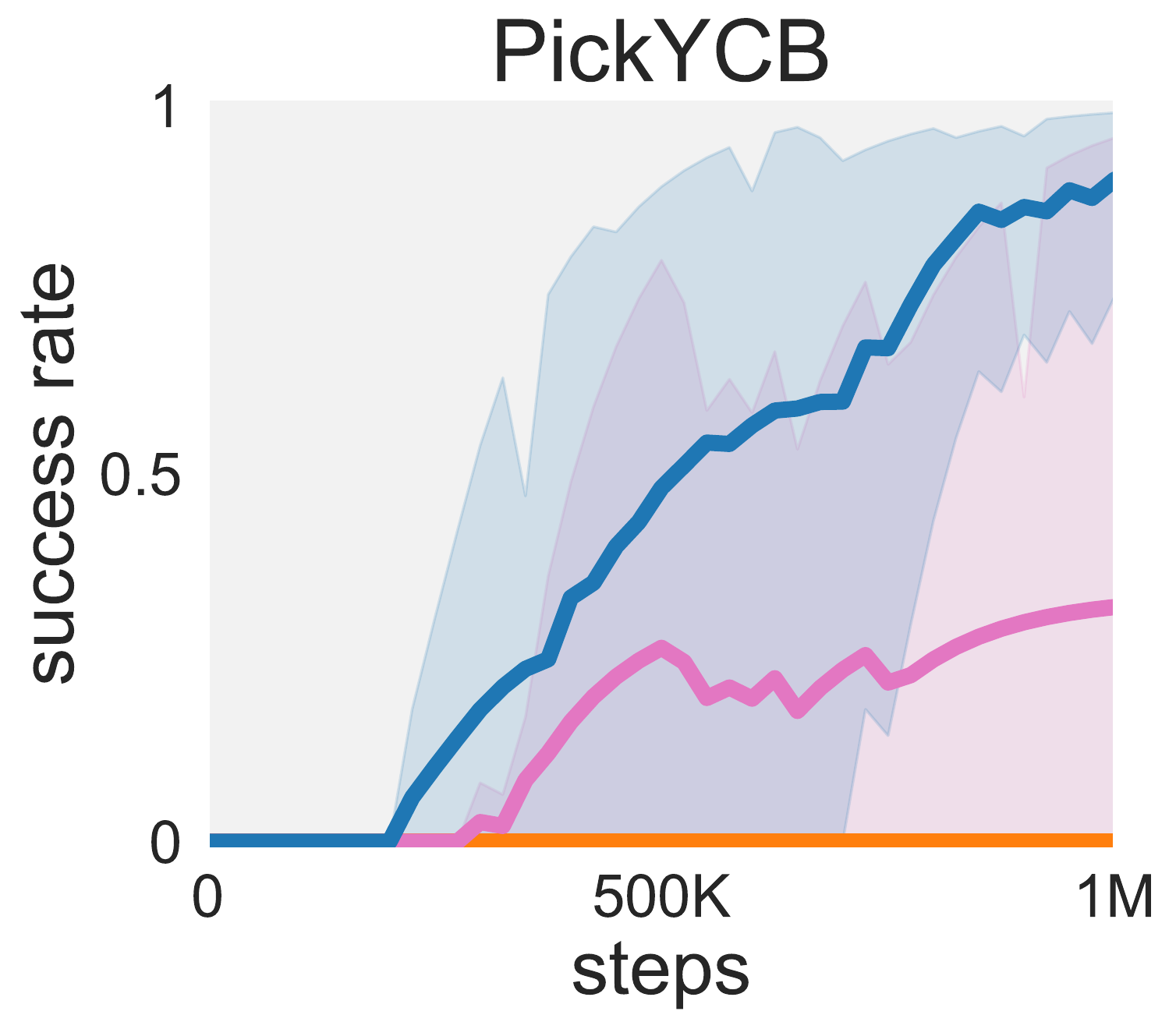}
    \\
    \includegraphics[width=0.3\textwidth]{Figures/3-legend_only.pdf}
    \caption{\textbf{ManiSkill2 tasks.} Success rate of \ourshort\ , SAC, TD3 on MainSkill2 benchmark tasks. Solid curves depict the mean of ten trials, and shaded regions correspond to the one standard deviation. }
    \label{fig:maniskill-res}
\end{figure}

\subsection{Evaluation on Shadow Dexterous Hand benchmark tasks}\label{section:shadowhand_benchmark}
\begin{figure}[h]
    \centering
    \includegraphics[width=3.6cm,keepaspectratio]{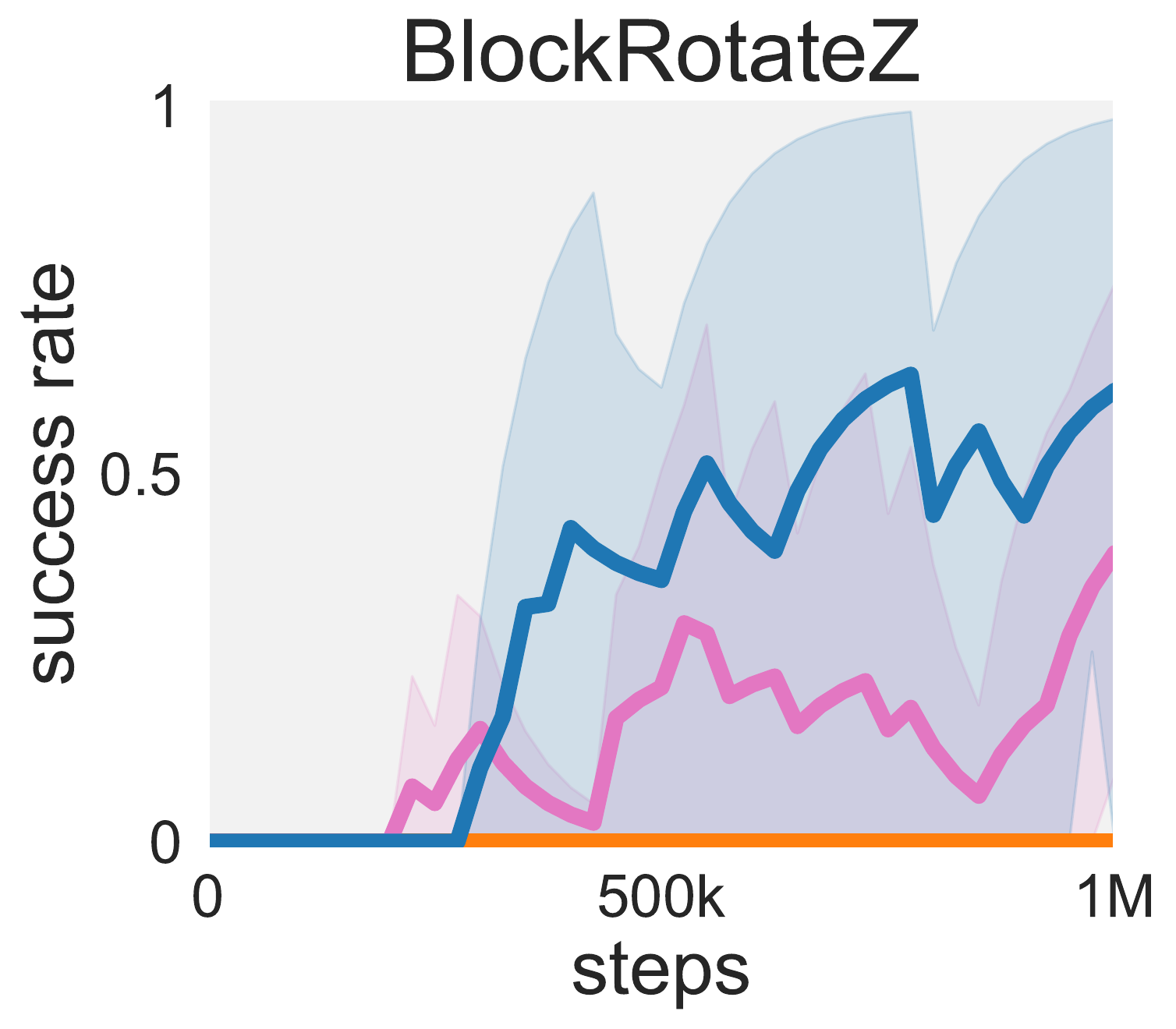}
    \includegraphics[width=3.6cm,keepaspectratio]{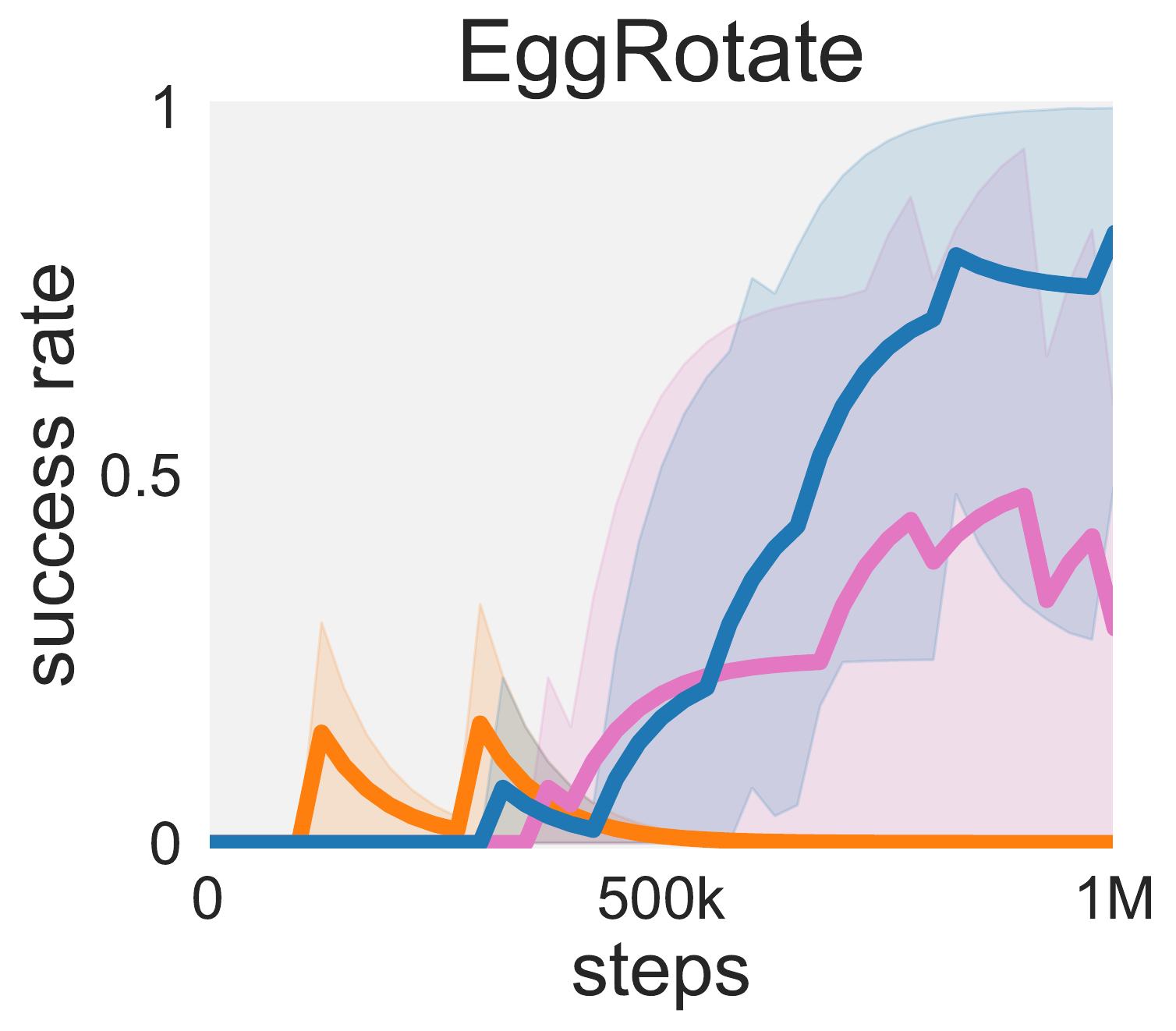}
    \includegraphics[width=3.6cm,keepaspectratio]{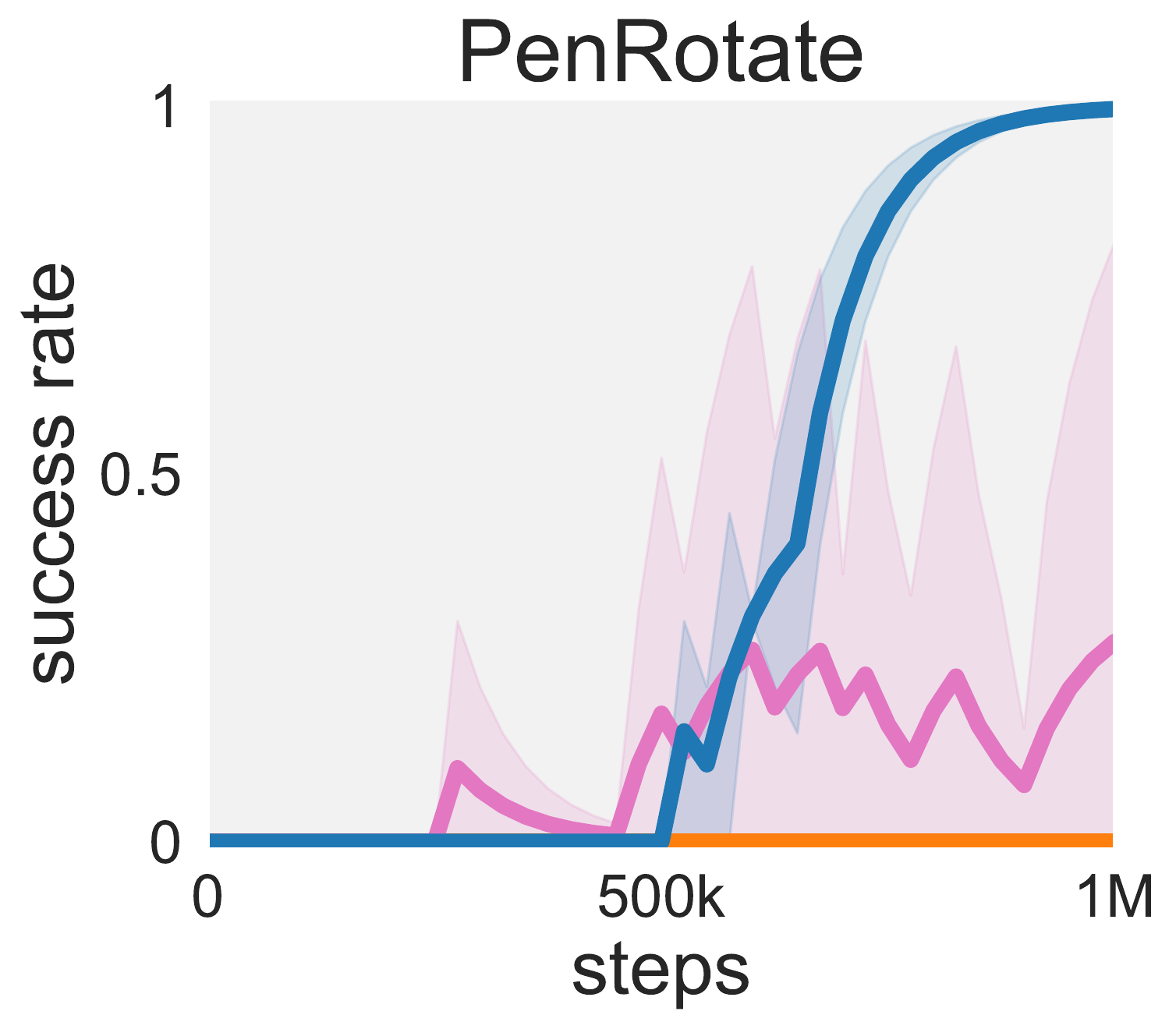}
    \\
    \includegraphics[width=0.3\textwidth]{Figures/3-legend_only.pdf}
    \caption{\textbf{Shadow Dexterous Hand tasks.} Success rate of \ourshort\ , SAC, TD3 on Shadow Dexterous Hand benchmark tasks. Solid curves depict the mean of ten trials, and shaded regions correspond to the one standard deviation. }
    \label{fig:shadowhand-res}
\end{figure}

\subsection{Evaluation on Swimmer benchmark tasks with an extremely high $\gamma$.}\label{ap:swimmer}
In our main paper, we adopt the same $\gamma=0.99$ parameters for all our tasks adhering to the community's advocacy for using a uniform set of parameters for all tasks. Simply applying the trick - setting $\gamma=0.9999$ recommended by~\citet{franceschetti2022making}, we observe BAC, SAC, and TD3 all surpass 300 before reaching 1M steps. Notably, BAC still outperforms. 
\begin{figure}[h]
    \centering
    \includegraphics[width=3.6cm,keepaspectratio]{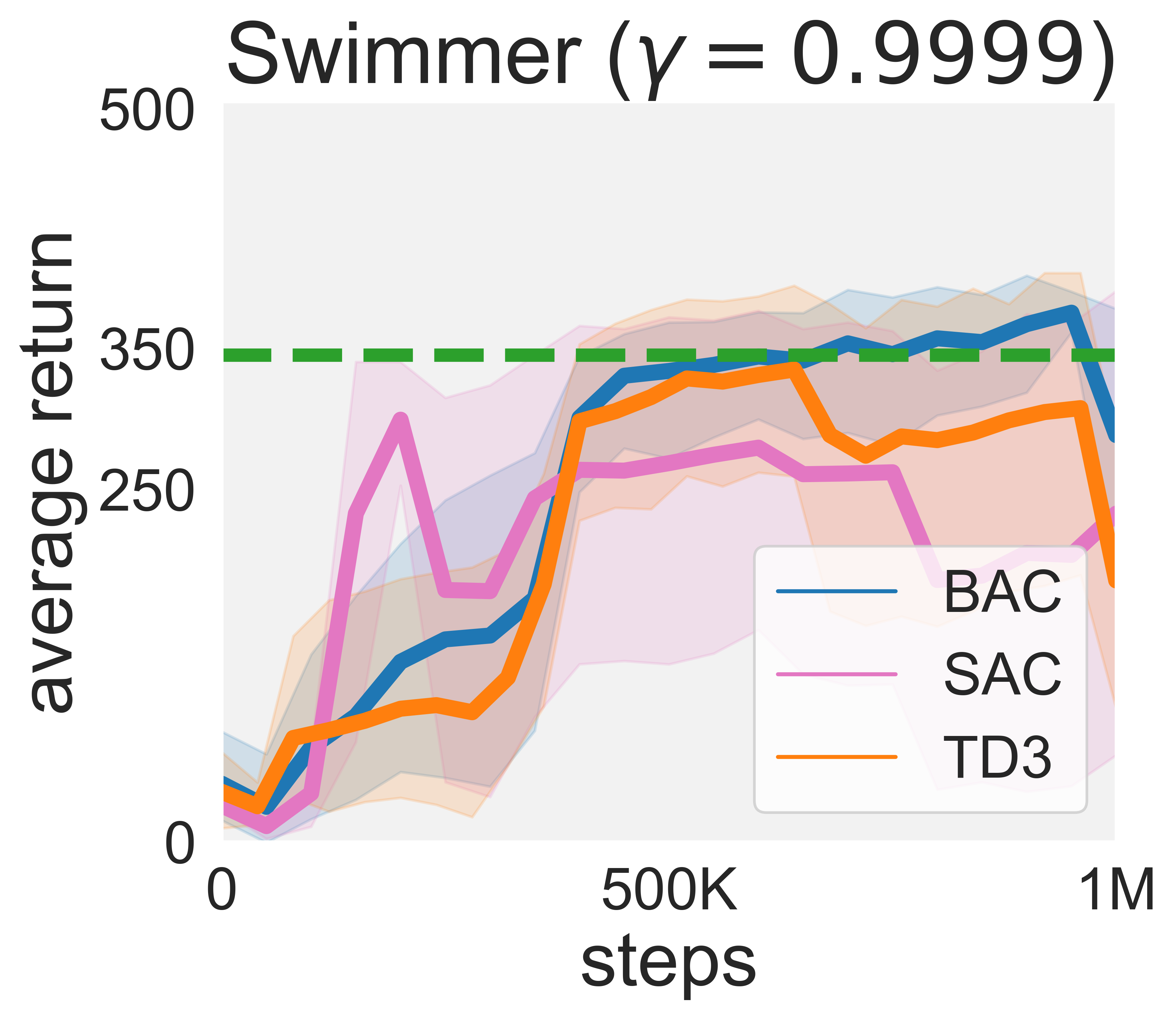}
    \caption{Performance curves of BAC, SAC, TD3 with the discount factor $\gamma=0.9999$ for Swimmer-v2 task. Run on 10 seeds. }
    \label{fig:swimmer-high-gamma-res}
\end{figure}

\end{document}